\documentclass[sigconf,balance,screen]{acmart}
\copyrightyear{2021}
\acmYear{2021}
\setcopyright{acmlicensed}\acmConference[FAccT '21]{Conference on Fairness, Accountability, and Transparency}{March 3--10, 2021}{Virtual Event, Canada}
\acmBooktitle{Conference on Fairness, Accountability, and Transparency (FAccT '21), March 3--10, 2021, Virtual Event, Canada}
\acmPrice{15.00}
\acmDOI{10.1145/3442188.3445910}
\acmISBN{978-1-4503-8309-7/21/03}


\startPage{1}

\PassOptionsToPackage{numbers, compress, sort}{natbib}
\PassOptionsToPackage{dvipsnames,table,prologue}{xcolor}
\usepackage[utf8]{inputenc} %
\usepackage[T1]{fontenc}    %
\usepackage{hyperref}       %
\usepackage{url}            %
\usepackage{booktabs}       %
\usepackage{amsfonts}       %
\usepackage{nicefrac}       %
\usepackage{microtype}      %
\usepackage{amsthm}
\usepackage{graphicx}
\usepackage{subcaption}
\usepackage{bbm}
\usepackage{amsmath}
\usepackage{cleveref}
\usepackage{wrapfig}
\usepackage{mathtools}
\DeclarePairedDelimiter\abs{\lvert}{\rvert}
\usepackage{amsmath}
\usepackage{multirow}

\usepackage{xspace}

\newcommand{\xhdr}[1]{\vspace{1.7mm}\noindent{{\bf #1.}}}
\newcommand{\RB}{\mathit{RB}}
\newcommand{\RBmeasure}{\RB}
\newcommand{\ERM}{\textsc{ERM}}

\newcommand{\ignore}[1]{}

\newtheorem{definition}{Definition}

\newcommand{\yhat}{\hat{y}}

\newcommand{\etc}{\emph{etc.}}
\newcommand{\ie}{\emph{i.e.}}
\newcommand{\eg}{\emph{e.g.}}

 \DeclareMathOperator{\sign}{sign}

\usepackage[font=small,labelfont=bf]{caption}

\newcommand{\jpd}[1]{ \textcolor{red}{[John: #1]} }

\title{Fairness Through Robustness:\\ Investigating Robustness Disparity in Deep Learning}

\begin{document}

\begin{abstract}
Deep neural networks (DNNs) are increasingly used in real-world applications (e.g. facial recognition). This has resulted in concerns about the fairness of decisions made by these models. Various notions and measures of fairness have been proposed to ensure that a decision-making system does not disproportionately harm (or benefit) particular subgroups of the population. In this paper, we argue that traditional notions of fairness that are only based on models' outputs are not sufficient when the model is vulnerable to adversarial attacks. We argue that in some cases, it may be easier for an attacker to target a particular subgroup, resulting in a form of \textit{robustness bias}. 
We show that measuring robustness bias is a challenging task for DNNs and propose two methods to measure this form of bias. We then conduct an empirical study on state-of-the-art neural networks on commonly used real-world datasets such as CIFAR-10, CIFAR-100, Adience, and UTKFace and show that in almost all cases there are subgroups (in some cases based on sensitive attributes like race, gender, etc) which are less robust and are thus at a disadvantage. We argue that this kind of bias arises due to both the data distribution and the highly complex nature of the learned decision boundary in the case of DNNs, thus making mitigation of such biases a non-trivial task. Our results show that robustness bias is an important criterion to consider while auditing real-world systems that rely on DNNs for decision making. Code to reproduce all our results can be found here: \url{https://github.com/nvedant07/Fairness-Through-Robustness}
\end{abstract}

\author{Vedant Nanda}
\authornote{Both authors contributed equally to the paper}
\email{vedant@cs.umd.edu}
\affiliation{
    University of Maryland
    \country{USA}}
\affiliation{MPI-SWS\country{Germany}}

\author{Samuel Dooley}
\authornotemark[1]
\email{sdooley1@cs.umd.edu}
\affiliation{
    University of Maryland
    \country{USA}}

\author{Sahil Singla}
\email{ssingla@cs.umd.edu}
\affiliation{
    University of Maryland
    \country{USA}}

\author{Soheil Feizi}
\email{sfeizi@cs.umd.edu}
\affiliation{
    University of Maryland
    \country{USA}}

\author{John P. Dickerson}
\email{john@cs.umd.edu}
\affiliation{
    University of Maryland
    \country{USA}}

\maketitle

\section{Introduction}
Automated decision-making systems that are driven by data are being used in a variety of different real-world applications. In many cases, these systems make decisions on data points that represent humans (\eg, targeted ads~\cite{speicher2018potential,ribeiro2019microtagging}, personalized recommendations~\cite{singh2018fairness, beiga2018equity}, hiring~\cite{schumann2019diverse, schumann2020hiring}, credit scoring~\cite{khandani2010consumer}, or recidivism prediction~\cite{Chouldechova17:Fair}). In such scenarios, there is often concern regarding the fairness of outcomes of the systems~\cite{barocas2016big,sainyam2017fairness}. This has resulted in a growing body of work from the nascent Fairness, Accountability, Transparency, and Ethics (FATE) community that---drawing on prior legal and philosophical doctrine---aims to define, measure, and (attempt to) mitigate manifestations of unfairness in automated systems~\cite{Chouldechova17:Fair,Feldman15:Certifying,Leben20:Normative,Binns17:Fairness}.

Most of the initial work on fairness in machine learning considered notions that were one-shot and considered the model and data distribution to be static~\cite{zafar2019constraints, zafar2017parity, Chouldechova17:Fair, barocas2016big, dwork2012fairness,  zemel2013learning}. Recently, there has been more work exploring notions of fairness that are dynamic and consider the possibility that the world (\ie, the model as well as data points) might change over time~\cite{heidari2019longterm, heidari2018preventing, hashimoto2018fairness, liu2018delayed}. Our proposed notion of robustness bias has subtle difference from existing one-shot and dynamic notions of fairness in that it requires each partition of the population be equally robust to imperceptible changes in the input (\eg, noise, adversarial perturbations, etc).

We propose a simple and intuitive notion of \textit{robustness bias} which requires subgroups of populations to be equally ``robust.'' Robustness can be defined in multiple different ways~\cite{szegedy2013intriguing, goodfellow2014explaining, papernot2015limitations}. We take a general definition which assigns points that are farther away from the decision boundary higher robustness. Our key contributions are as follows:
\begin{itemize}
    \item We define a simple, intuitive notion of \textbf{\textit{robustness bias}} that requires all partitions of the dataset to be equally robust. We argue that such a notion is especially important when the decision-making system is a deep neural network (DNN) since these have been shown to be susceptible to various attacks~\cite{carlini2017towards,moosavi2016deepfool}. Importantly, our notion depends not only on the outcomes of the system, but also on the distribution of distances of data-points from the decision boundary, which in turn is a characteristic of \emph{both} the data distribution and the learning process.
    \item We propose different methods to \textbf{\textit{measure this form of bias}}. Measuring the exact distance of a point from the decision boundary is a challenging task for deep neural networks which have a highly non-convex decision boundary. This makes the measurement of robustness bias a non-trivial task. In this paper we leverage the literature on adversarial machine learning and show that we can efficiently approximate robustness bias by using adversarial attacks and randomized smoothing to get estimates of a point's distance from the decision boundary. 
    \item We do an in-depth analysis of \textit{robustness bias} on popularly used datasets and models. Through \textit{\textbf{extensive empirical evaluation}} we show that unfairness can exist due to different partitions of a dataset being at different levels of robustness for many state-of-the art models that are trained on common classification datasets. We argue that this form of unfairness can happen due to both the data distribution and the learning process and is an important criterion to consider when auditing models for fairness.
\end{itemize}

\subsection{Related Work}

\xhdr{Fairness in ML} Models that learn from historic data have been shown to exhibit unfairness, \ie, they disproportionately benefit or harm certain subgroups (often a sub-population that shares a common sensitive attribute such as race, gender \etc) of the population~\cite{barocas2016big, Chouldechova17:Fair, khandani2010consumer}. This has resulted in a lot of work on quantifying, measuring and to some extent also mitigating unfairness~\cite{dwork2012fairness, dwork2018fairness, zemel2013learning, zafar2019constraints, zafar2017parity, hardt2016equality, nina2018beyond, tameem2019one, wadsworth2018achieving,Saha20:Measuring, donini2018empirical, calmon2017optimized, kusner2017counterfactual, Kilbertus17:Avoiding, Pleiss17:Fairness, Wang2020Towards}. Most of these works consider notions of fairness that are one-shot---that is, they do not consider how these systems would behave over time as the world (\ie, the model and data distribution) evolves. Recently more works have taken into account the dynamic nature of these decision-making systems and consider fairness definitions and learning algorithms that fare well across multiple time steps~\cite{heidari2019longterm, heidari2018preventing, hashimoto2018fairness, liu2018delayed}. We take inspiration from both the one-shot and dynamic notions, but take a slightly different approach by requiring all subgroups of the population to be equally robust to minute changes in their features. These changes could either be random (\eg natural noise in measurements)  or carefully crafted adversarial noise. This is closely related to~\citet{heidari2019longterm}'s effort-based notion of fairness; however, their notion has a very specific use case of societal scale models whereas our approach is more general and applicable to all kinds of models. Our work is also closely related to and inspired by Zafar et al.'s use of a regularized loss function which captures fairness notions and reduces disparity in outcomes~\cite{zafar2019constraints}. There are major differences in both the {\it approach} and {\it application} between our work and that of Zafar et al's. Their disparate impact formulation aims to equalize the average distance of points to the decision boundary, $\mathbb{E}[d(x)]$; our approach, instead, aims to equalize the number of points that are ``safe'', i.e.,  $\mathbb{E}[\mathbbm{1}\{d(x)>\tau\}]$ (see section~\ref{sec:robustness_bias} for a detailed description). Our proposed metric is preferable for applications of adversarial attack or noisy data, the focus of our paper; whereas the metric of Zafar et al is more applicable for an analysis of the consequence of a decision in a classification setting.

\xhdr{Robustness} Deep Neural Networks (DNNs) have been shown to be susceptible to carefully crafted adversarial perturbations which---imperceptible to a human---result in a misclassification by the model~\cite{szegedy2013intriguing, goodfellow2014explaining, papernot2015limitations}. In the context of our paper, we use adversarial attacks to approximate the distance of a data point to the decision boundary. For this we use state-of-the-art white-box attacks proposed by~\citet{moosavi2016deepfool} and~\citet{carlini2017towards}. Due to the many works on adversarial attacks, there have been many recent works on provable robustness to such attacks. The high-level goal of these works is to estimate a (tight) lower bound on the distance of a point from the decision boundary~\cite{cohen2019smoothing, salman2019provable, singla2020curvature}. We leverage these methods to estimate distances from the decision boundary which helps assess robustness bias (defined formally in Section~\ref{sec:robustness_bias}).

\xhdr{Fairness and Robustness} Recent works have proposed  poisoning attacks on fairness~\cite{solans2020poisoning, mehrabi2020exacerbating}.~\citet{khani2019noise} analyze why noise in features can cause disparity in error rates when learning a regression. We believe that our work is the very first to show that different subgroups of the population can have different levels of robustness which can lead to unfairness. We hope that this will lead to more work at the intersection of these two important sub fields of ML.

\begin{figure}[b!]
    \centering
        \includegraphics[width=\linewidth]{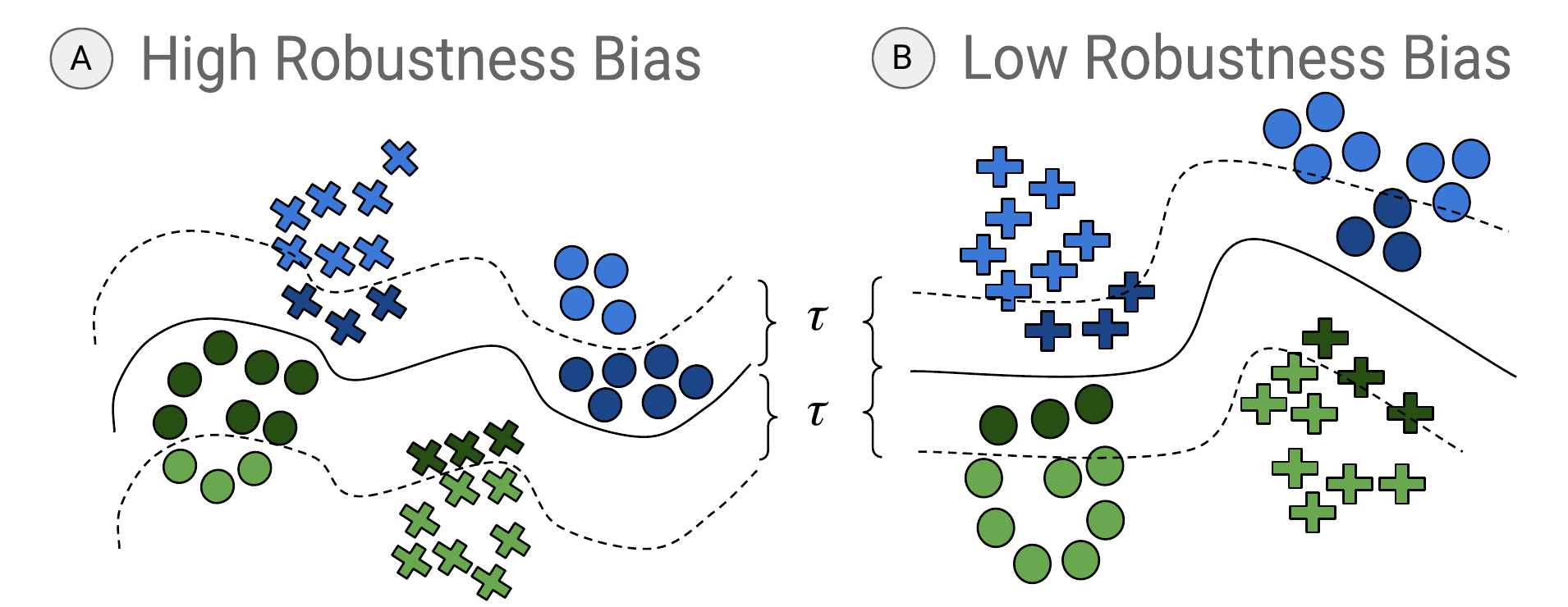}
        \caption{A toy example showing robustness bias. A.) the classifier (solid line) has $100\%$ accuracy for blue and green points. However for a budget $\tau$ (dotted lines), $70\%$ of points belonging to the ``round'' subclass (showed by dark blue and dark green) will get attacked while only $30\%$ of points in the ``cross'' subclass will be attacked. This shows a clear bias against the ``round'' subclass which is less robust in this case. B.) shows a different classifier for the same data points also with $100\%$ accuracy. However, in this case, with the same budget $\tau$, $30\%$ of both ``round'' and ``cross'' subclass will be attacked, thus being less biased.}
        \label{fig:fig1}
\vspace{-3mm}
\end{figure}

\begin{figure*}[t!]
    \centering
    \begin{subfigure}[t]{0.475\textwidth}
        \centering
        \includegraphics[width=.7\linewidth]{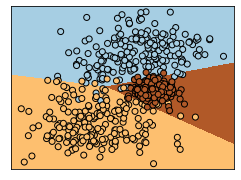}
        \caption{Three-class classification problem for randomly generated data.}
        \label{MLR-example}
    \end{subfigure}%
    ~ 
    \begin{subfigure}[t]{0.5\textwidth}
        \centering
        \includegraphics[width=.7\linewidth]{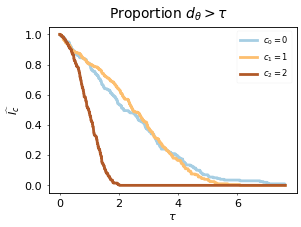}
        \caption{Proportion samples which are greater than $\tau$ away from a decision boundary.}
        \label{MLR-prop}
    \end{subfigure}
    \caption{An example of multinomial logistic regression.}
    \label{fig:MLR}
\end{figure*}

\section{Heterogeneous Robustness}

In a classification setting, a learner is given data $\mathcal{D} = \{(x_i,y_i)\}_{i=1}^N$ consisting of inputs $x_i\in\mathbb{R}^d$ and outputs $y_i\in\mathcal{C}$ which are labels in some set of classes $\mathcal{C}=\{c_1,\dots,c_k\}$. These classes form a partition on the dataset such that $\mathcal{D} = \bigsqcup_{c\in\mathcal{C}} \{(x_i,y_i) \mid y_i=c_j\}$. The goal of learning in decision boundary-based optimization is to draw delineations between points in feature space which sort the data into groups according to their class label. The learning generally tries to maximize the classification accuracy of the decision boundary choice. A learner chooses some loss function $\mathcal{L}$ to minimize on a training dataset, parameterized by parameters $\theta$, while maximizing the classification accuracy on a test dataset.

Of course there are other aspects to classification problems that have recently become more salient in the machine learning community. Considerations about the fairness of classification decisions, for example, are one such way in which additional constraints are brought into a learner's optimization strategy. In these settings, the data $\mathcal{D}=\{(x_i,y_i,s_i)\}_{i=1}^N$ is imbued with some metadata which have a sensitive attribute $\mathcal{S}=\{s_1,\dots,s_t\}$ associated with each point. Like the classes above, these sensitive attributes form a partition on the data such that $\mathcal{D} = \bigsqcup_{s\in\mathcal{S}} \{(x_i,y_i,s_i) \mid s_i = s\}$. Without loss of generality, we assume a single sensitive attribute. Generally speaking, learning with fairness in mind considers the output of a classifier based off of the partition of data by the sensitive attribute, where some objective behavior, like minimizing disparate impact or treatment~\cite{zafar2019constraints}, is integrated into the loss function or learning procedure to find the optimal parameters $\theta$.

There is not a one-to-one correspondence between decision boundaries and classifier performance. For any given performance level on a test dataset, there are infinitely many decision boundaries which produce the same performance, see Figure \ref{fig:fig1}. This raises the question: \emph{if we consider all decision boundaries or model parameters which achieve a certain performance, how do we choose among them? What are the properties of a desirable, high-performing decision boundary?} As the community has discovered, one \emph{undesirable} characteristic of a decision boundary is its proximity to data which might be susceptible to adversarial attack~\cite{goodfellow2014explaining, szegedy2013intriguing, papernot2015limitations}.  This provides intuition that we should prefer boundaries that are as far away as possible from example data ~\cite{suykens1999least, boser1992svm}.

Let us look at how this plays out in a simple example. In multinomial logistic regression, the decision boundaries are well understood and can be written in closed form. This makes it easy for us to compute how close each point is to a decision boundary. Consider for example a dataset and learned classifier as in Figure \ref{MLR-example}. For this dataset, we observe that the brown class, as a whole, is closer to a decision boundary than the yellow or blue classes. We can quantify this by plotting the proportion of data that are greater than a distance $\tau$ away from a decision boundary, and then varying $\tau$. Let $d_\theta(x)$ be the minimal distance between a point $x$ and a decision boundary corresponding to parameters $\theta$. For a given partition $\mathcal{P}$ of a dataset, $\mathcal{D}$, such that $\mathcal{D} = \bigsqcup_{P\in\mathcal{P}} P$, we define the function:  
\[\widehat{I_P}(\tau)=\frac{|\{(x,y)\in P \mid d_\theta(x) > \tau, y = \yhat\} |}{|P|}\]
If each element of the partition is uniquely defined by an element, say a class label, $c$, or a sensitive attribute label, $s$, we equivalently will write $\widehat{I_{c}}(\tau)$ or $\widehat{I_s}(\tau)$ respectively. We plot this over a range of $\tau$ in Figure \ref{MLR-prop} for the toy classification problem in Figure \ref{MLR-example}. Observe that the function for the brown class decreases significantly faster than the other two classes, quantifying how much closer the brown class is to the decision boundary.

\begin{figure*}[t!]
    \centering
        \includegraphics[width=0.75\linewidth]{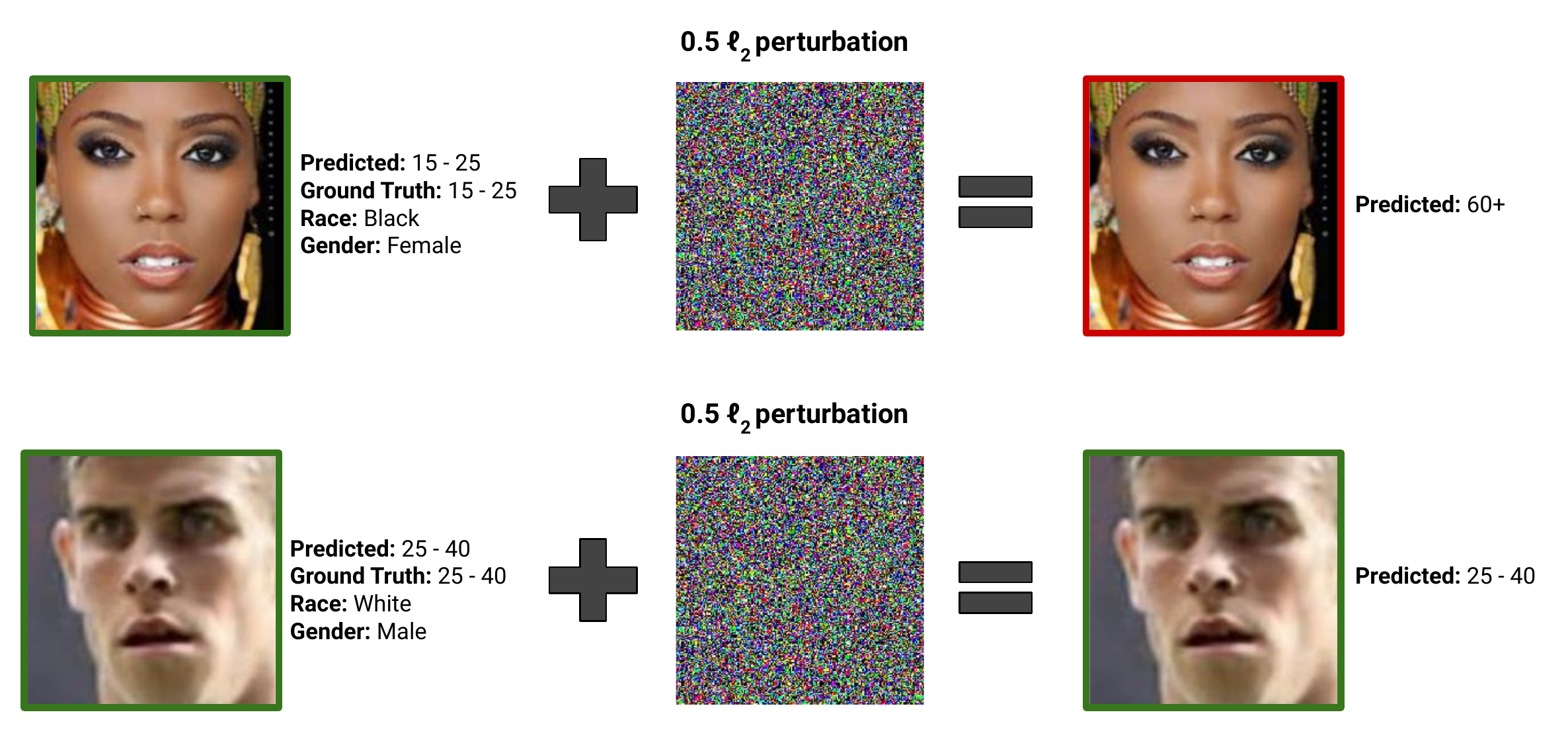}
        \caption{An example of robustness bias in the UTKFace dataset. A model trained to predict age group from faces is fooled for an inputs belonging to certain subgroups (black and female in this example) for a given perturbation, but is robust for inputs belonging to other subgroups (white and male in this example) for the \textit{same magnitude} of perturbation. We use the UTKFace dataset to make a broader point that robustness bias can cause harms. In the specific case of UTKFace (and similar datasets), the task definition of predicting age from faces itself is flawed, as has been noted in many previous studies~\cite{cramer2019challenges,crawford2019excavating,buolamwini2018gender}.}
        \label{fig:real_world_robustness_bias}
\end{figure*}

From a strictly classification accuracy point of view, the brown class being significantly closer to the decision boundary is not of concern; all three classes achieve similar classification accuracy. However, when we move away from this toy problem and into neural networks on real data, this difference between the classes could become a potential vulnerability to exploit, particularly when we consider adversarial examples.

\section{Robustness Bias}\label{sec:robustness_bias}

Our goal is to understand how susceptible different classes are to perturbations (e.g., natural noise, adversarial perturbations). Ideally, no one class would be more susceptible than any other, but this may not be possible.  We have observed that for the same dataset, there may be some classifiers which have differences between the distance of that partition to a decision boundary; and some which do not. There may also be one partition $\mathcal{P}$ which exhibits this discrepancy, and another partition $\mathcal{P}'$ which does not. Therefore, we make the following statement about robustness bias:

\begin{definition}\label{def:robustness-bias}
A dataset $\mathcal{D}$ with a partition $\mathcal{P}$ and a classifier parameterized by $\theta$ exhibits {\bf robustness bias} if there exists an element $P\in\mathcal{P}$ for which the elements of $P$ are either significantly closer to (or significantly farther from) a decision boundary than elements not in $P$.
\end{definition}

A partition $\mathcal{P}$ may be based on sensitive attributes such as race, gender, or ethnicity---or other class labels. For example, given a classifier and dataset with sensitive attribute ``race'', we might say that classifier exhibits robustness bias if, partitioning on that sensitive attribute, for some value of ``race'' the average distance of members of that particular racial value are substantially closer to the decision boundary than other members.

We might say that a dataset, partition, and classifier do not exhibit robustness bias 
if for all $P,P'\in\mathcal{P}$ and all $\tau>0$

\begin{equation}
\begin{split}
    \mathbb{P}_{(x,y)\in\mathcal{D}}\{d_\theta (x) > \tau \mid x\in P , y = \yhat \} \approx \\
    \mathbb{P}_{(x,y)\in\mathcal{D}}\{d_\theta (x) > \tau \mid x\in P', y = \yhat \}. 
\end{split}
\end{equation}

Intuitively, this definition requires that for a given perturbation budget $\tau$ and a given partition $P$, one should not have any incentive to perturb data points from $P$ over points that do not belong to $P$.
Even when examining this criteria, we can see that this might be particularly hard to satisfy. Thus, we want to quantify the disparate susceptibility of each element of a partition to adversarial attack, i.e., how much farther or closer it is to a decision boundary when compared to all other points. We can do this with the following function for a dataset $\mathcal{D}$ with partition element $P\in\mathcal{P}$ and classifier parameterized by $\theta$:

\begin{equation}
\begin{split}
    \RBmeasure{(P,\tau)} = \, | \, \mathbb{P}_{x\in\mathcal{D}}\{ d_\theta (x) > \tau \mid x\in P, y = \yhat\} - \\ \mathbb{P}_{x\in\mathcal{D}}\{d_\theta (x) > \tau \mid x\notin P, y = \yhat\} \, | 
\end{split}
\end{equation}

Observe that $\RBmeasure{(P,\tau)}$ is a large value if and only if the elements of $P$ are much more (or less) adversarially robust than elements not in $P$. 
We can then quantify this for each element $P\in\mathcal{P}$---but a more pernicious variable to handle is $\tau$. We propose to look at the area under the curve $\widehat{I_P}$ for all $\tau$:

\begin{equation}\label{eq:sigma}
    \sigma(P) = \frac{AUC(\widehat{I_P}) - AUC(\sum_{P'\neq P}\widehat{I_{P'}})}{AUC(\sum_{P'\neq P}\widehat{I_{P'}})}
\end{equation}

Note that these notions take into account the distances of data points from the decision boundary and hence are orthogonal and complementary to other traditional notions of bias or fairness (\eg, disparate impact/disparate mistreatment~\cite{zafar2019constraints}, etc). This means that having lower robustness bias does not necessarily come at the cost of fairness as measured by these notions. Consider the motivating example shown in Figure~\ref{fig:fig1}: the decision boundary on the right has lower robustness bias but preserves all other common notions (\eg~\cite{hardt2016equality, dwork2012fairness, zafar2017parity}) as both classifiers maintain $100\%$ accuracy.

\begin{figure*}[t!]
    \centering

        \includegraphics[width=.9\linewidth]{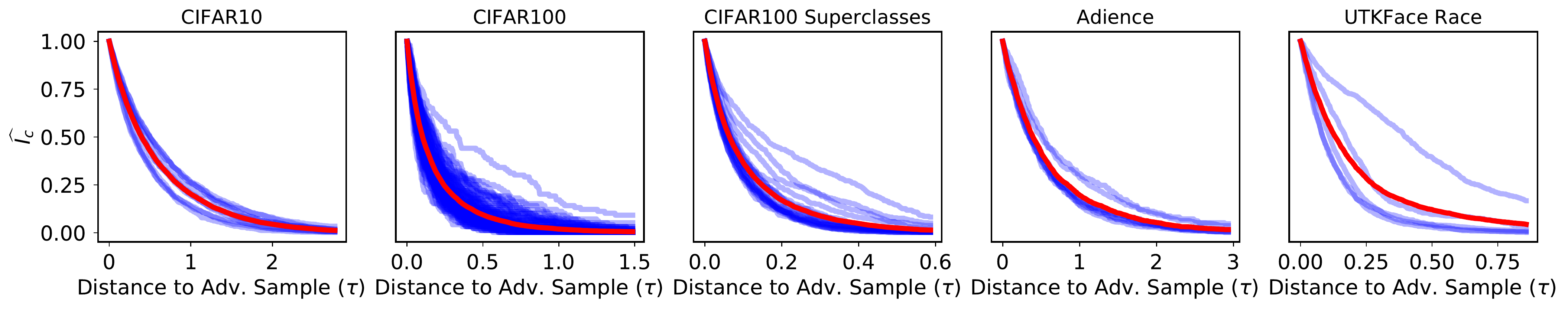}
        \caption{For each dataset, we plot $\widehat{I_c}(\tau)$ for each class $c$ in each dataset. Each blue line  represents one class. The red line represents the mean of the blue lines, i.e., $\sum_{c\in\mathcal{C}} \widehat{I_c}(\tau)$ for each $\tau$. }
        \label{fig:MLR}
\end{figure*}%
\begin{figure*}[t!]
        \centering
        \includegraphics[width=.9\linewidth]{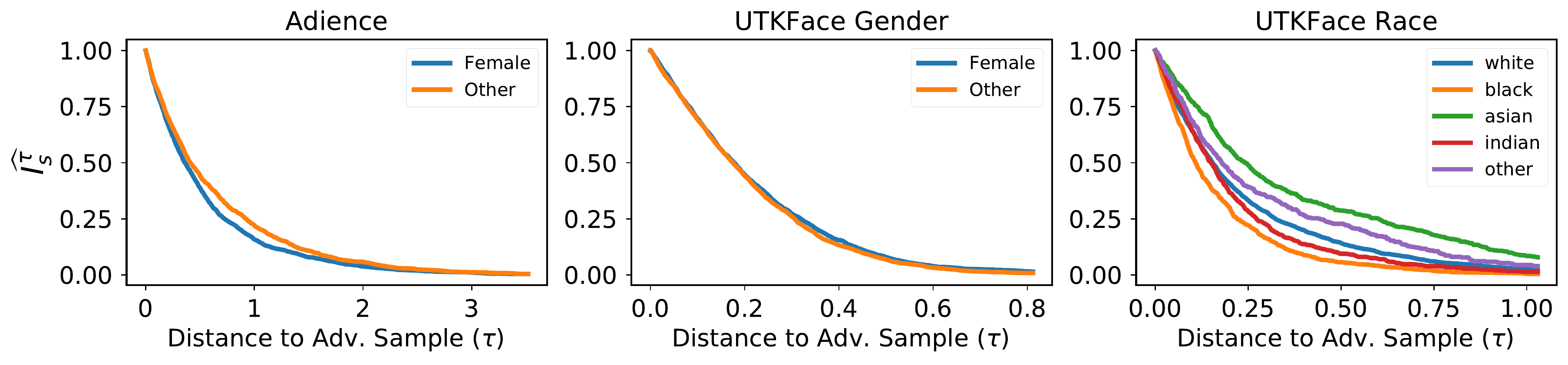}
        \caption{For each dataset, we plot $\widehat{I_s^\tau}$ for each sensitive attribute $s$ in each dataset. Figures for other models can be found in Appendix \ref{Conv results}.}
        \label{fig:MLR-sens}
\end{figure*}

\subsection{Real-world Implications: Degradation of Quality of Service}\label{subsec:robustness_bias_implications}

Deep neural networks are the core of many real world applications, for example, facial recognition, object detection, etc. In such cases, perturbations in the input can occur due to multiple factors such as noise due to the environment or malicious intent by an adversary. Previous works have highlighted how harms can be caused due to the degradation in quality of service for certain sub-populations~\cite{cramer2019challenges, holstein2019improving}. Figure~\ref{fig:real_world_robustness_bias} shows an example of inputs from the UTKFace dataset where an $\ell_2$ perturbation of $0.5$ could change the predicted label for an input with race ``black'' and gender ``female'' but an input with race ``white'' and gender ``male'' was robust to the same magnitude of perturbation. In such a case, the system worked better for a certain sub-group (white, male) thus resulting in unfairness. It is important to note that we use datasets such as Adience and UTKFace (described in detail in section~\ref{sec:experiments}) only to demonstrate the importance of having unbiased robustness. As noted in previous works, the very task of predicting age from a person's face is a flawed task definition with many ethical concerns~\cite{cramer2019challenges, buolamwini2018gender, crawford2019excavating}.

\section{Measuring Robustness Bias}

Robustness bias as defined in the previous section requires a way to measure the distance between a point and the (closest) decision boundary. For deep neural networks in use today, a direct computation of $d_\theta(x)$ is not feasible due to their highly complicated and non-convex decision boundary. However, we show that we can leverage existing techniques from the literature on adversarial attacks to efficiently approximate $d_\theta(x)$. We describe these in more detail in this section.

\subsection{Adversarial Attacks (Upper Bound)}\label{subsec:upper_bounds}

For a given input and model, one can compute an {\it upper bound} on $d_{\theta}(x)$ by performing an optimization which alters the input image slightly so as to place the altered image into a different category than the original. Assume for a given data point $x$, we are able to compute an adversarial image $\tilde{x}$, then the distance between these two images provides an upper bound on distance to a decision boundary, i.e, $\|x-\tilde{x}\| \geq d_\theta(x)$. 

We evaluate two adversarial attacks: DeepFool \cite{moosavi2016deepfool} and CarliniWagner's L2 attack \cite{carlini2017towards}. 
We extend $\widehat{I_P}$ for DeepFool and CarliniWagner as 
\begin{equation}
  \label{eq:measure_df}
  \widehat{I^{DF}_P} = \frac{| \{(x,y)\in P | \tau < \|x-\tilde{x}\|, y=\yhat\} |} { |P|}
\end{equation}
and
\begin{equation}
  \label{eq:measure_cw}
  \widehat{I^{CW}_P} = \frac{| \{(x,y)\in P | \tau < \|x-\tilde{x}\|, y=\yhat\} |} { |P|}
\end{equation}
respectively.
We use similar notation to define $\sigma^{DF}(P)$, 
and $\sigma^{CW}(P)$ ($\sigma$ as defined in Eq~\ref{eq:sigma}).  
While these methods are guaranteed to yield upper bounds on $d_\theta(x)$, they need not yield similar behavior to $\widehat{I_P}$ or $\sigma(P)$. We perform an evaluation of this in Section \ref{eval_of_measures}.

\subsection{Randomized Smoothing (Lower Bound)}\label{subsec:lower_bounds}
Alternatively one can compute a lower bound on $d_\theta(x)$ using techniques from recent works on training provably robust classifiers~\cite{salman2019provable, cohen2019smoothing}. For each input, these methods calculate a radius in which the prediction of $x$ will not change (i.e. the robustness certificate). In particular, we use the {\it randomized smoothing} method \cite{cohen2019smoothing,salman2019provable} since it is scalable to large and deep neural networks and leads to the state-of-the-art in provable defenses. Randomized smoothing transforms the base classifier $f$ to a new smooth classifier $g$ by averaging the output of $f$ over noisy versions of $x$. This new classifier $g$ is more robust to perturbations while also having accuracy on par to the original classifier. It is also possible to calculate the radius $\delta_x$ (in the $\ell_2$ distance) in which, with high probability, a given input's prediction remains the same for the smoothed classifier (i.e. $d_{\theta}(x)\geq \delta_x$). A given input $x$ is then said to be provably robust, with high probability, for a $\delta_x$ $\ell_2$-perturbation where $\delta_x$ is the robustness certificate of $x$.

For each point we use its $\delta_x$, calculated using the method proposed by~\cite{salman2019provable}, as a proxy for $d_{\theta}(x)$. The magnitude of $\delta_x$ for an input is a measure of how robust an input is. Inputs with higher $\delta_x$ are more robust than inputs with smaller $\delta_x$. Again, we extend $\widehat{I_P}$ for Randomized Smoothing as
\begin{equation}
  \label{eq:measure_rs}
  \widehat{I^{RS}_P} = \frac{| \{(x,y)\in P | \tau < \delta_x, y=\yhat\} |} { |P|}
\end{equation}

We use similar notation to define $\sigma^{RS}(P)$ (see Eq~\ref{eq:sigma}).

\section{Empirical Evidence of Robustness Bias in the Wild}\label{sec:experiments}

We hypothesize that there exist datasets and model architectures which exhibit robustness bias. To investigate this claim, we examine several image-based classification datasets and common model architectures.

\xhdr{Datasets and Model Architectures:} We perform these tests of the datasets {\bf CIFAR-10} \cite{krizhevsky2009learning}, {\bf CIFAR-100} \cite{krizhevsky2009learning} (using both 100 classes and 20 super classes), {\bf Adience} \cite{eidinger2014age}, and {\bf UTKFace} \cite{zhifei2017cvpr}. The first two are widely accepted benchmarks in image classification, while the latter two provide significant metadata about each image, permitting various partitions of the data by final classes and sensitive attributes. More details can be found in Appendix~\ref{Conv results}.

Our experiments were performed using PyTorch's torchvision module~\cite{pytorch}. 
We first explore a simple {\bf Multinomial Logistic Regression} model which could be fully analyzed with direct computation of the distance to the nearest decision boundary. For convolutional neural networks, we focus on {\bf Alexnet} \cite{krizhevsky2014one}, {\bf VGG19} \cite{simonyan2014very}, {\bf ResNet50} \cite{he2016deep}, {\bf DenseNet121} \cite{huang2017densely}, and \textbf{ Squeezenet1\_0} \cite{iandola2016squeezenet} which are all available through torchvision. 
We use these models since these are widely used for a variety of tasks. We achieve performance that is comparable to state of the art performance on these datasets for these models.\footnote{See Appendix~\ref{Conv results} Table \ref{tab:conv_results_tbl} for model performances, additional quantitative results and supporting figures.} Additionally we also train some other popularly used dataset specific architectures like a deep convolutional neural network (we call this \textbf{Deep CNN})\footnote{\url{http://torch.ch/blog/2015/07/30/cifar.html}} and \textbf{PyramidNet} ($\alpha=64$, depth=110, no bottleneck)~\cite{han2017pyramidnet} for CIFAR-10. We re-implemented Deep CNN in pytorch and used the publicly available repo to train PyramidNet\footnote{\url{https://github.com/dyhan0920/PyramidNet-PyTorch}}. We use another deep convolutional neural network (which we refer to as \textbf{Deep CNN CIFAR100\footnote{https://github.com/aaron-xichen/pytorch-playground/blob/master/cifar/model.py}} and \textbf{PyramidNet} ($\alpha=48$, depth=164, with bottleneck) for CIFAR-100 and CIFAR-100Super. For Adience and UTKFace we additionally take simple deep convolutional neural networks with multiple convolutional layers each of which is followed by a ReLu activation, dropout and maxpooling. As opposed to architectures from torchvision (which are pre-trained on ImageNet) these architectures are trained from scratch on the respective datasets. We refer to them as \textbf{UTK Classifier} and \textbf{Adience Classifier} respectively. These simple models serve two purposes: they form reasonable baselines for comparison with pre-trained ImageNet models finetuned on the respective datasets, and they allow us to  analyze robustness bias when models are trained from scratch. 

In sections~\ref{sec:adv_attacks} and~\ref{sec:randomized_smoothing} we audit these datasets and the listed models for robustness bias. In section~\ref{sec:exact_computation}, we train logistic regression on all the mentioned datasets and evaluate robustness bias using an exact computation. We then show in section~\ref{sec:adv_attacks} and~\ref{sec:randomized_smoothing} that robustness bias can be efficiently approximated using the techniques mentioned in \ref{subsec:upper_bounds} and \ref{subsec:lower_bounds} respectively for much more complicated models, which are often used in the real world. We also provide a thorough analysis of the types of robustness biases exhibited by some of the popularly used models on these datasets.

\section{Exact Computation in a Simple Model: Multinomial Logistic Regression}\label{sec:exact_computation}

We begin our analysis by studying the behavior of multinomial logistic regression. Admittedly, this is a simple model compared to modern deep-learning-based approaches; however, it enables is to explicitly compute the exact distance to a decision boundary, $d_\theta(x)$. We fit a regression to each of our vision datasets to their native classes and plot $\widehat{I_c}(\tau)$ for each dataset. Figure~\ref{fig:MLR} shows the distributions of $\widehat{I_c}(\tau)$, from which we observe three main phenomena: (1) the general shape of the curves are similar for each dataset, (2) there are classes which are significant outliers from the other classes, and (3) the range of support of the $\tau$ for each dataset varies significantly. We discuss each of these individually.

First, we note that the shape of the curves for each dataset is qualitatively similar. Since the form of the decision boundaries in multinomial logistic regression are linear delineations in the input space, it is fair to assume that this similarity in shape in Figure \ref{fig:MLR} can be attributed to the nature of the classifier. 

Second, there are classes $c$ which indicate disparate treatment under $\widehat{I_c}(\tau)$. The treatment disparities are most notable in UTKFace, the superclass version CIFAR-100, and regular CIFAR-100. This suggests that, when considering the dataset as a whole, these outlier classes are less suceptible to adversarial attack than other classes. Further, in UTKFace, there are some classes that are considerably more susceptible to adversarial attack because a larger proportion of that class is closer to the decision boundaries. 

We also observe that the median distance to decision boundary can vary based on the dataset. The median distance to a decision boundary for each dataset is: 0.40 for CIFAR-10;
0.10 for CIFAR-100; 0.06 for the superclass version of CIFAR-100; 0.38 for Adience; and 0.12 for UTKFace.
This is no surprise as $d_\theta(x)$ depends both on the location of the data points (which are fixed and immovable in a learning environment) and the choice of architectures/parameters.  \ignore{\jpd{Do I know if 0.40 is big or small at this point?  Can I compare distances across architectures and/or datasets?}}

Finally, we consider another partition of the datasets. Above, we consider the partition of the dataset which occurs by the class labels. With the Adience and UTKFace datasets, we have an additional partition by sensitive attributes. Adience admits partitions based off of gender; UTKFace admits partition by gender and ethnicity. We note that Adience and UTKFace use categorical labels for these multidimensional and socially complex concepts. We know this to be reductive and serves to minimize the contextualization within which race and gender derive their meaning \cite{hanna2020towards,buolamwini2018gender}. Further, we acknowledge the systems and notions that were used to reify such data partitions and the subsequent implications and conclusions draw therefrom. We use these socially and systemically-laden partitions to demonstrate that the functions we define, $\widehat{I_P}$ and $\sigma$ depend upon how the data are divided for analysis. To that end, the function $\widehat{I_P}$ is visualized in Figure \ref{fig:MLR-sens}. We observe that the Adience dataset, which exhibited some adversarial robustness bias in the partition on $\mathcal{C}$ only exhibits minor adversarial robustness bias in the partition on $\mathcal{S}$ for the attribute `Female'. On the other hand, UTKFace which had signifiant adversarial robustness bias does exhibit the phenomenon for the sensitive attribute `Black' but not for the sensitive attribute `Female'. 

This emphasizes that adversarial robustness bias is dependant upon the dataset and the partition. We will demonstrate later that it is also dependant on the choice of classifier. First, we talk about ways to approximate $d_\theta(x)$ for more complicated models.

\begin{figure*}
    \centering
    \begin{subfigure}[b]{0.23\textwidth}
        \includegraphics[trim={0cm 0cm 0cm 0cm},clip,width=1\textwidth]{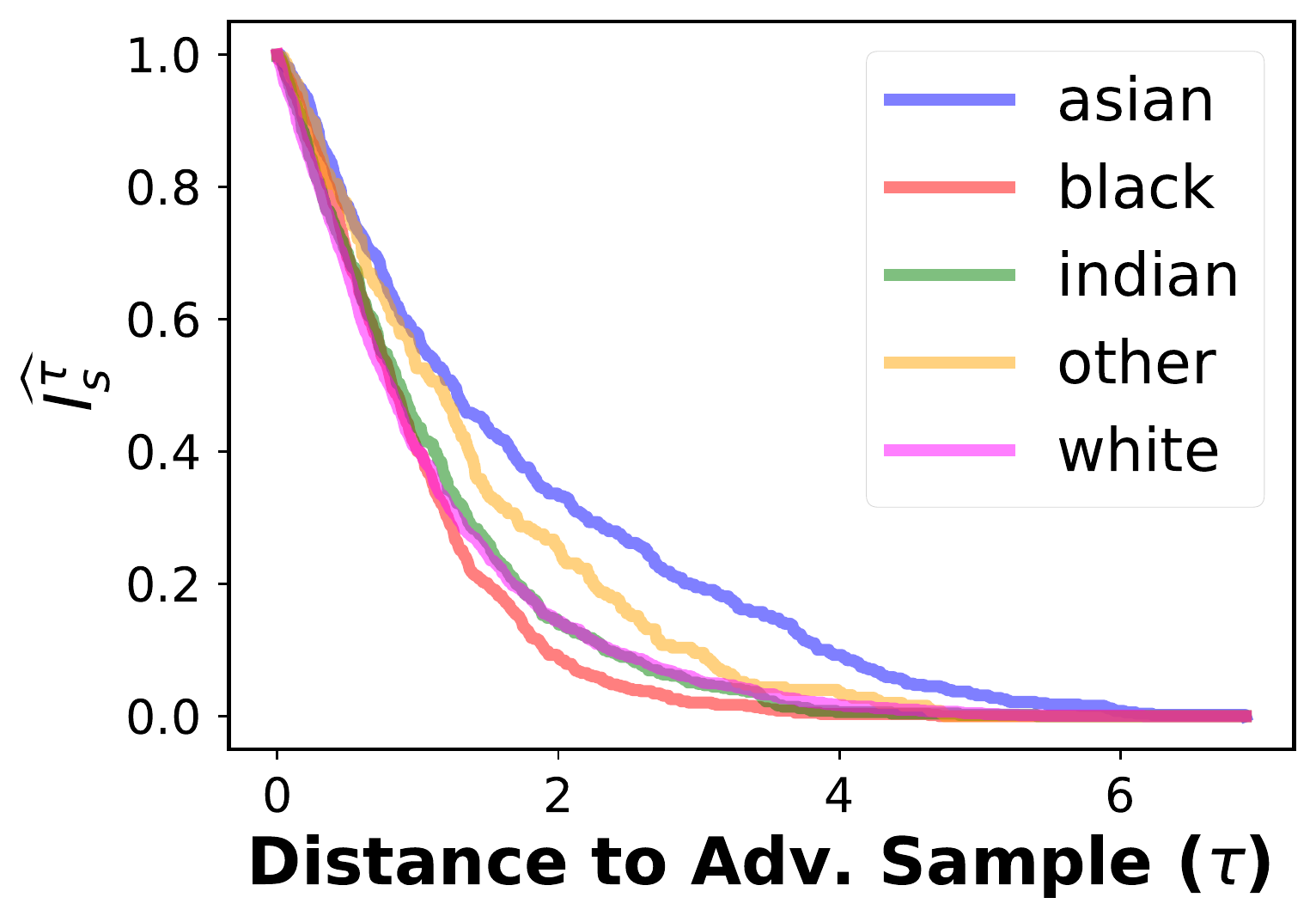}
        \caption{UTK Classifier: DeepFool}
        \label{fig:utkface_race_utk_classifier_df}
    \end{subfigure}
    \begin{subfigure}[b]{0.23\textwidth}
        \includegraphics[trim={0cm 0cm 0cm 0cm},clip,width=1\textwidth]{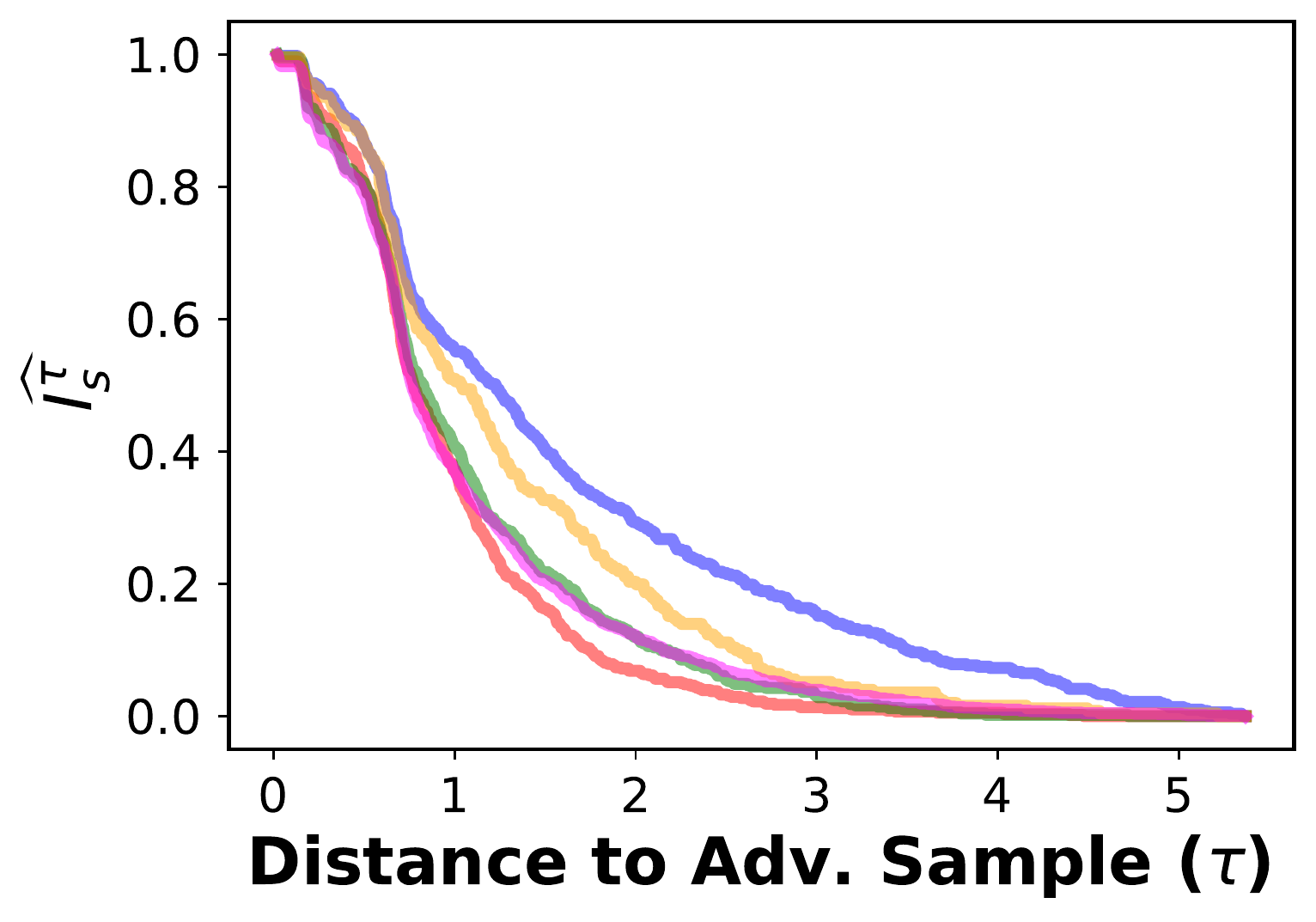}
        \caption{UTK Classifier: CarliniWagner}
        \label{fig:utkface_race_utk_classifier_cw}
    \end{subfigure}
    \begin{subfigure}[b]{0.23\textwidth}
        \includegraphics[trim={0cm 0cm 0cm 0cm},clip,width=1\textwidth]{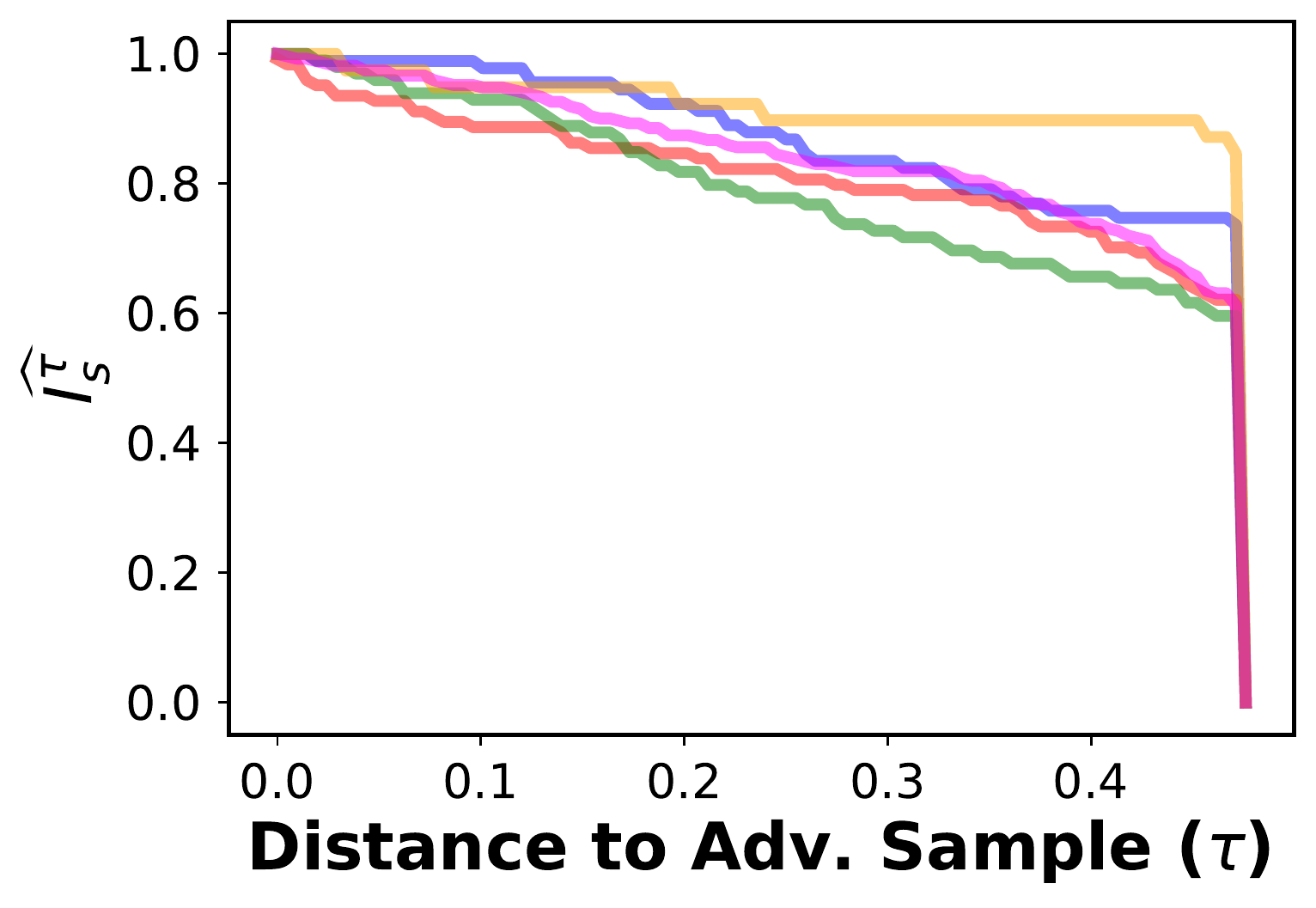}
        \caption{UTK Classifier: Rand. Smoothing}
        \label{fig:lb_utkface_race_utk_classifier}
    \end{subfigure}
    
    \begin{subfigure}[b]{0.23\textwidth}
        \includegraphics[trim={0cm 0cm 0cm 0cm},clip,width=1\textwidth]{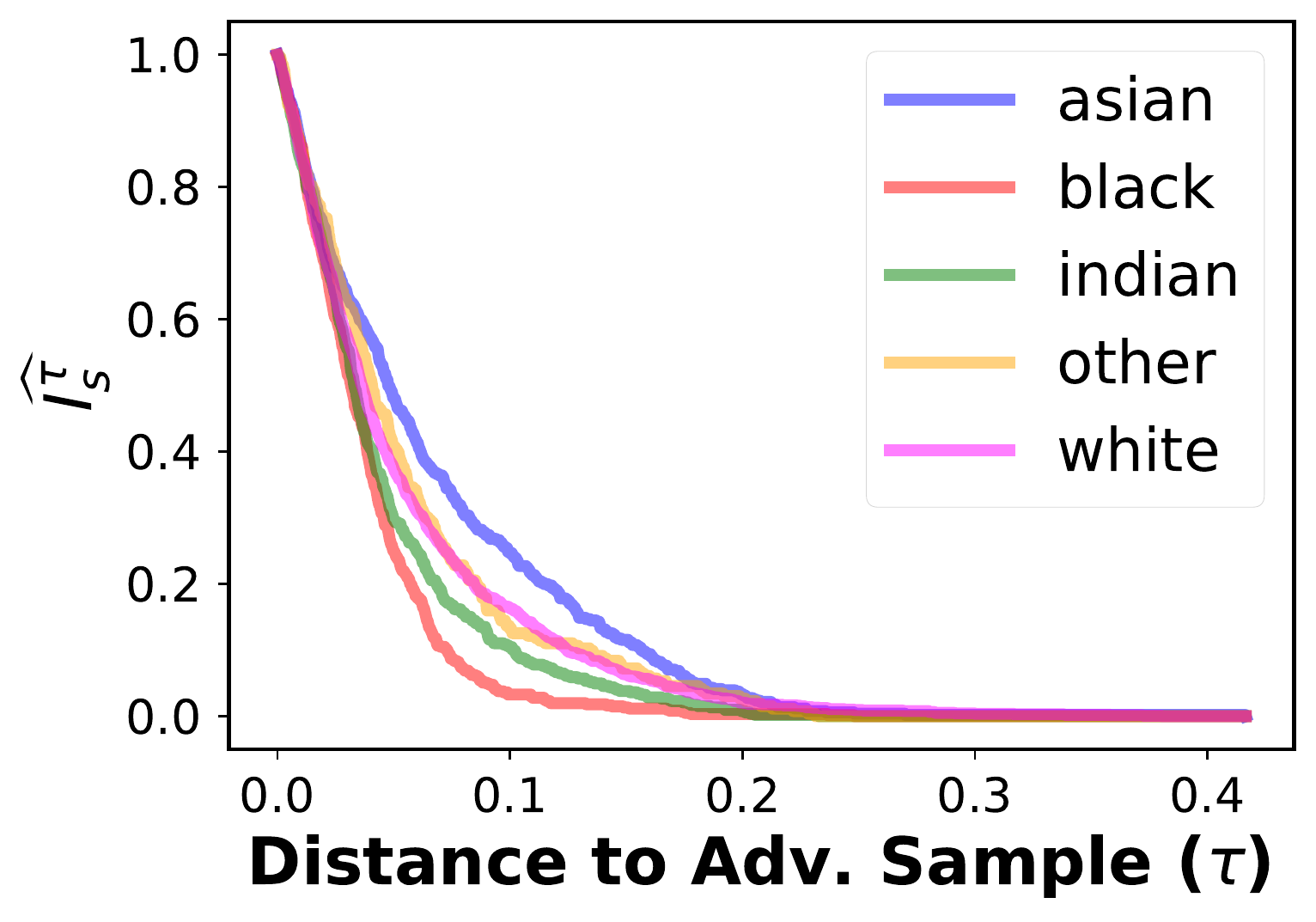}
        \caption{ResNet50: DeepFool}
        \label{fig:utkface_race_resnet_df}
    \end{subfigure}
    \begin{subfigure}[b]{0.23\textwidth}
        \includegraphics[trim={0cm 0cm 0cm 0cm},clip,width=1\textwidth]{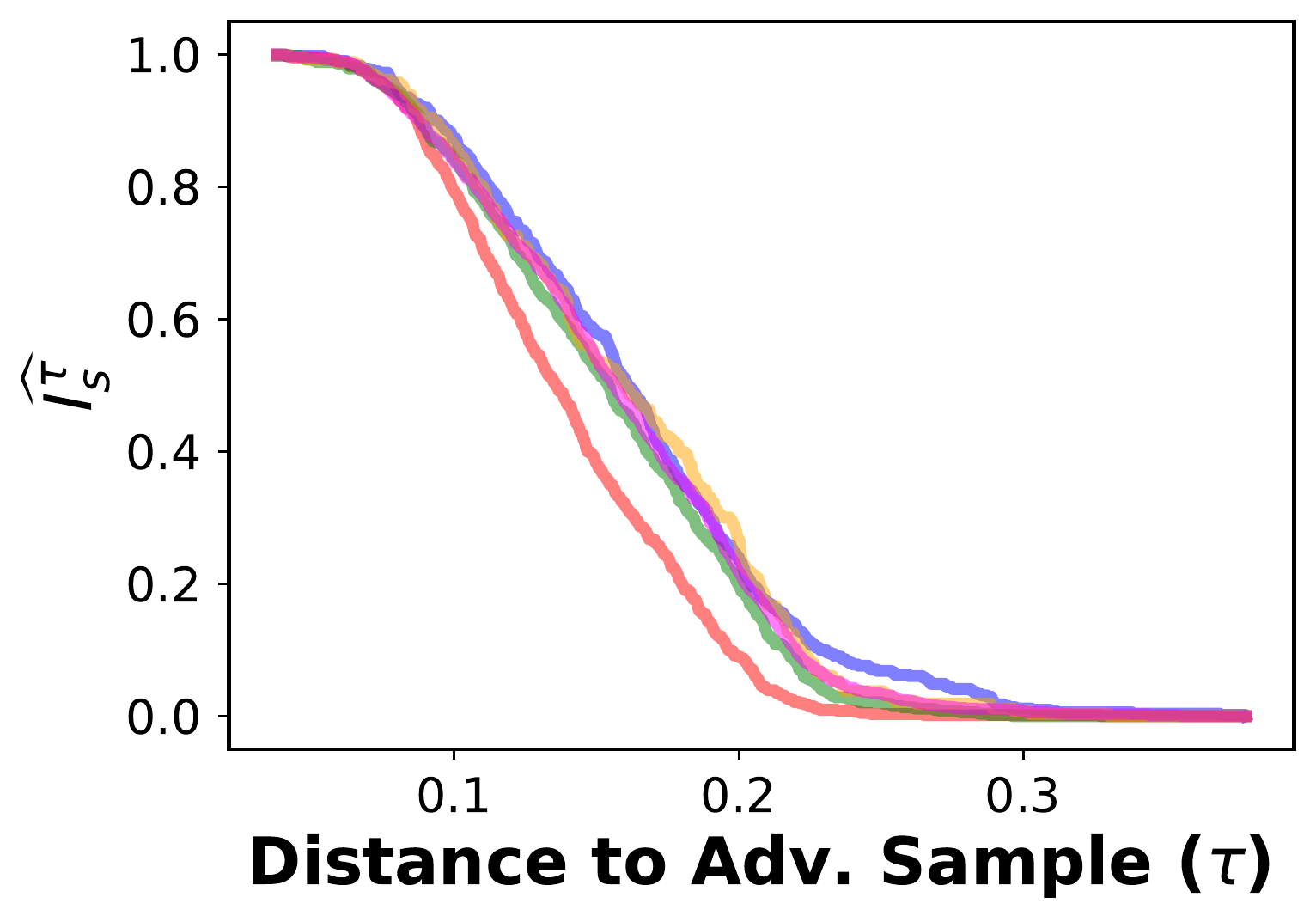}
        \caption{ResNet50: CarliniWagner}
        \label{fig:utkface_race_resnet_cw}
    \end{subfigure}
    \begin{subfigure}[b]{0.23\textwidth}
        \includegraphics[trim={0cm 0cm 0cm 0cm},clip,width=1\textwidth]{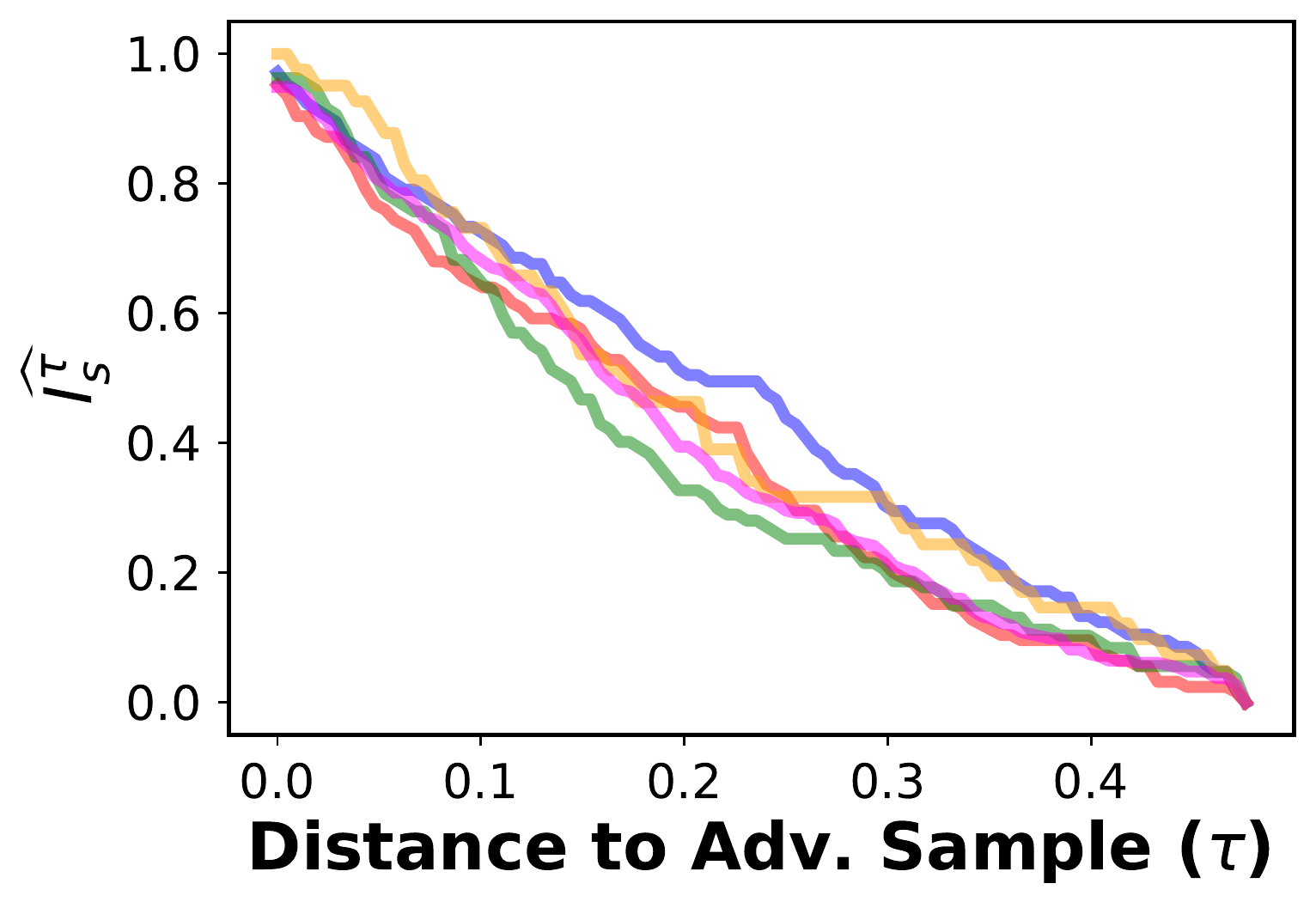}
        \caption{ResNet50: Rand. Smoothing}
        \label{fig:lb_utkface_race_resnet}
    \end{subfigure}
    
    \begin{subfigure}[b]{0.23\textwidth}
        \includegraphics[trim={0cm 0cm 0cm 0cm},clip,width=1\textwidth]{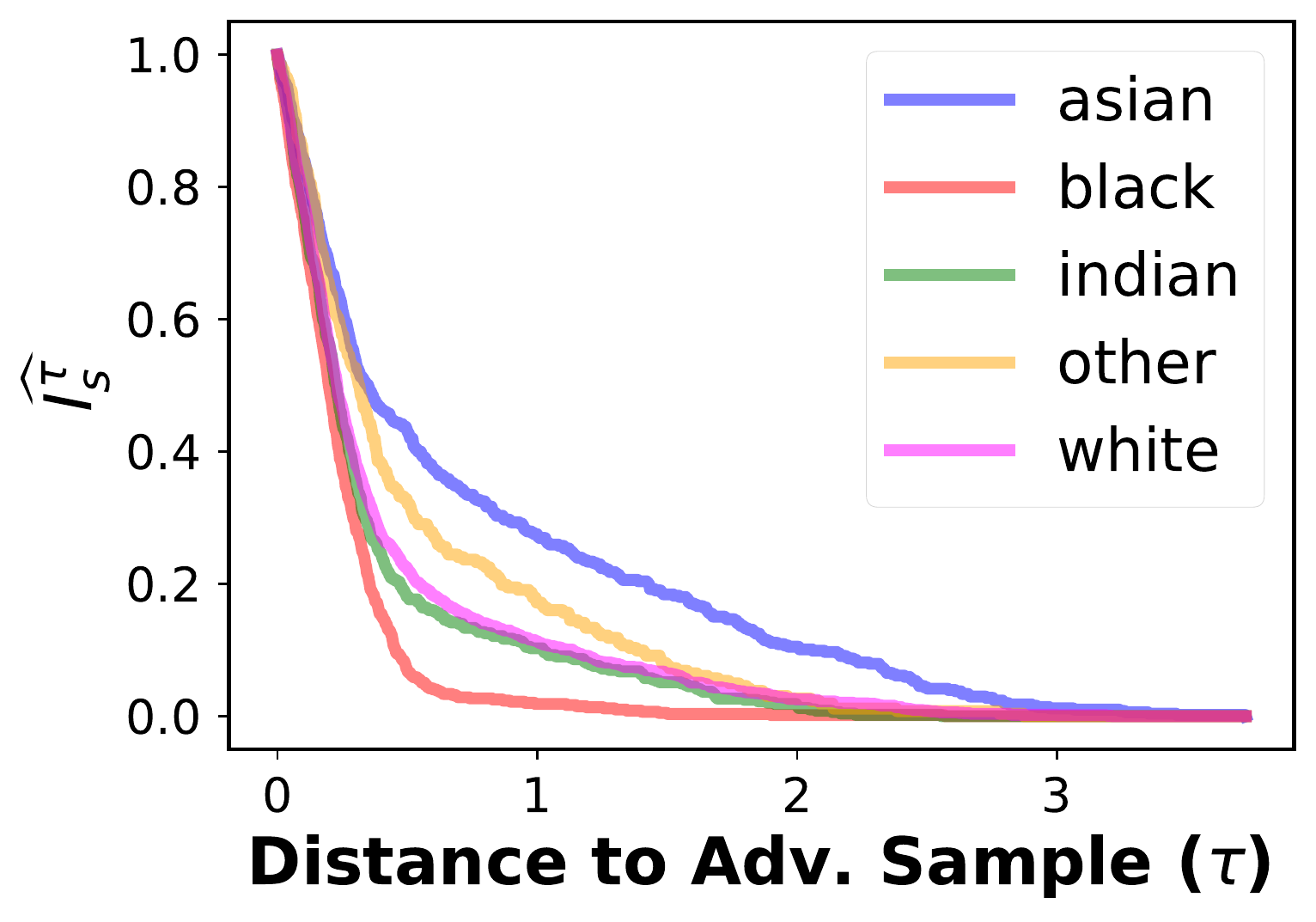}
        \caption{Alexnet: DeepFool}
        \label{fig:utkface_race_alexnet_df}
    \end{subfigure}
    \begin{subfigure}[b]{0.23\textwidth}
        \includegraphics[trim={0cm 0cm 0cm 0cm},clip,width=1\textwidth]{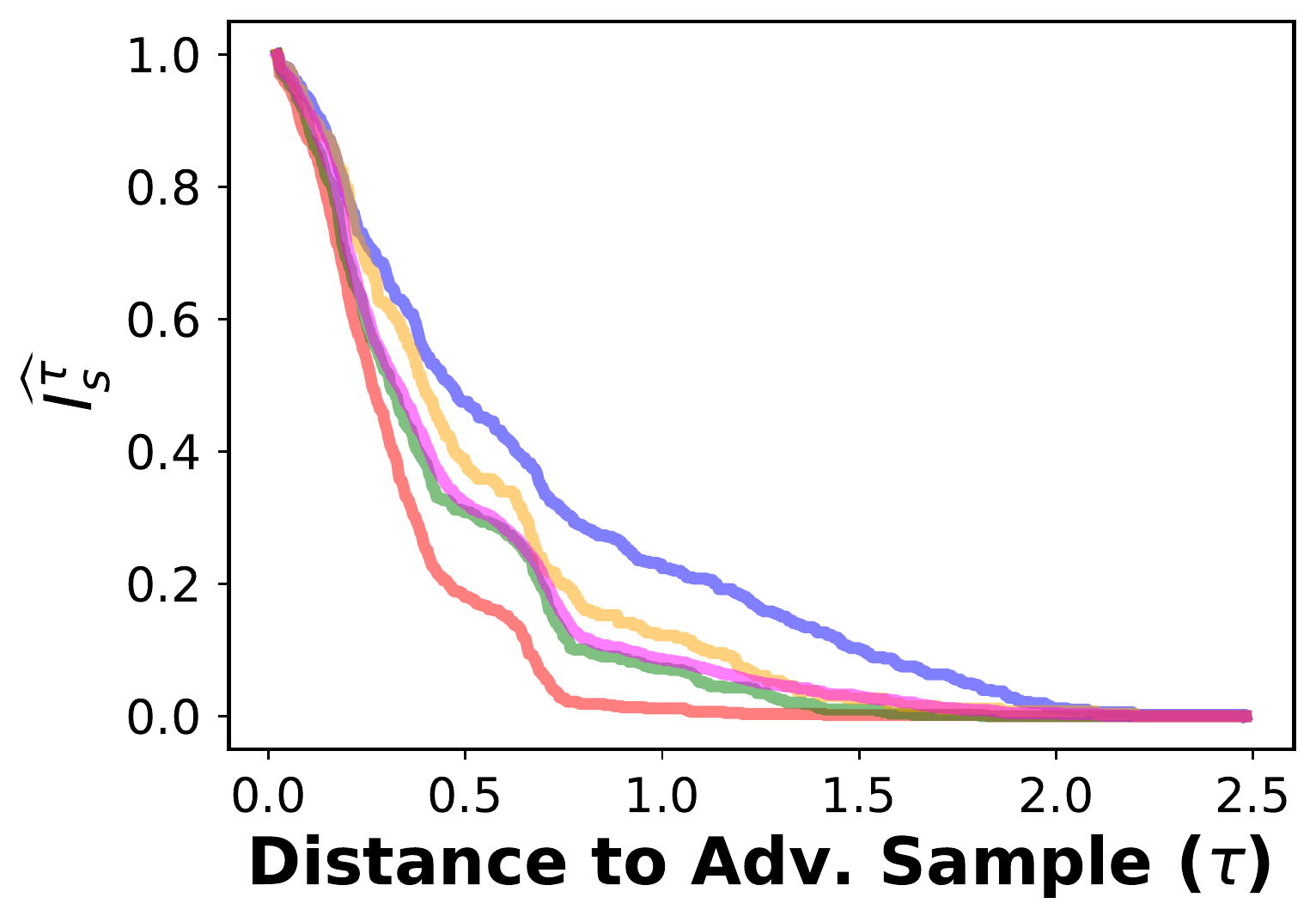}
        \caption{Alexnet: CarliniWagner}
        \label{fig:utkface_race_alexnet_cw}
    \end{subfigure}
    \begin{subfigure}[b]{0.23\textwidth}
        \includegraphics[trim={0cm 0cm 0cm 0cm},clip,width=1\textwidth]{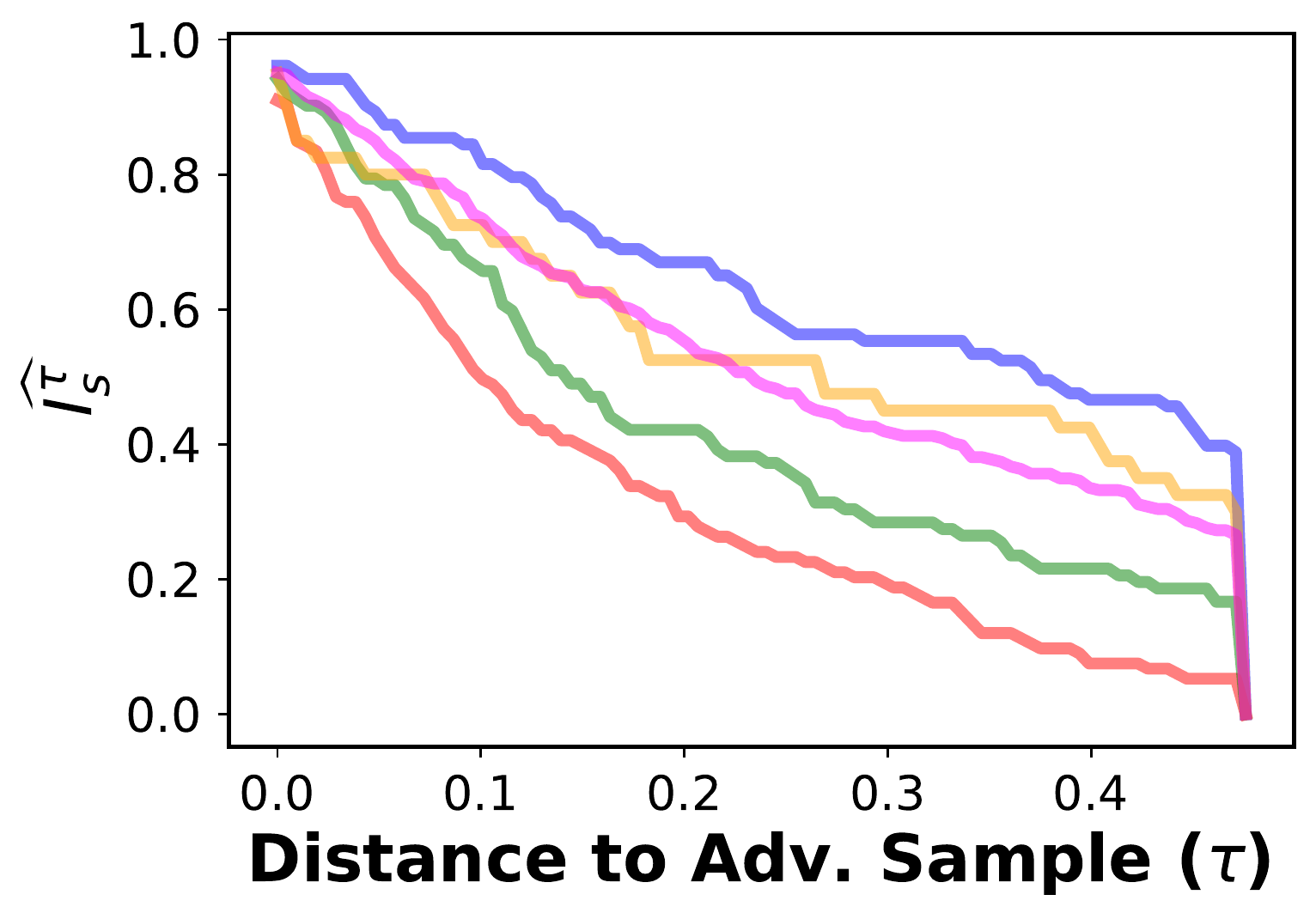}
        \caption{Alexnet: Rand. Smoothing}
        \label{fig:lb_utkface_race_alexnet}
    \end{subfigure}
    
    \begin{subfigure}[b]{0.23\textwidth}
        \includegraphics[trim={0cm 0cm 0cm 0cm},clip,width=1\textwidth]{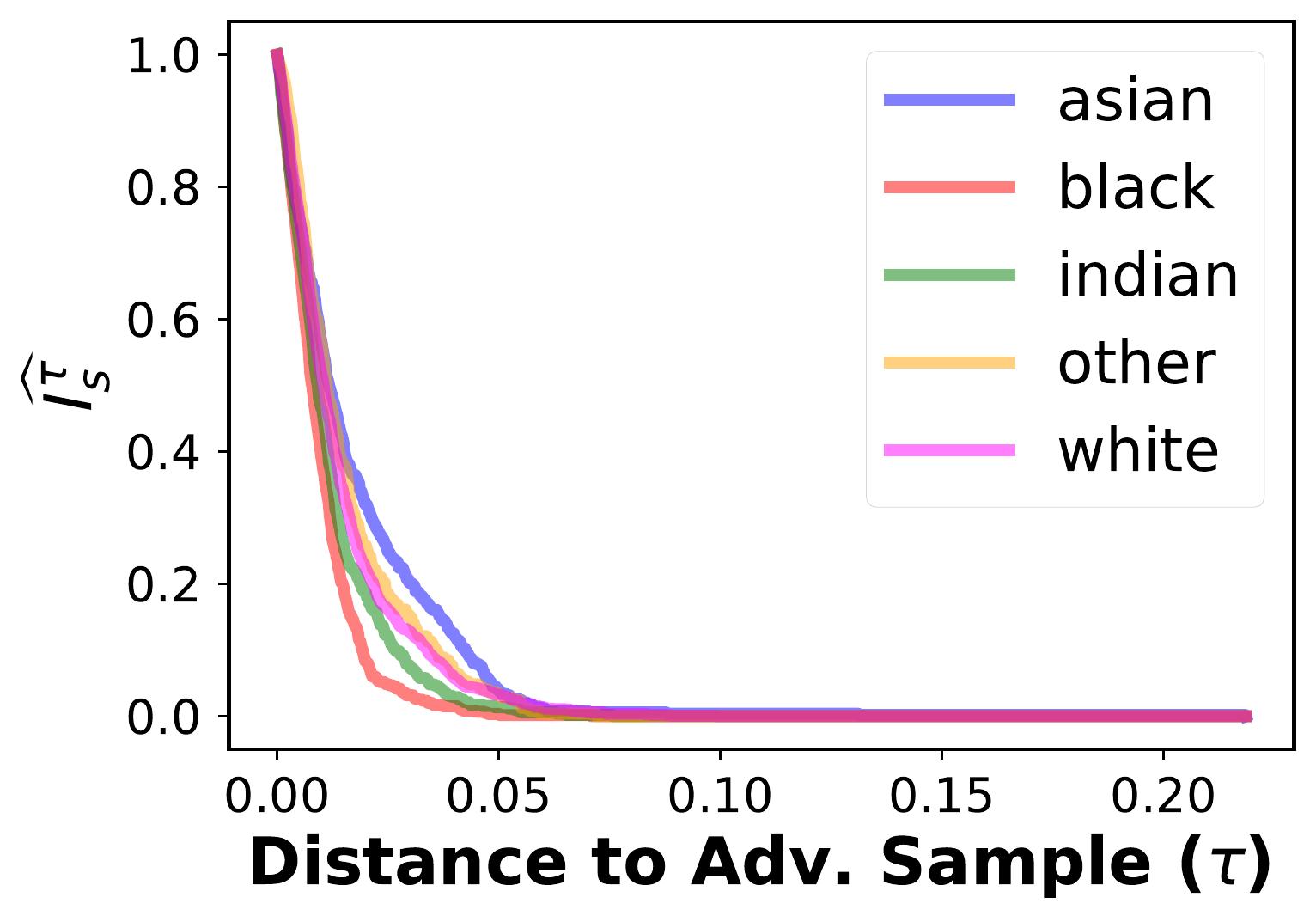}
        \caption{VGG-19: DeepFool}
        \label{fig:utkface_race_vgg_df}
    \end{subfigure}
    \begin{subfigure}[b]{0.23\textwidth}
        \includegraphics[trim={0cm 0cm 0cm 0cm},clip,width=1\textwidth]{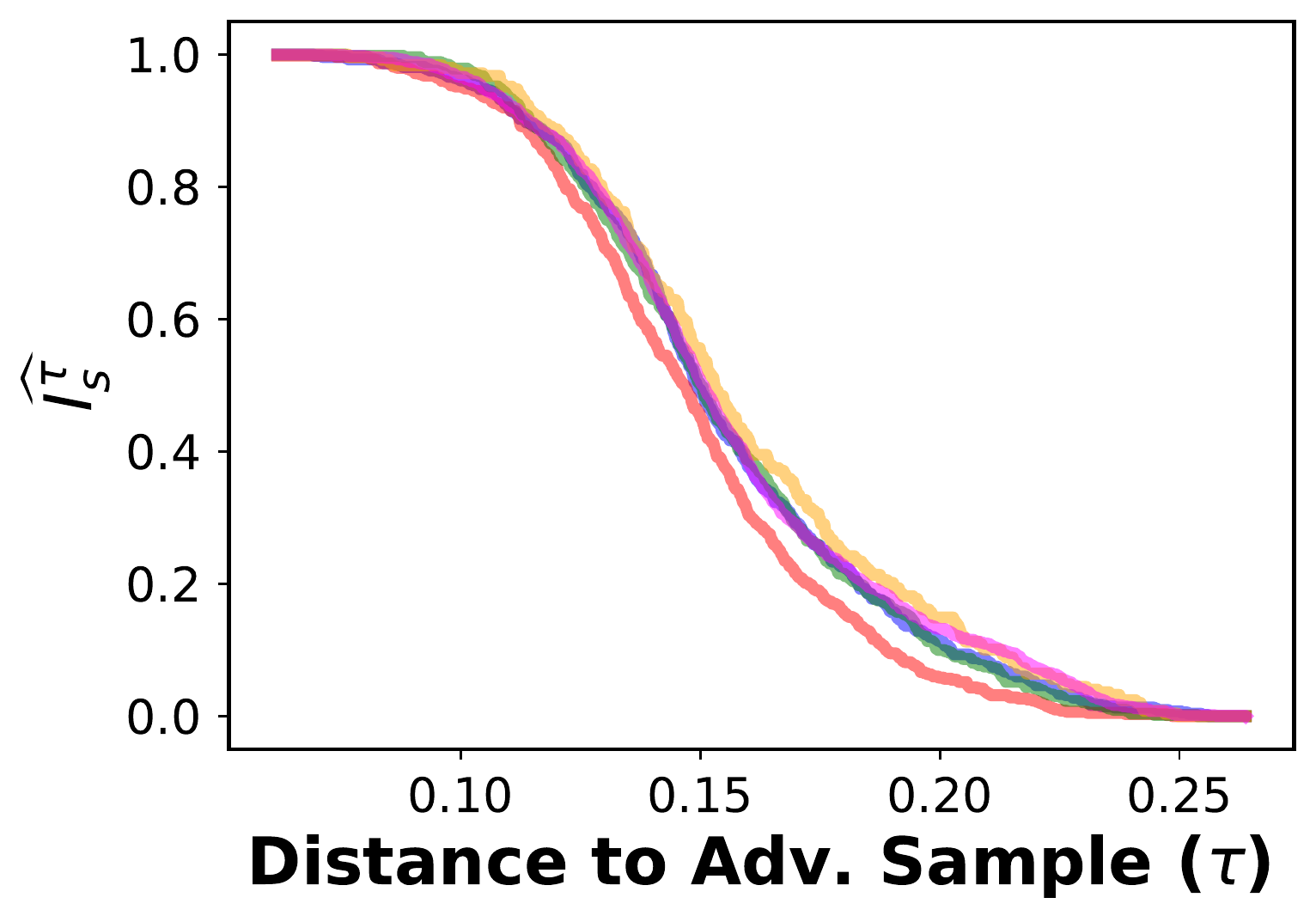}
        \caption{VGG-19: CarliniWagner}
        \label{fig:utkface_race_vgg_cw}
    \end{subfigure}
    \begin{subfigure}[b]{0.23\textwidth}
        \includegraphics[trim={0cm 0cm 0cm 0cm},clip,width=1\textwidth]{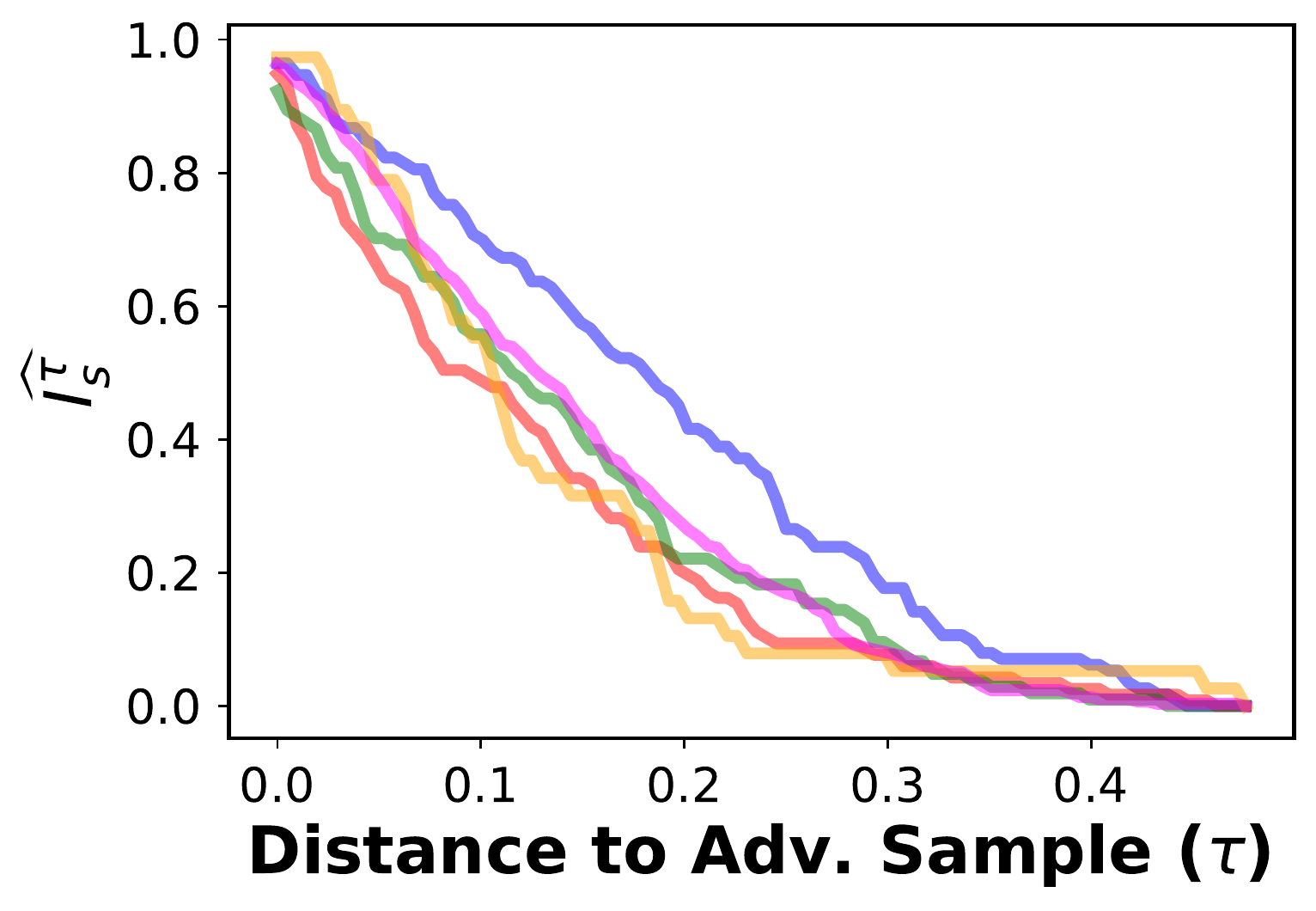}
        \caption{VGG-19: Rand. Smoothing}
        \label{fig:lb_utkface_race_vgg}
    \end{subfigure}
    
    \begin{subfigure}[b]{0.23\textwidth}
        \includegraphics[trim={0cm 0cm 0cm 0cm},clip,width=1\textwidth]{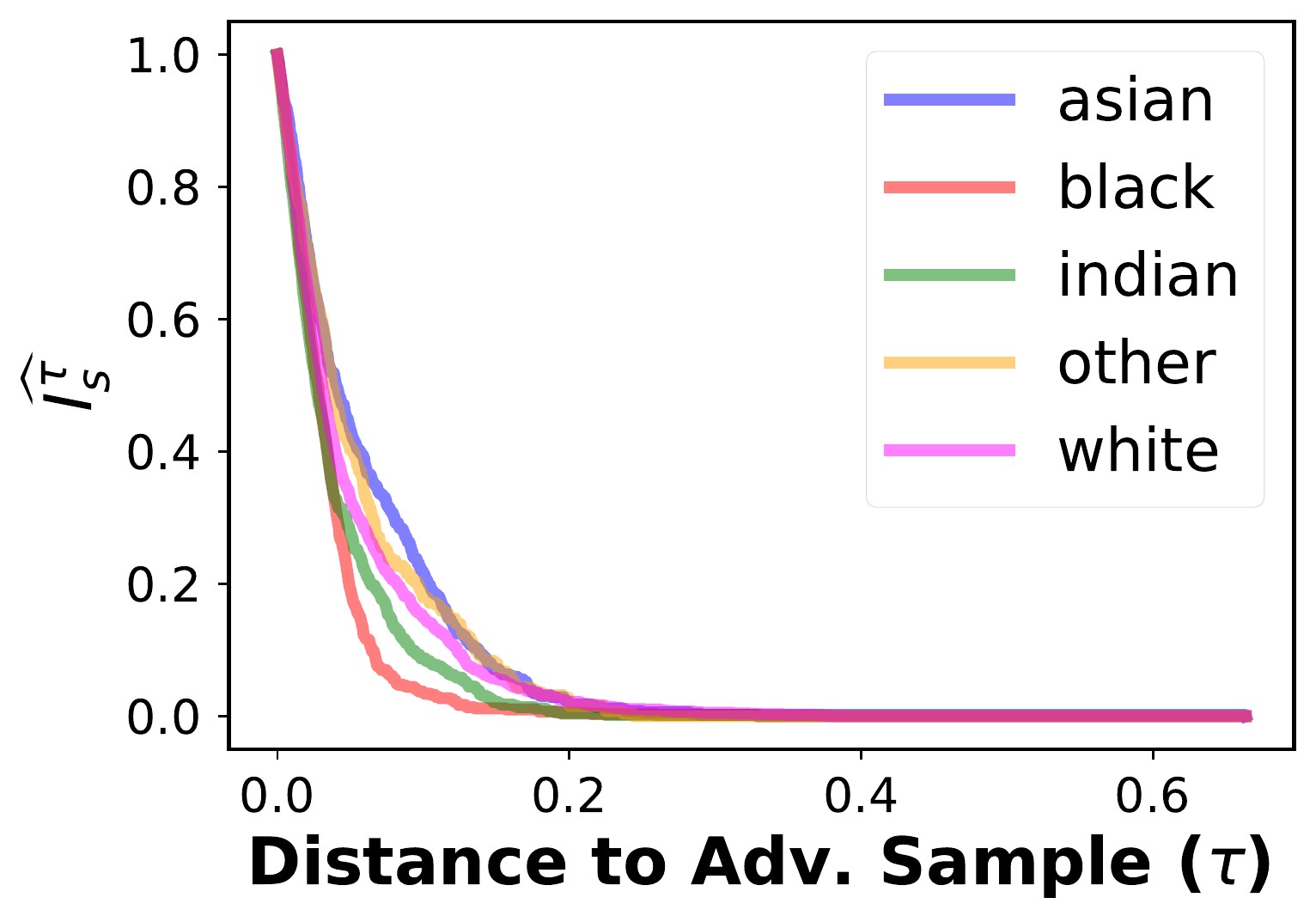}
        \caption{Densenet: DeepFool}
        \label{fig:utkface_race_densenet_df}
    \end{subfigure}
    \begin{subfigure}[b]{0.23\textwidth}
        \includegraphics[trim={0cm 0cm 0cm 0cm},clip,width=1\textwidth]{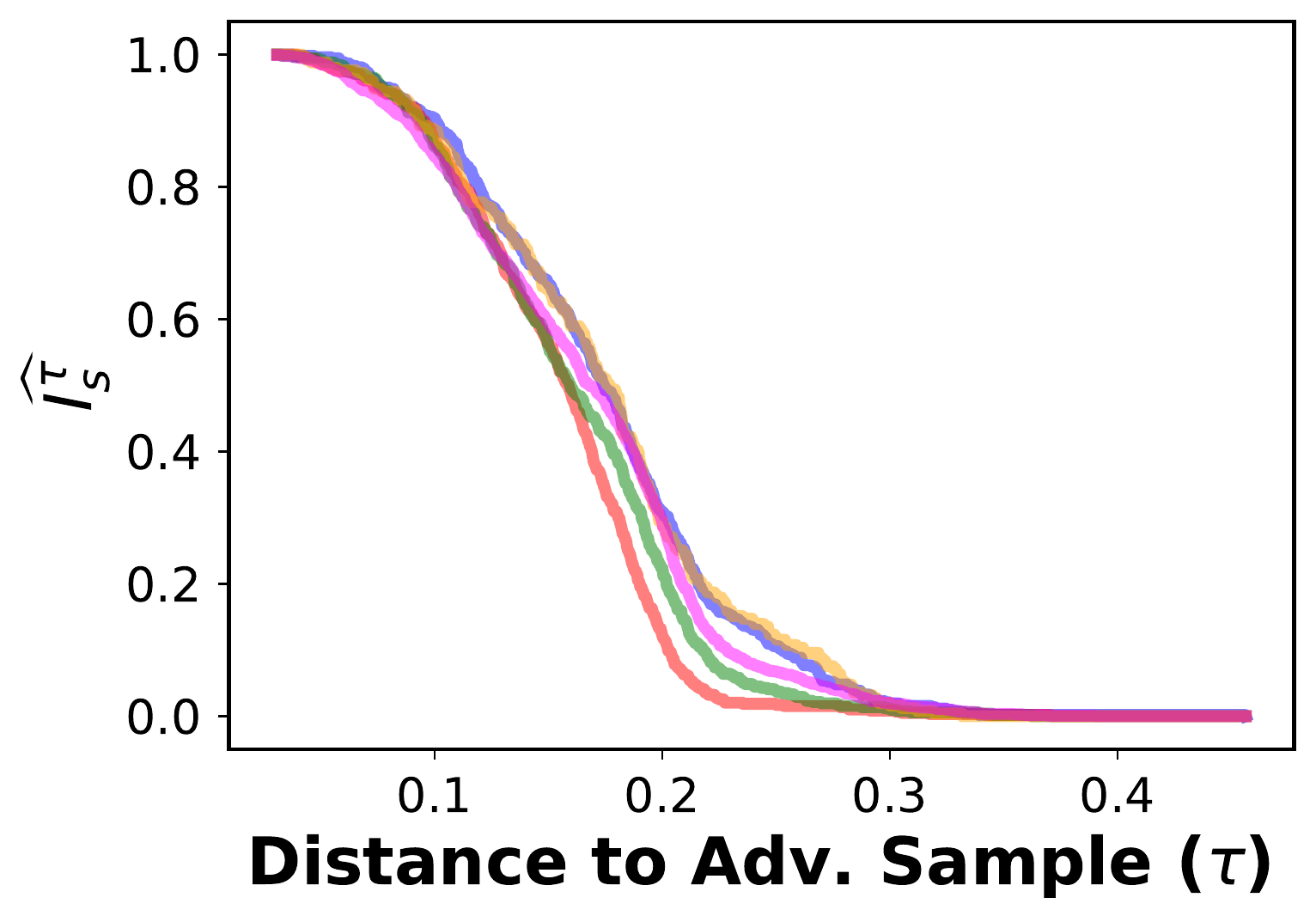}
        \caption{Densenet: CarliniWagner}
        \label{fig:utkface_race_densenet_cw}
    \end{subfigure}
    \begin{subfigure}[b]{0.23\textwidth}
        \includegraphics[trim={0cm 0cm 0cm 0cm},clip,width=1\textwidth]{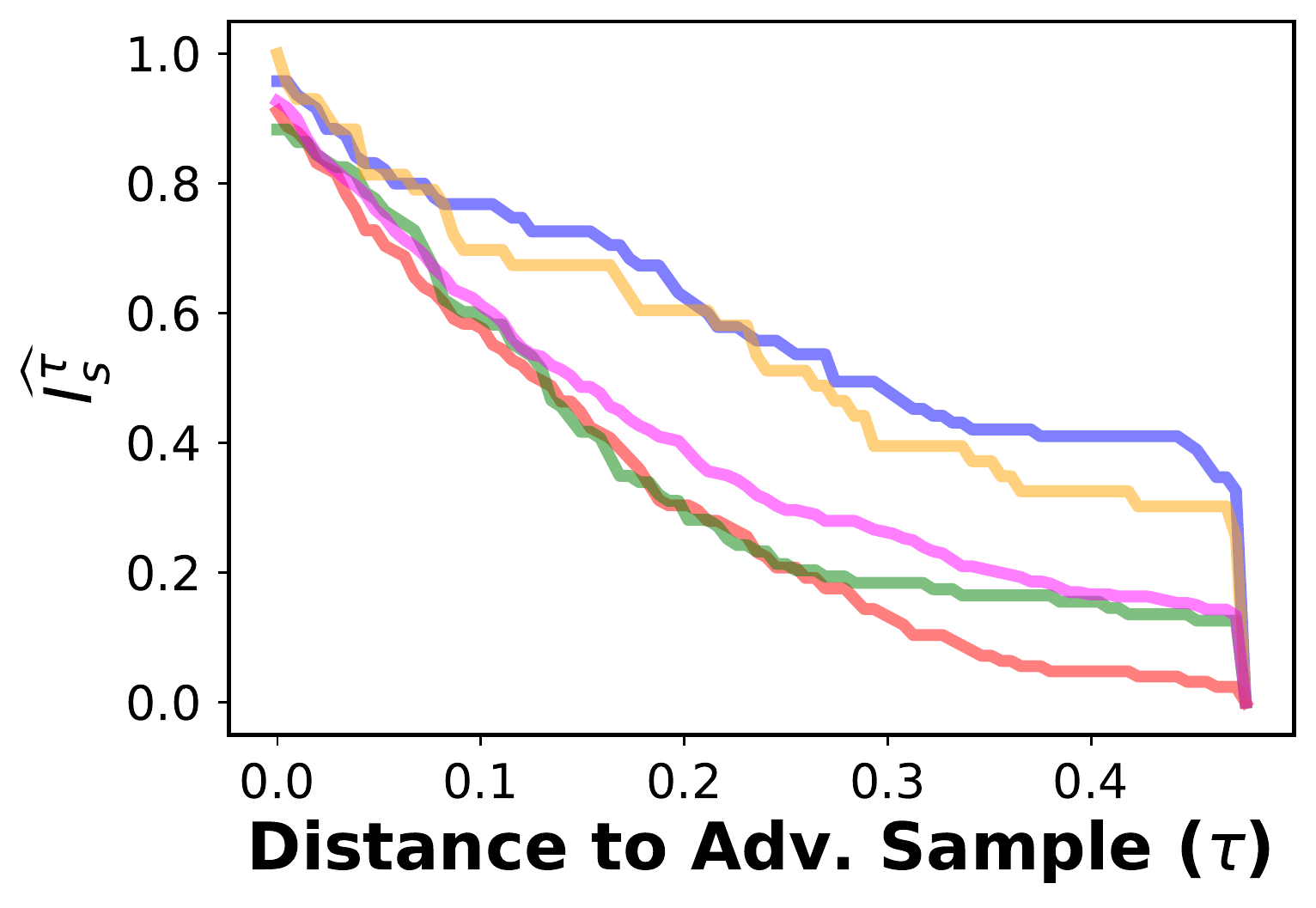}
        \caption{Densenet: Rand. Smoothing}
        \label{fig:lb_utkface_race_densenet}
    \end{subfigure}
    
    \begin{subfigure}[b]{0.23\textwidth}
        \includegraphics[trim={0cm 0cm 0cm 0cm},clip,width=1\textwidth]{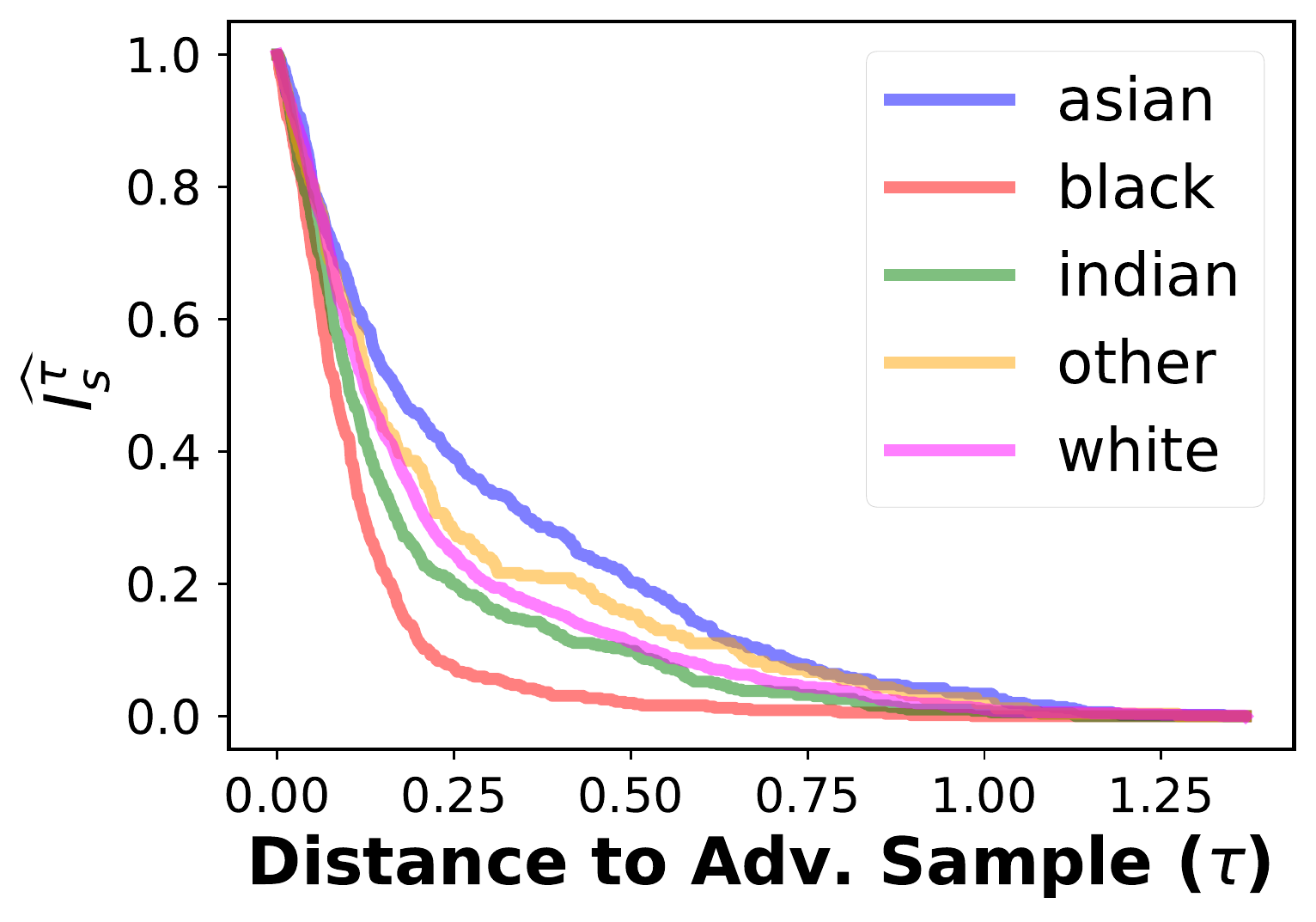}
        \caption{Squeezenet: DeepFool}
        \label{fig:utkface_race_squeezenet_df}
    \end{subfigure}
    \begin{subfigure}[b]{0.23\textwidth}
        \includegraphics[trim={0cm 0cm 0cm 0cm},clip,width=1\textwidth]{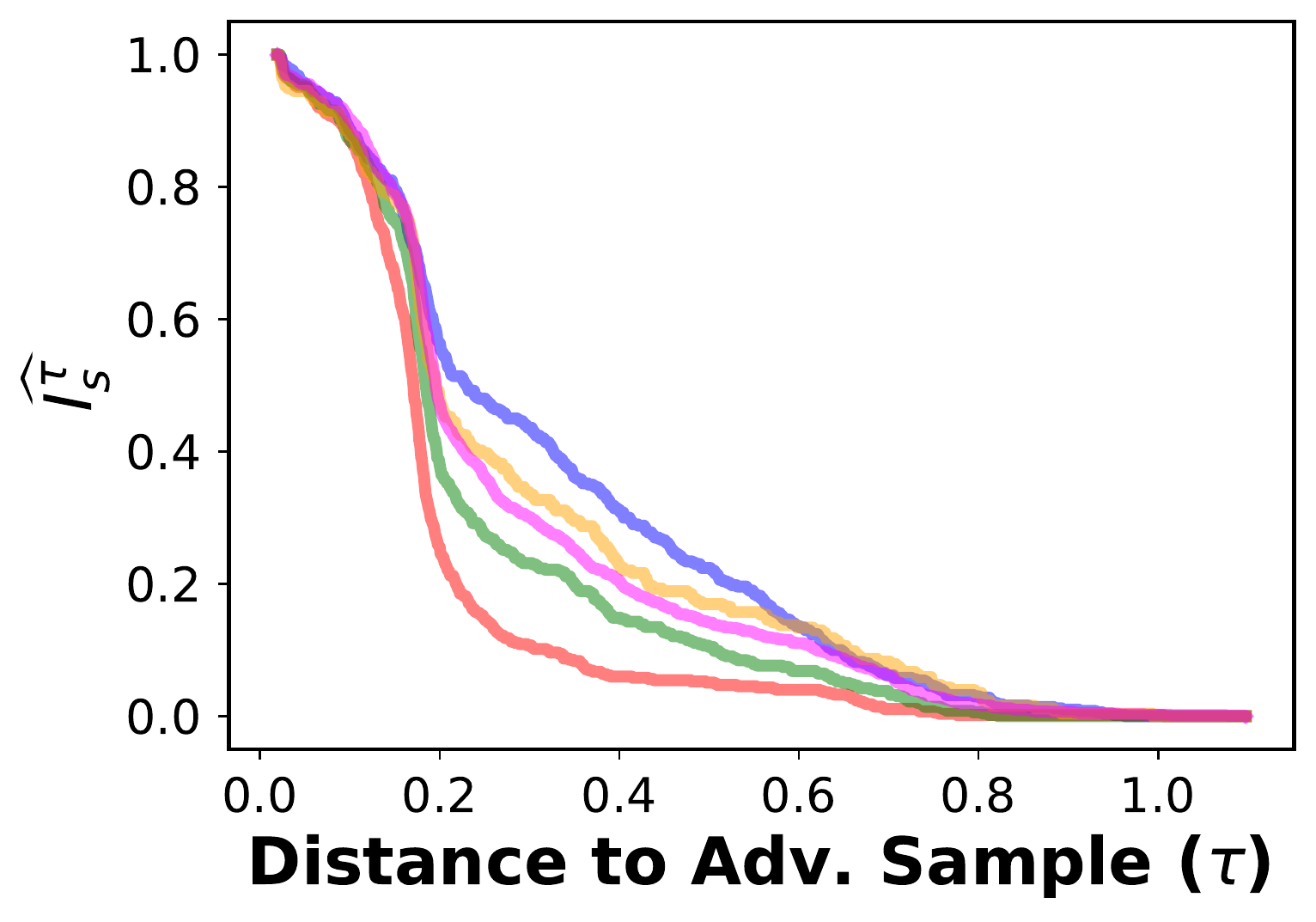}
        \caption{Squeezenet: CarliniWagner}
        \label{fig:utkface_race_squeezenet_cw}
    \end{subfigure}
    \begin{subfigure}[b]{0.23\textwidth}
        \includegraphics[trim={0cm 0cm 0cm 0cm},clip,width=1\textwidth]{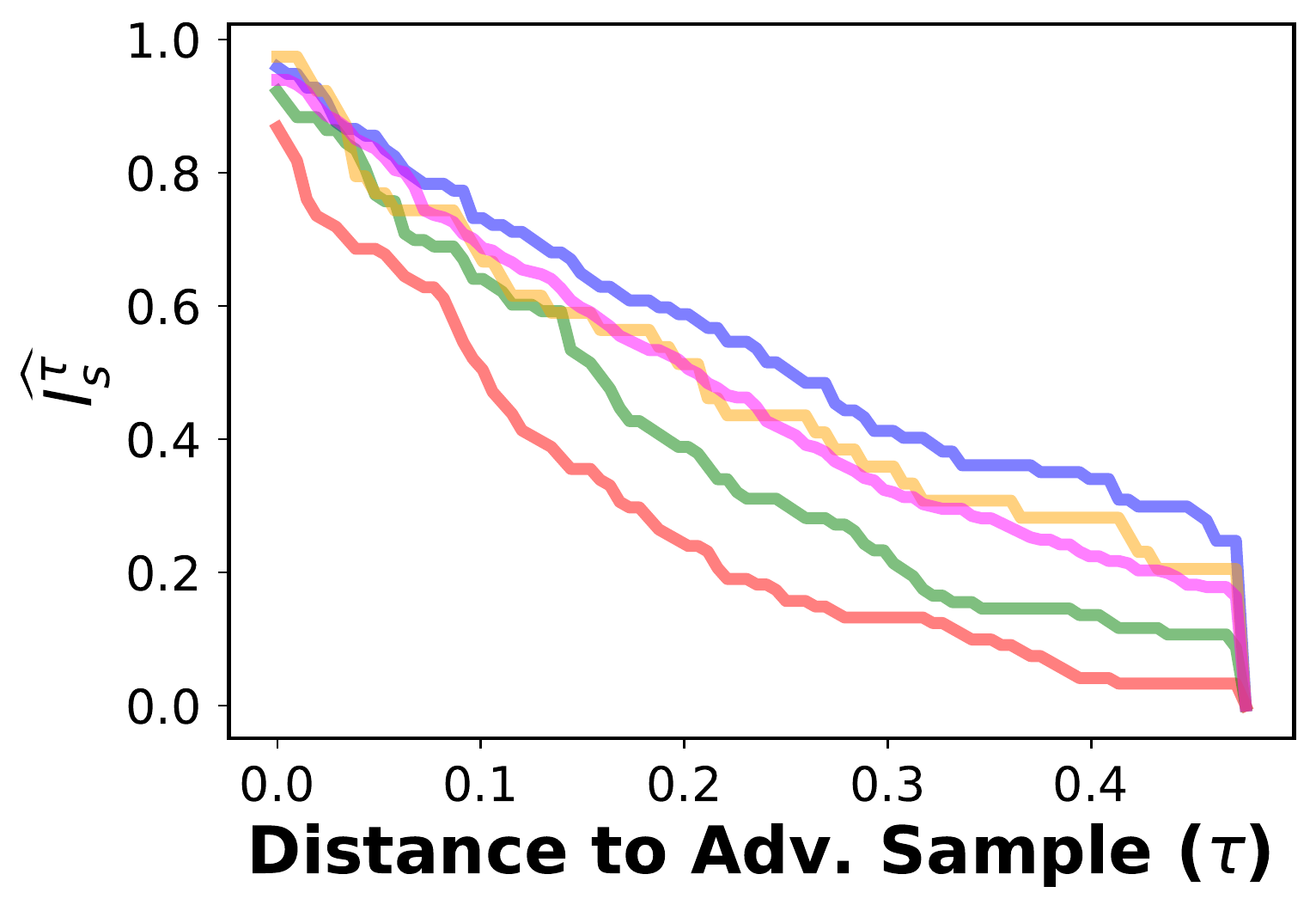}
        \caption{Squeezenet: Rand. Smoothing}
        \label{fig:lb_utkface_race_squeezenet}
    \end{subfigure}
    
\caption{ UTKFace partitioned by race. We can see that across models, that different populations are at different levels of robustness as calculated by different proxies (DeepFool on the left, CarliniWagner in the middle and Randomized Smoothing on the right). This suggests that robustness bias is an important criterion to consider when auditing models for fairness.
} 
\label{fig:utkface_race}
\end{figure*}

\section{Evaluation of Robustness Bias using Adversarial Attacks}\label{sec:adv_attacks}

As described in Section~\ref{subsec:upper_bounds}, we argued that adversarial attacks can be used to obtain upper bounds on $d_\theta(x)$ which can then be used to measure robustness bias. In this section we audit some popularly used models on datasets mentioned in Section~\ref{sec:experiments} for robustness bias as measured using the approximation given by adversarial attacks.

\subsection{Evaluation of $\widehat{I_P^{DF}}$ and $\widehat{I_P^{CW}}$}\label{eval_of_measures}

To compare the estimate of $d_\theta(x)$ by DeepFool and CarliniWagner, we first look at the signedness of $\sigma(P)$, $\sigma^{DF}(P)$, and $\sigma^{CW}(P)$. For a given partition $P$, $\sigma(P)$ captures the disparity in robustness between points in $P$ relative to points not in $P$ (see Eq~\ref{eq:sigma}).  Considering all 151 possible partitions (based on class labels and sensitive attributes, where available) for all five datasets, both CarliniWagner and DeepFool agree with the signedness of the direct computation 125 times, i.e., 
$\mathbbm{1}_P\left[ \sign(\sigma(P)) = \sign(\sigma^{DF}(P))\right] = 125 = \mathbbm{1}_P\left[ \sign(\sigma(P)) = \sign(\sigma^{CW}(P))\right]$. 
Further, the mean difference between $\sigma(P)$ and $\sigma^{CW}(P)$ or $\sigma^{DF}(P)$, i.e., $(\sigma(P) - \sigma^{DF}(P))$, is 0.17 for DeepFool and 0.19 for CarliniWagner with variances of 0.07 and 0.06 respectively.  

There is 83\% agreement between the direct computation and the DeepFool and CarliniWagner estimates of $\widehat{I_P}$. This behavior provides evidence that adversarial attacks provide meaningful upper bounds on $d_\theta(x)$ in terms of the behavior of identifying instances of robustness bias.

\subsection{Audit of Commonly Used Models}\label{subsec:audit_adv_attacks}

\begin{figure*}[h!]
    \centering
    \includegraphics[width = .95\linewidth]{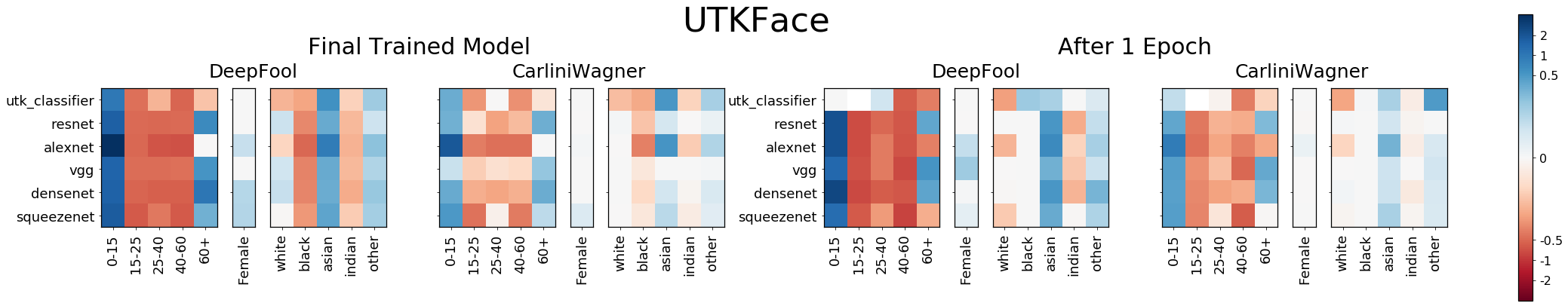}
    \caption{Depiction of $\sigma_P^{DF}$ and $\sigma_P^{CW}$ for the UTKFace dataset with partitions corresponding to the (1) class labels $\mathcal{C}$ and the, (2) gender, and (3) race/ethnicity. These values are reported for all five convolutional models both at the beginning of their training (after one epoch) and at the end. We observe that, largely, the signedness of the functions are consistent between the five models and also across the training cycle.}
    \label{fig:utkface}
\end{figure*}

We now evaluate five commonly-used convolutional neural networks (CNNs): Alexnet, VGG, ResNet, DenseNet, and Squeezenet. We trained these networks using PyTorch with standard stochastic gradient descent. We achieve comparable performance to documented state of the art for these models on these datasets. A full table of performance on the test data are described in Table \ref{tab:conv_results_tbl} (Appendix). After training each model on each dataset, we generated adversarial examples using both methods and computed $\sigma(P)$ for each possible partition of the dataset. An example of the results for the UTKFace dataset can be see in Figure~\ref{fig:utkface}.\footnote{Our full slate of approximation results are available in~\Cref{Conv results}}.

With evidence from Section~\ref{eval_of_measures} that DeepFool and CarliniWagner can approximate the robustness bias behavior of direct computations of $d_\theta$, we first ask if there are any major differences between the two methods. \emph{If DeepFool exhibits adversarial robustness bias for a dataset and a model and a class, does CarliniWagner exhibit the same? and vice versa?} Since there are 5 different convolutional models, we have $151\cdot5 = 755$ different comparisons to make. Again, we first look at the signedness of $\sigma^{DF}(P)$ and $\sigma^{CW}(P)$ and we see that $\mathbbm{1}_P\left[ \sign(\sigma^{DF}(P)) = \sign(\sigma^{CW}(P))\right] = 708$. This means there is 94\% agreement between DeepFool and CarliniWagner about the direction of the adversarial robustness bias.

To investigate if this behavior is exhibited earlier in the training cycle than at the final, fully-trained model, we compute $\sigma^{CW}(P)$ and $\sigma^{DF}(P)$ for the various models and datasets for trained models after 1 epoch and the middle epoch. For the first epoch, 637 of the 755 partitions were internally consistent, i.e., the signedness of $\sigma$ was the same in the first and last epoch, and 621  were internally consistent. We see that at the middle epoch, 671 of the 755 partitions were internally consistent for DeepFool and 665 were internally consistent for CarliniWagner. Unsurprisingly, this implies that as the training progresses, so does the behavior of the adversarial robustness bias. However, it is surprising that much more than 80\% of the final behavior is determined after the first epoch, and there is a slight increase in agreement by the middle epoch. 

We note that, of course, adversarial robustness bias is not necessarily an intrinsic value of a dataset; it may be exhibited by some models and not by others. However, in our studies, we see that the UTKFace dataset partition on Race/Ethnicity does appear to be significantly prone to adversarial attacks given its comparatively low $\sigma^{DF}(P)$ and $\sigma^{CW}(P)$ values across all models.

\section{Evaluation of Robustness Bias using Randomized Smoothing}\label{sec:randomized_smoothing}

In Section~\ref{subsec:lower_bounds}, we argued that randomized smoothing can be used to obtain lower bounds on $d_\theta(x)$ which can then be used to measure robustness bias. In this section we audit popular models on a variety of datasets (described in detail in Section~\ref{sec:experiments}) for robustness bias, as measured using the approximation given by randomzied smoothing.

\begin{figure*}[h!]
    \centering
    \includegraphics[width = .95\linewidth]{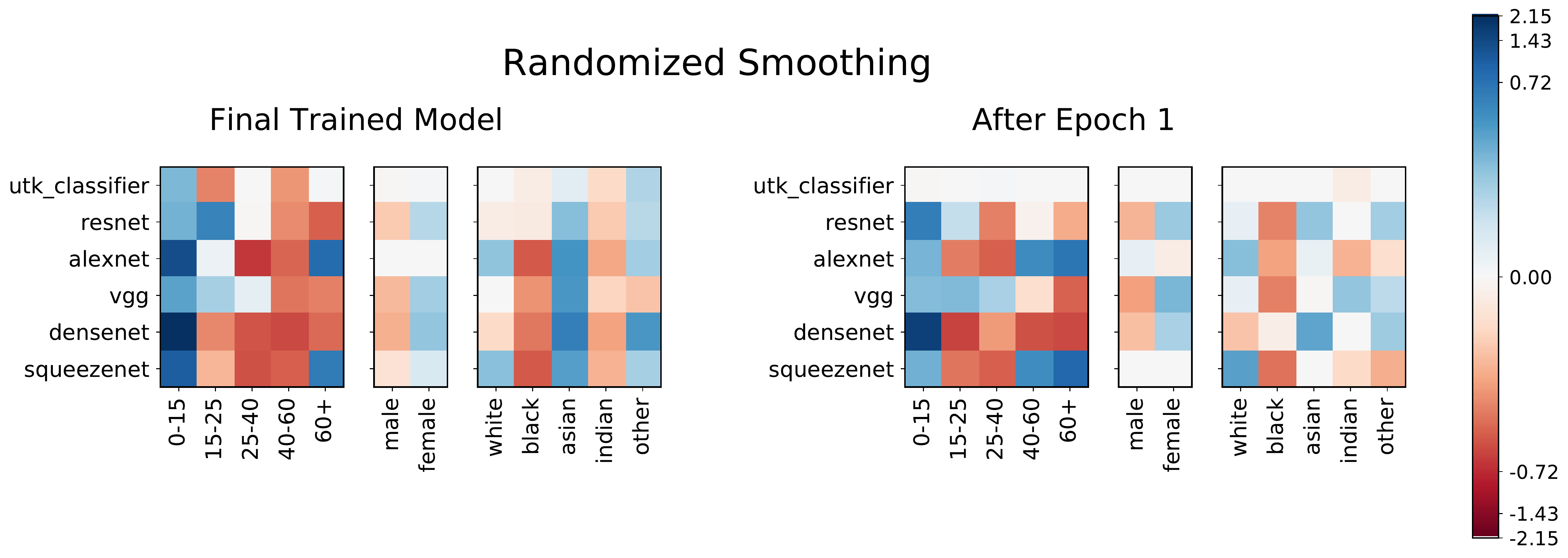}
    \caption{Depiction of $\sigma_P^{RS}$ for the UTKFace dataset with partitions corresponding to the (1) class labels $\mathcal{C}$ and the, (2) gender, and (3) race/ethnicity. A more negative value indicates less robustness bias for the partition. Darker regions indicate high robustness bias. We observe that the trend is largely consistent amongst models and also similar to the trend observed when using adversarial attacks to measure robustness bias (see Figure~\ref{fig:utkface}).}
    \label{fig:utkface_lb}
\end{figure*}

\subsection{Evaluation of $\widehat{I_P^{RS}}$}\label{eval_of_measures_rs}

To assess whether the estimate of $d_\theta(x)$ by randomized smoothing is an appropriate measure of robustness bias, we compare the signedness of $\sigma(P)$ and $\sigma^{RS}(P)$. When $\sigma(P)$ has positive sign, higher magnitude indicates a higher robustness of members of partition $P$ as compared to members not included in that partition $P$; similarly, when $\sigma(P)$ is negatively signed, higher magnitude corresponds to lesser robustness for those members of partition $P$ (see Eq~\ref{eq:sigma}). We may interpret shared signedness of both $\sigma(P)$ (where $d_\theta(x)$ is deterministic) and $\sigma^{RS}(P)$ (where $d_\theta(x)$ is measured by randomized smoothing as described in Section~\ref{subsec:lower_bounds}) as positive support for the $\widehat{I_P^{RS}}$ measure.

Similar to Section~\ref{eval_of_measures}, we consider all possible 151 partitions across CIFAR-10, CIFAR-100, CIFAR-100Super, UTKFace and Adience. For each of these partitions, we compare $\sigma^{RS}(P)$ to the corresponding $\sigma(P)$. We find that their sign agrees 101 times, \ie, $\mathbbm{1}_P\left[ \sign(\sigma(P)) = \sign(\sigma^{RS}(P))\right] = 101$, thus giving a $66.9\%$ agreement. Furthermore, the mean difference between $\sigma(P)$ and $\sigma^{RS}(P)$, \ie, $(\sigma(P) - \sigma^{RS}(P))$ is $0.08$ with a variance of $0.19$. 

This provides evidence that randomized smoothing can also provide a meaningful estimate on $d_\theta(x)$ in terms of measuring robustness bias.

\subsection{Audit of Commonly Used Models}
We now evaluate the same models and all the datasets for robustness bias as measured by randomized smoothing.  Our comparison is analogous to the one performed in Section~\ref{subsec:audit_adv_attacks} using adversarial attacks. Figure~\ref{fig:utkface_lb} shows results for all models on the UTKFace dataset. Here we plot $\sigma_{P}^{RS}$ for each partition of the dataset (on x-axis) and for each model (y-axis). A darker color in the heatmap indicates high robustness bias (darker red indicates that the partition is \textit{less} robust than others, whereas a darker blue indicates that the partition is \textit{more} robust). We can see that some partitions, for example, the partition based on class label ``40-60'' and the partition based on race ``black'' tend to be less robust in the final trained model, for all models (indicated by a red color across all models). Similarly there are partitions that are more robust, for example, the partition based on class ``0-15'' and race ``asian'' end up being robust across different models (indicated by a blue color). Figure~\ref{fig:utkface_race} takes a closer look at the distribution of distances for the UTKFace dataset when partitioned by race, showing that for different models different races can be more or less robust. Figures~\ref{fig:utkface_race},~\ref{fig:utkface} and \ref{fig:utkface_lb} (we see similar trends for CIFAR-10, CIFAR-100, CIFAR-100Super and Adience which we report exhaustively in Appendix~\ref{Conv results}) lead us to the following key conclusions:

\xhdr{Dependence on data distribution} The presence of certain partitions that show similar robustness trends as discussed above (\eg see final trained model in Figs~\ref{fig:utkface_lb} and \ref{fig:utkface}, the partitions by class ``0-15'' and race ``asian'' are more robust, whereas the class ``40-60'' and race ``black'' are less robust across \textit{all models}) point to some intrinsic property of the data distribution that results in that partition being more (or less) robust regardless of the type decision boundary. Thus we conclude that robustness bias may depend in part on the data distribution of various sub-populations.

\xhdr{Dependence on model} There are also certain partitions of the dataset (e.g., based on the classes ``15-25'' and ``60+'' as per Fig~\ref{fig:utkface_lb}) that show varying levels of robustness across different models. Moreover, even partitions that have same sign of $\sigma^{RS}(P)$ across different models have very different values of $\sigma^{RS}(P)$. This is also evident from Fig~\ref{fig:utkface_race} which shows that the distributions of $d_\theta(x)$ (as approximated by all our proposed methods) for different races can be very different for different models. Thus, we conclude that robustness bias is also dependent on the learned model.

\xhdr{Role of pre-training} 
We now explore the role of pre-training on our measures of robustness bias.  Specifically, we pre-train five of the six models (Resnet, Alexnet, VGG, Densenet, and Squeezenet) on ImageNet and then fine-tune on UTKFace.  We also train a UTK classifier from scratch on UTKFace.  Figures~\ref{fig:utkface_lb} and \ref{fig:utkface} shows robustness bias scores after the first epoch and in the final, fully-trained model.  At epoch 1, we mostly see no robustness bias (indicated by close-to-zero values of $\sigma^{RS}(P)$) for UTK Classifier. This is because the model has barely trained by that first epoch and predictions are roughly equivalent to random guesses. In contrast, the other five models already have pre-trained ImageNet weights, and hence we see certain robustness biases that already exist in the model, even after the first epoch of training. Thus, we conclude that pre-trained models bring in biases due to the distributions of the data on which they were pre-trained and the resulting learned decision boundary after pre-training. We additionally see that these biases can persist even after fine-tuning.

\subsection{Comparison of Randomized Smoothing and Upper Bounds}

We have now presented two ways of measuring robustness bias: via upper bounds and via randomized smoothing. While there are important distinctions between the two methods, it is worth comparing them. To do this, we compare the sign of the randomized smoothing method and the upper bounds as
$$\mathbbm{1}_P\left[ \sign(\sigma^{RS}(P)) = \sign(\sigma^{DF}(P))\right]$$
and
$$\mathbbm{1}_P\left[ \sign(\sigma^{RS}(P)) = \sign(\sigma^{CW}(P))\right].$$
We see that there is some evidence that the two methods agree. The Adience, UTKFace, and CIFAR-10 dataset have strong agreement (at or above 75\%) between the randomized smoothing for both types of upper bounds (DeepFool and CariliniWagner), while the CIFAR-100 dataset has a much weaker agreement (above but closer to 50\%) and CIFAR-100Super has an approximately 66\% agreement.  %

It is important to point out that it is not entirely appropriate to perform a comparison in this way. Recall that the upper bounds provide estimates of $d_\theta$ using a trained model. However, the randomized smoothing method estimates $d_\theta$ not directly with the trained model --- instead it first modifies (smooths) the model of interest and then performs an estimation.
Since the upper bounds and randomized smoothing methods are so different in practice, there may be no truly appropriate way to compare the results therefrom.
Therefore, too much credence should not be placed on the comparison of these two methods.
Both methods indicate the existence of the robustness bias phenomenon and can be useful in distinct settings.

We present the full results table in Appendix \ref{sec:lower_bounds_experiments}.

\section{An ``Obvious'' Mitigation Strategy}\label{sec:mitigation}

Having demonstrated the existence of this robustness bias phenomenon, it is natural to look ahead at common machine learning techniques to address it.  In Appendix~\ref{app:regularizer}, we have done just that by adding to the objective function a regularizer term which penalizes for large distances in the treatment of a minority and majority group. We write the empiric estimate of $\RBmeasure{(P,\tau)}$ as $\tilde{\RBmeasure}{(P,\tau)}$; a full derivation can be found in \Cref{app:regularizer-derivation}.  Formally,
\vspace{-2.5mm}
{\small
\begin{multline*}
    \tilde{\RBmeasure}{(P,\tau)} = \Bigg|  \frac{1}{\displaystyle \sum_{x\notin P} \mathbbm{1}\{ y = \yhat \}} \sum_{\substack{x \notin P\\y = \yhat}} \mathbbm{1}\{ d_\theta(x) > \tau \} -   \\
    \frac{1}{\displaystyle\sum_{x\in P} \mathbbm{1}\{ y = \yhat \}} \sum_{\substack{x \in P\\y = \yhat}} \mathbbm{1}\{ d_\theta(x) > \tau \}  \Bigg|
\end{multline*}

}

Details of a full implementation of this loss function and the results can be found in Appendix~\ref{app:regularizer}.  Experimental results based on that implemntation, reported in Appendix~\ref{app:regularizer-experiments}, support the idea that regularization---a typical approach taken by the fairness in machine learning community---can reduce measures of robustness bias.
However, we do believe that this type of experimentation belies the larger point of the present, largely descriptive, work: that robustness bias is a real and significant artifact of popular and commonly-used datasets and models.
Surely there are ways to mitigate some of the effects or manifestations of this bias (as we show with our fairly standard regularization-based mitigation technique).
However, we believe that any type of mitigation should be taken in concert with the contextualization of these technical systems in the social world, and thus leave ``mitigation'' research to, ideally, application-specific future work involving both machine learning practictioners and the stakeholders of particular systems.
\section{Discussion and Conclusion}\label{sec:broader}
We propose a unique definition of fairness which requires all partitions of a population to be equally robust to minute (often adversarial) perturbations, and give experimental evidence that this phenomenon can exist in some commonly-used models trained on real-world datasets. Using these observations, we argue that this can result in a potentially unfair circumstance where, in the presence of an adversary, a certain partition might be more susceptible (\ie, less secure). Susceptibility is prone to known issues with adversarial robustness such as sensitivity to hyperparameters~\cite{tramer2020adaptive}. Thus, we call for extra caution while deploying deep neural nets in the real world since this form of unfairness might go unchecked when auditing for notions that are based on just the model outputs and ground truth labels. We then show that this form of bias can be mitigated to some extent by using a regularizer that minimizes our proposed measure of robustness bias. However, we do not claim to ``solve'' unfairness; rather, we view analytical approaches to bias detection and optimization-based approaches to bias mitigation as potential pieces in a much larger, multidisciplinary approach to addressing these issues in fielded systems.

Indeed, we view our work as largely observational---we \emph{observe} that, on many commonly-used models trained on many commonly-used datasets, a particular notion of bias, \emph{robustness bias}, exists.  We show that some partitions of data  are more susceptible to two state-of-the-art and commonly-used adversarial attacks.  This knowledge could be used for \emph{attack} or to design \emph{defenses}, both of which could have potential positive or negative societal impacts depending on the parties involved and the reasons for attacking and/or defending.  We have also \emph{defined} a notion of bias as well as a corresponding notion of fairness, and by doing that we admittedly toe a morally-laden line.  Still, while we do use ``fairness'' as both a higher-level motivation and a lower-level quantitative tool, we have tried to remain ethically neutral in our presentation and have eschewed making normative judgements to the best of our ability.

\section*{Acknowledgments}
Dickerson, Dooley, and Nanda were supported in part by NSF CAREER Award IIS-1846237, NIST MSE Award \#20126334, DARPA GARD \#HR00112020007, DARPA SI3-CMD \#S4761, DoD WHS Award \#HQ003420F0035, and a Google Faculty Research Award.
Feizi and Singla were supported in part by the NSF CAREER award 1942230, Simons Fellowship on Deep Learning Foundations, AWS Machine Learning Research Award, NIST award 60NANB20D134 and award HR001119S0026-GARD-FP-052. The authors would like to thank Juan Luque and Aviva Prins for fruitful discussions in earlier stages of the project, and reviewers at NeurIPS-20 for detailed discussion of an earlier draft of the paper.

\bibliographystyle{plainnat}
\bibliography{refs}

\appendix

\clearpage

\section{Additional Model and Dataset Details}\label{Conv results}

\xhdr{CIFAR-10, CIFAR-100, CIFAR100Super} These are standard deep learning benchmark datasets. Both CIFAR-10 and CIFAR-100 contain $60,000$ images in total which are split into $50,000$ train and $10,000$ test images. The task is to classify a given image. Images are mean normalized with mean and std of $(0.5, 0.5, 0.5)$.

\xhdr{UTKFace} Contains images of a people labeled with race, gender and age. We split the dataset into a random $80:20$ train:test split to get $4,742$ test and $18,966$ train samples. We bin the age into 5 age groups and convert this into a 5-class classification problem. Images are normalized with mean and std of 0 and 1 respectively.

\xhdr{Adience}  Contains images of a people labeled with gender and age group. Task is to classify a given image into one of 8 age groups. We split the dataset into a random 80:20 train:test split to get $14,007$ train and $3,445$ test samples. Images are normalized with mean and std of $(0.485, 0.456, 0.406)$ and $(0.229, 0.224, 0.225)$ respectively.

\begin{figure*}[t!]
    \centering
    \includegraphics[width = .95\linewidth]{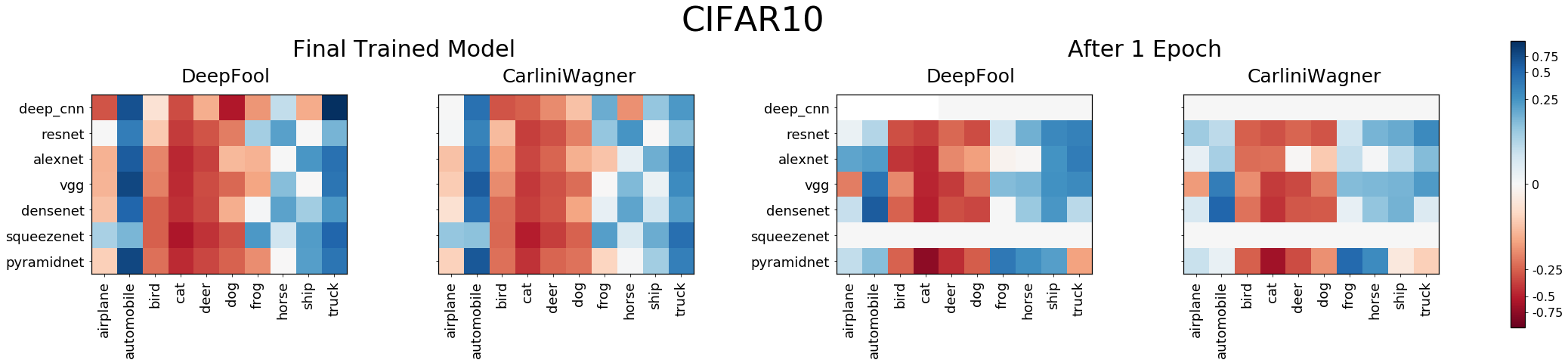}
    \caption{Depiction of $\sigma_P^{DF}$ and $\sigma_P^{CW}$ for the CIFAR10 dataset with partitions corresponding to the class labels $\mathcal{C}$. These values are reported for all five convolutional models both at the beginning of their training (after one epoch) and at the end. We observe that, largely, the signedness of the functions are consistent between the five models and also across the training cycle.}
    \label{fig:cifar10_supp}
\end{figure*}

Accuracy of models trained on these datasets can be found in Table~\ref{tab:conv_results_tbl}. Figures~\ref{fig:cifar10_supp},~\ref{fig:cifar100_supp},~\ref{fig:cifar100_super_supp}, and~\ref{fig:adience_supp} show the measures of robustness bias as approximated by CarliniWagner ($\sigma_P^{CW}$) and DeepFool ($\sigma_P^{DF}$) across different partitions of all the datasets.

\begin{figure*}[t!]
    \centering
    \includegraphics[width = .95\linewidth]{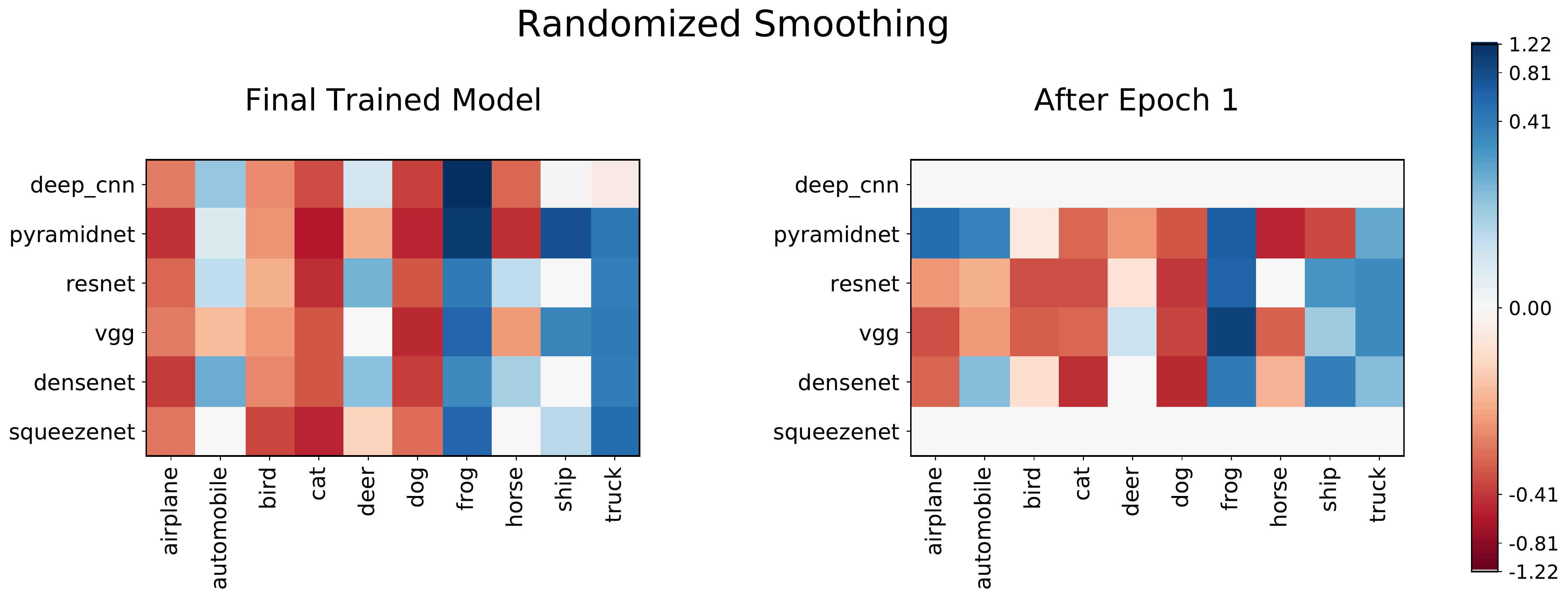}
    \caption{Depiction of $\sigma_P^{RS}$ for the CIFAR10 dataset with partitions corresponding to the class labels $\mathcal{C}$.}
    \label{fig:cifar10_lbs}
\end{figure*}

\begin{figure*}[t!]
    \centering
    \includegraphics[width = .95\linewidth]{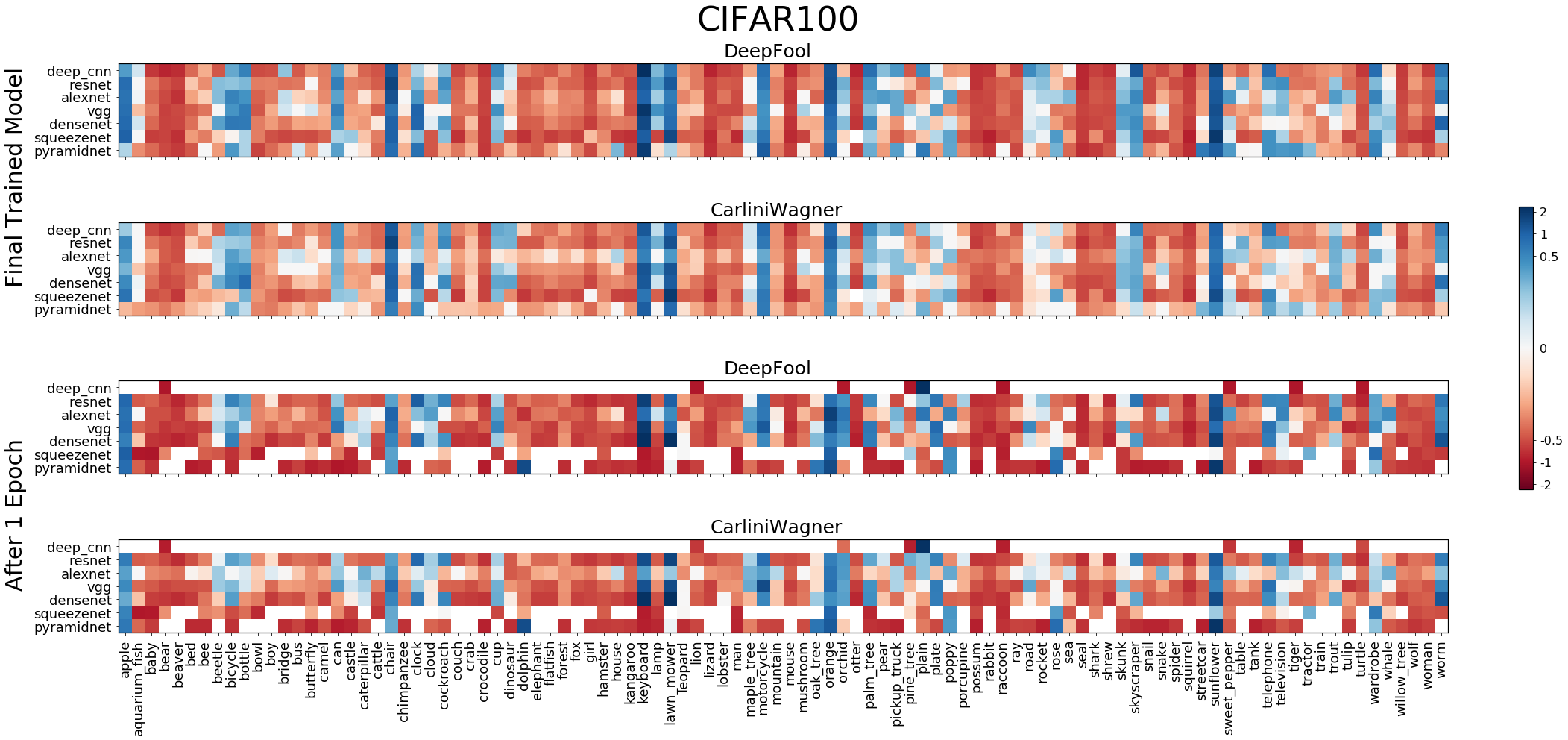}
    \caption{Depiction of $\sigma_P^{DF}$ and $\sigma_P^{CW}$ for the CIFAR100 dataset with partitions corresponding to the class labels $\mathcal{C}$. These values are reported for all five convolutional models both at the beginning of their training (after one epoch) and at the end. We observe that, largely, the signedness of the functions are consistent between the five models and also across the training cycle.}
    \label{fig:cifar100_supp}
\end{figure*}
\begin{figure*}[t!]
    \centering
    \includegraphics[width = .95\linewidth]{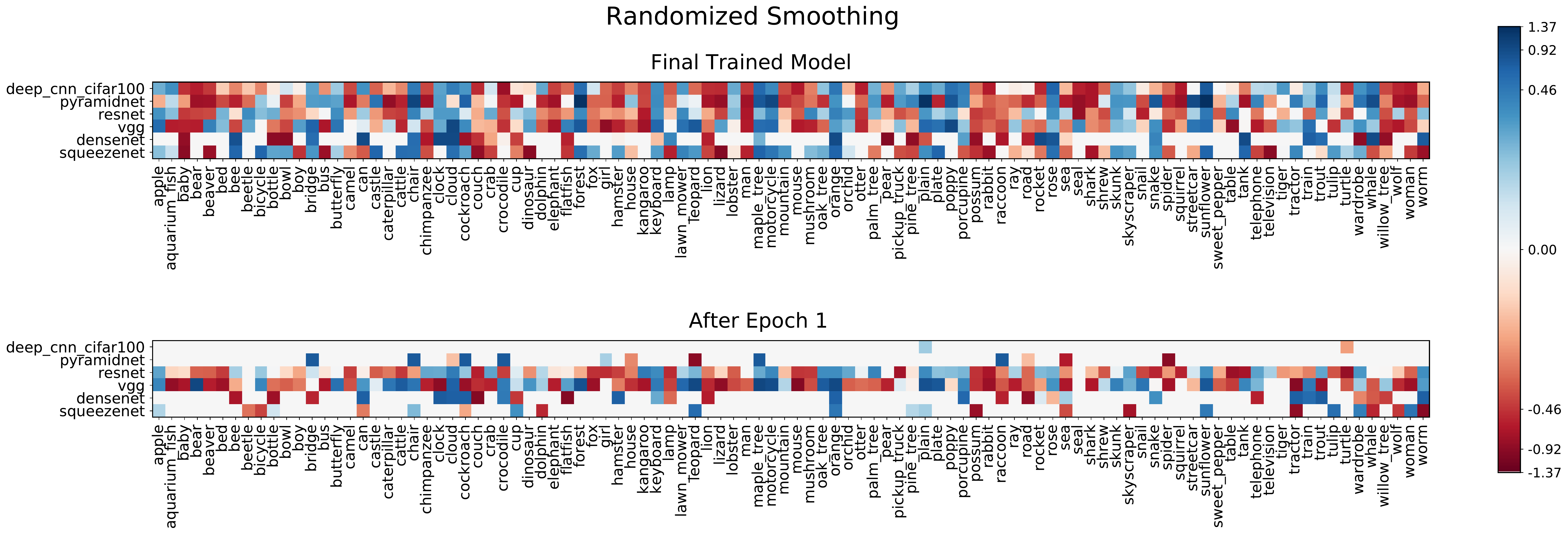}
    \caption{Depiction of $\sigma_P^{RS}$ for the CIFAR100 dataset with partitions corresponding to the class labels $\mathcal{C}$.}
    \label{fig:cifar100_lb}
\end{figure*}

\begin{figure*}[t!]
    \centering
    \includegraphics[width = .95\linewidth]{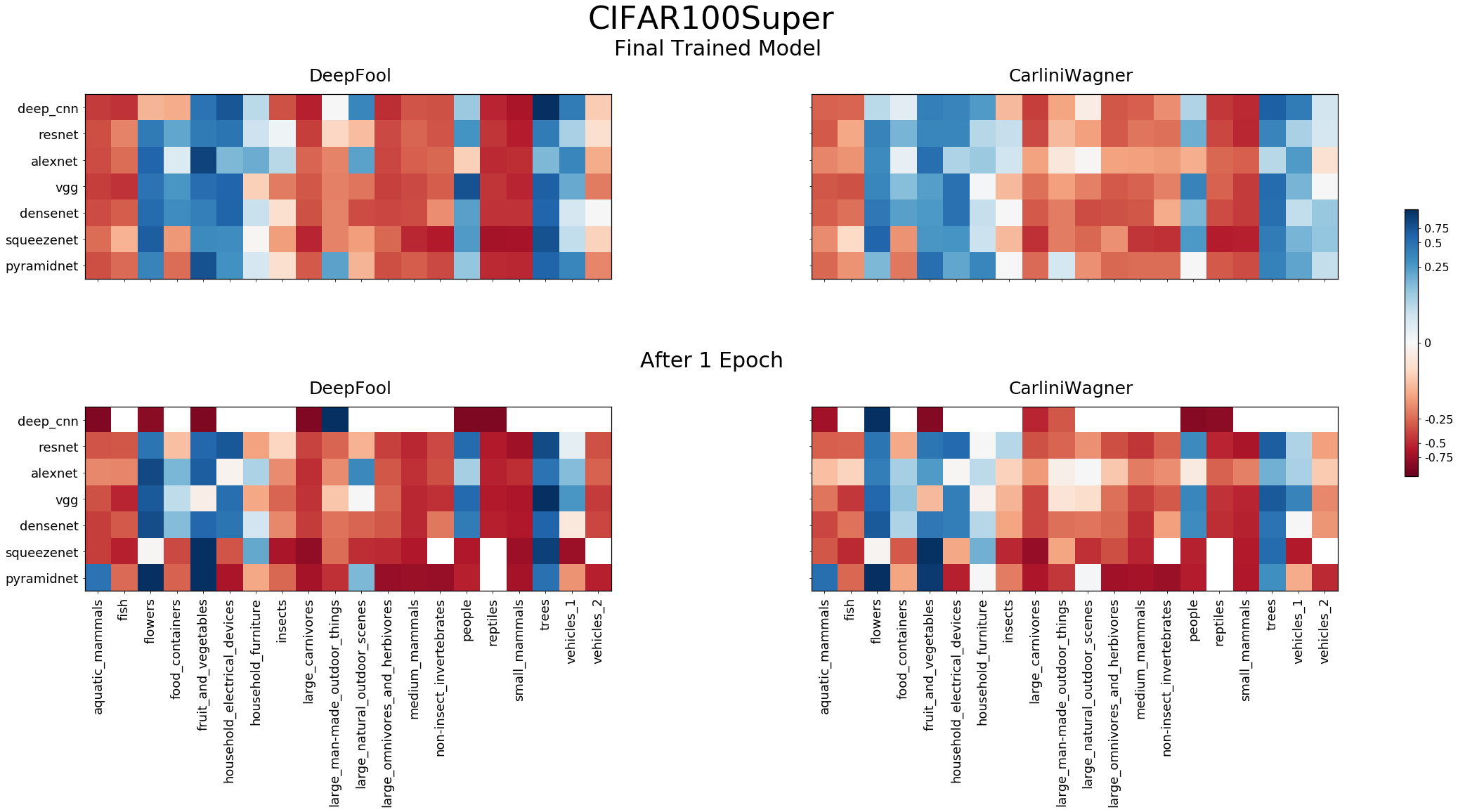}
    \caption{Depiction of $\sigma_P^{DF}$ and $\sigma_P^{CW}$ for the CIFAR100super dataset with partitions corresponding to the class labels $\mathcal{C}$. These values are reported for all five convolutional models both at the beginning of their training (after one epoch) and at the end. We observe that, largely, the signedness of the functions are consistent between the five models and also across the training cycle.}
    \label{fig:cifar100_super_supp}
\end{figure*}
\begin{figure*}[t!]
    \centering
    \includegraphics[width = .95\linewidth]{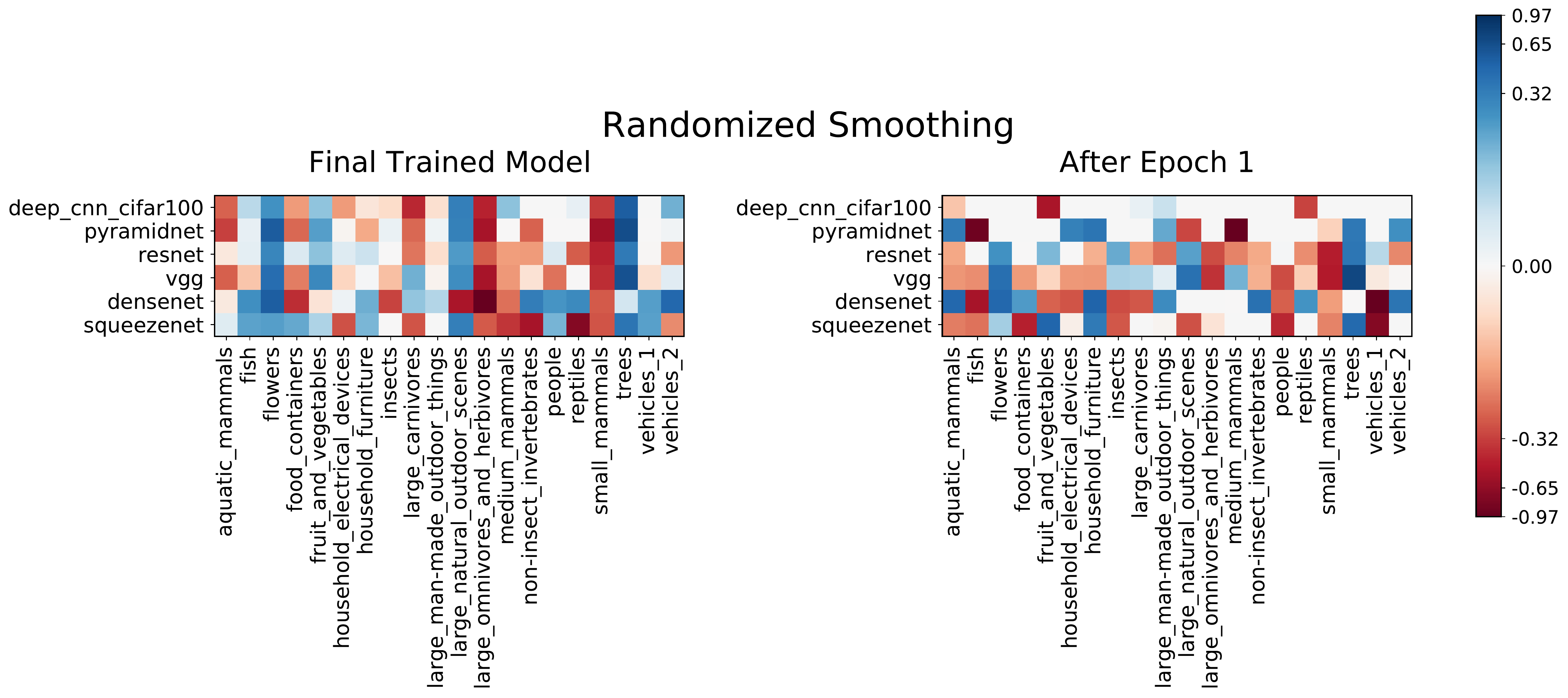}
    \caption{Depiction of $\sigma_P^{RS}$ for the CIFAR100super dataset with partitions corresponding to the class labels $\mathcal{C}$. }
    \label{fig:cifar100_super_lb}
\end{figure*}

\begin{figure*}[t!]
    \centering
    \includegraphics[width = .95\linewidth]{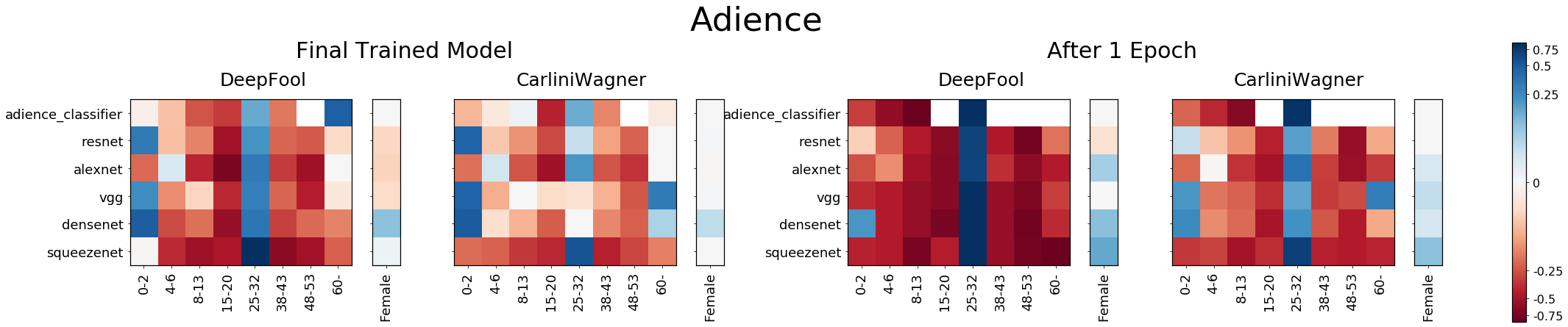}
    \caption{Depiction of $\sigma_P^{DF}$ and $\sigma_P^{CW}$ for the Adience dataset with partitions corresponding to the (1) class labels $\mathcal{C}$ and the and (2) gender. These values are reported for all five convolutional models both at the beginning of their training (after one epoch) and at the end. We observe that, largely, the signedness of the functions are consistent between the five models and also across the training cycle.}
    \label{fig:adience_supp}
\end{figure*}
\begin{figure*}[t!]
    \centering
    \includegraphics[width = .95\linewidth]{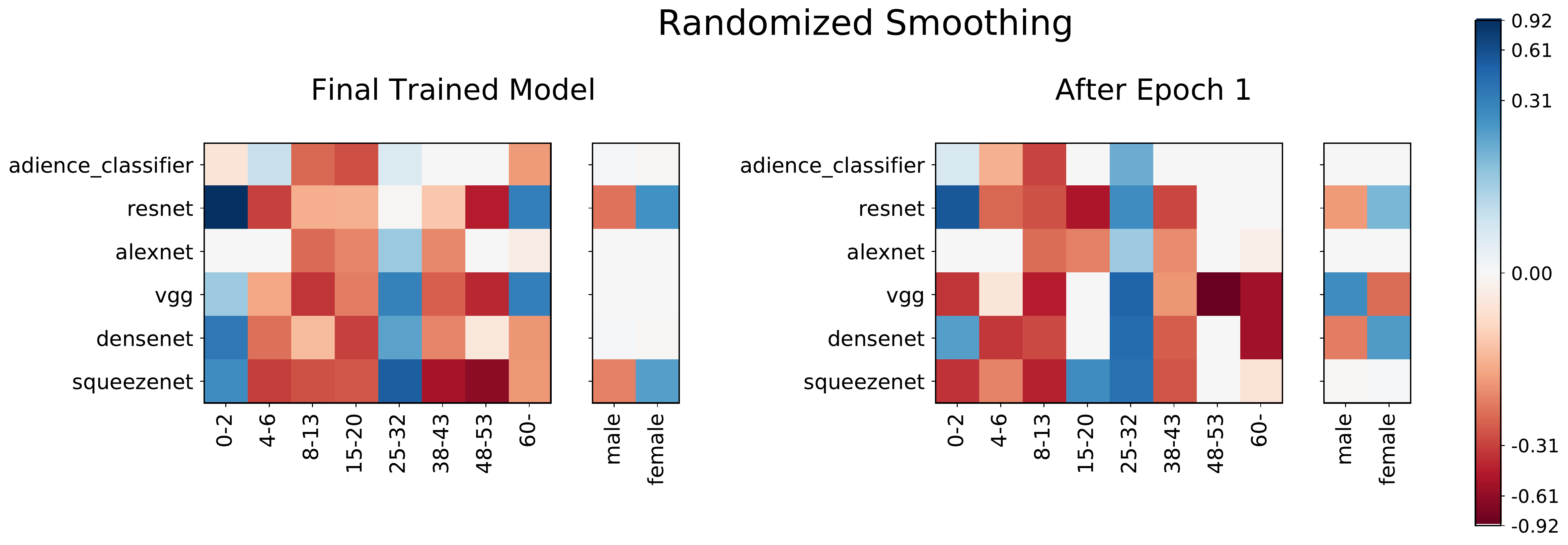}
    \caption{Depiction of $\sigma_P^{RS}$ for the Adience dataset with partitions corresponding to the (1) class labels $\mathcal{C}$ and the and (2) gender.}
    \label{fig:adience_lb}
\end{figure*}

\section{Regularization Results}\label{app:regularizer}

In this section we first derive the regularization term introduced in section~\ref{sec:mitigation} in the main paper. Then using this regularization term we re-train models to see if we can mitigate robustness bias. As we saw in Figures~\ref{fig:adience_lb} and~\ref{fig:adience_supp} in the main paper, Adience barely showed any robustness bias when partitioned on the sensitive attribute (gender), so in this section we only focus on CIFAR-10 and UTKFace (partitioned by race). We show an in-depth analysis Resnet50, VGG19 and Deep CNN on CIFAR-10 and Resnet50, VGG19 and UTK Classifier on UTKFace.

\subsection{Formal Derivation of the Regularization Term}\label{app:regularizer-derivation}
In this section, we provide a step-by-step derivation of the regularization term used in Section~\ref{sec:mitigation} in the main paper.  Recall the traditional Empirical Risk Minimization objective, $\ERM{} := l_{cls} (f_\theta(X), Y)$, where $l_{cls}$ is cross entropy loss. Now we wish to model our measure of fairness (\S\ref{sec:robustness_bias}) and minimize for it alongside $\ERM{}$. To model our measure, we first evaluate the following cumulative distribution functions:

\begin{align*}
\mathbb{P}_{x\in\mathcal{D}}\{d_\theta (x) > \tau \mid x\in P, y = \yhat\} = \\ \frac{1}{\displaystyle \sum_{x\notin P} \mathbbm{1}\{ y = \yhat \}} \sum_{\substack{x \notin P\\y = \yhat}} \mathbbm{1}\{ d_\theta(x) > \tau \}
\\
\mathbb{P}_{x\in\mathcal{D}}\{d_\theta (x) > \tau \mid x\notin P, y = \yhat\} = \\ \frac{1}{\displaystyle\sum_{x\in P} \mathbbm{1}\{ y = \yhat \}} \sum_{\substack{x \in P\\y = \yhat}} \mathbbm{1}\{ d_\theta(x) > \tau \}
\end{align*}

This gives us the empirical estimate of the robustness bias term $\RBmeasure{(P,\tau)}$, parameterized by partition $P$ and threshold $\tau$, defined as Equation~\ref{eq:rb_raw} below.

\begin{equation}\label{eq:rb_raw}
\begin{split}
    \tilde{\RBmeasure}{(P,\tau)} = \abs[\Big] { \frac{1}{\displaystyle \sum_{x\notin P} \mathbbm{1}\{ y = \yhat \}} \sum_{\substack{x \notin P\\y = \yhat}} \mathbbm{1}\{ d_\theta(x) > \tau \}  - \\ \frac{1}{\displaystyle\sum_{x\in P} \mathbbm{1}\{ y = \yhat \}} \sum_{\substack{x \in P\\y = \yhat}} \mathbbm{1}\{ d_\theta(x) > \tau \}  }
\end{split}
\end{equation}

Now, for $\tilde{\RBmeasure}$ as defined in Equation~\ref{eq:rb_raw} to be computed and used, for example, during training, we must approximate a closed form expression of $d_\theta$. To formulate this, we take inspiration from the way adversarial inputs are created using DeepFool~\cite{moosavi2016deepfool}. Just like in DeepFool, we also approximate distance from $f_\theta$ considering $f_\theta$ to be linear (even though it may be a highly non-linear Deep Neural Network). Thus, we get,

\begin{align}\label{eq:d_theta}
    d_\theta(x) = \left| \frac{f_\theta(x)}{|| \nabla_x f_\theta(x) ||} \right| .
\end{align}

By combining Equations~\ref{eq:rb_raw} and~\ref{eq:d_theta}, we recover $\tilde{\RBmeasure}$, a computable estimate of the robustness bias term $\RBmeasure$, as follows.

\begin{equation}\label{eq:rb_complete}
\begin{split}
    \tilde{\RBmeasure}{(P,\tau)} = \abs[\Big]{  \frac{1}{\displaystyle \sum_{x\notin P} \mathbbm{1}\{ y = \yhat \}} \sum_{\substack{x \notin P\\y = \yhat}} \mathbbm{1}\{ | \,  \frac{f_\theta(x)}{|| \nabla_x f_\theta(x) ||}  \, | > \tau \}  - \\ \frac{1}{\displaystyle\sum_{x\in P} \mathbbm{1}\{ y = \yhat \}} \sum_{\substack{x \in P\\y = \yhat}} \mathbbm{1}\{ | \,  \frac{f_\theta(x)}{|| \nabla_x f_\theta(x) ||}  \, | > \tau \}  }
\end{split}
\end{equation}

Finally, given scalar $\alpha$, we minimize for the new objective function, \textit{AdvERM}, as follows:

\begin{align*}
    \mathit{AdvERM} := l_{\mathit{cls}} (f_\theta(X), Y) + \alpha \cdot \tilde{\RBmeasure}{(P,\tau)} .
\end{align*}

\subsection{Additional Regularization Results}\label{app:regularizer-experiments}

\begin{figure*}[h]
    \centering
    \begin{subfigure}[b]{0.22\textwidth}
        \includegraphics[trim={0cm 0cm 0cm 0cm},clip,width=1\textwidth]{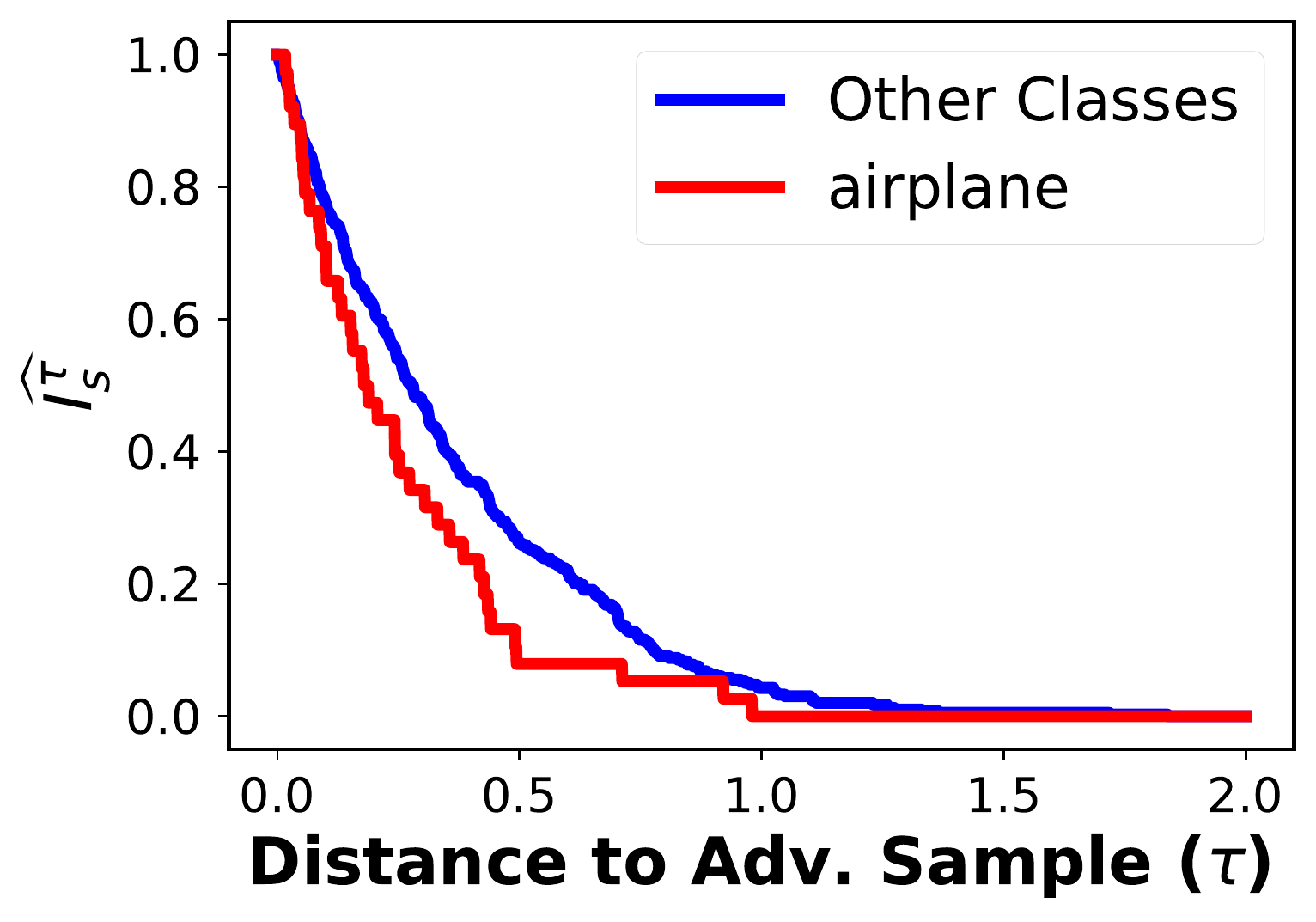}
        \caption{Airplane}
        \label{fig:cifar10_airplane}
    \end{subfigure}
    \begin{subfigure}[b]{0.22\textwidth}
        \includegraphics[trim={0cm 0cm 0cm 0cm},clip,width=1\textwidth]{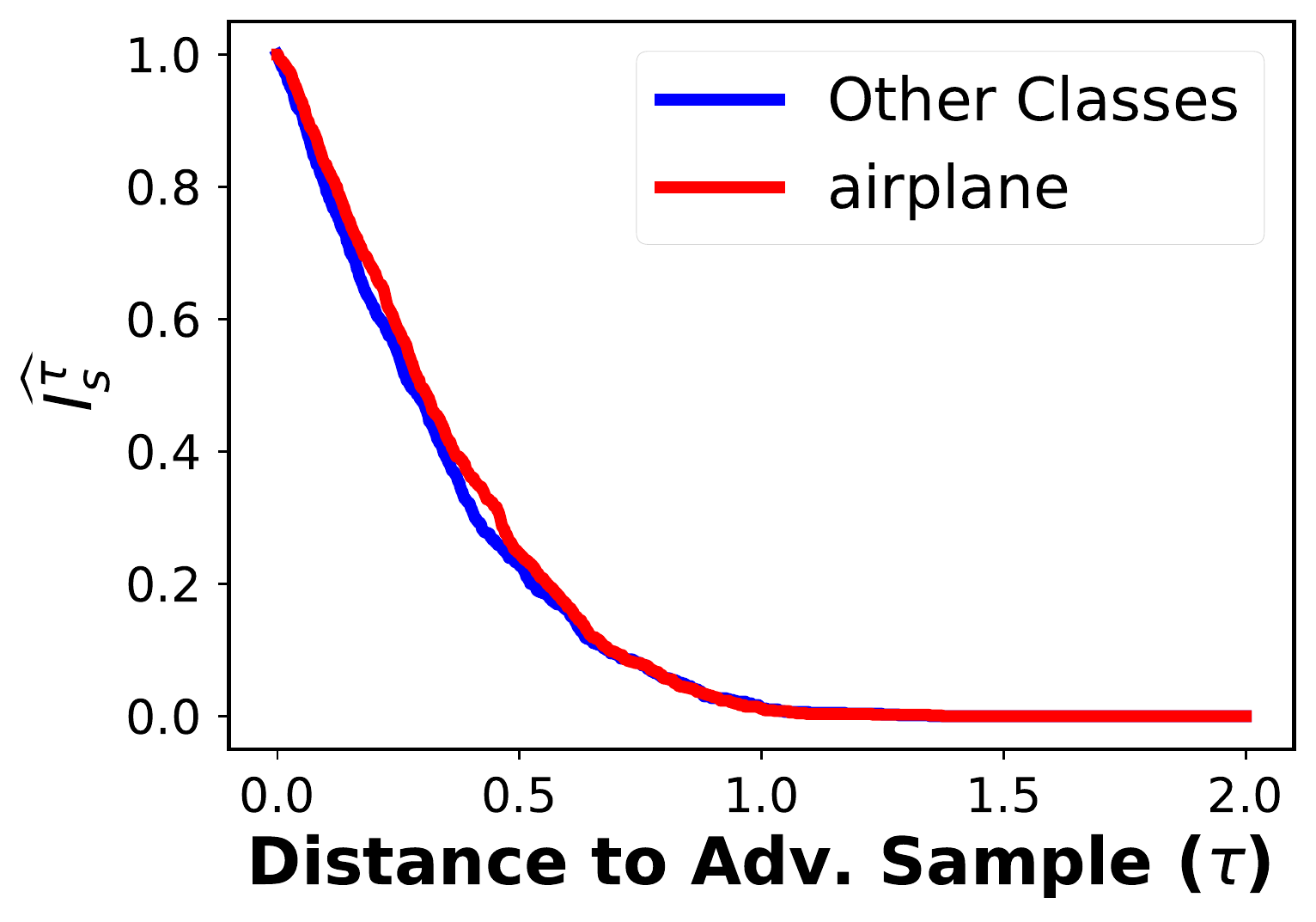}
        \caption{Reg. Airplane}
        \label{fig:cifar10_airplane_reg}
    \end{subfigure}
    \begin{subfigure}[b]{0.22\textwidth}
        \includegraphics[trim={0cm 0cm 0cm 0cm},clip,width=1\textwidth]{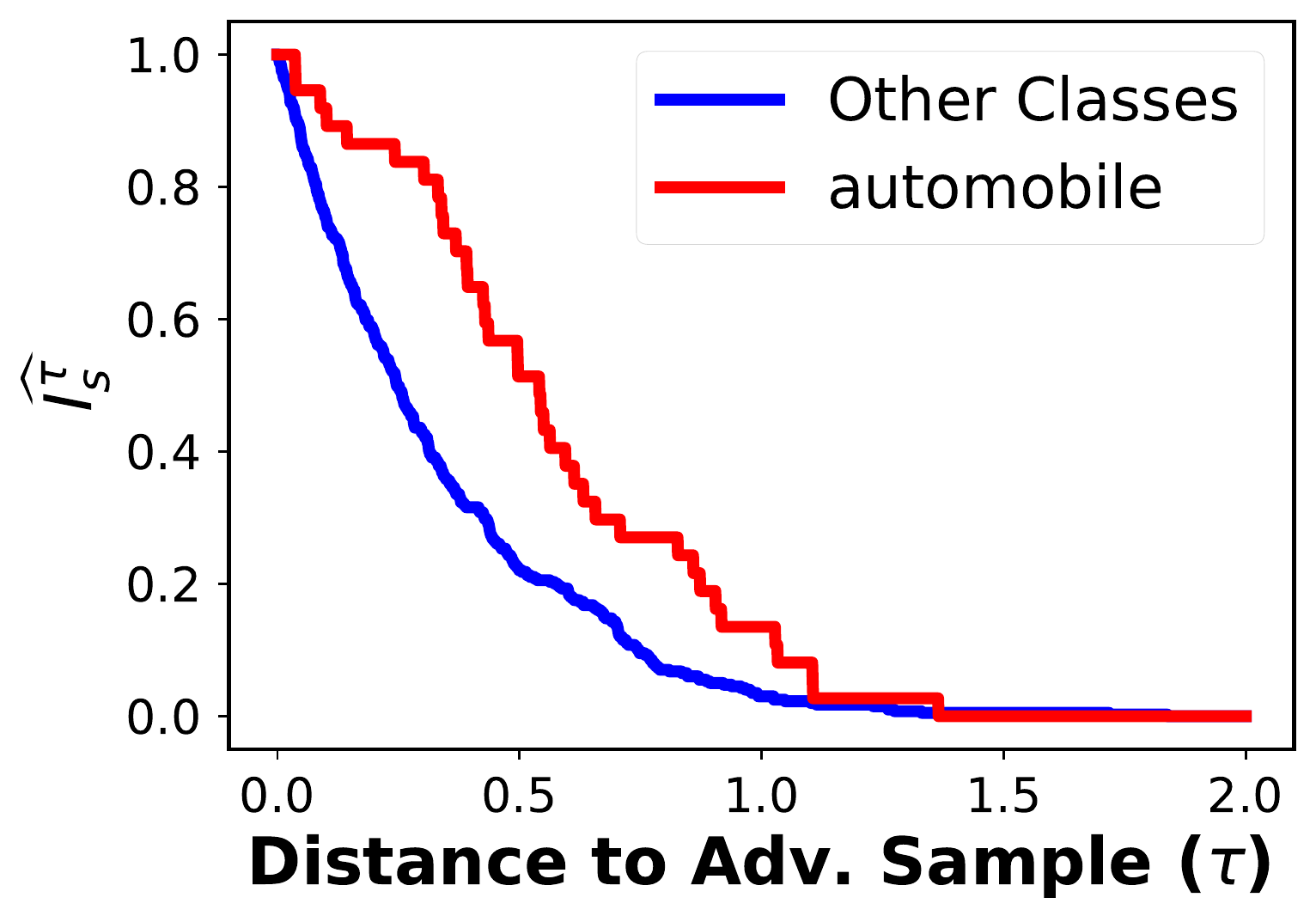}
        \caption{Automobile}
        \label{fig:cifar10_automobile}
    \end{subfigure}
    \begin{subfigure}[b]{0.22\textwidth}
        \includegraphics[trim={0cm 0cm 0cm 0cm},clip,width=1\textwidth]{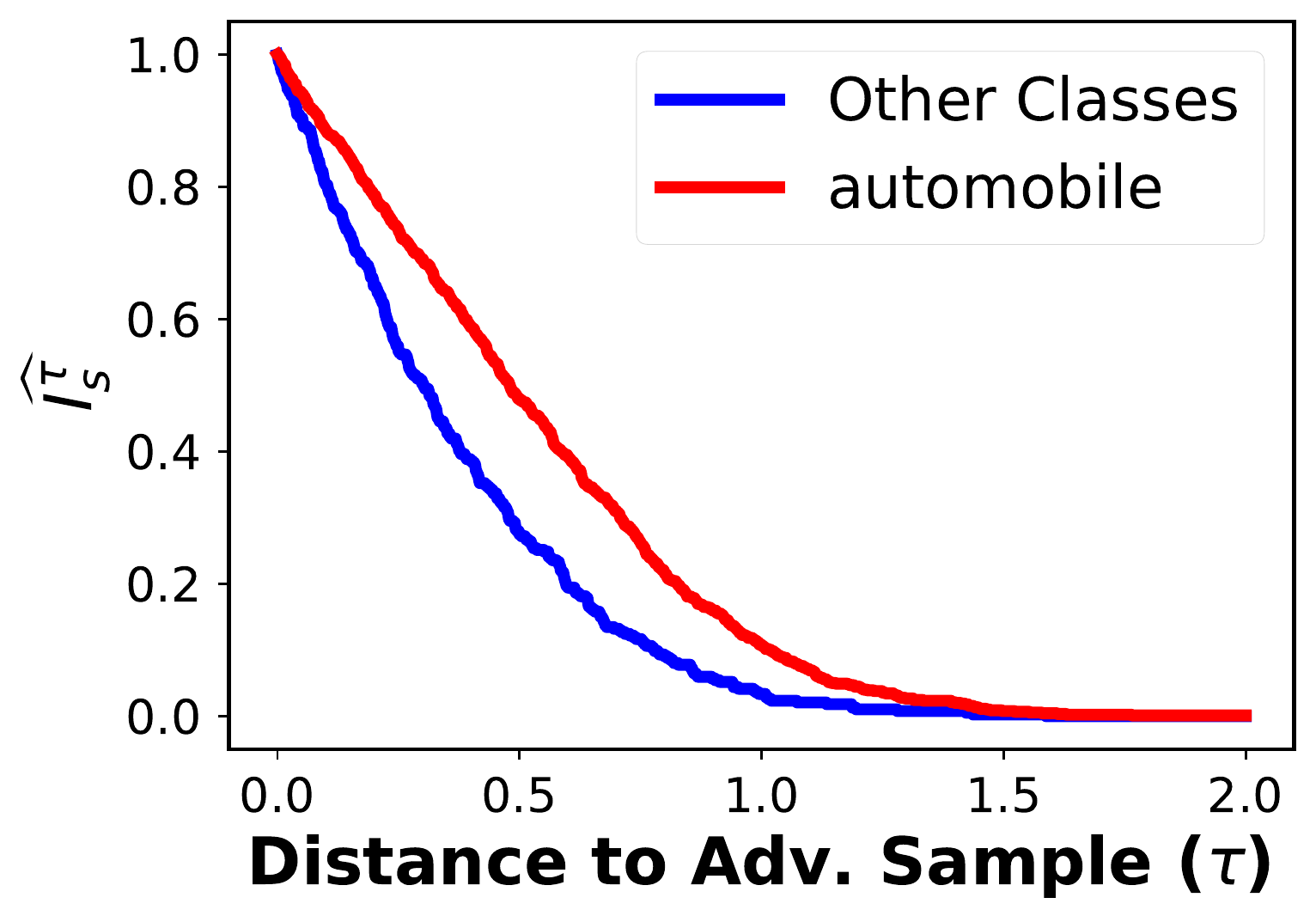}
        \caption{Reg. Automobile}
        \label{fig:cifar10_automobile_reg}
    \end{subfigure}
    
    \begin{subfigure}[b]{0.22\textwidth}
        \includegraphics[trim={0cm 0cm 0cm 0cm},clip,width=1\textwidth]{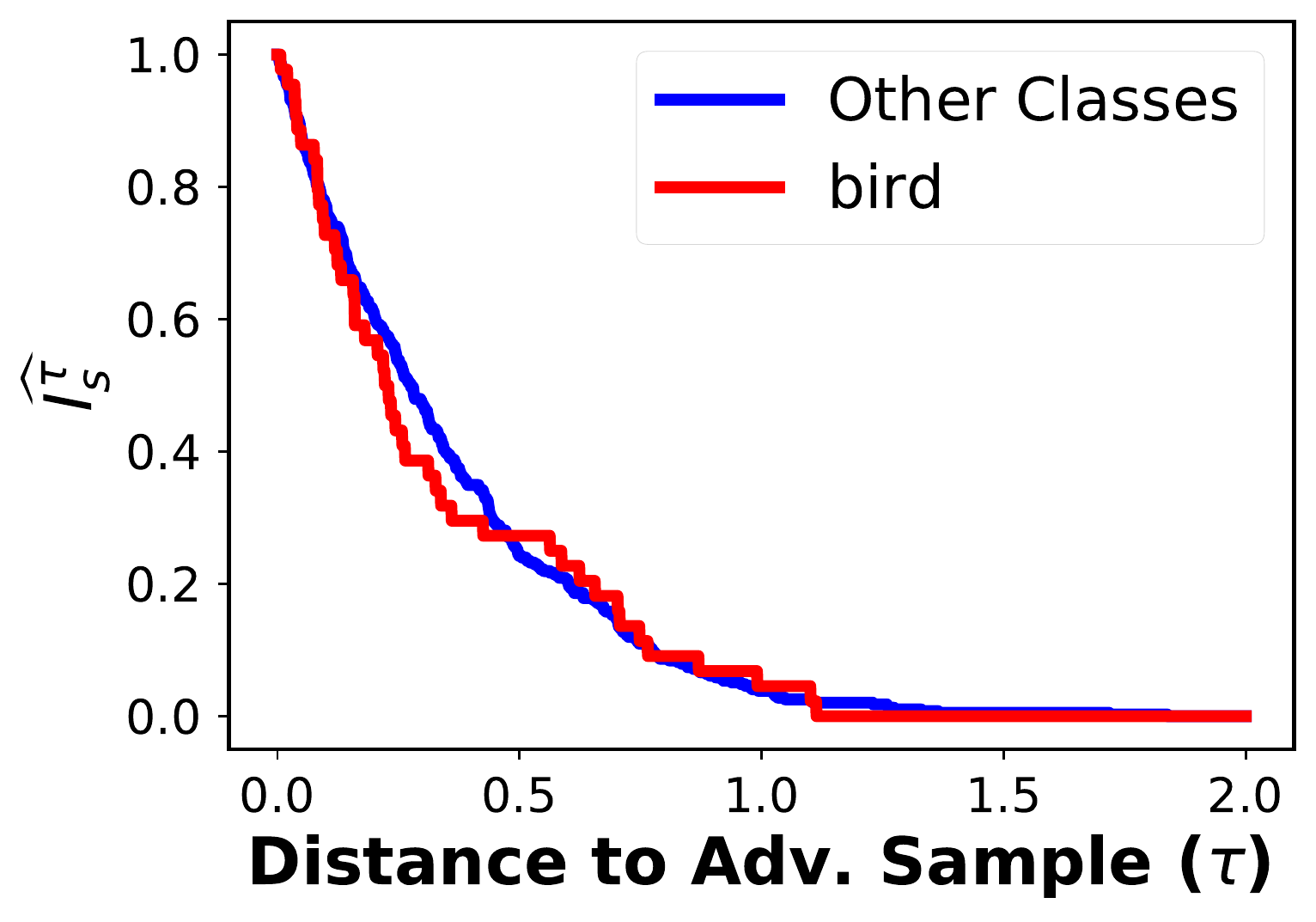}
        \caption{Bird}
        \label{fig:cifar10_bird}
    \end{subfigure}
    \begin{subfigure}[b]{0.22\textwidth}
        \includegraphics[trim={0cm 0cm 0cm 0cm},clip,width=1\textwidth]{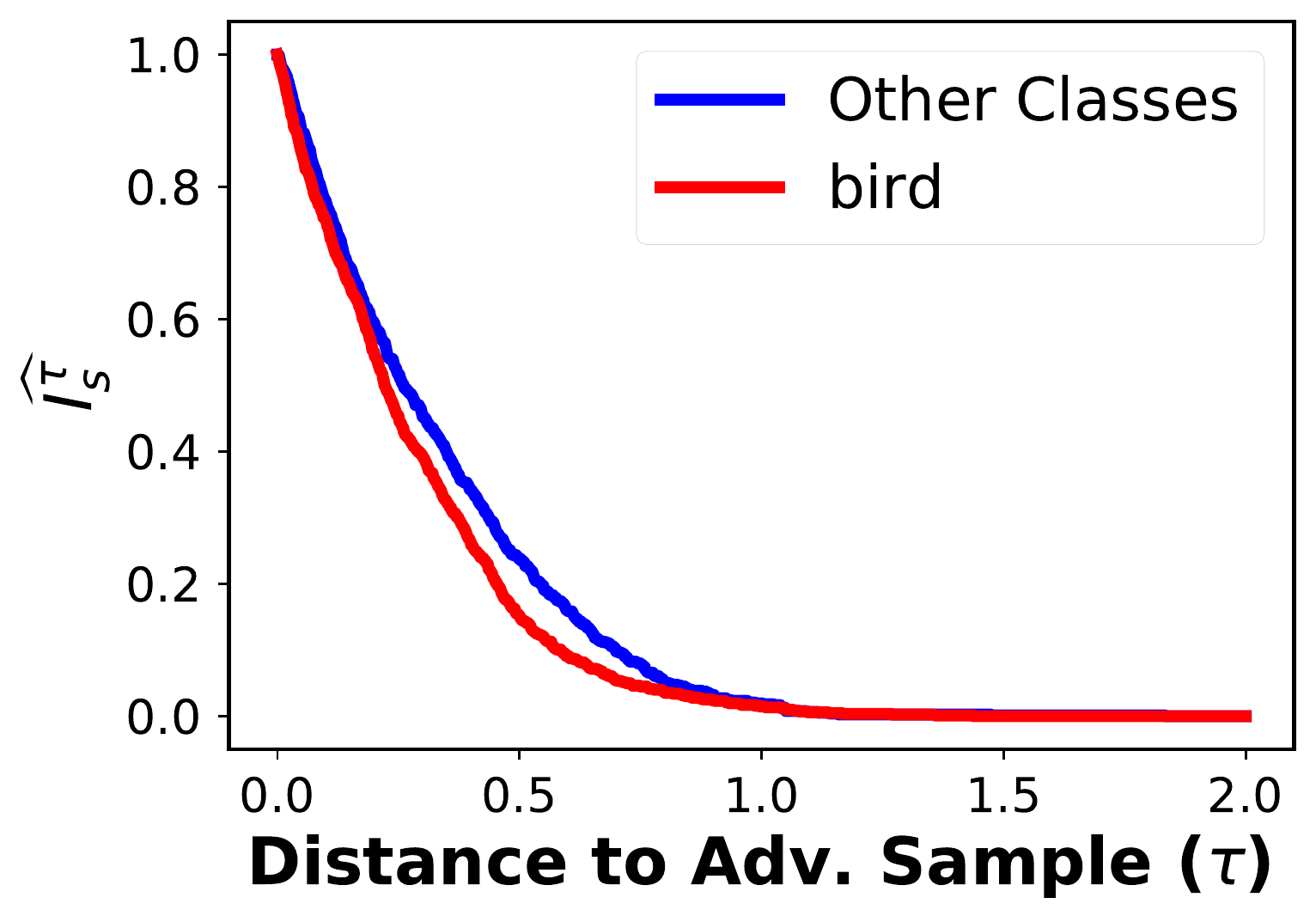}
        \caption{Reg. Bird}
        \label{fig:cifar10_bird_reg}
    \end{subfigure}
    \begin{subfigure}[b]{0.22\textwidth}
        \includegraphics[trim={0cm 0cm 0cm 0cm},clip,width=1\textwidth]{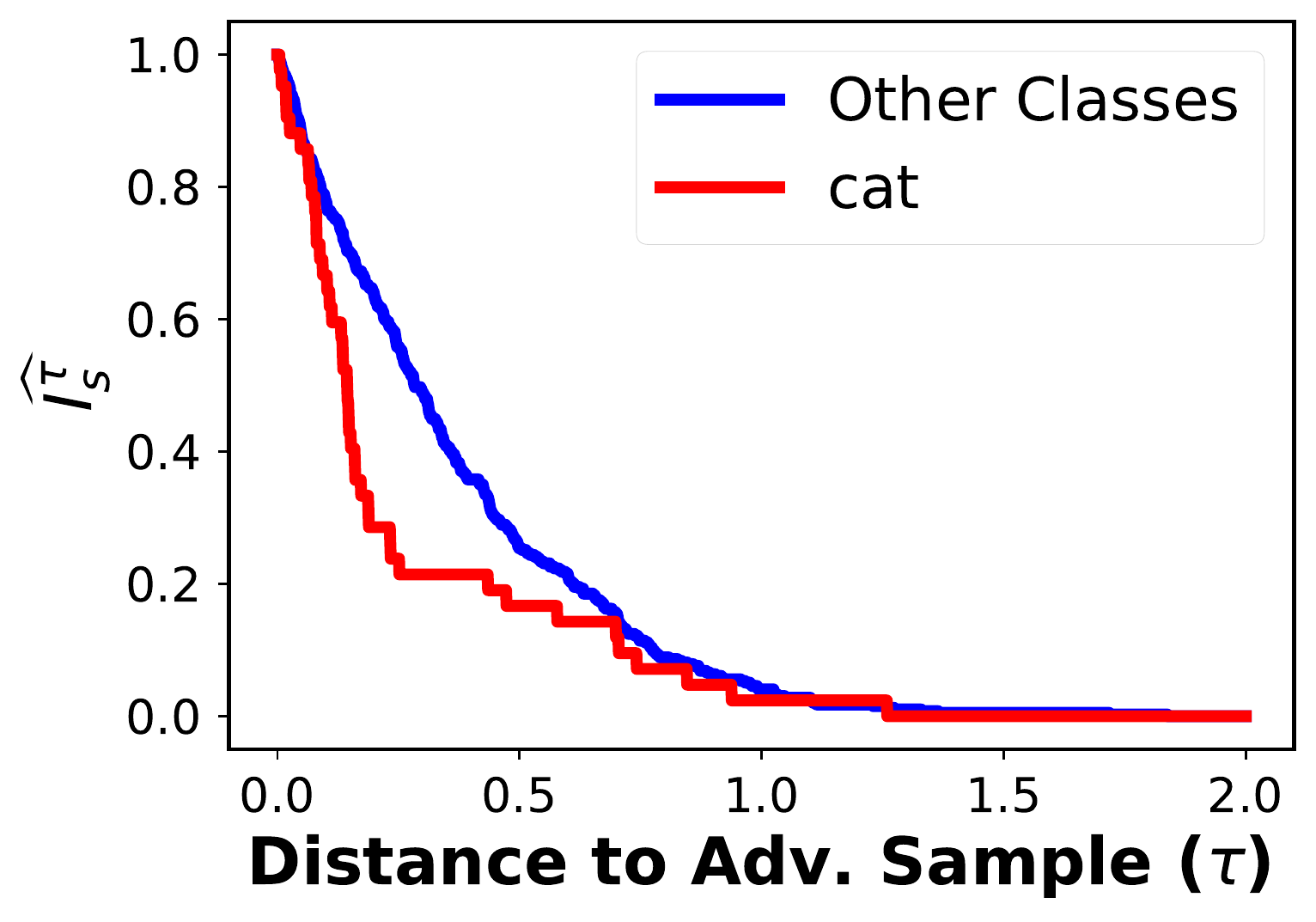}
        \caption{Cat}
        \label{fig:cifar10_cat}
    \end{subfigure}
    \begin{subfigure}[b]{0.22\textwidth}
        \includegraphics[trim={0cm 0cm 0cm 0cm},clip,width=1\textwidth]{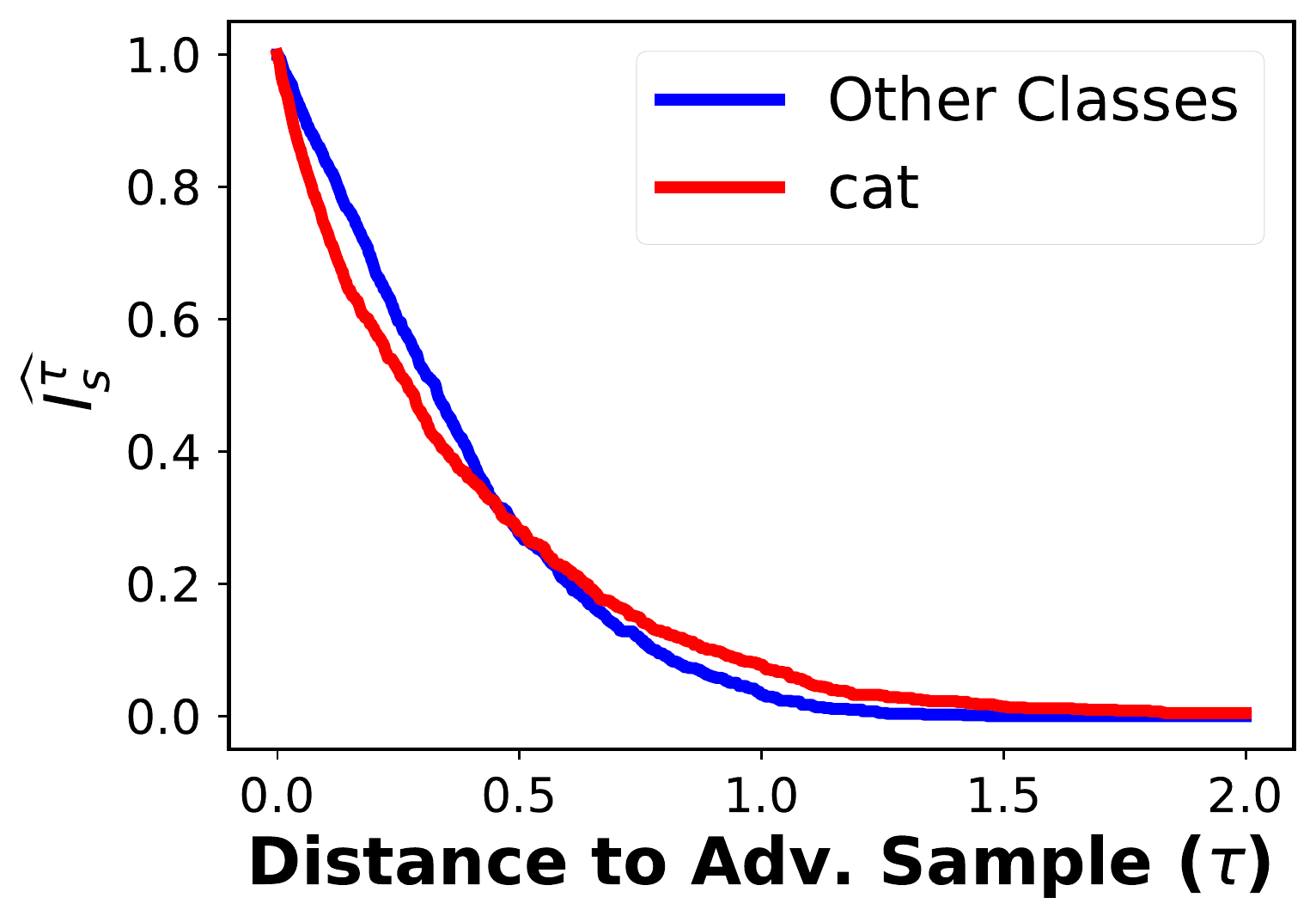}
        \caption{Reg. Cat}
        \label{fig:cifar10_cat_reg}
    \end{subfigure}
    
    \begin{subfigure}[b]{0.22\textwidth}
        \includegraphics[trim={0cm 0cm 0cm 0cm},clip,width=1\textwidth]{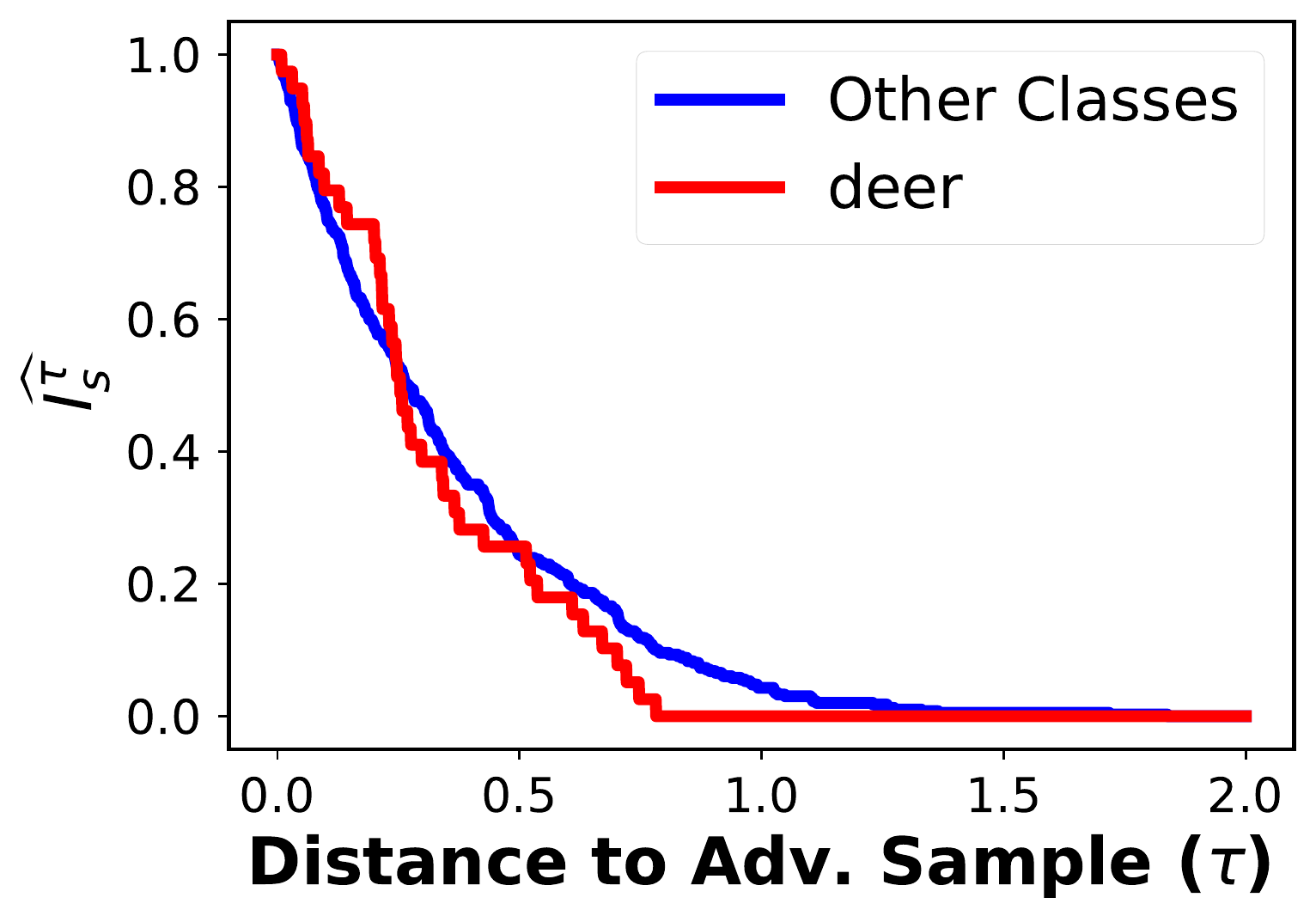}
        \caption{Deer}
        \label{fig:cifar10_deer}
    \end{subfigure}
    \begin{subfigure}[b]{0.22\textwidth}
        \includegraphics[trim={0cm 0cm 0cm 0cm},clip,width=1\textwidth]{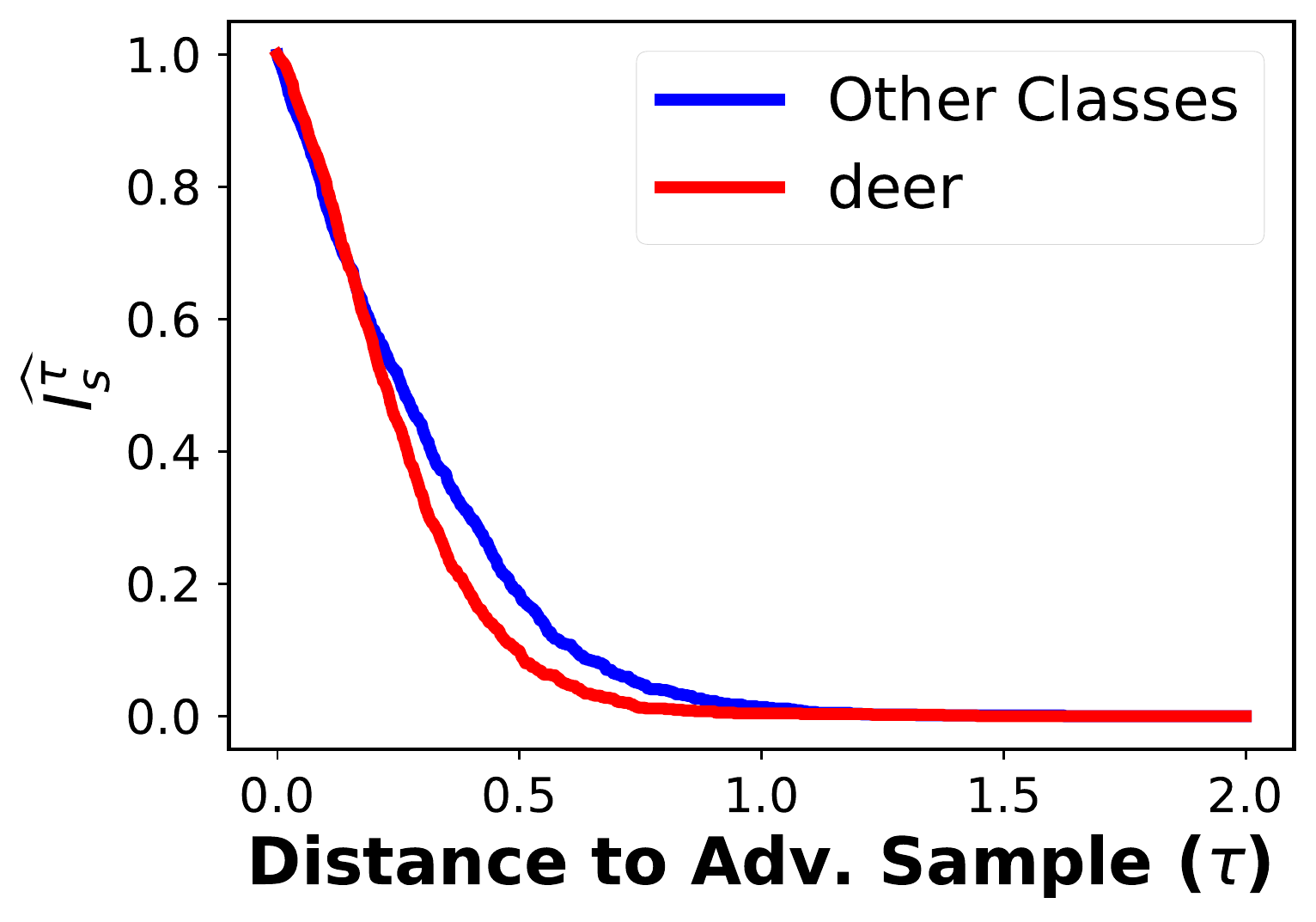}
        \caption{Reg. Deer}
        \label{fig:cifar10_deer_reg}
    \end{subfigure}
    \begin{subfigure}[b]{0.22\textwidth}
        \includegraphics[trim={0cm 0cm 0cm 0cm},clip,width=1\textwidth]{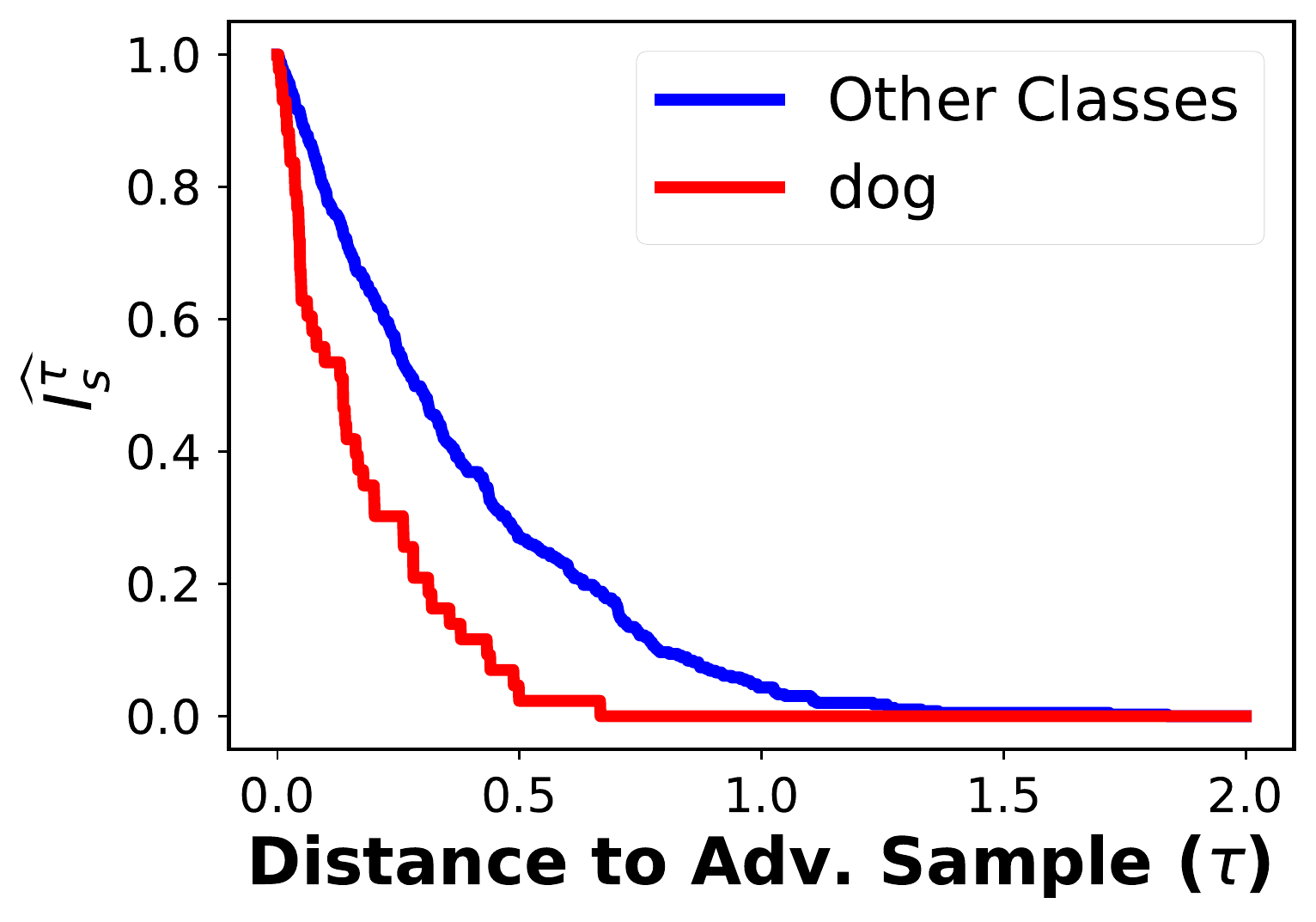}
        \caption{Dog}
        \label{fig:cifar10_dog}
    \end{subfigure}
    \begin{subfigure}[b]{0.22\textwidth}
        \includegraphics[trim={0cm 0cm 0cm 0cm},clip,width=1\textwidth]{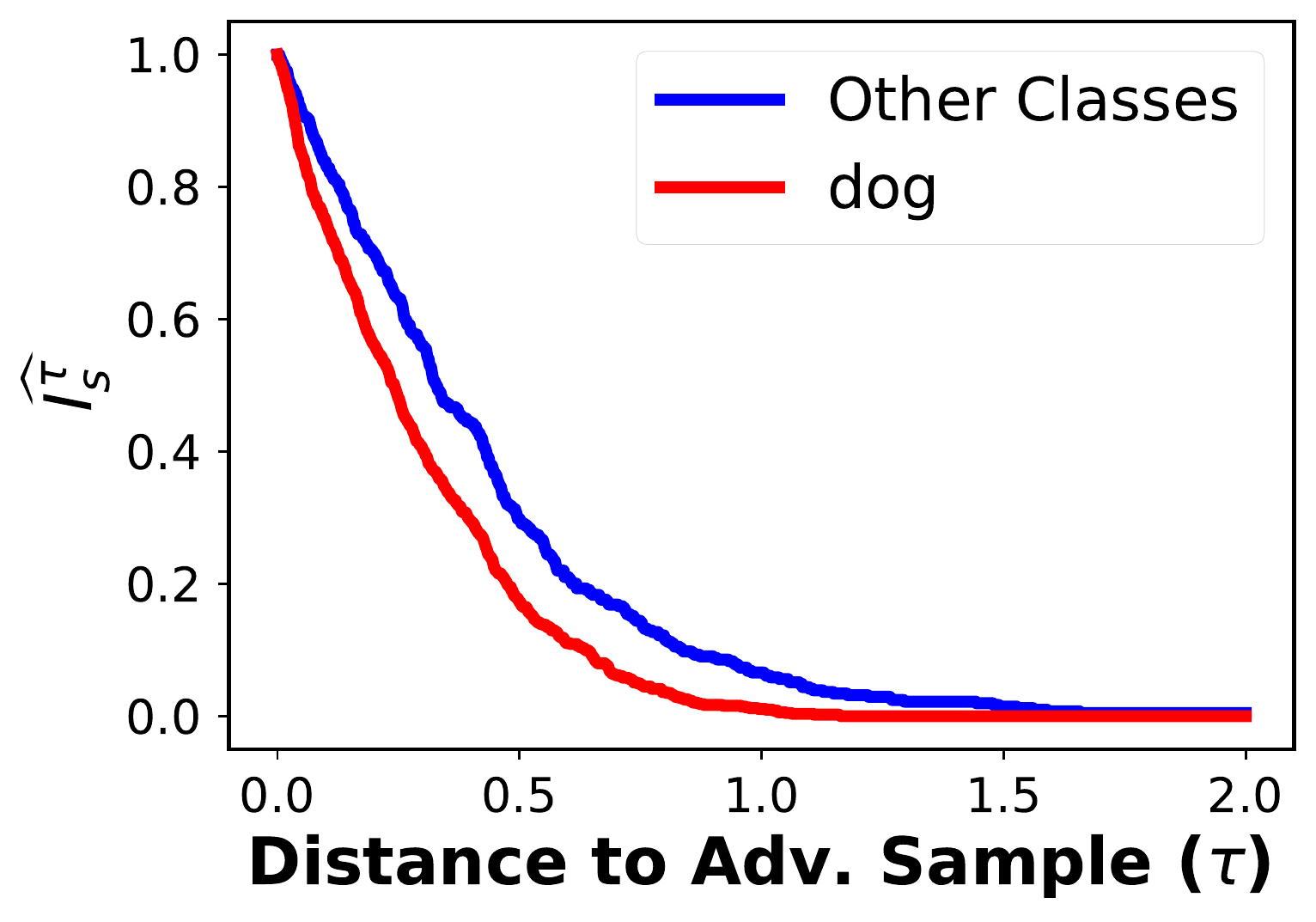}
        \caption{Reg. Dog}
        \label{fig:cifar10_dog_reg}
    \end{subfigure}
    
    \begin{subfigure}[b]{0.22\textwidth}
        \includegraphics[trim={0cm 0cm 0cm 0cm},clip,width=1\textwidth]{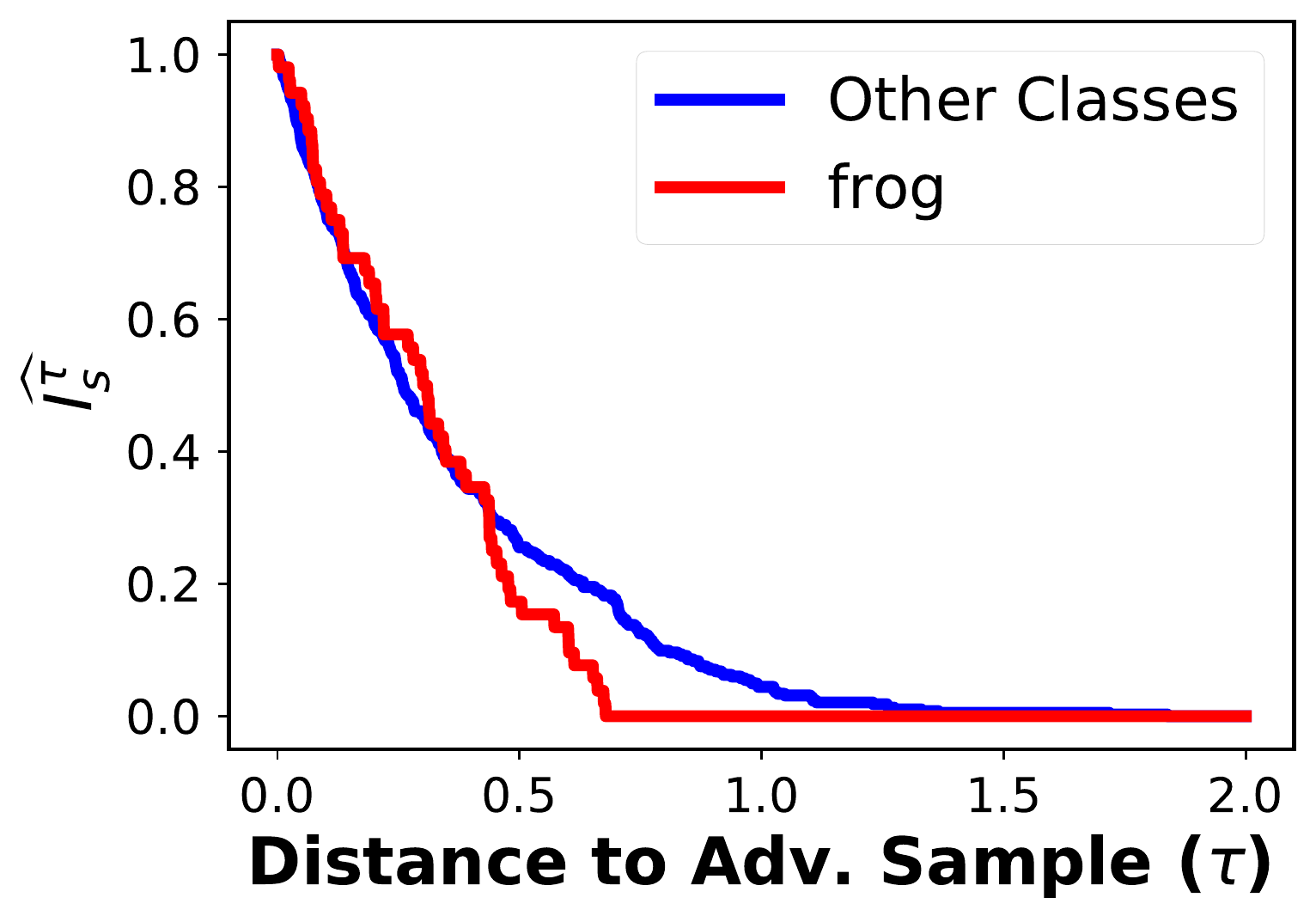}
        \caption{Frog}
        \label{fig:cifar10_frog}
    \end{subfigure}
    \begin{subfigure}[b]{0.22\textwidth}
        \includegraphics[trim={0cm 0cm 0cm 0cm},clip,width=1\textwidth]{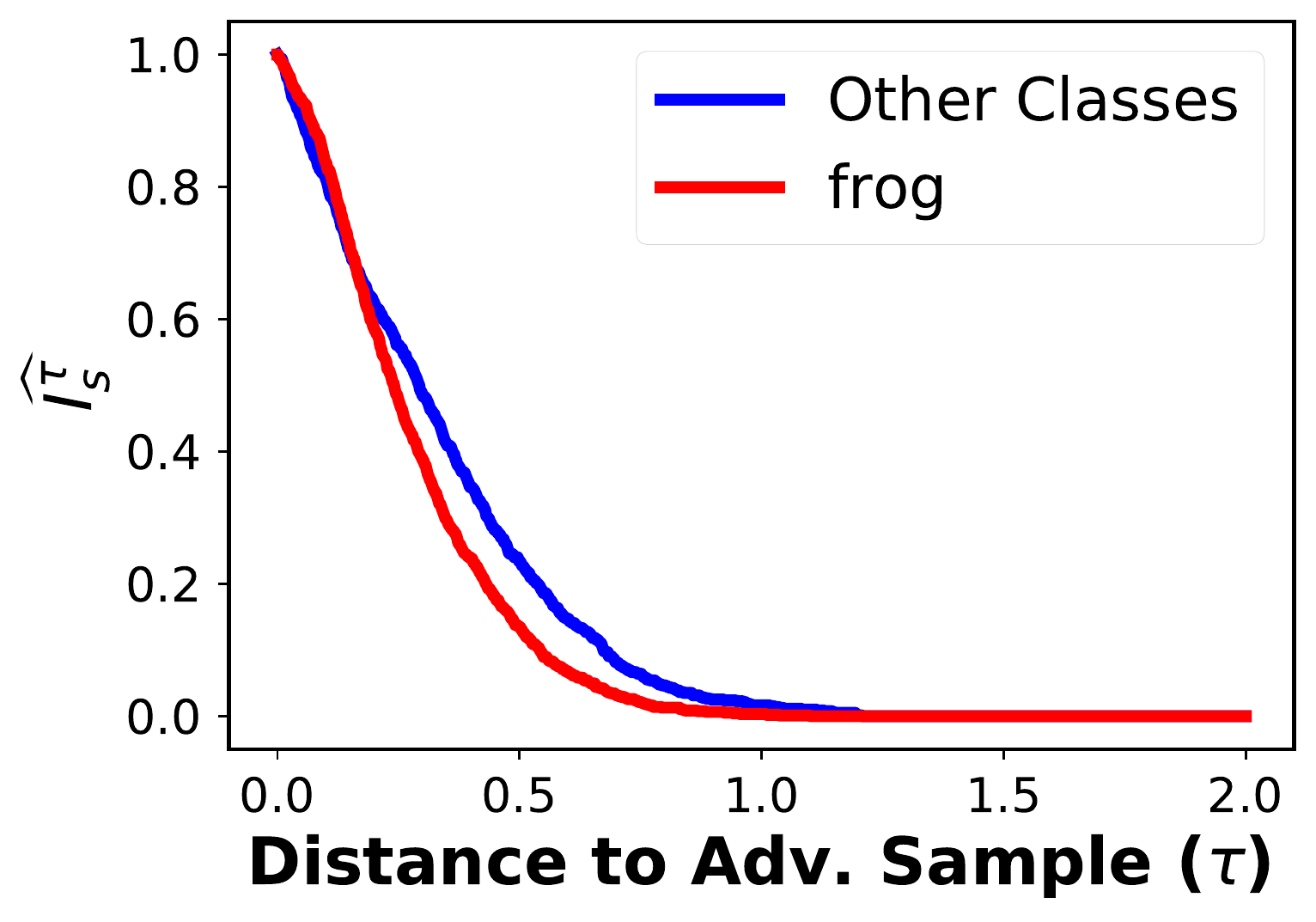}
        \caption{Reg. Frog}
        \label{fig:cifar10_frog_reg}
    \end{subfigure}
    \begin{subfigure}[b]{0.22\textwidth}
        \includegraphics[trim={0cm 0cm 0cm 0cm},clip,width=1\textwidth]{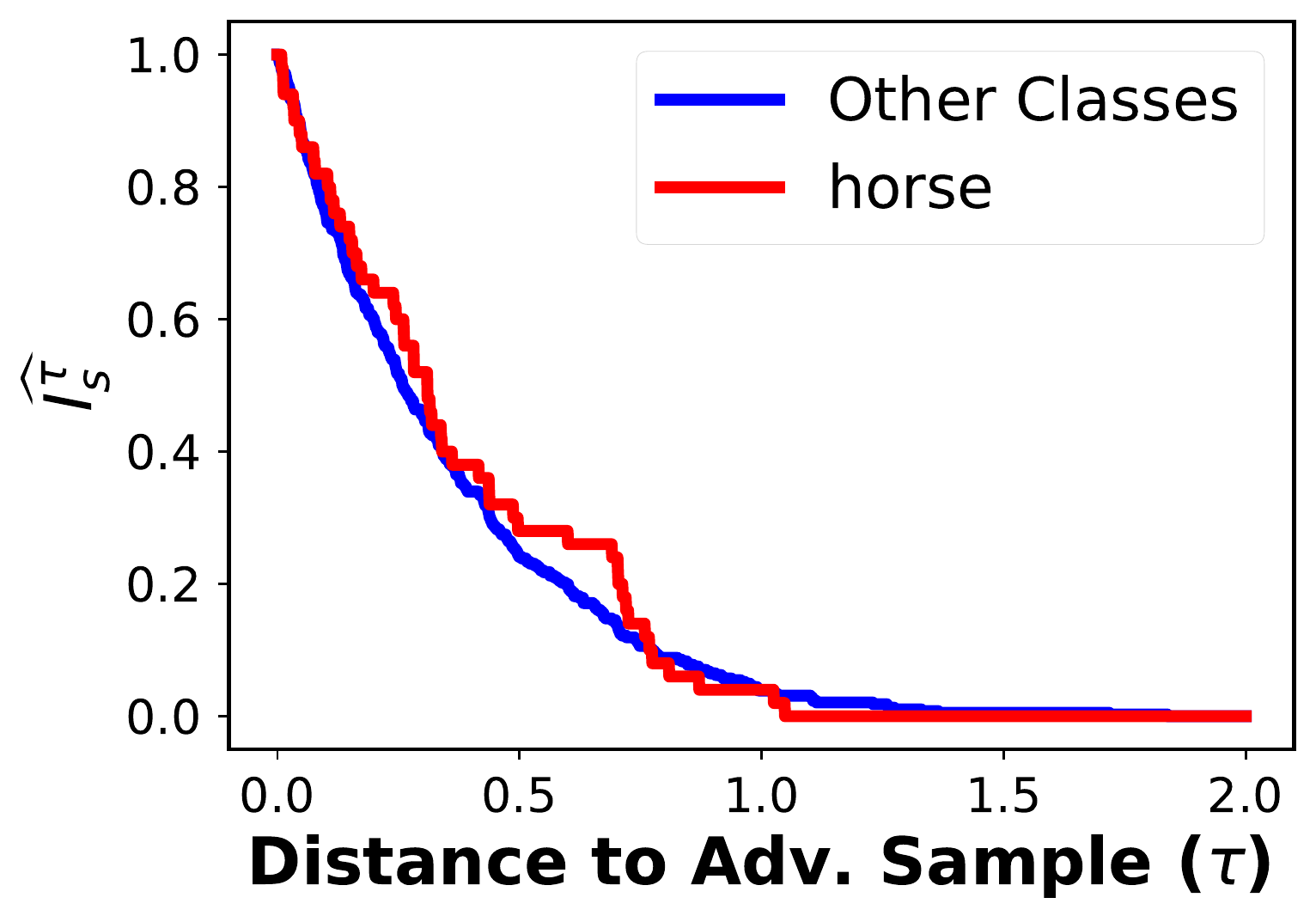}
        \caption{Horse}
        \label{fig:cifar10_horse}
    \end{subfigure}
    \begin{subfigure}[b]{0.22\textwidth}
        \includegraphics[trim={0cm 0cm 0cm 0cm},clip,width=1\textwidth]{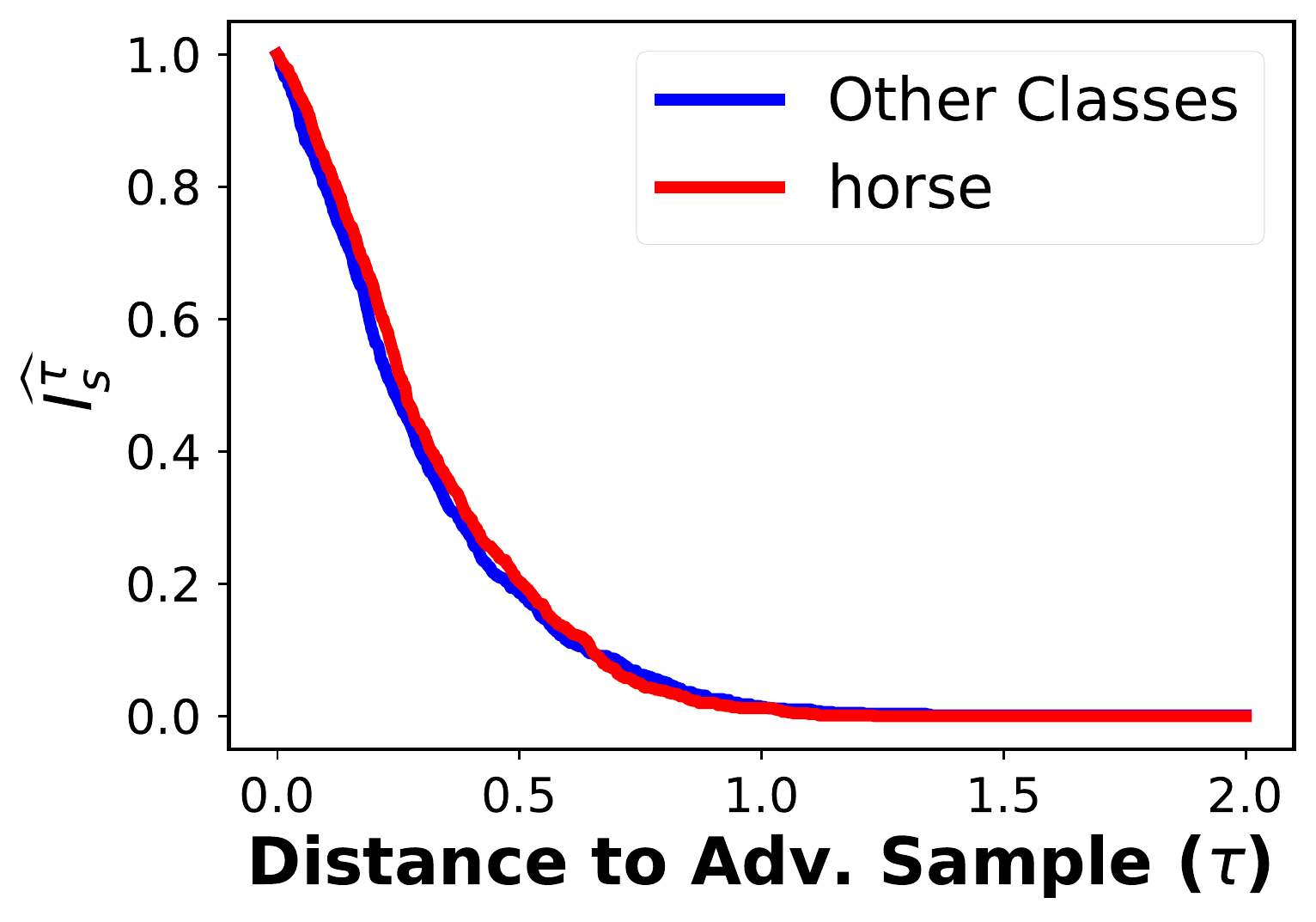}
        \caption{Reg. Horse}
        \label{fig:cifar10_horse_reg}
    \end{subfigure}
    
    \begin{subfigure}[b]{0.22\textwidth}
        \includegraphics[trim={0cm 0cm 0cm 0cm},clip,width=1\textwidth]{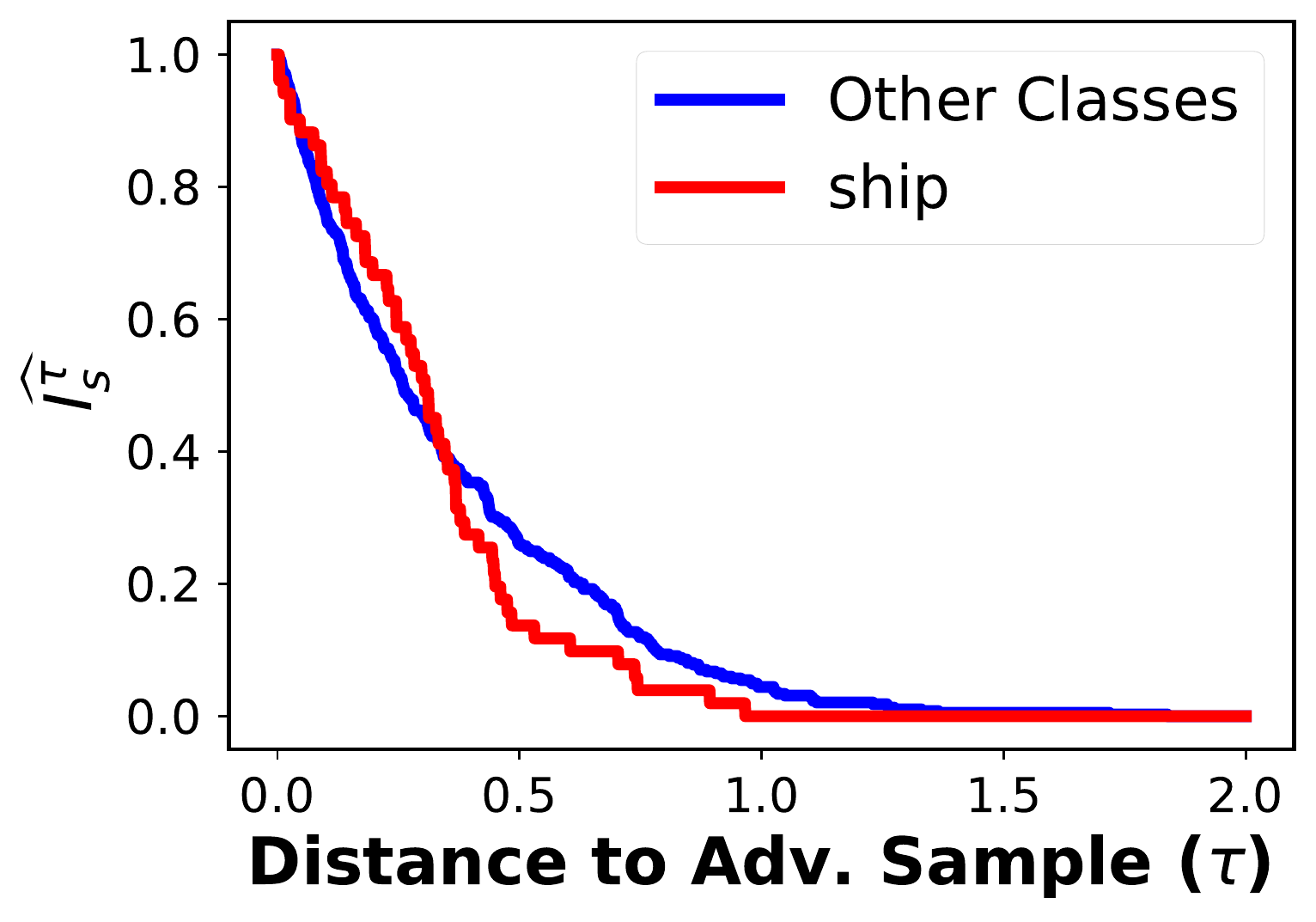}
        \caption{Ship}
        \label{fig:cifar10_ship}
    \end{subfigure}
    \begin{subfigure}[b]{0.22\textwidth}
        \includegraphics[trim={0cm 0cm 0cm 0cm},clip,width=1\textwidth]{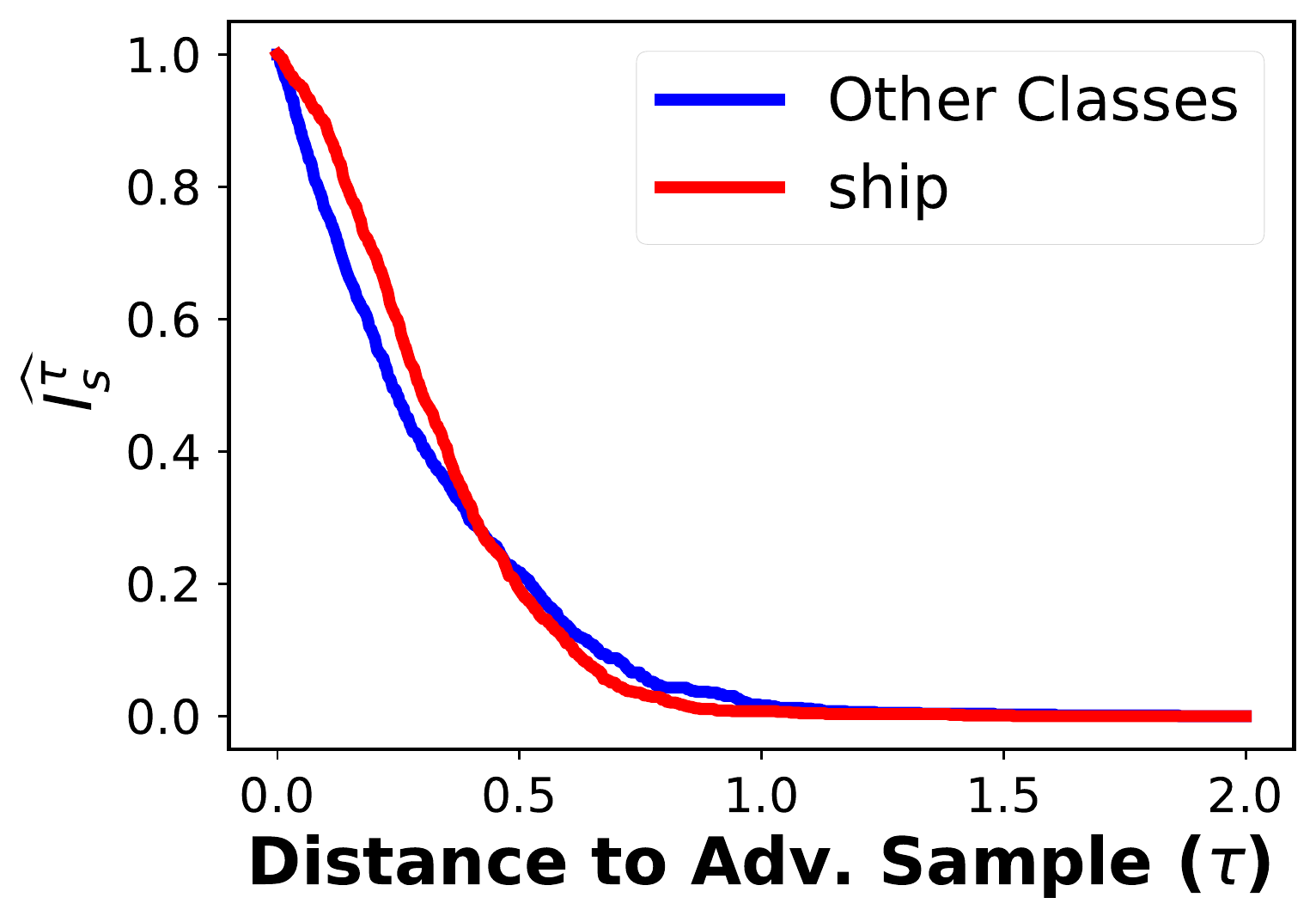}
        \caption{Reg. Ship}
        \label{fig:cifar10_ship_reg}
    \end{subfigure}
    \begin{subfigure}[b]{0.22\textwidth}
        \includegraphics[trim={0cm 0cm 0cm 0cm},clip,width=1\textwidth]{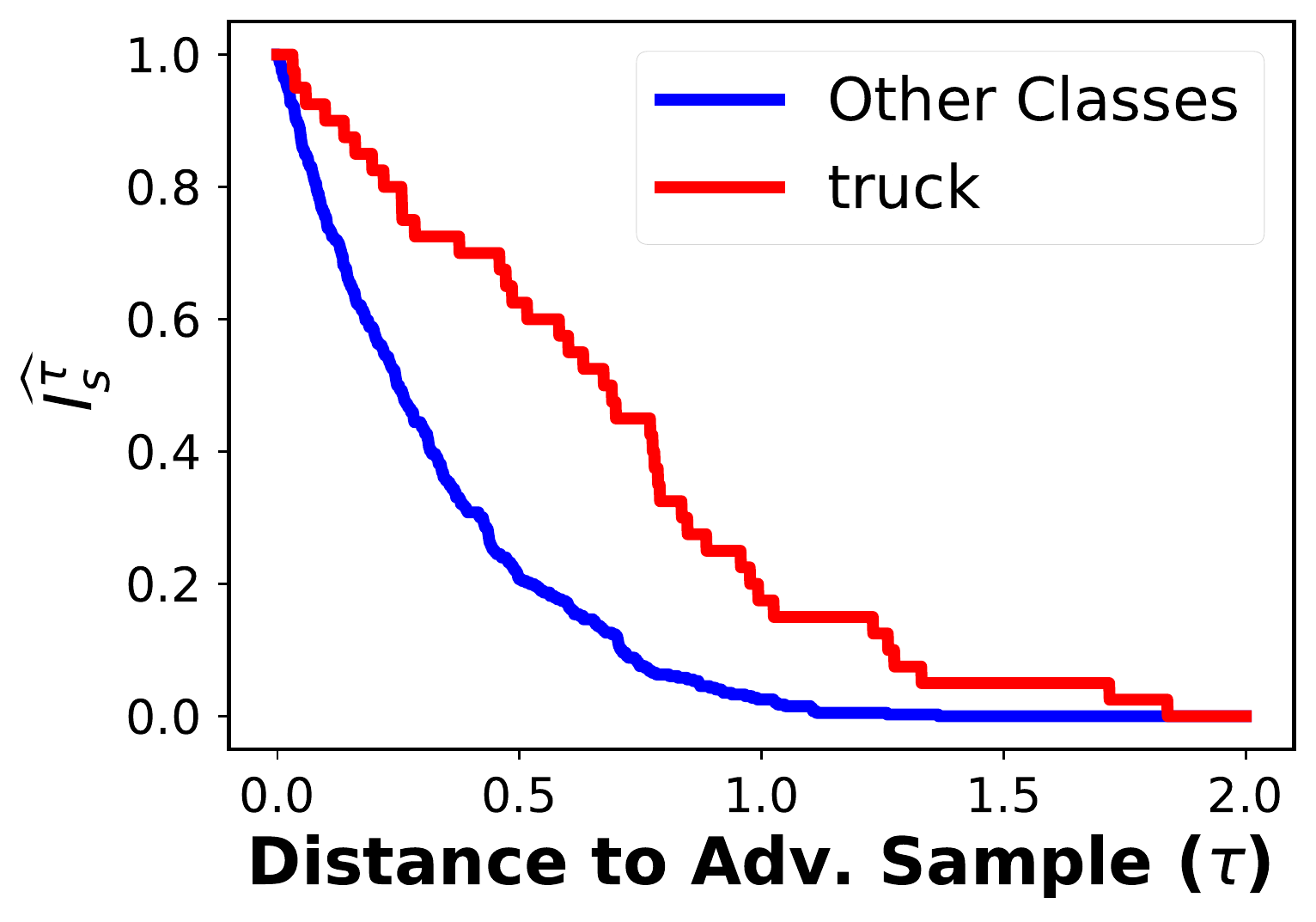}
        \caption{Truck}
        \label{fig:cifar10_truck}
    \end{subfigure}
    \begin{subfigure}[b]{0.22\textwidth}
        \includegraphics[trim={0cm 0cm 0cm 0cm},clip,width=1\textwidth]{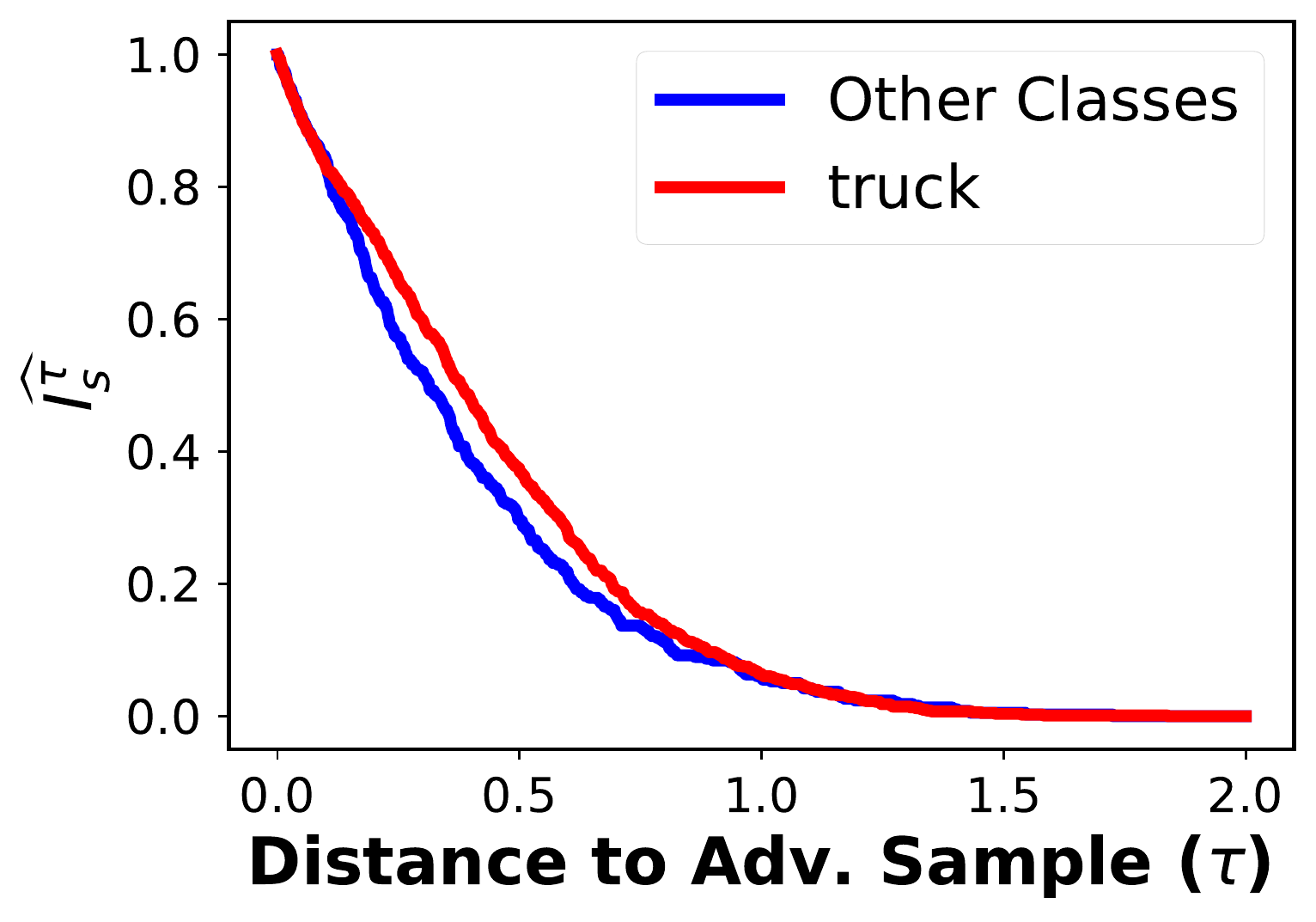}
        \caption{Reg. Truck}
        \label{fig:cifar10_truck_reg}
    \end{subfigure}

\caption{ [Regularization] CIFAR10 - Deep CNN
} 
\label{fig:reg_cifar10_deep_cnn}
\end{figure*}

\begin{figure*}[h]
    \centering
    \begin{subfigure}[b]{0.22\textwidth}
        \includegraphics[trim={0cm 0cm 0cm 0cm},clip,width=1\textwidth]{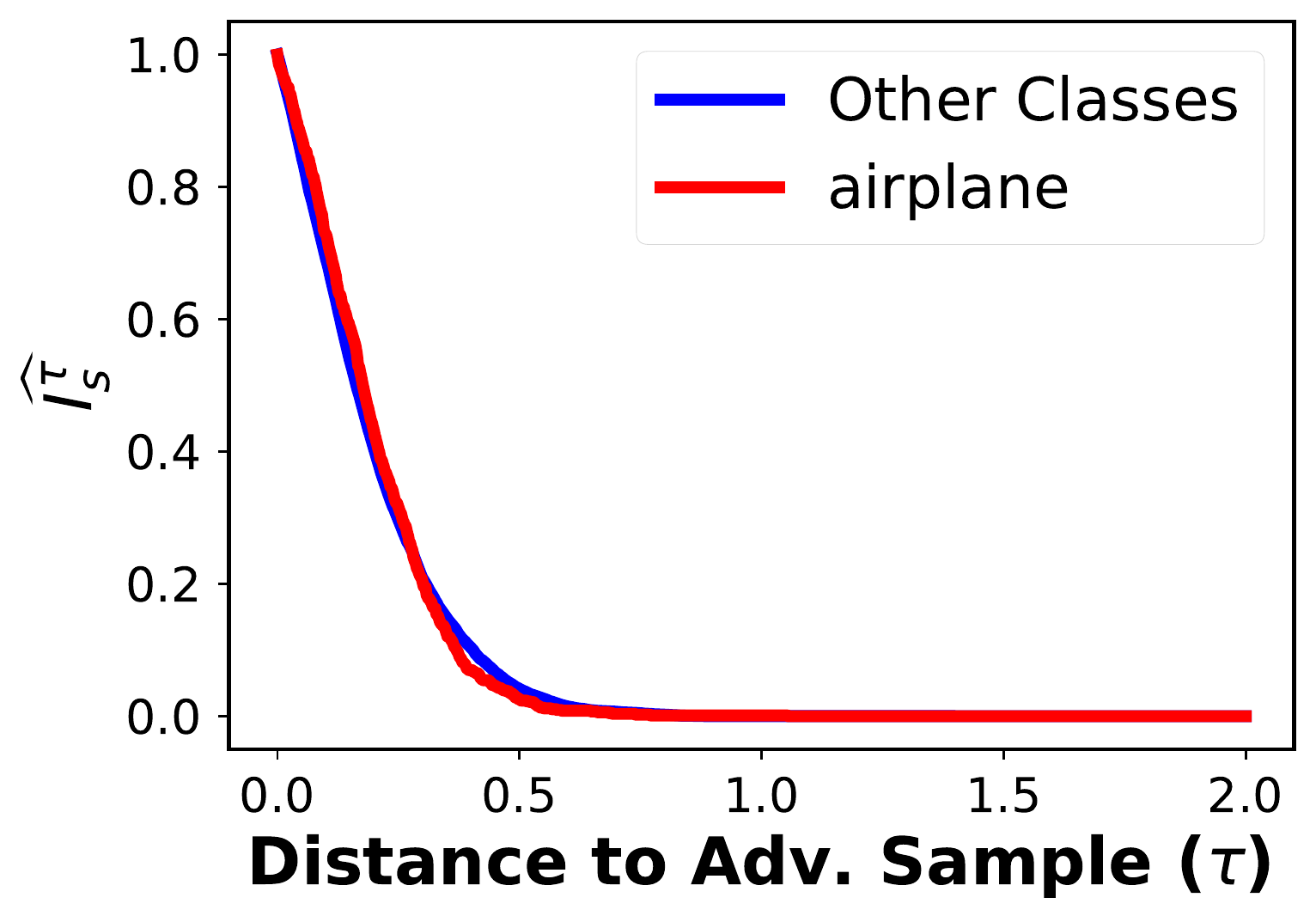}
        \caption{Airplane}
        \label{fig:cifar10_airplane_resnet}
    \end{subfigure}
    \begin{subfigure}[b]{0.22\textwidth}
        \includegraphics[trim={0cm 0cm 0cm 0cm},clip,width=1\textwidth]{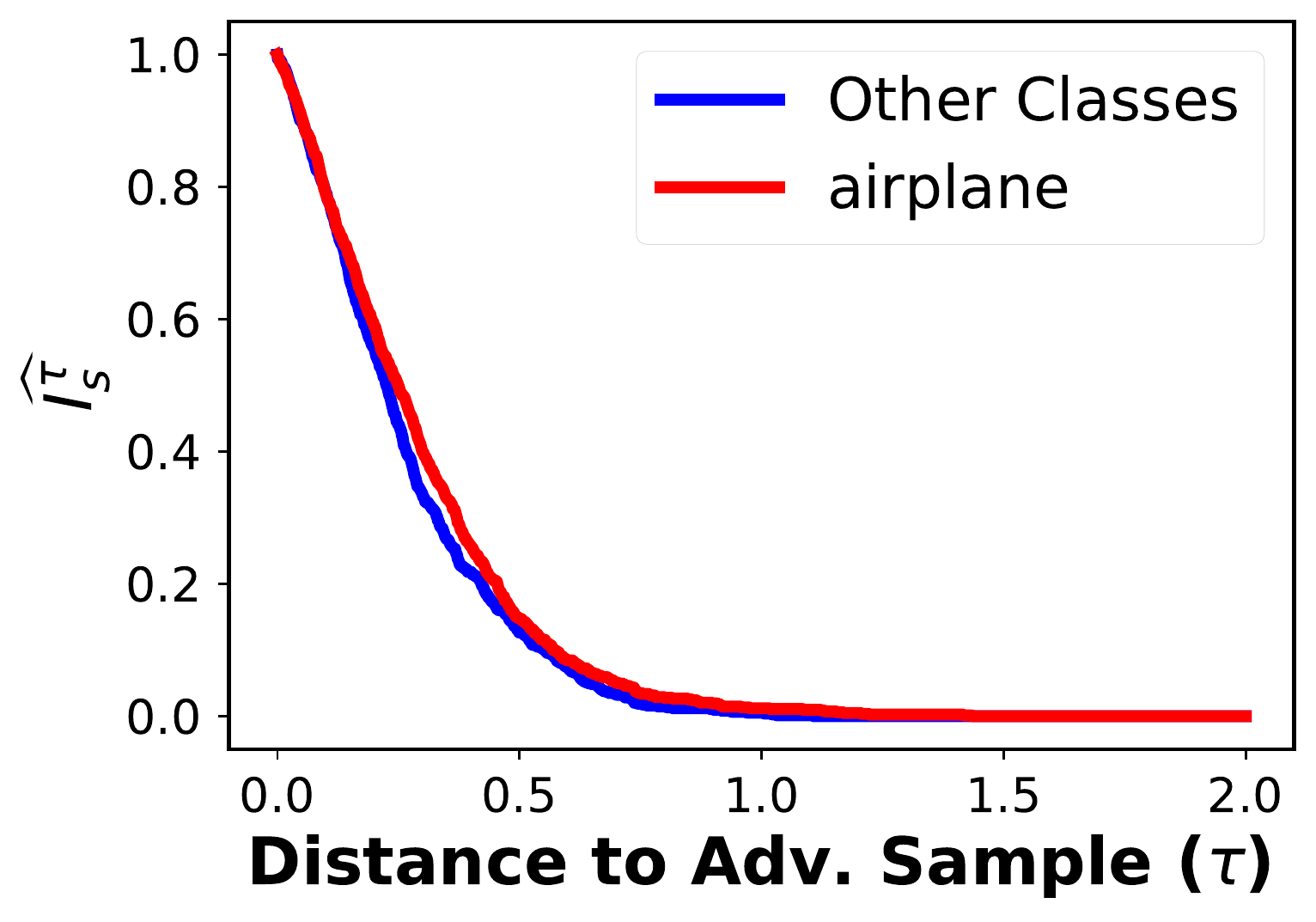}
        \caption{Reg. Airplane}
        \label{fig:cifar10_airplane_reg_resnet}
    \end{subfigure}
    \begin{subfigure}[b]{0.22\textwidth}
        \includegraphics[trim={0cm 0cm 0cm 0cm},clip,width=1\textwidth]{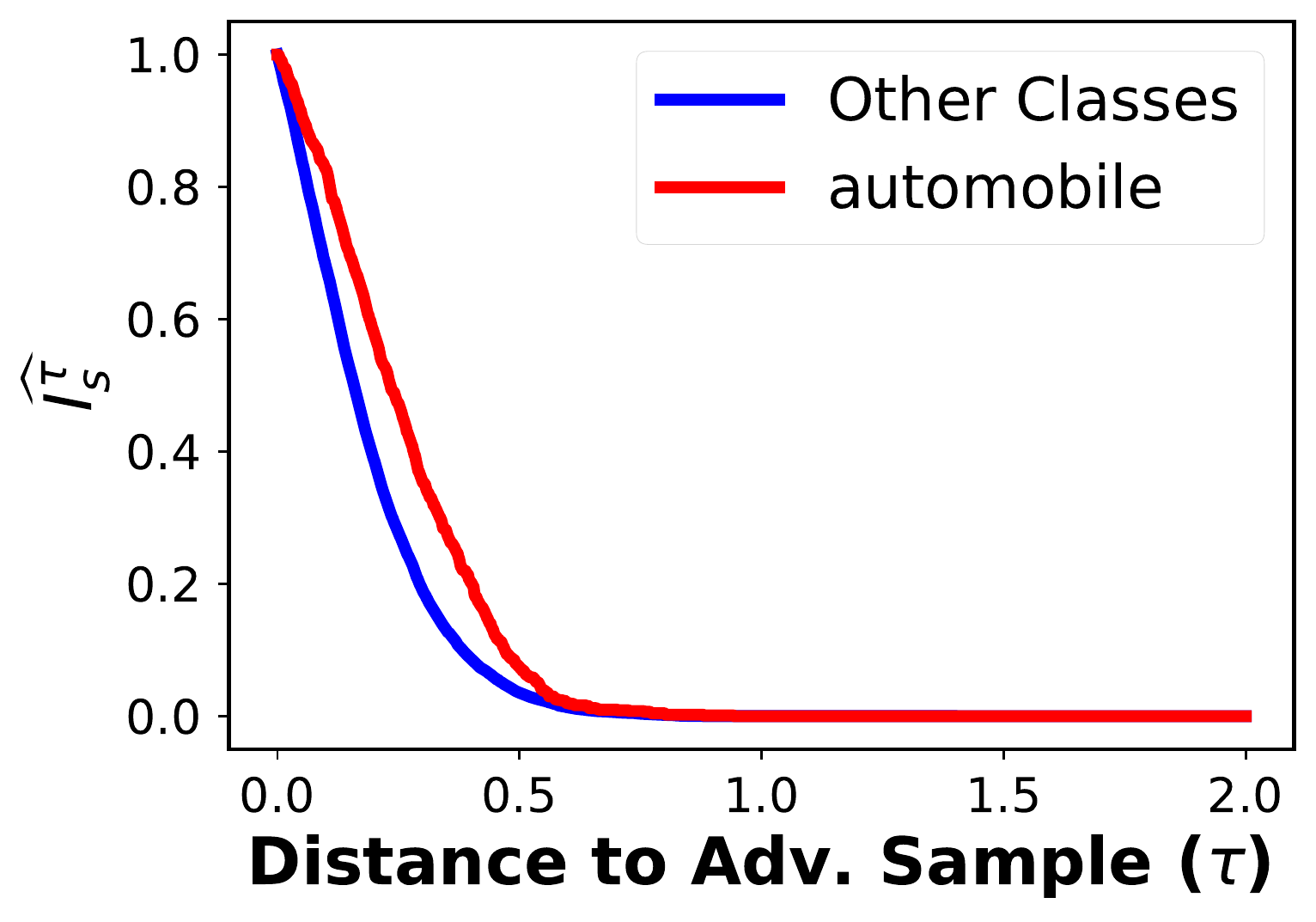}
        \caption{Automobile}
        \label{fig:cifar10_automobile_resnet}
    \end{subfigure}
    \begin{subfigure}[b]{0.22\textwidth}
        \includegraphics[trim={0cm 0cm 0cm 0cm},clip,width=1\textwidth]{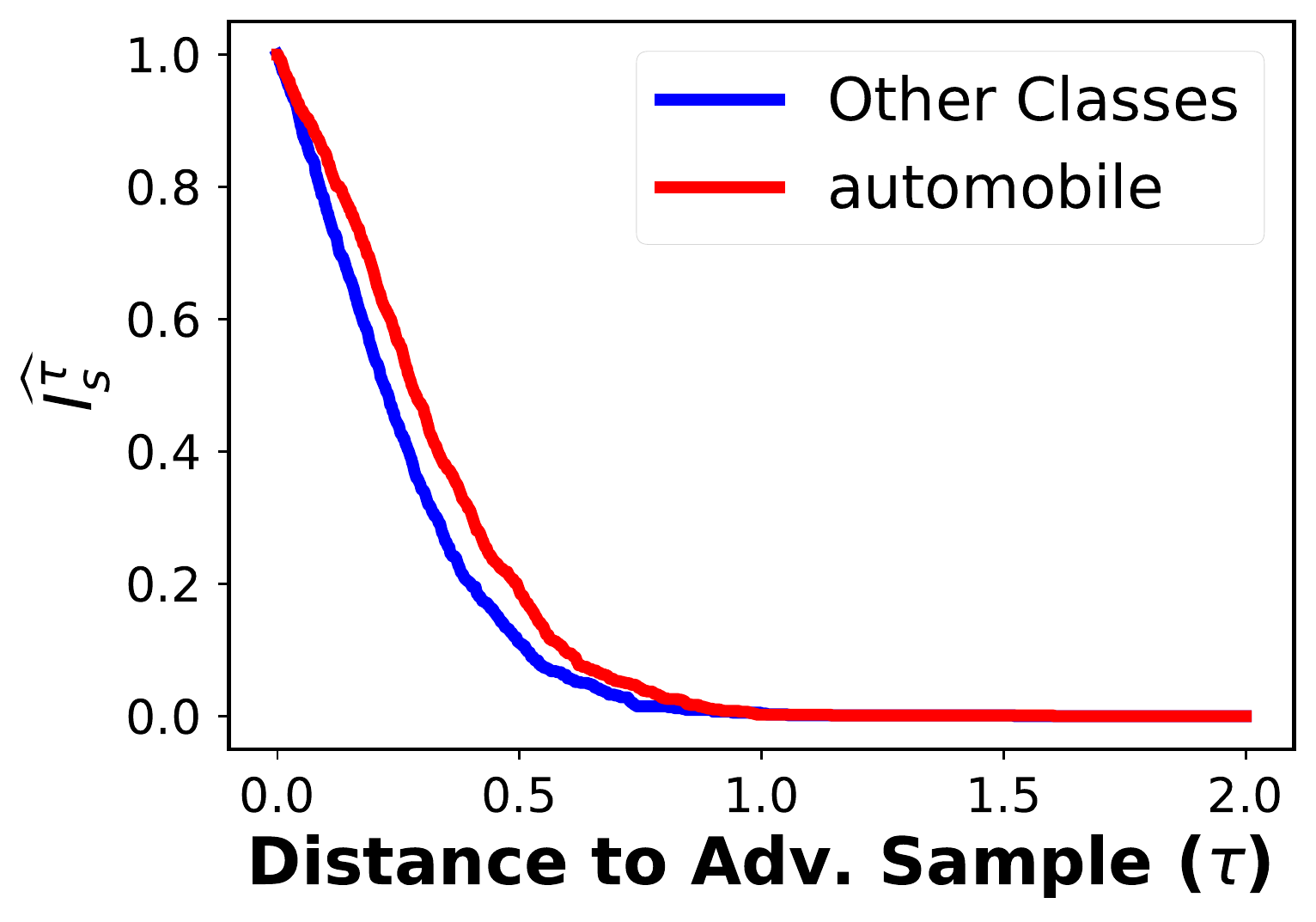}
        \caption{Reg. Automobile}
        \label{fig:cifar10_automobile_reg_resnet}
    \end{subfigure}
    
    \begin{subfigure}[b]{0.22\textwidth}
        \includegraphics[trim={0cm 0cm 0cm 0cm},clip,width=1\textwidth]{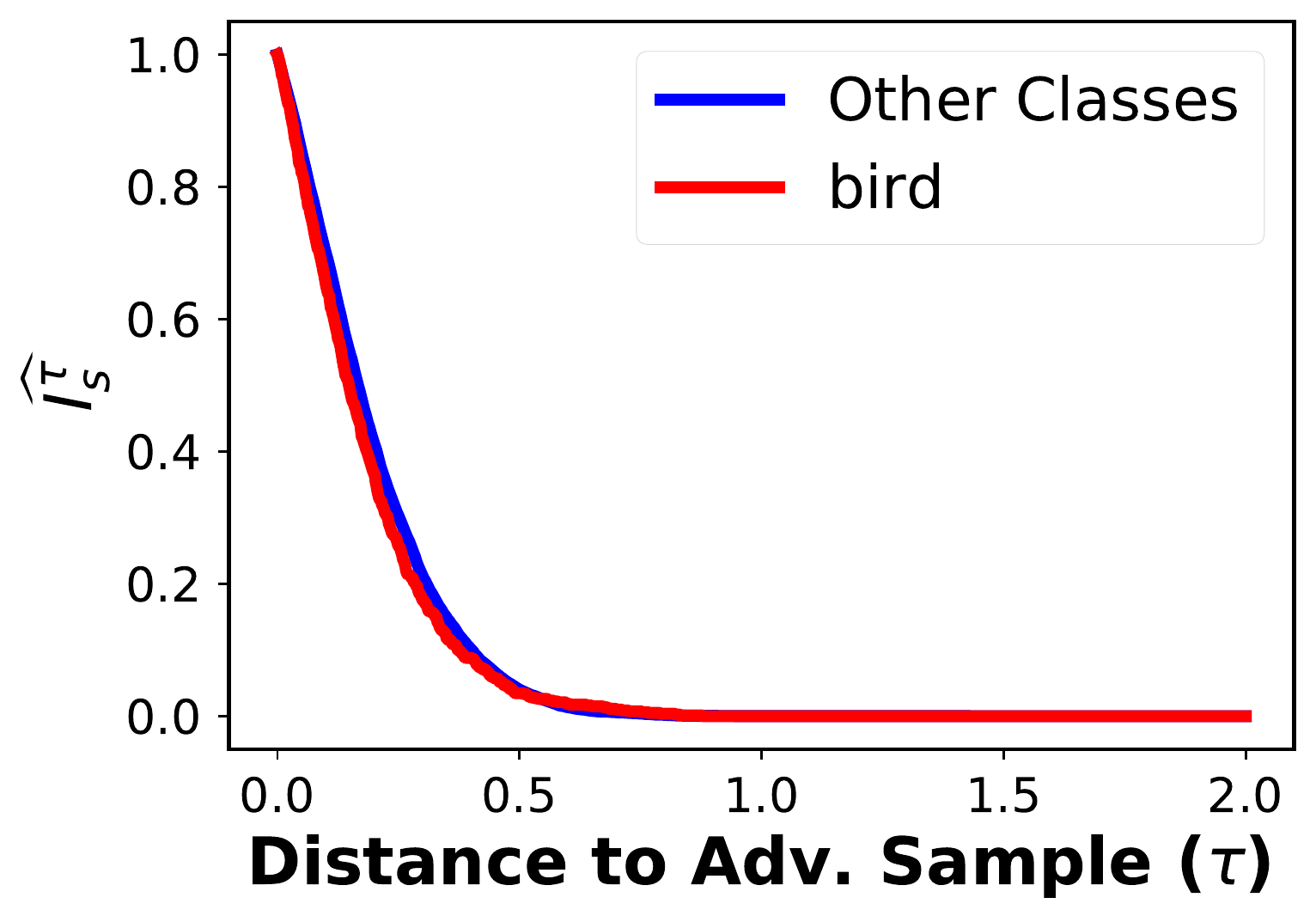}
        \caption{Bird}
        \label{fig:cifar10_bird_resnet}
    \end{subfigure}
    \begin{subfigure}[b]{0.22\textwidth}
        \includegraphics[trim={0cm 0cm 0cm 0cm},clip,width=1\textwidth]{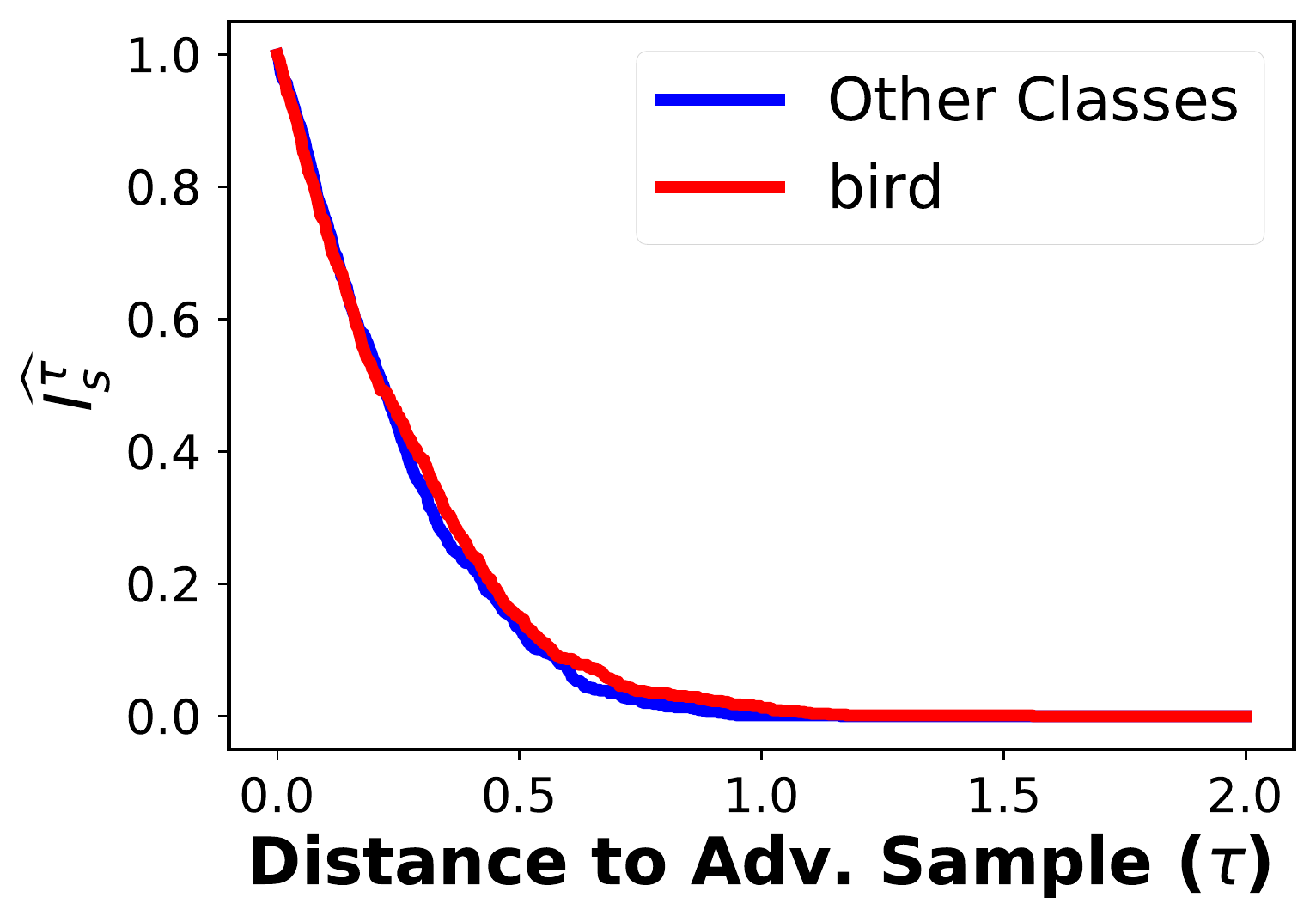}
        \caption{Reg. Bird}
        \label{fig:cifar10_bird_reg_resnet}
    \end{subfigure}
    \begin{subfigure}[b]{0.22\textwidth}
        \includegraphics[trim={0cm 0cm 0cm 0cm},clip,width=1\textwidth]{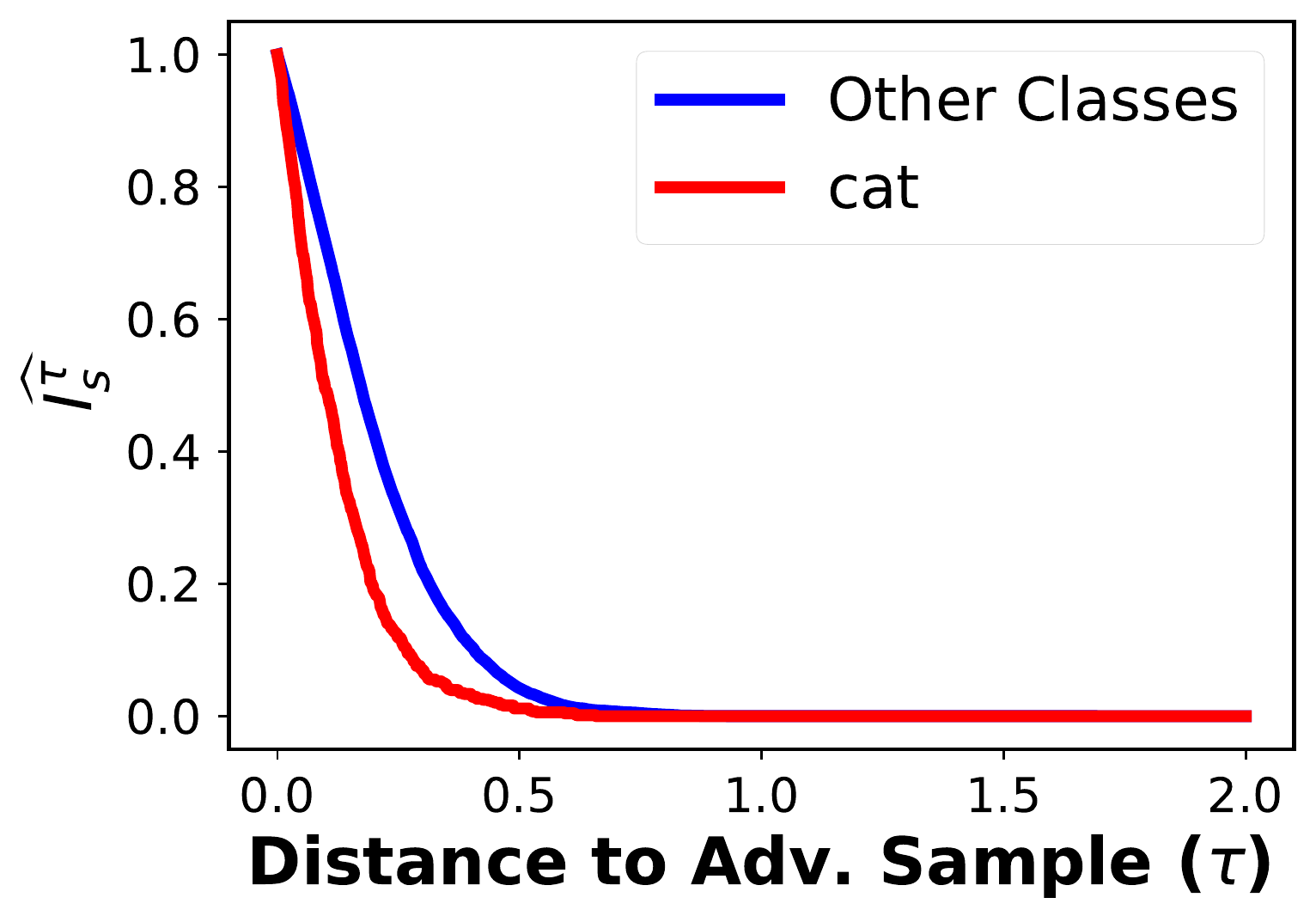}
        \caption{Cat}
        \label{fig:cifar10_cat_resnet}
    \end{subfigure}
    \begin{subfigure}[b]{0.22\textwidth}
        \includegraphics[trim={0cm 0cm 0cm 0cm},clip,width=1\textwidth]{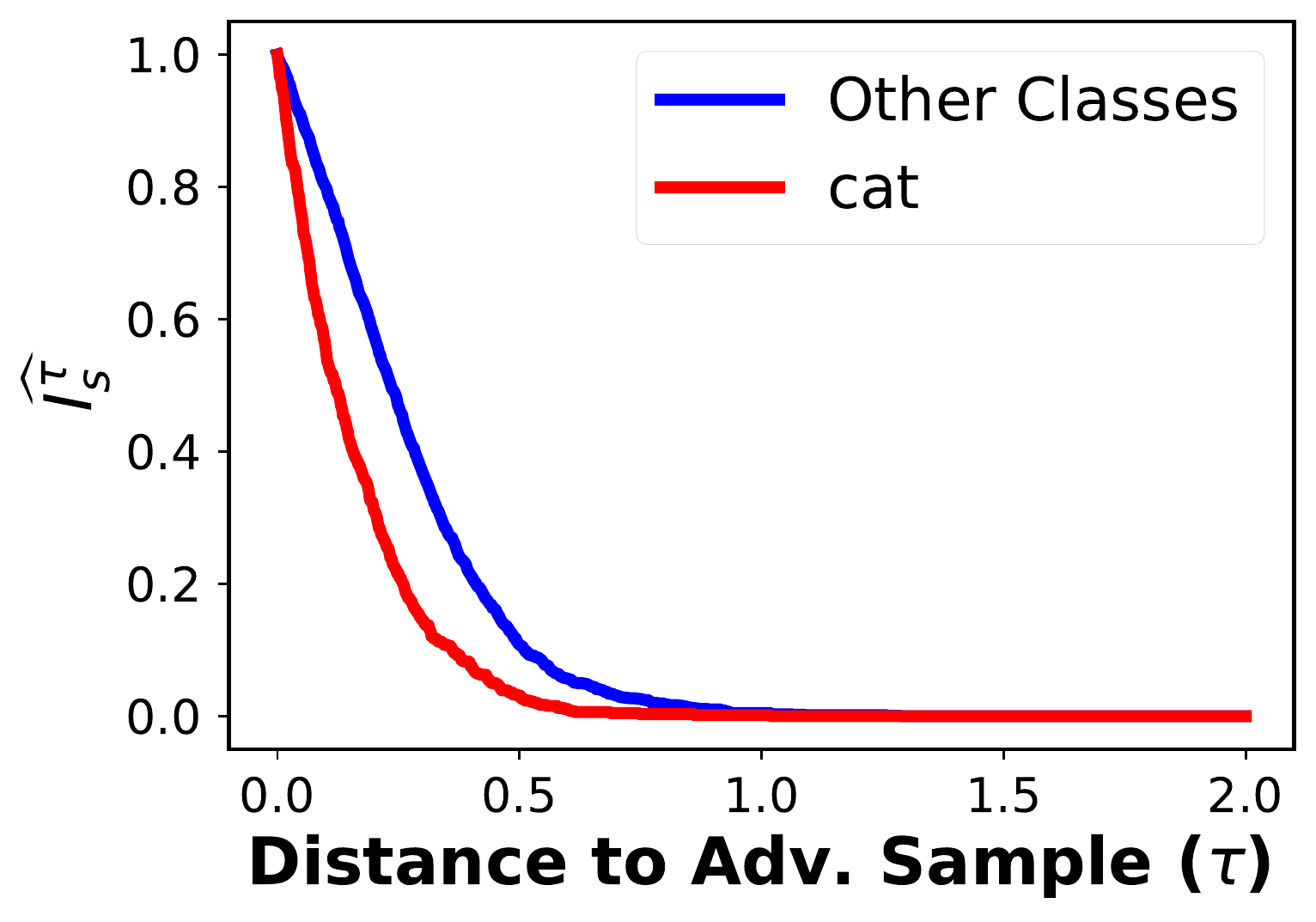}
        \caption{Reg. Cat}
        \label{fig:cifar10_cat_reg_resnet}
    \end{subfigure}
    
    \begin{subfigure}[b]{0.22\textwidth}
        \includegraphics[trim={0cm 0cm 0cm 0cm},clip,width=1\textwidth]{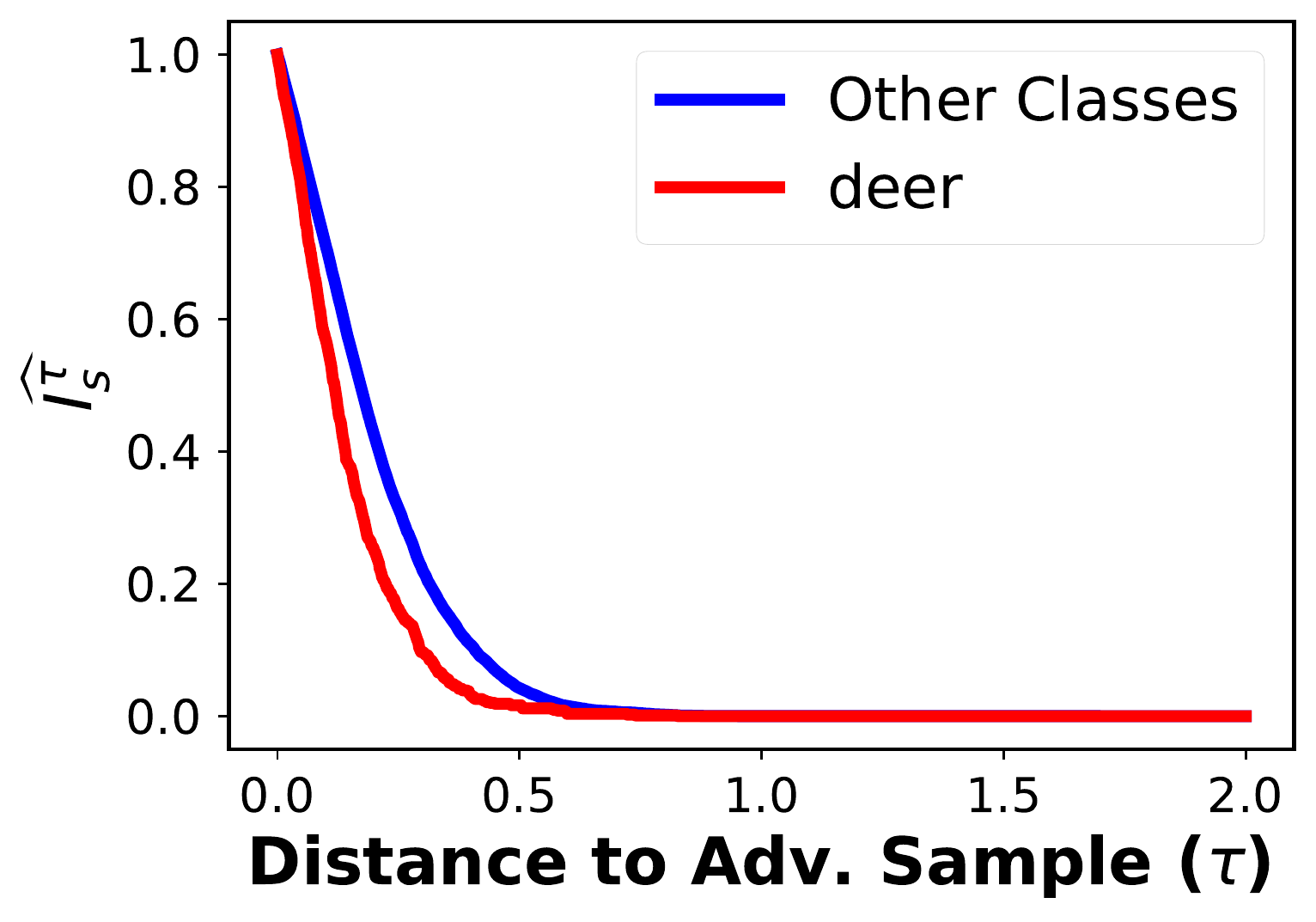}
        \caption{Deer}
        \label{fig:cifar10_deer_resnet}
    \end{subfigure}
    \begin{subfigure}[b]{0.22\textwidth}
        \includegraphics[trim={0cm 0cm 0cm 0cm},clip,width=1\textwidth]{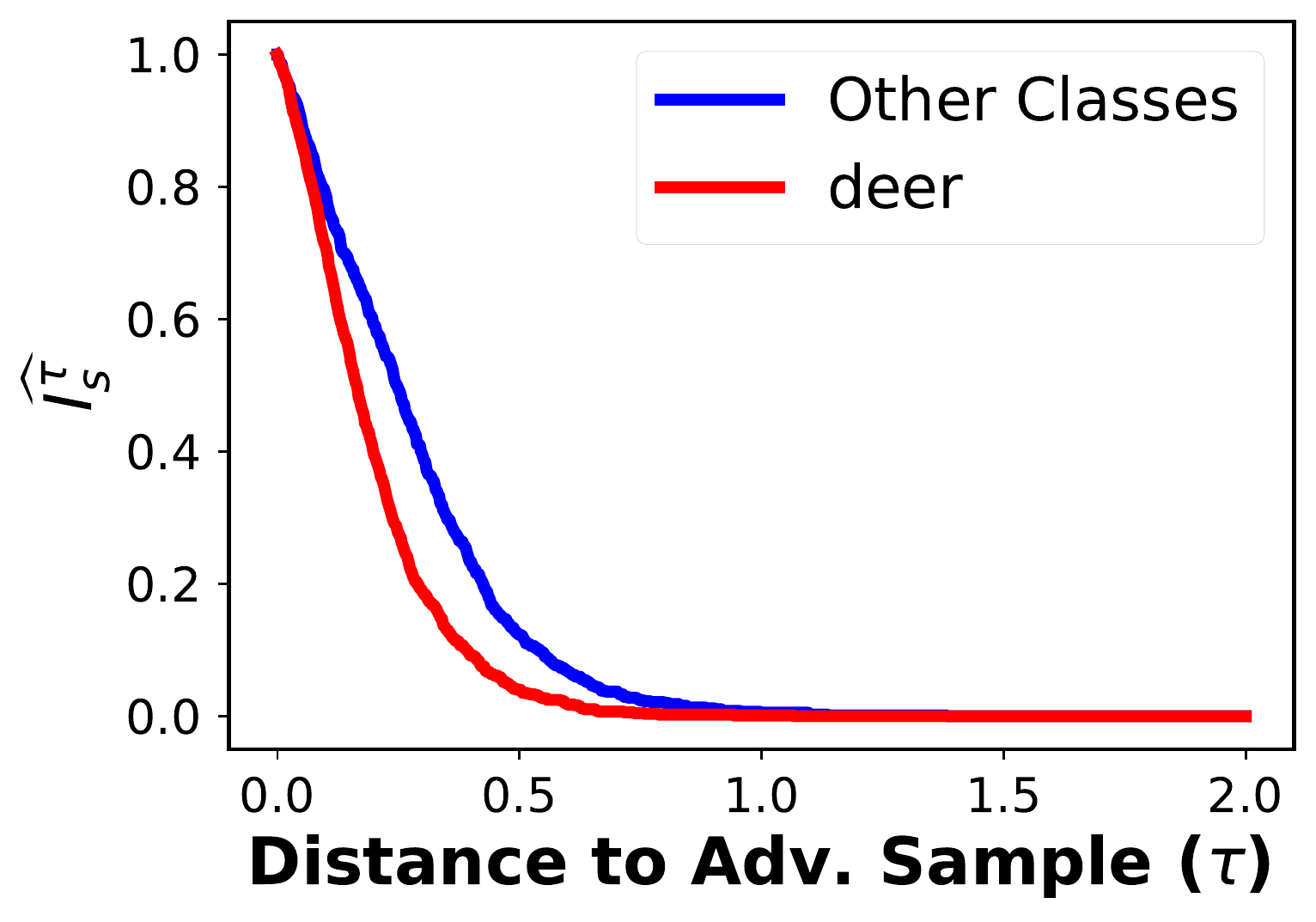}
        \caption{Reg. Deer}
        \label{fig:cifar10_deer_reg_resnet}
    \end{subfigure}
    \begin{subfigure}[b]{0.22\textwidth}
        \includegraphics[trim={0cm 0cm 0cm 0cm},clip,width=1\textwidth]{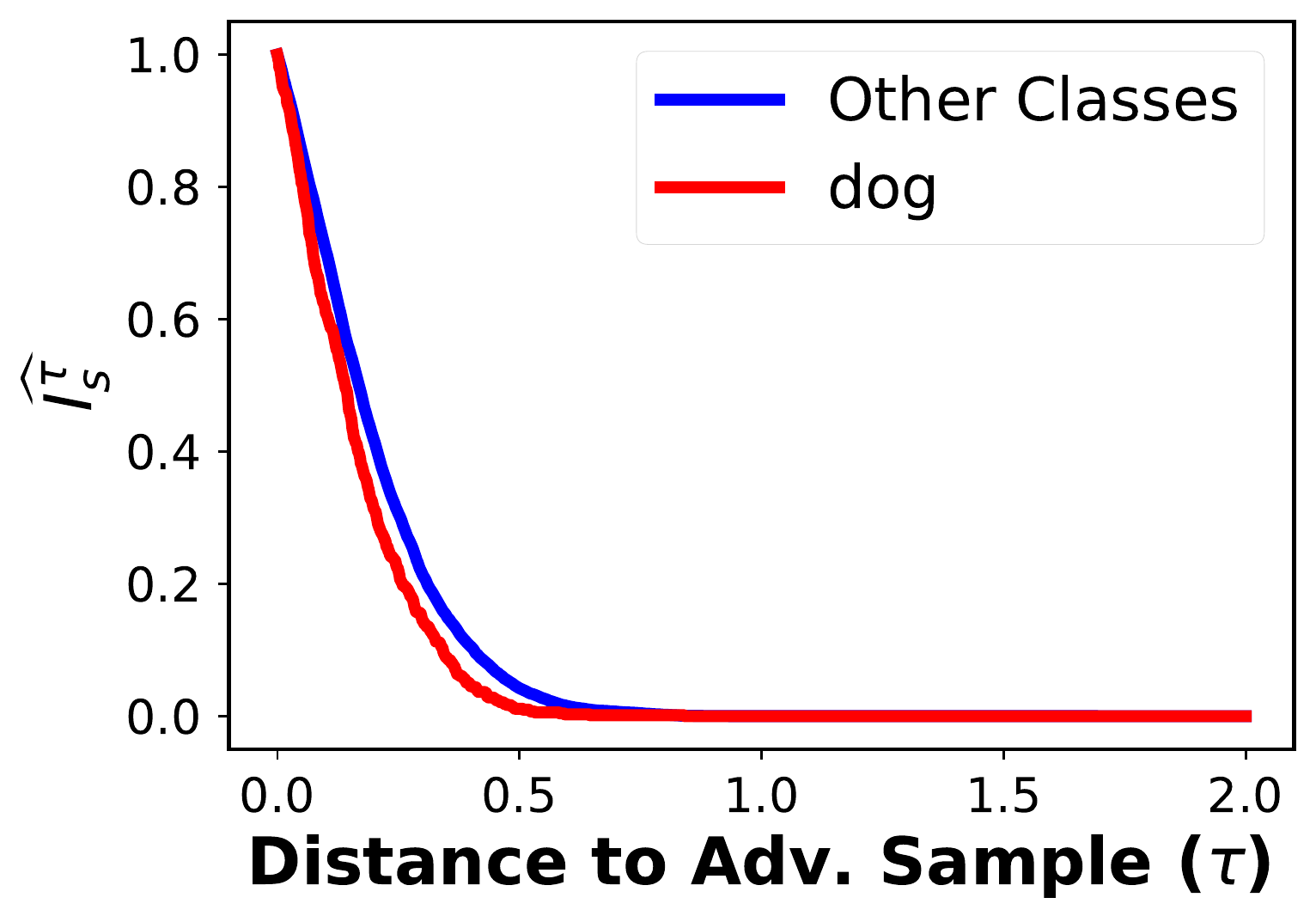}
        \caption{Dog}
        \label{fig:cifar10_dog_resnet}
    \end{subfigure}
    \begin{subfigure}[b]{0.22\textwidth}
        \includegraphics[trim={0cm 0cm 0cm 0cm},clip,width=1\textwidth]{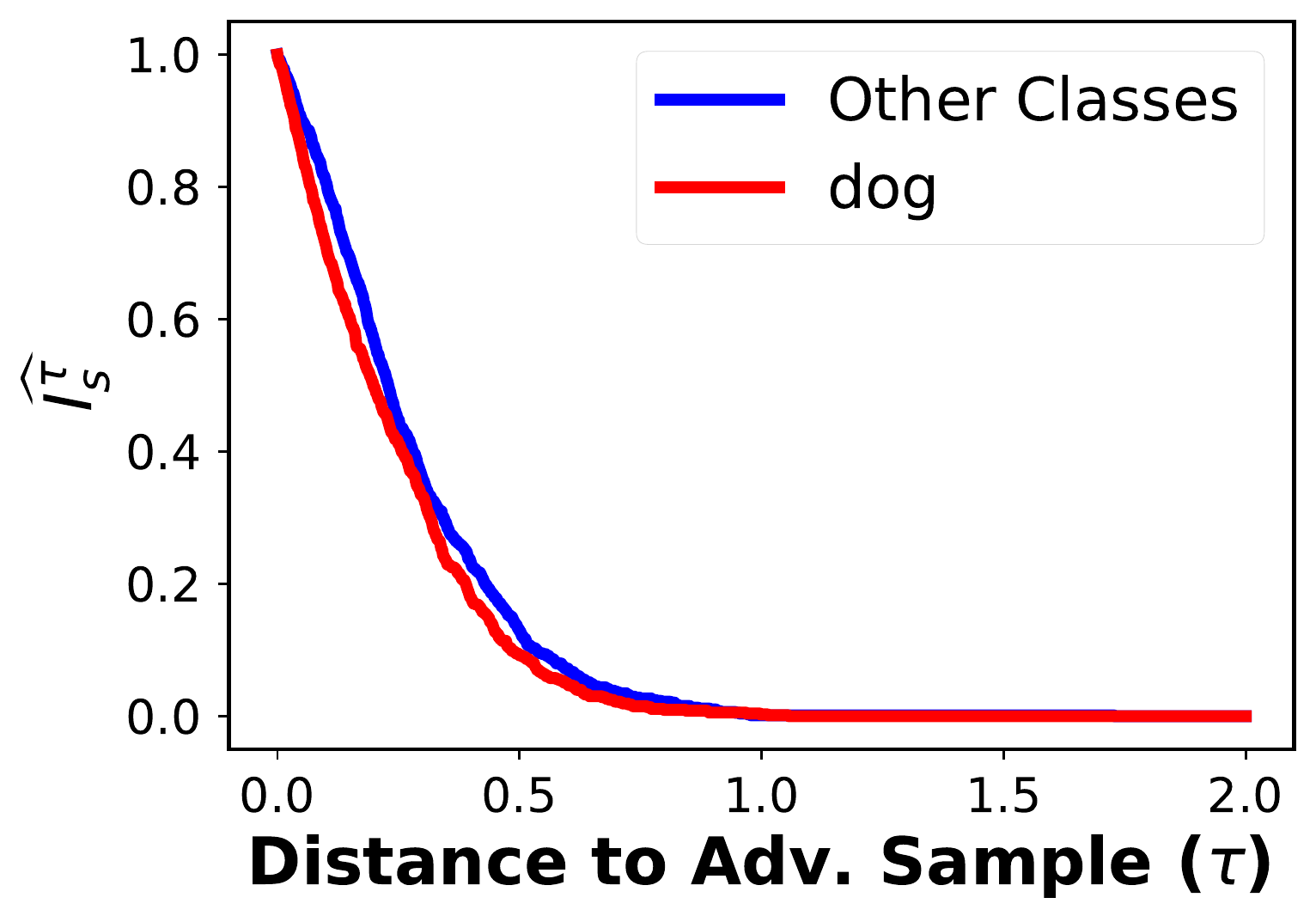}
        \caption{Reg. Dog}
        \label{fig:cifar10_dog_reg_resnet}
    \end{subfigure}
    
    \begin{subfigure}[b]{0.22\textwidth}
        \includegraphics[trim={0cm 0cm 0cm 0cm},clip,width=1\textwidth]{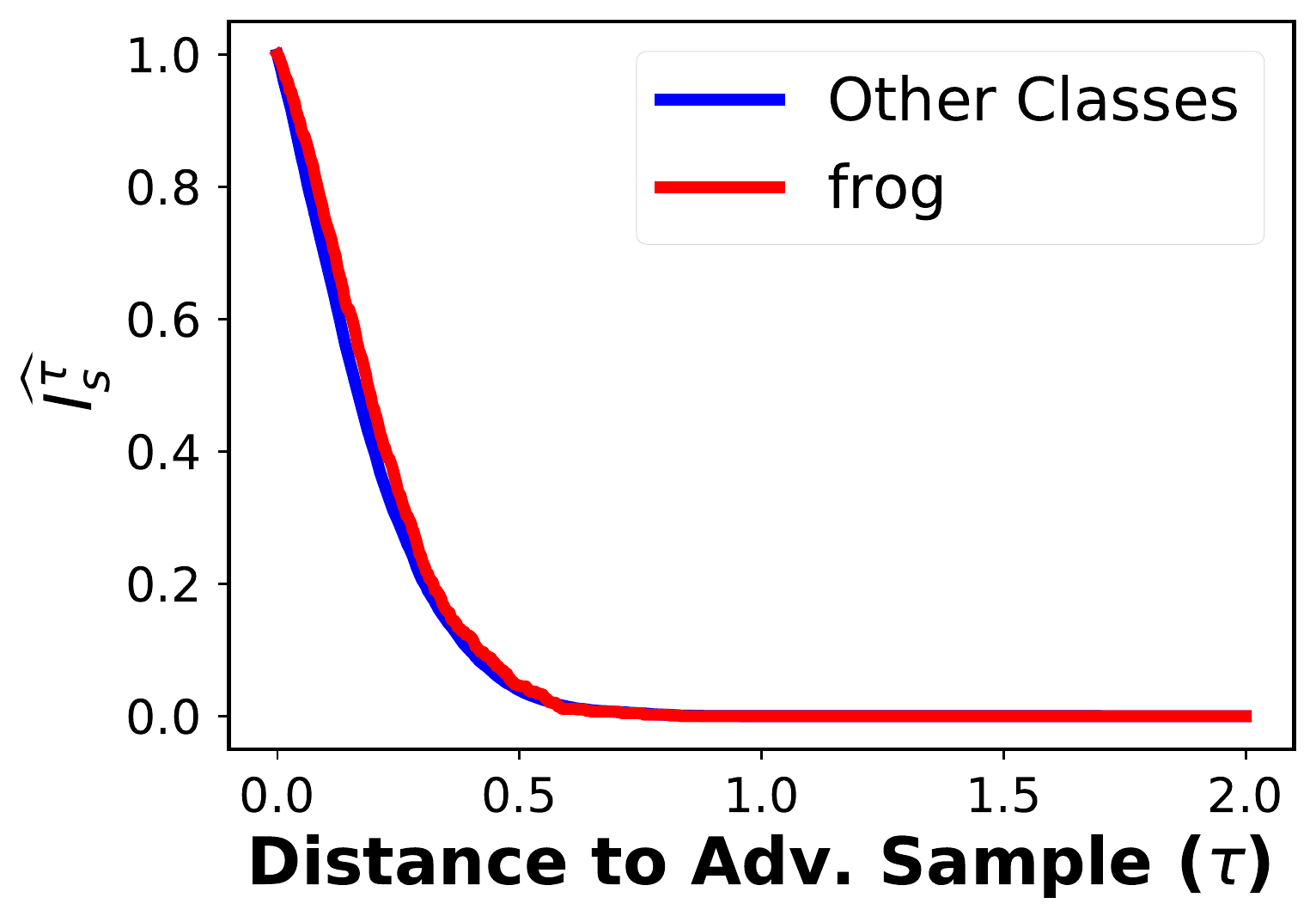}
        \caption{Frog}
        \label{fig:cifar10_frog_resnet}
    \end{subfigure}
    \begin{subfigure}[b]{0.22\textwidth}
        \includegraphics[trim={0cm 0cm 0cm 0cm},clip,width=1\textwidth]{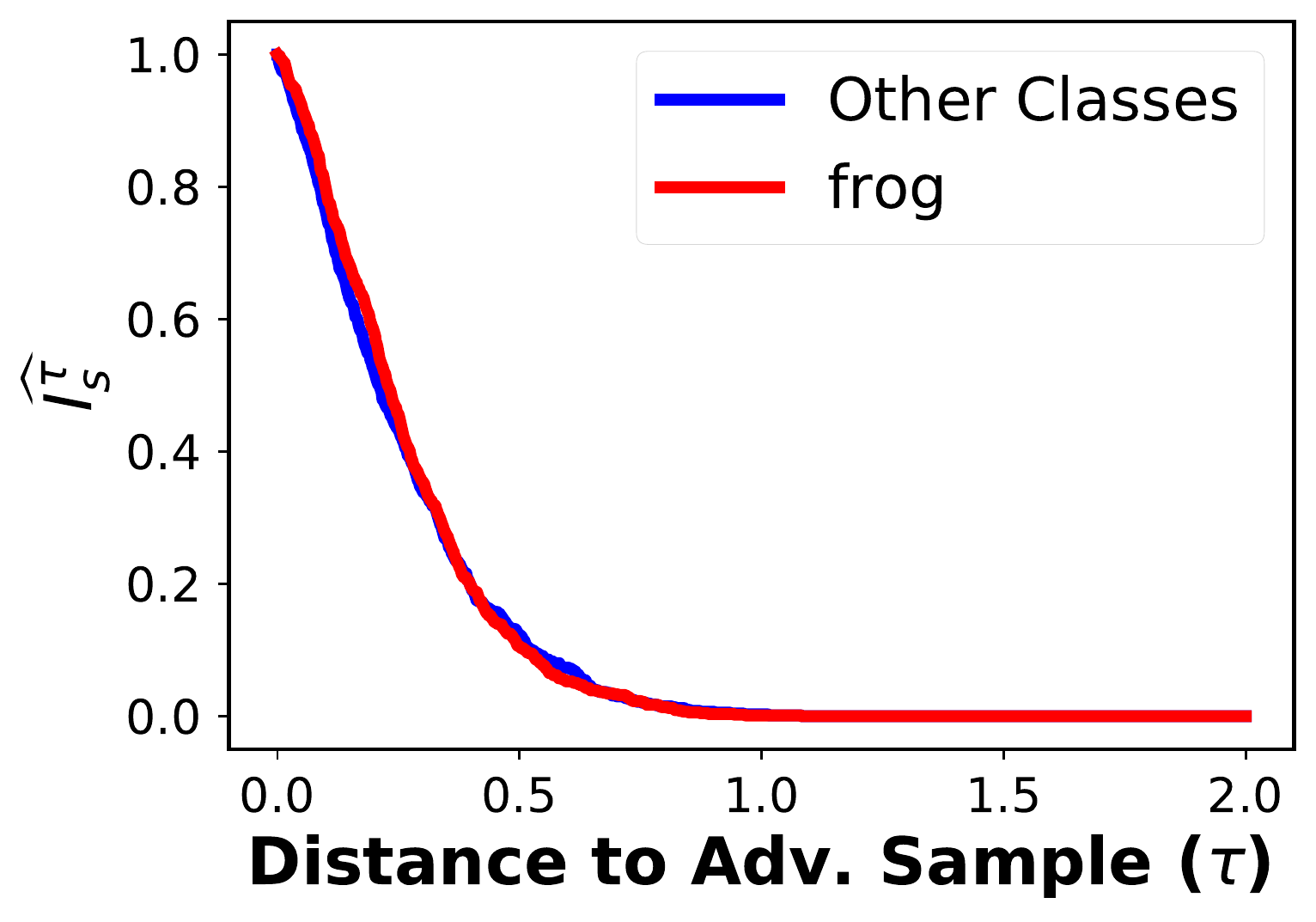}
        \caption{Reg. Frog}
        \label{fig:cifar10_frog_reg_resnet}
    \end{subfigure}
    \begin{subfigure}[b]{0.22\textwidth}
        \includegraphics[trim={0cm 0cm 0cm 0cm},clip,width=1\textwidth]{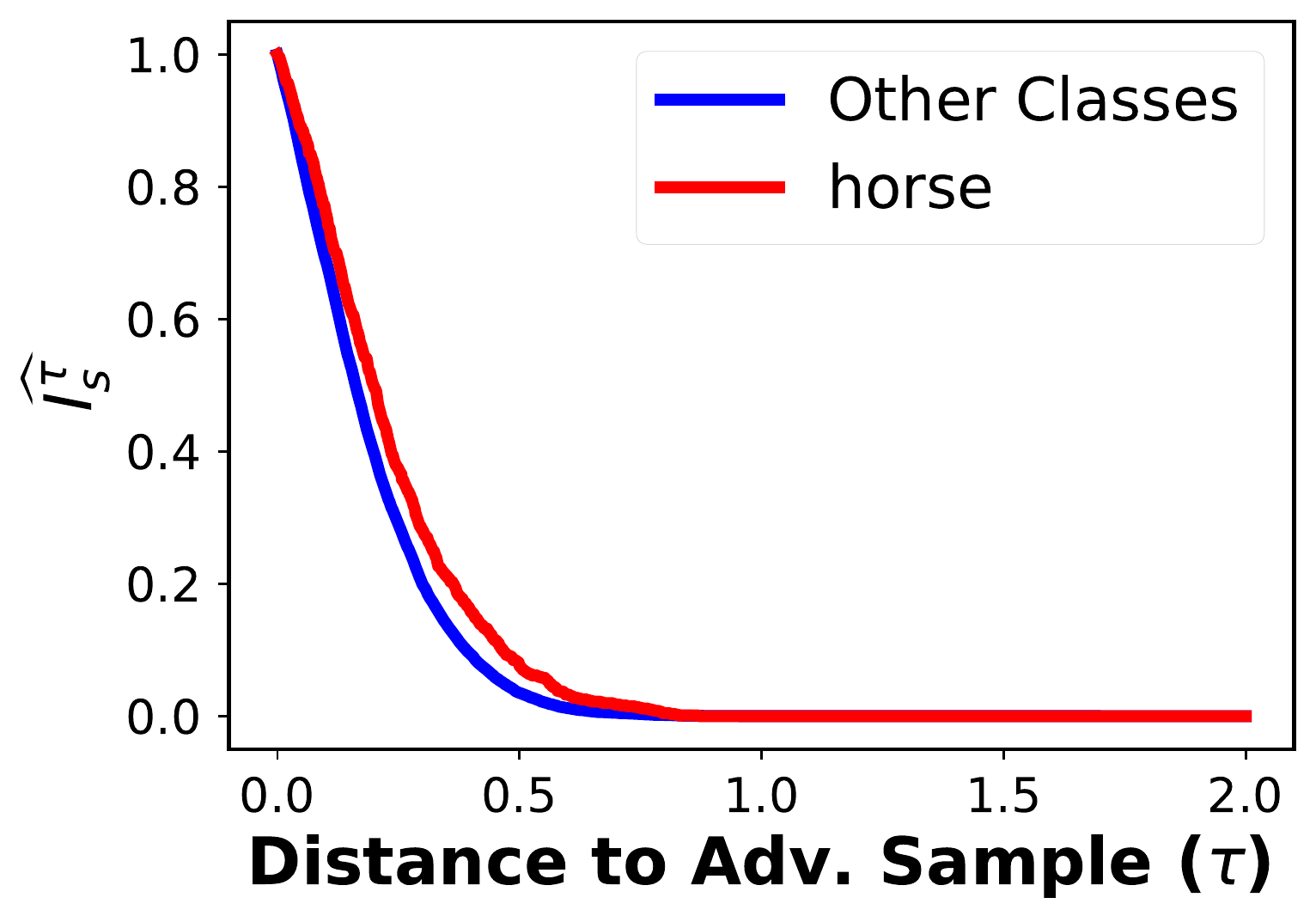}
        \caption{Horse}
        \label{fig:cifar10_horse_resnet}
    \end{subfigure}
    \begin{subfigure}[b]{0.22\textwidth}
        \includegraphics[trim={0cm 0cm 0cm 0cm},clip,width=1\textwidth]{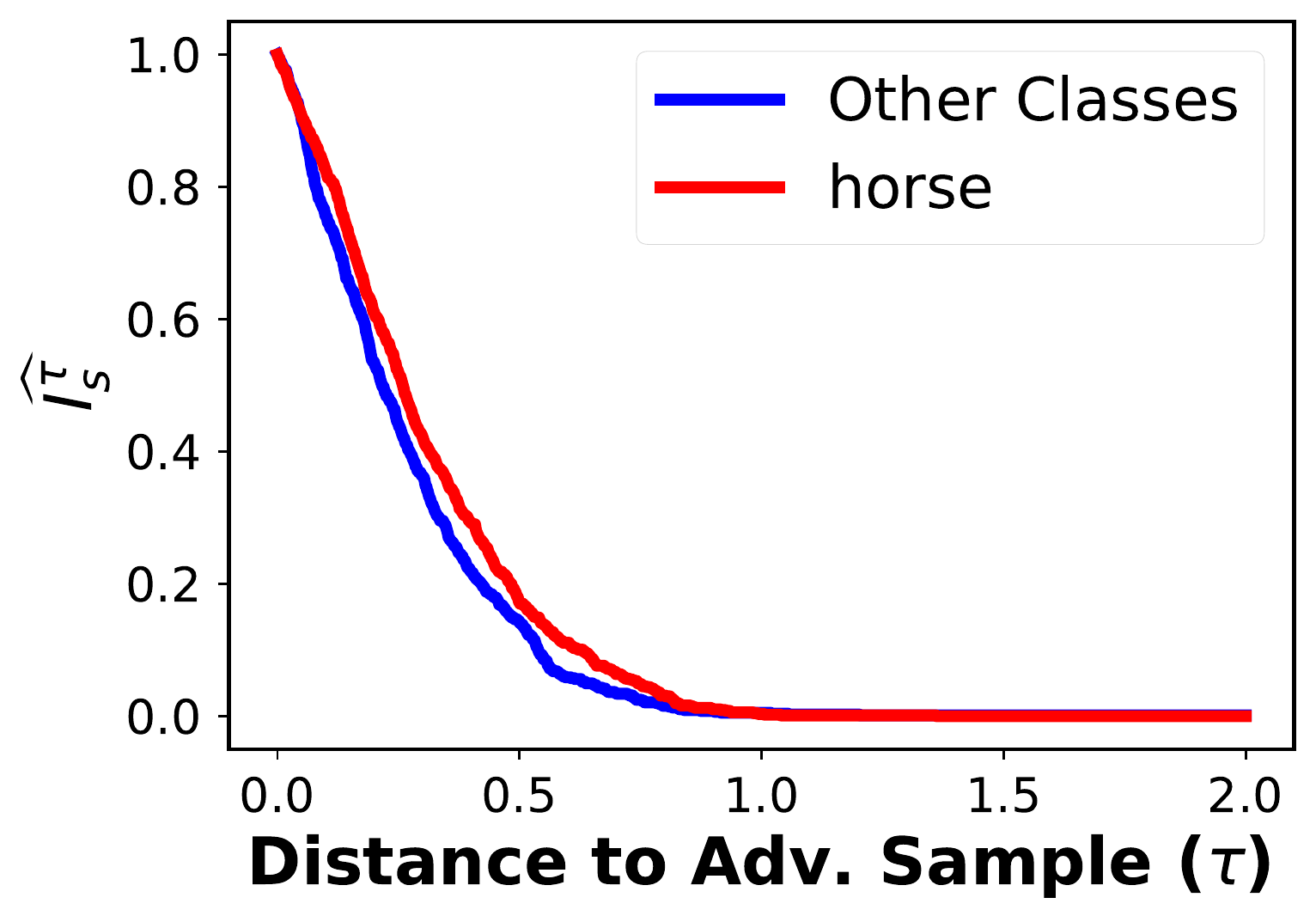}
        \caption{Reg. Horse}
        \label{fig:cifar10_horse_reg_resnet}
    \end{subfigure}
    
    \begin{subfigure}[b]{0.22\textwidth}
        \includegraphics[trim={0cm 0cm 0cm 0cm},clip,width=1\textwidth]{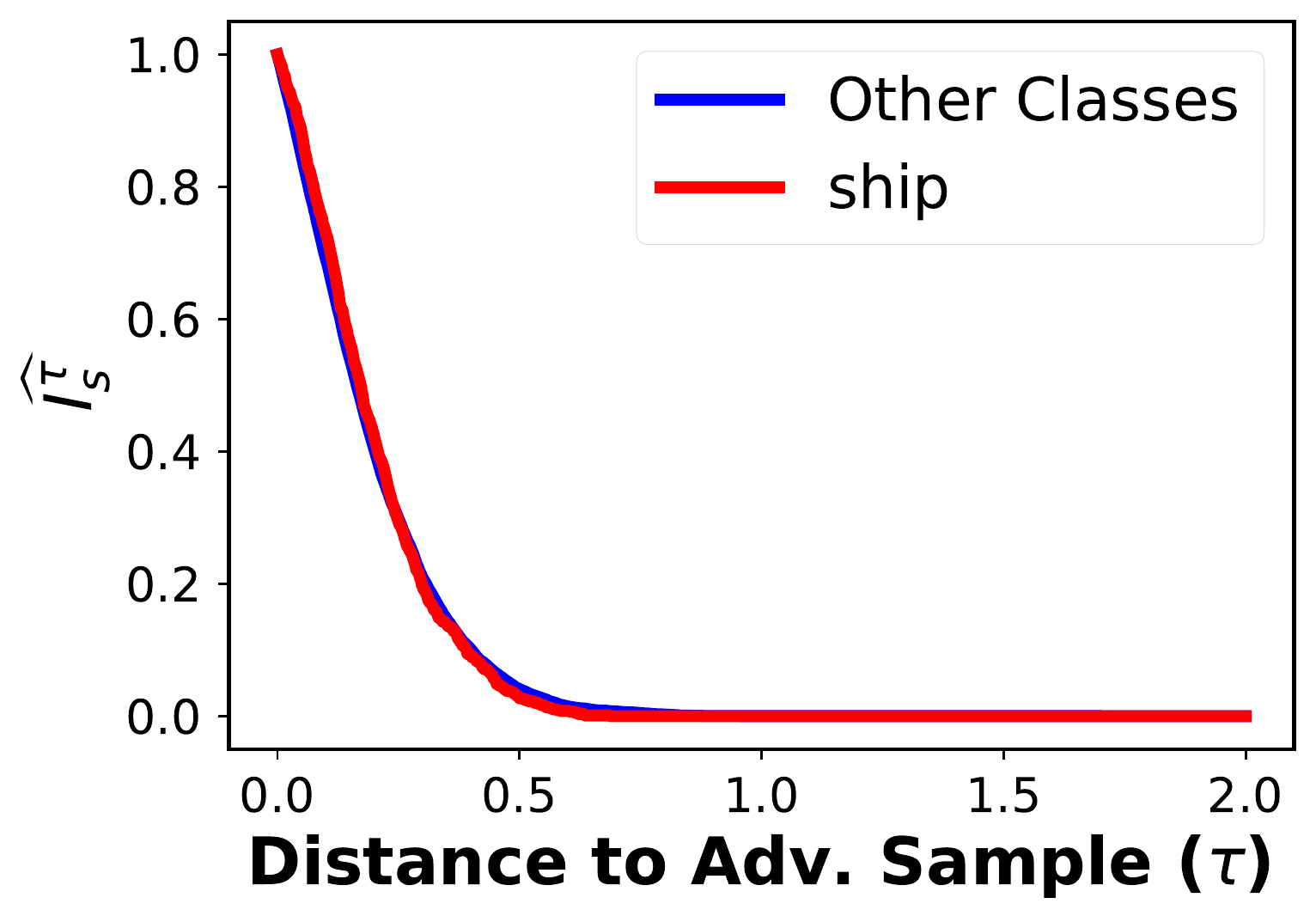}
        \caption{Ship}
        \label{fig:cifar10_ship_resnet}
    \end{subfigure}
    \begin{subfigure}[b]{0.22\textwidth}
        \includegraphics[trim={0cm 0cm 0cm 0cm},clip,width=1\textwidth]{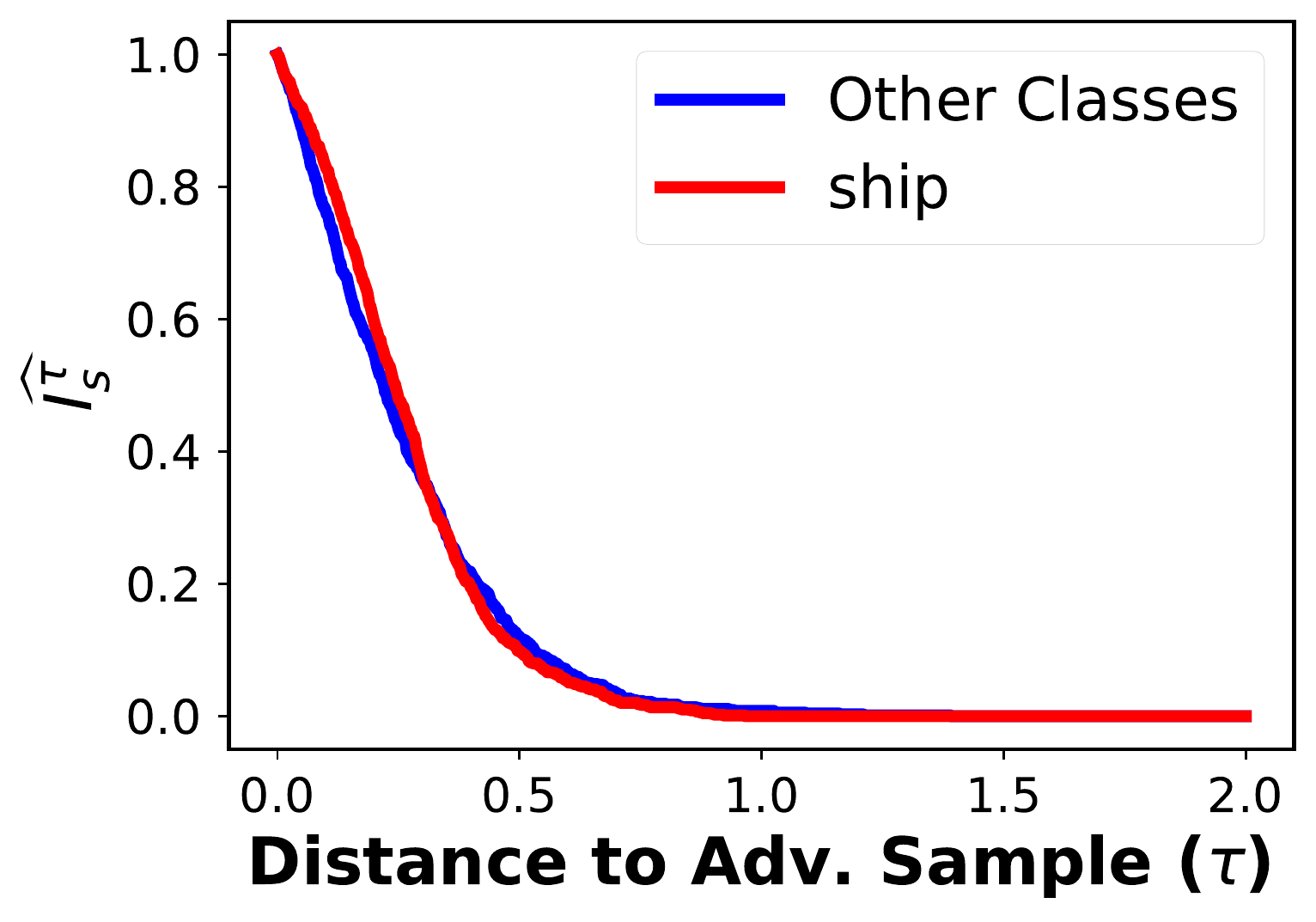}
        \caption{Reg. Ship}
        \label{fig:cifar10_ship_reg_resnet}
    \end{subfigure}
    \begin{subfigure}[b]{0.22\textwidth}
        \includegraphics[trim={0cm 0cm 0cm 0cm},clip,width=1\textwidth]{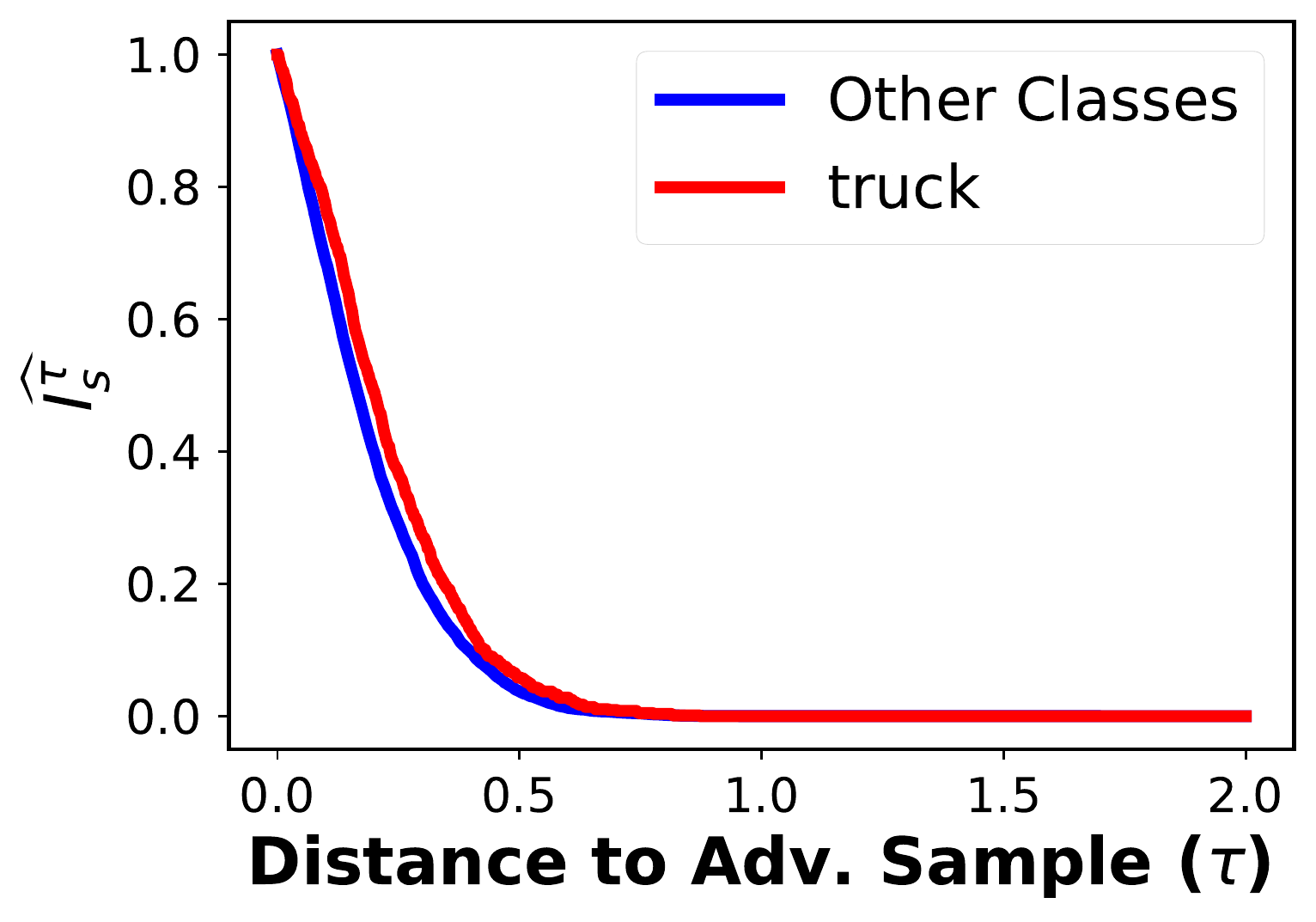}
        \caption{Truck}
        \label{fig:cifar10_truck_resnet}
    \end{subfigure}
    \begin{subfigure}[b]{0.22\textwidth}
        \includegraphics[trim={0cm 0cm 0cm 0cm},clip,width=1\textwidth]{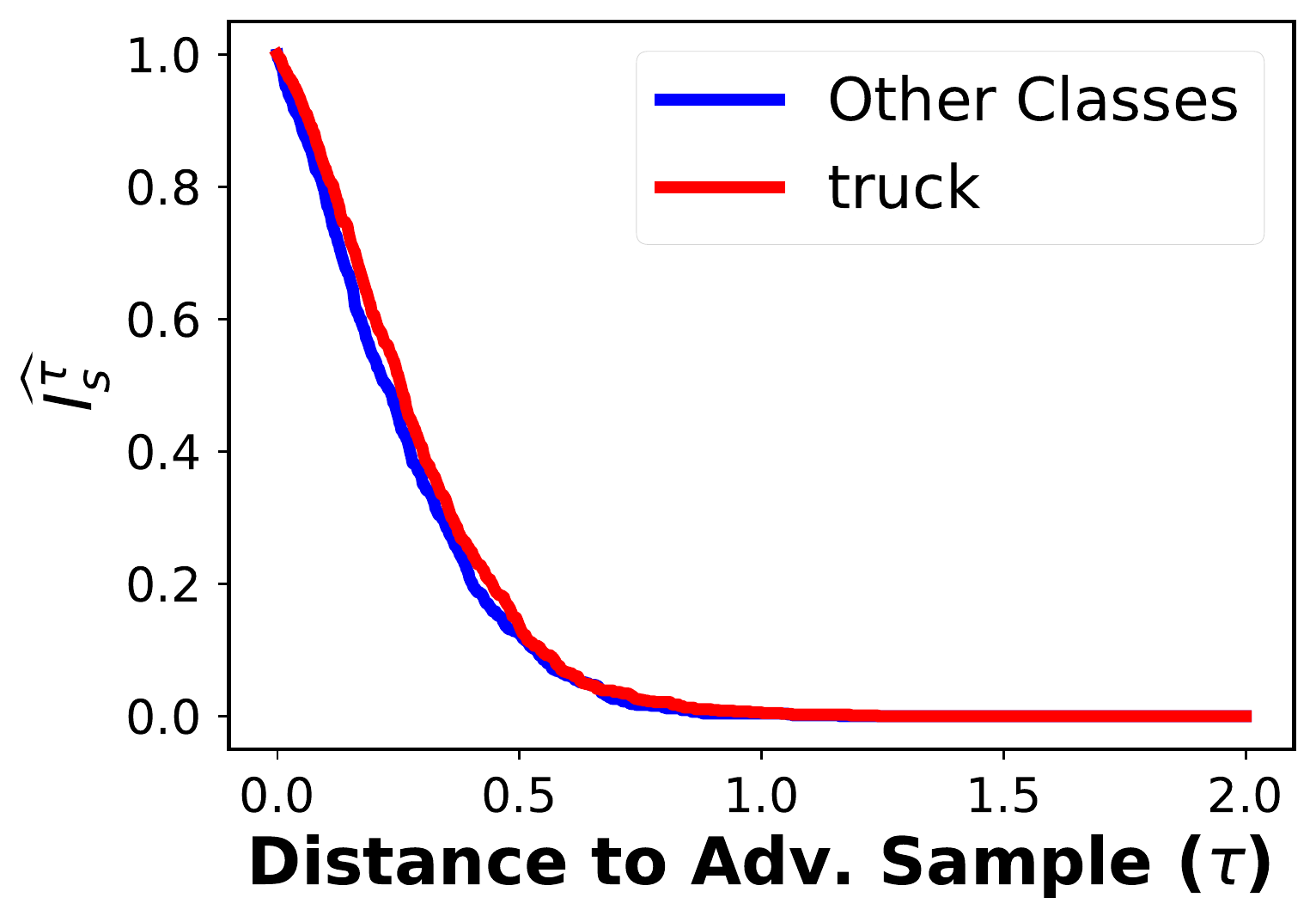}
        \caption{Reg. Truck}
        \label{fig:cifar10_truck_reg_resnet}
    \end{subfigure}

\caption{ [Regularization] CIFAR10 - Resnet50
} 
\label{fig:reg_cifar10_resnet}
\end{figure*}

\begin{figure*}[h]
    \centering
    \begin{subfigure}[b]{0.22\textwidth}
        \includegraphics[trim={0cm 0cm 0cm 0cm},clip,width=1\textwidth]{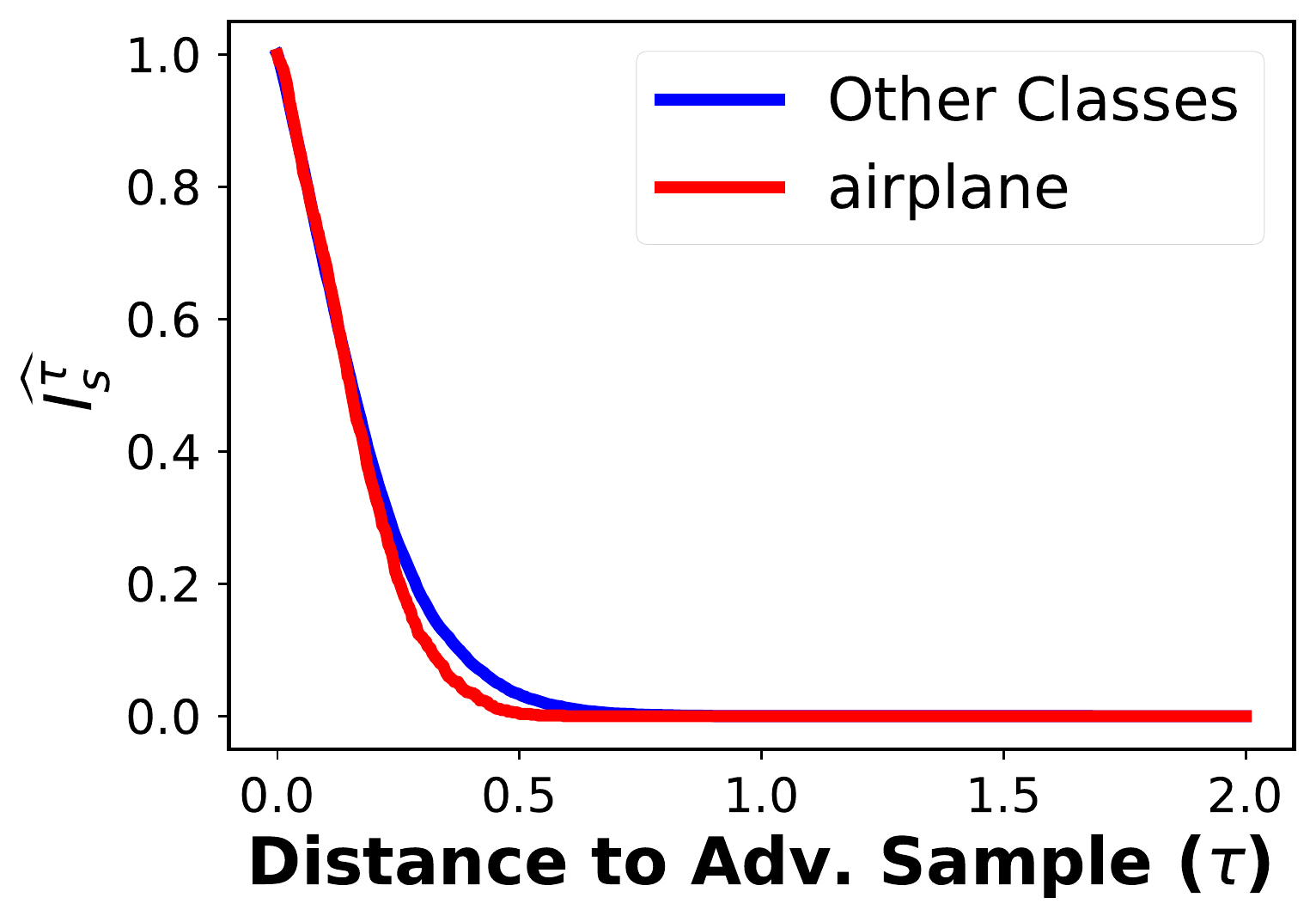}
        \caption{Airplane}
        \label{fig:cifar10_airplane_vgg}
    \end{subfigure}
    \begin{subfigure}[b]{0.22\textwidth}
        \includegraphics[trim={0cm 0cm 0cm 0cm},clip,width=1\textwidth]{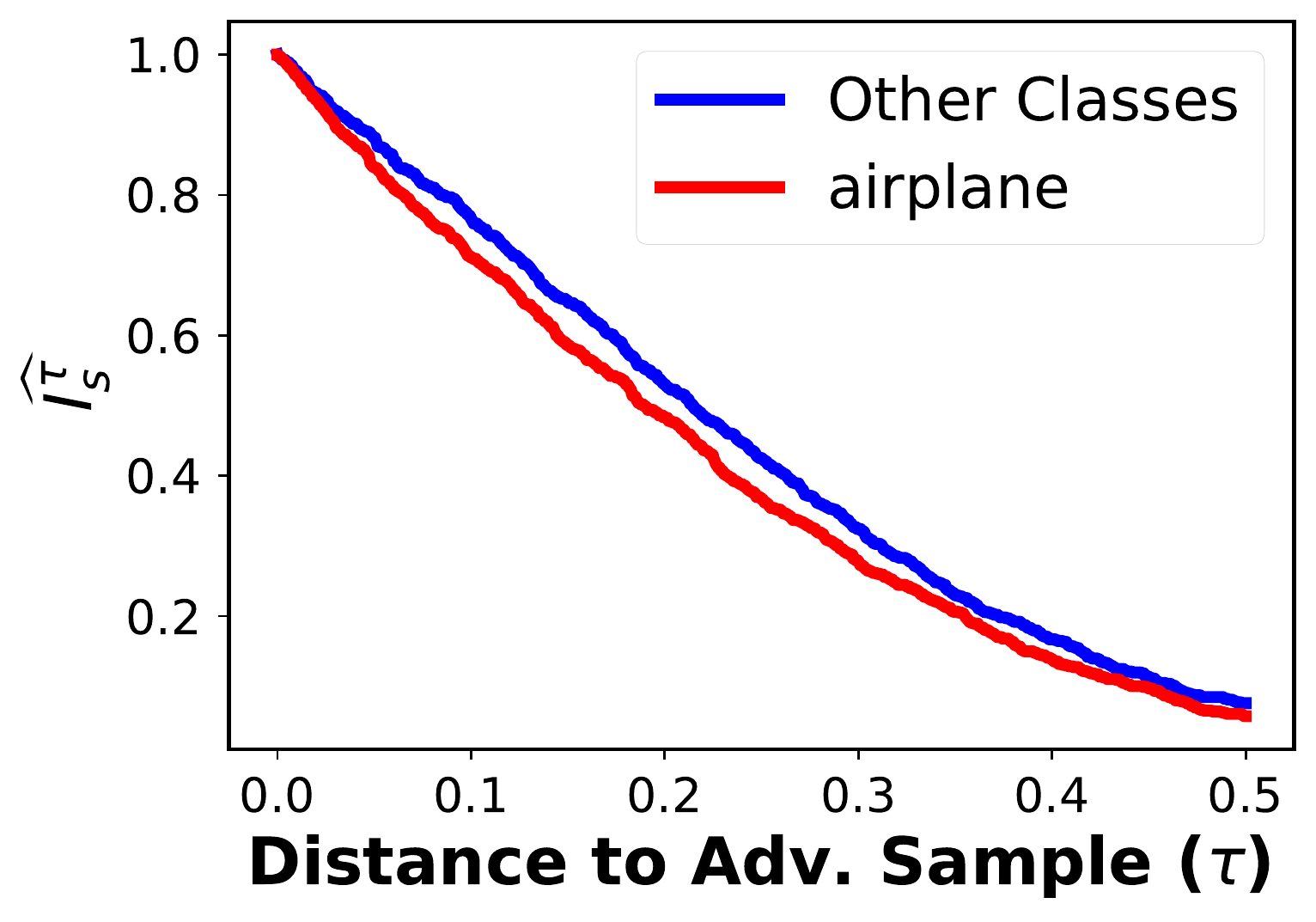}
        \caption{Reg. Airplane}
        \label{fig:cifar10_airplane_reg_vgg}
    \end{subfigure}
    \begin{subfigure}[b]{0.22\textwidth}
        \includegraphics[trim={0cm 0cm 0cm 0cm},clip,width=1\textwidth]{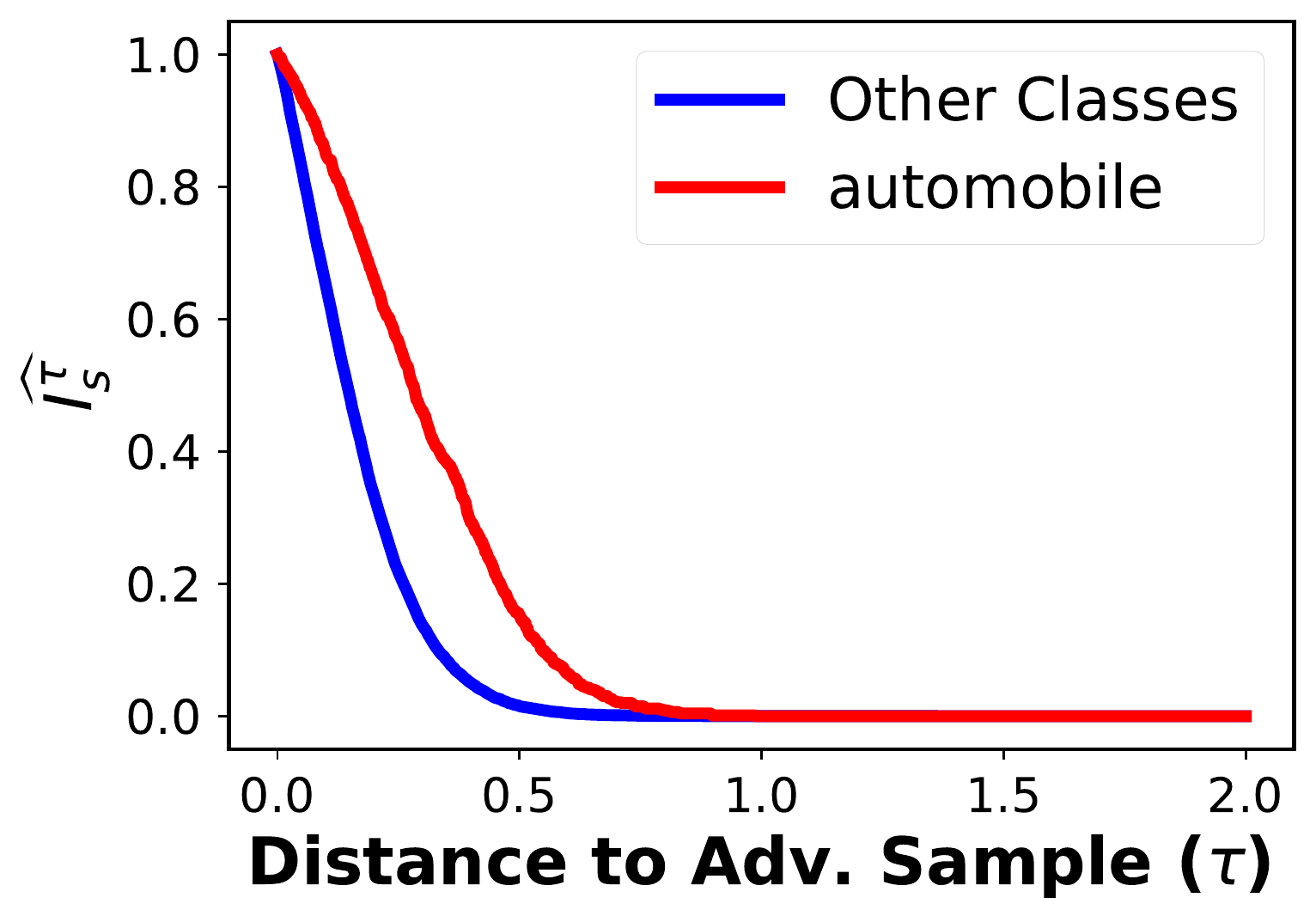}
        \caption{Automobile}
        \label{fig:cifar10_automobile_vgg}
    \end{subfigure}
    \begin{subfigure}[b]{0.22\textwidth}
        \includegraphics[trim={0cm 0cm 0cm 0cm},clip,width=1\textwidth]{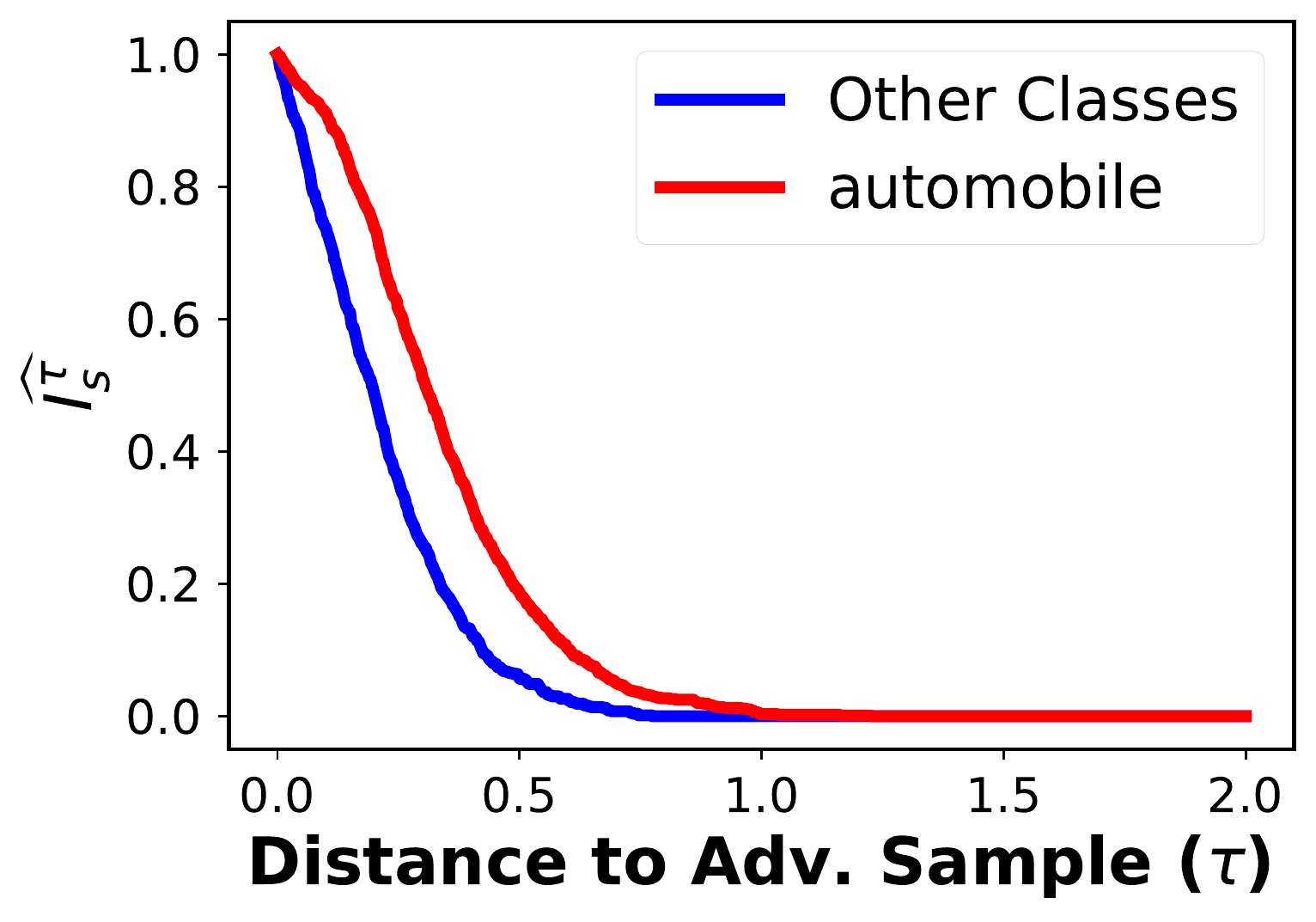}
        \caption{Reg. Automobile}
        \label{fig:cifar10_automobile_reg_vgg}
    \end{subfigure}
    
    \begin{subfigure}[b]{0.22\textwidth}
        \includegraphics[trim={0cm 0cm 0cm 0cm},clip,width=1\textwidth]{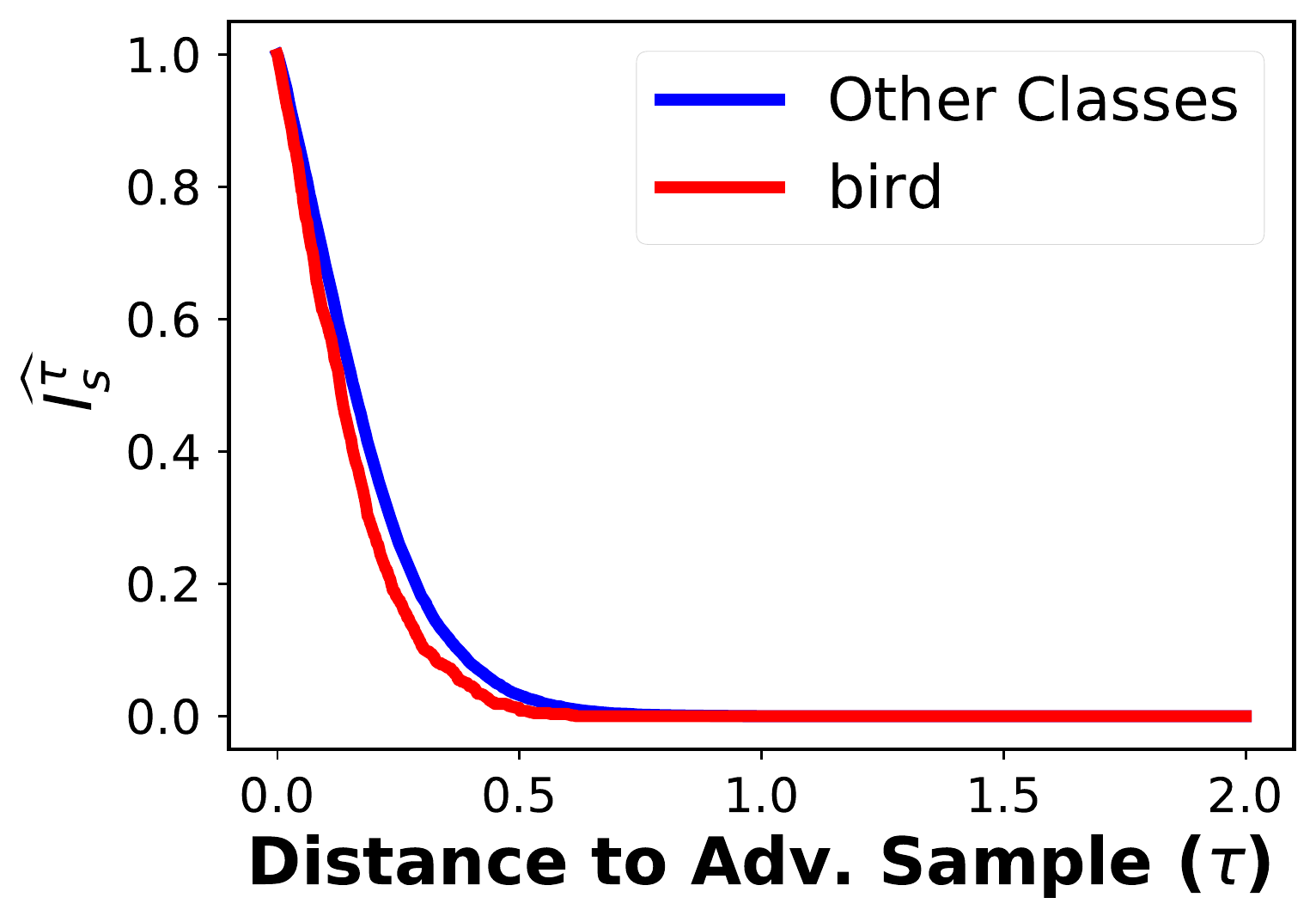}
        \caption{Bird}
        \label{fig:cifar10_bird_vgg}
    \end{subfigure}
    \begin{subfigure}[b]{0.22\textwidth}
        \includegraphics[trim={0cm 0cm 0cm 0cm},clip,width=1\textwidth]{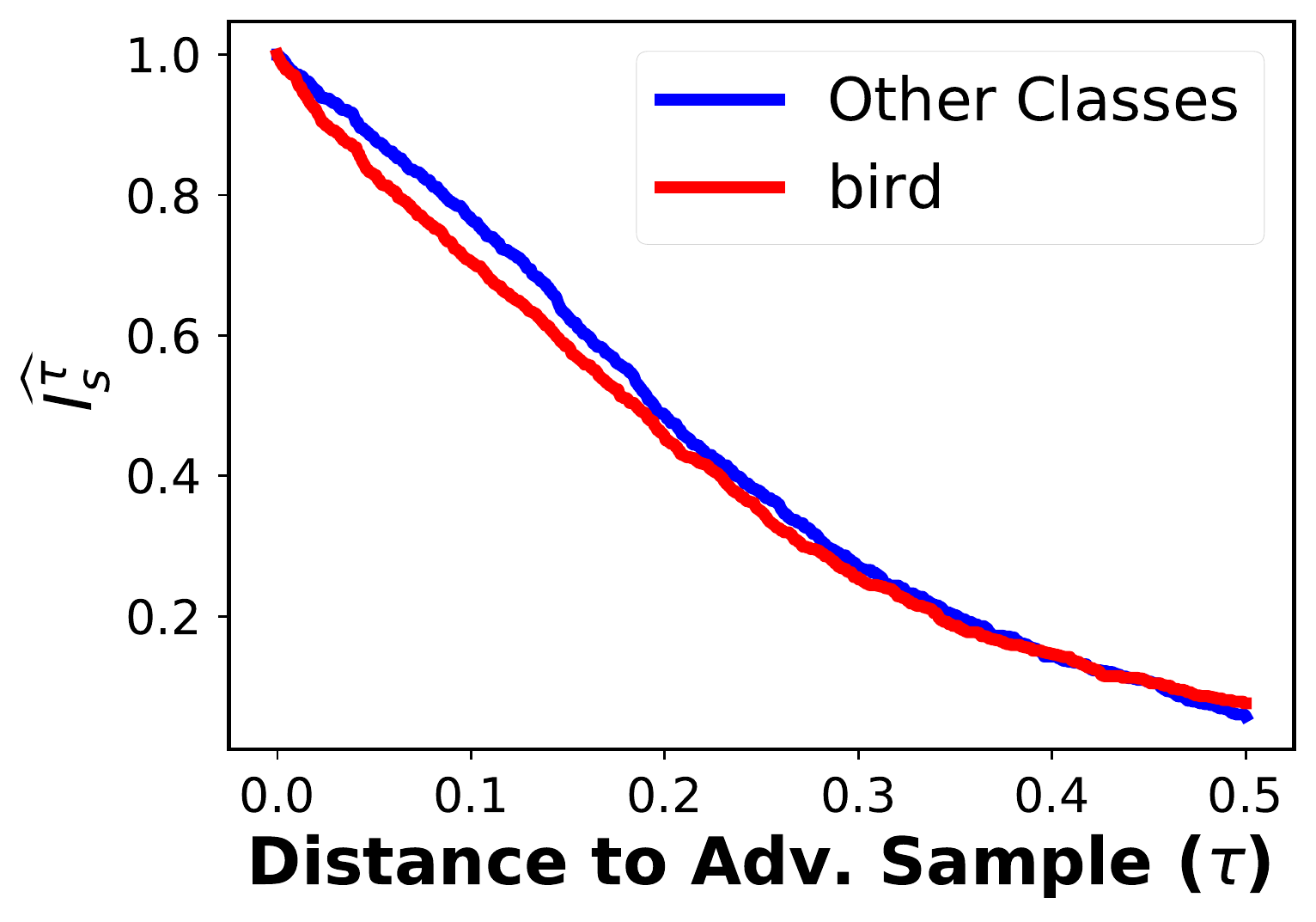}
        \caption{Reg. Bird}
        \label{fig:cifar10_bird_reg_vgg}
    \end{subfigure}
    \begin{subfigure}[b]{0.22\textwidth}
        \includegraphics[trim={0cm 0cm 0cm 0cm},clip,width=1\textwidth]{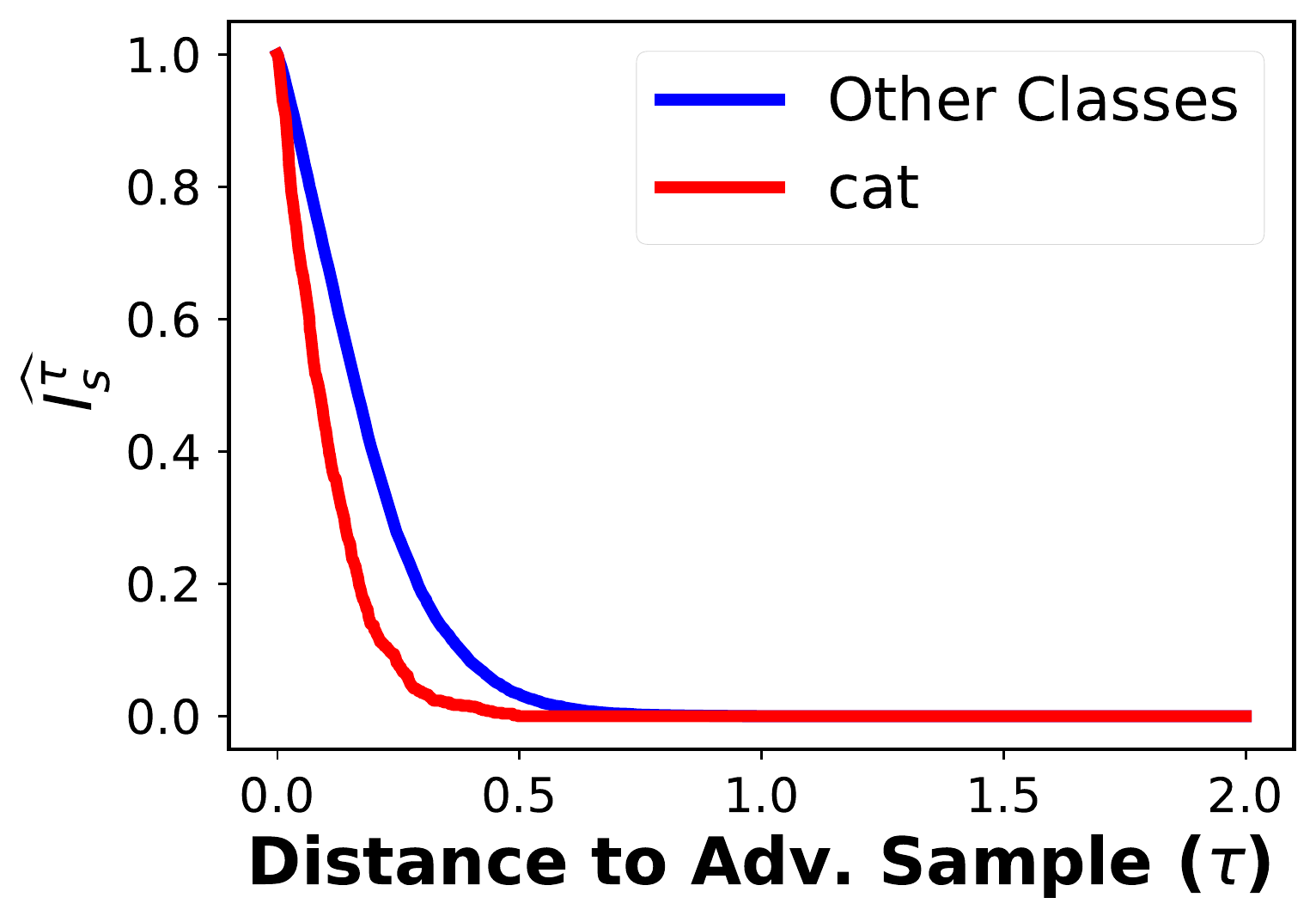}
        \caption{Cat}
        \label{fig:cifar10_cat_vgg}
    \end{subfigure}
    \begin{subfigure}[b]{0.22\textwidth}
        \includegraphics[trim={0cm 0cm 0cm 0cm},clip,width=1\textwidth]{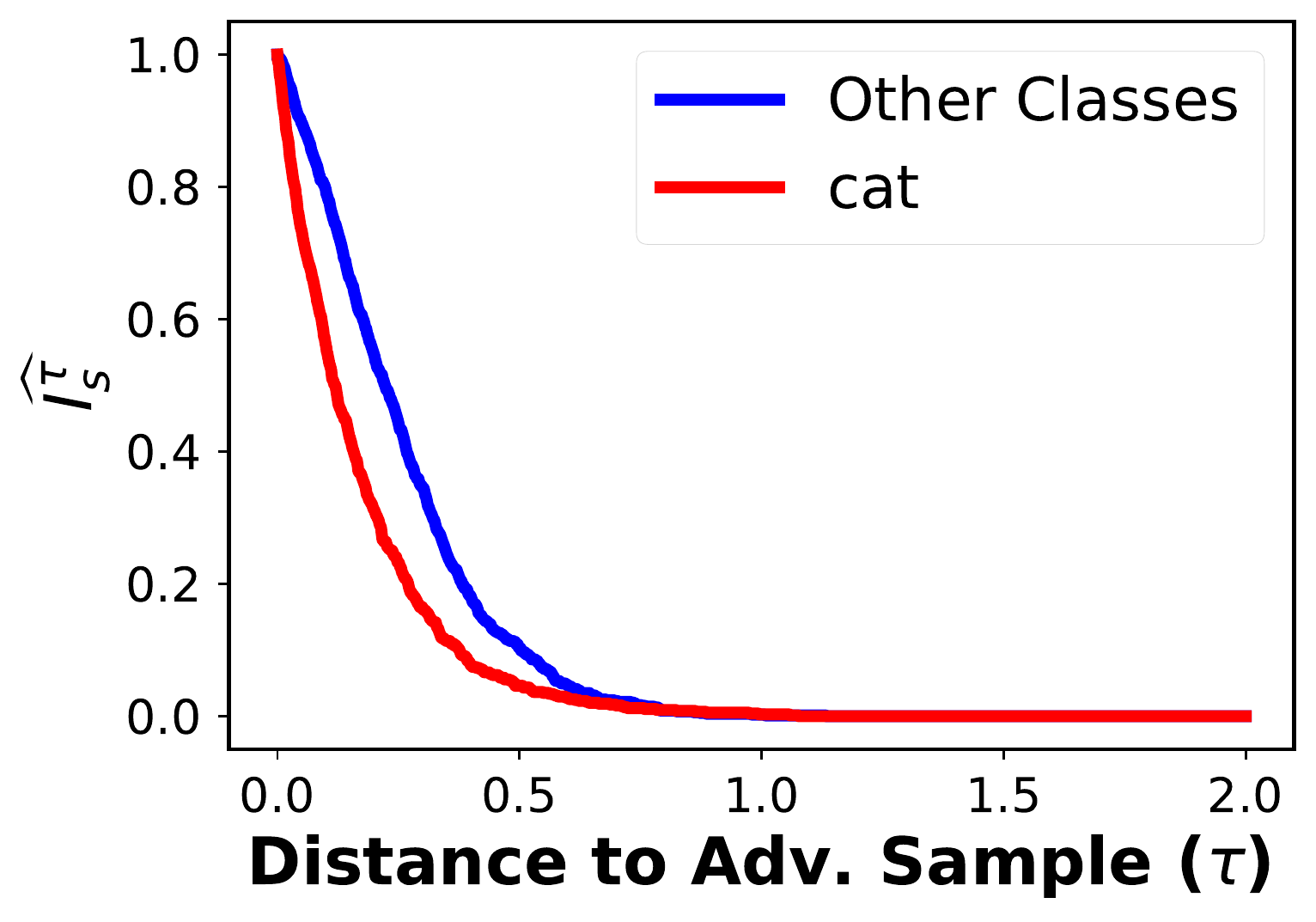}
        \caption{Reg. Cat}
        \label{fig:cifar10_cat_reg_vgg}
    \end{subfigure}
    
    \begin{subfigure}[b]{0.22\textwidth}
        \includegraphics[trim={0cm 0cm 0cm 0cm},clip,width=1\textwidth]{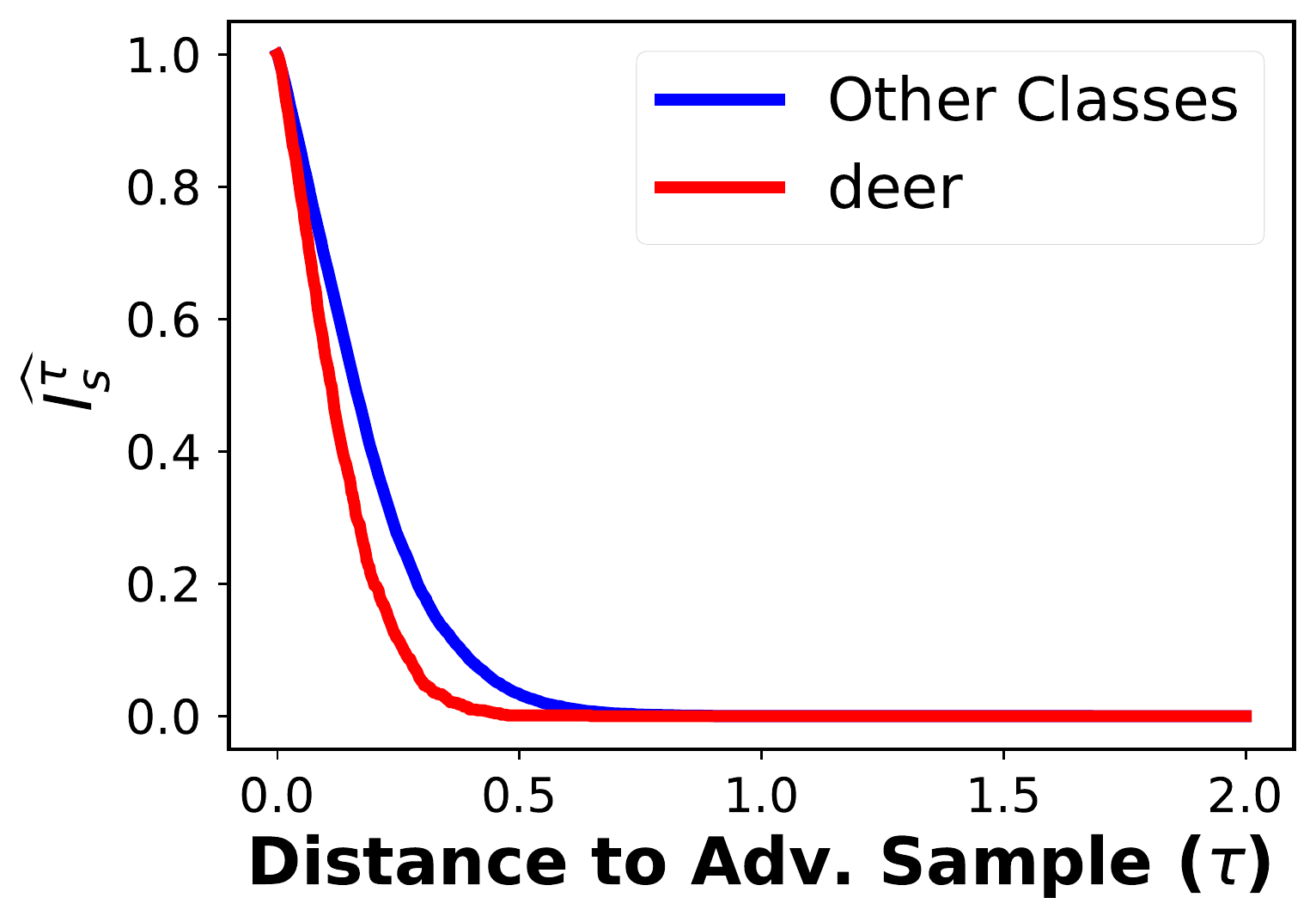}
        \caption{Deer}
        \label{fig:cifar10_deer_vgg}
    \end{subfigure}
    \begin{subfigure}[b]{0.22\textwidth}
        \includegraphics[trim={0cm 0cm 0cm 0cm},clip,width=1\textwidth]{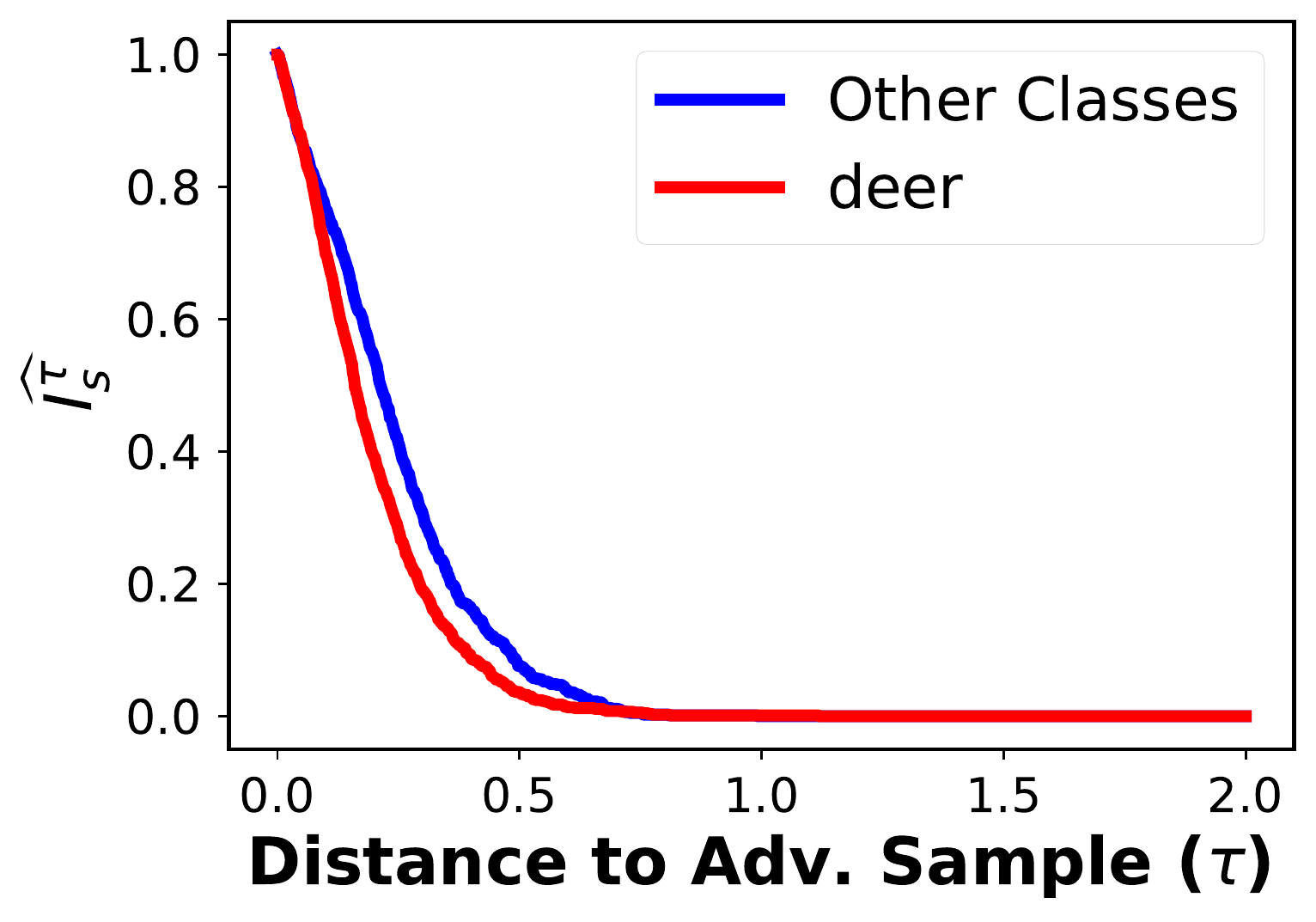}
        \caption{Reg. Deer}
        \label{fig:cifar10_deer_reg_vgg}
    \end{subfigure}
    \begin{subfigure}[b]{0.22\textwidth}
        \includegraphics[trim={0cm 0cm 0cm 0cm},clip,width=1\textwidth]{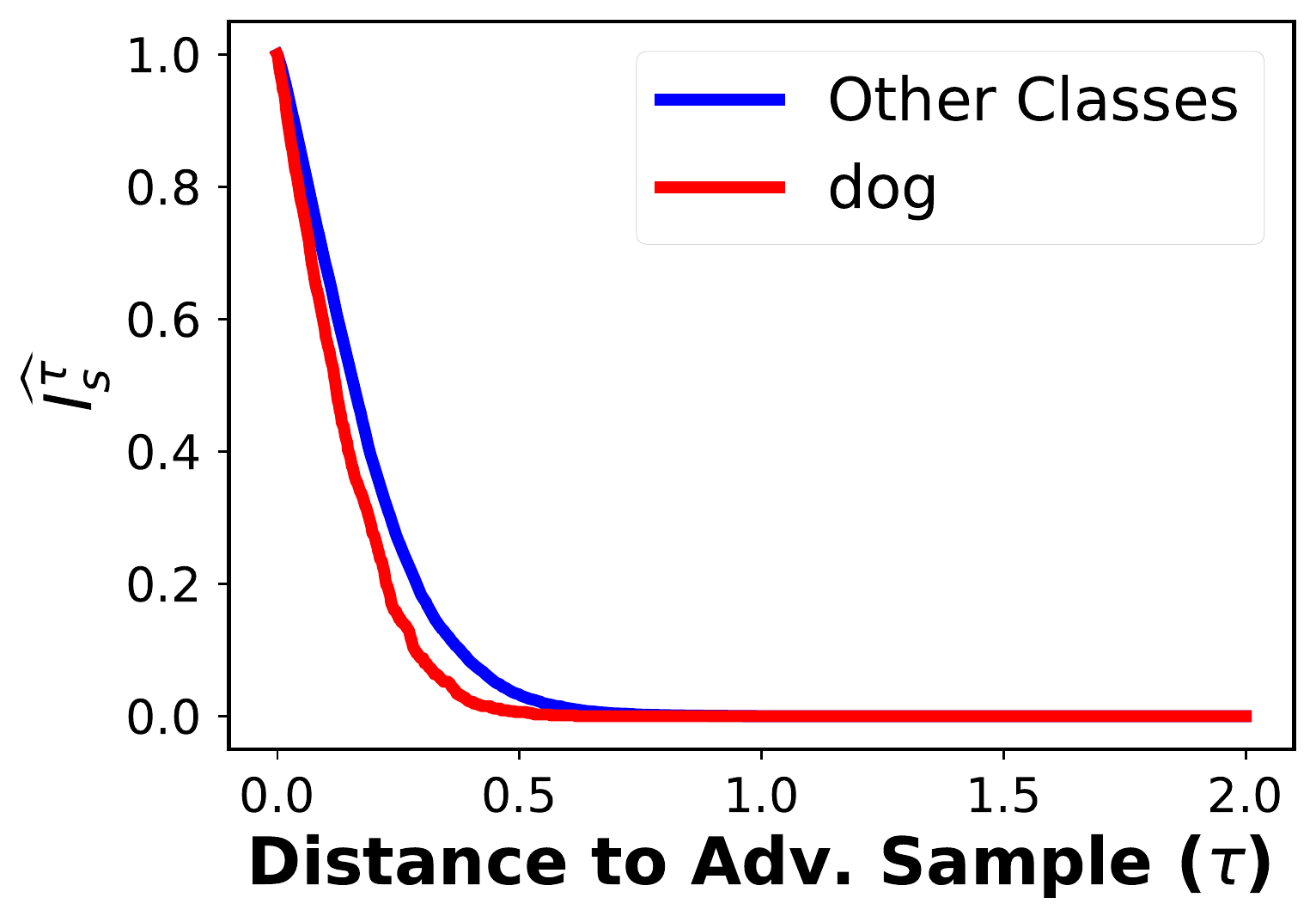}
        \caption{Dog}
        \label{fig:cifar10_dog_vgg}
    \end{subfigure}
    \begin{subfigure}[b]{0.22\textwidth}
        \includegraphics[trim={0cm 0cm 0cm 0cm},clip,width=1\textwidth]{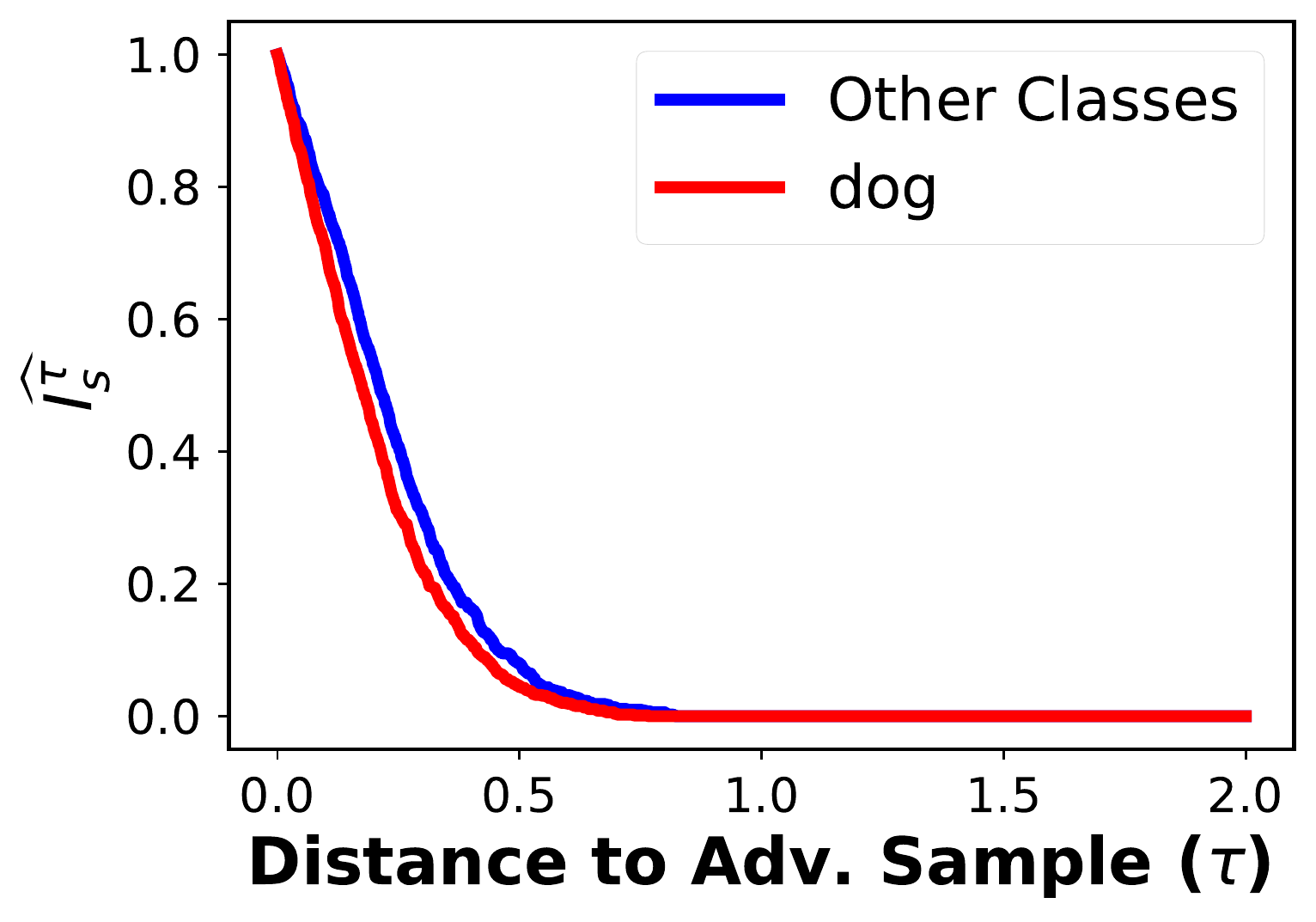}
        \caption{Reg. Dog}
        \label{fig:cifar10_dog_reg_vgg}
    \end{subfigure}
    
    \begin{subfigure}[b]{0.22\textwidth}
        \includegraphics[trim={0cm 0cm 0cm 0cm},clip,width=1\textwidth]{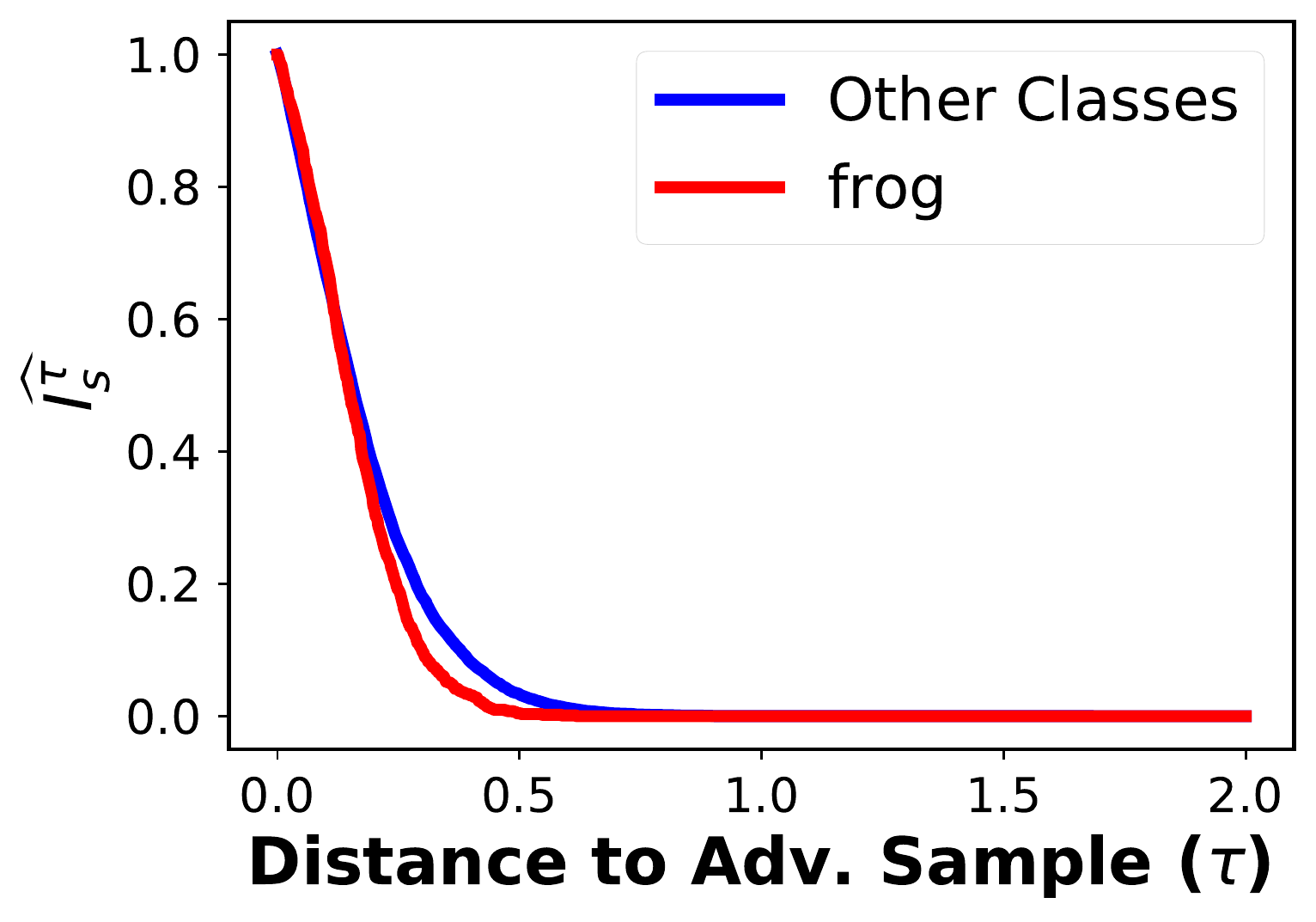}
        \caption{Frog}
        \label{fig:cifar10_frog_vgg}
    \end{subfigure}
    \begin{subfigure}[b]{0.22\textwidth}
        \includegraphics[trim={0cm 0cm 0cm 0cm},clip,width=1\textwidth]{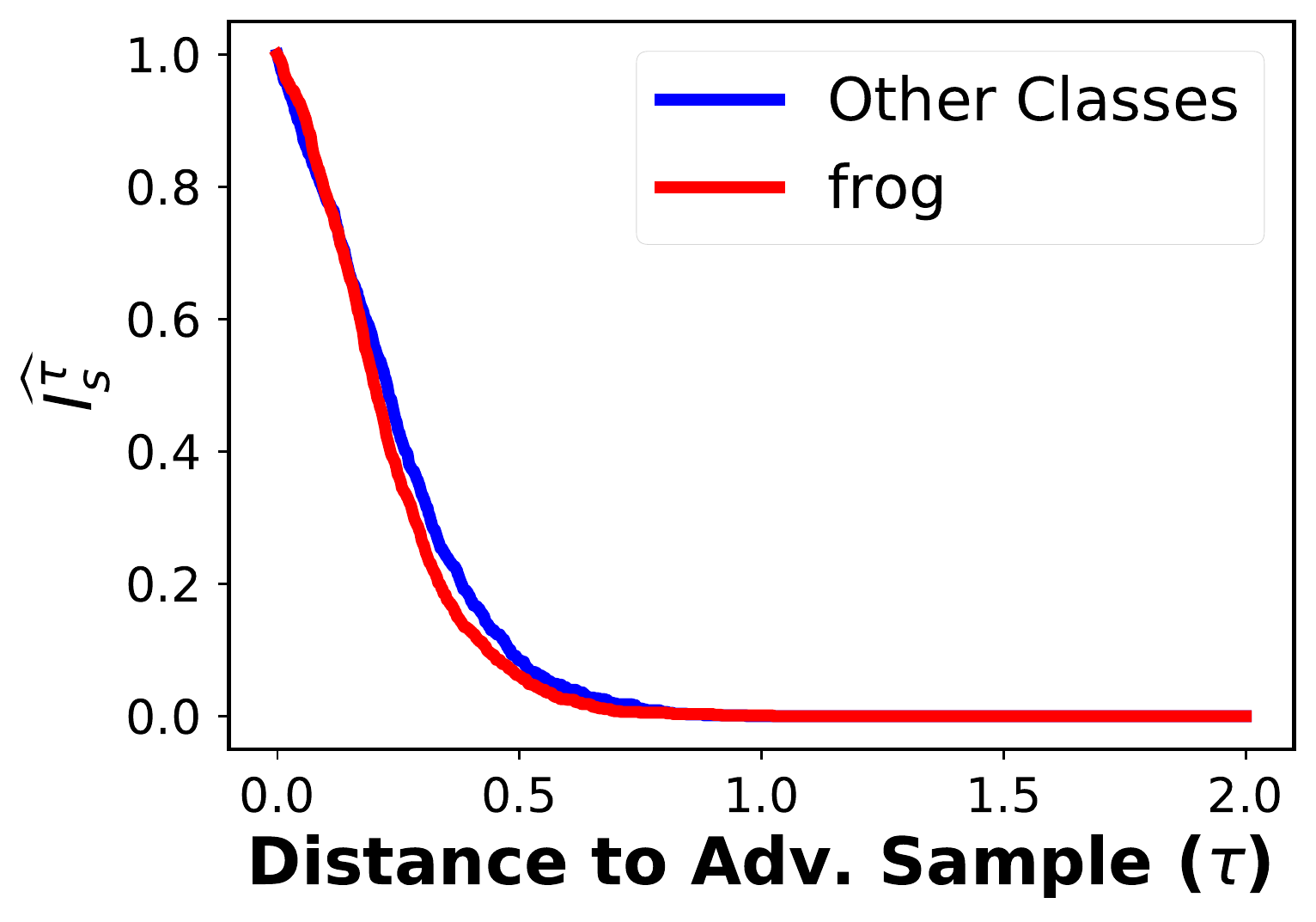}
        \caption{Reg. Frog}
        \label{fig:cifar10_frog_reg_vgg}
    \end{subfigure}
    \begin{subfigure}[b]{0.22\textwidth}
        \includegraphics[trim={0cm 0cm 0cm 0cm},clip,width=1\textwidth]{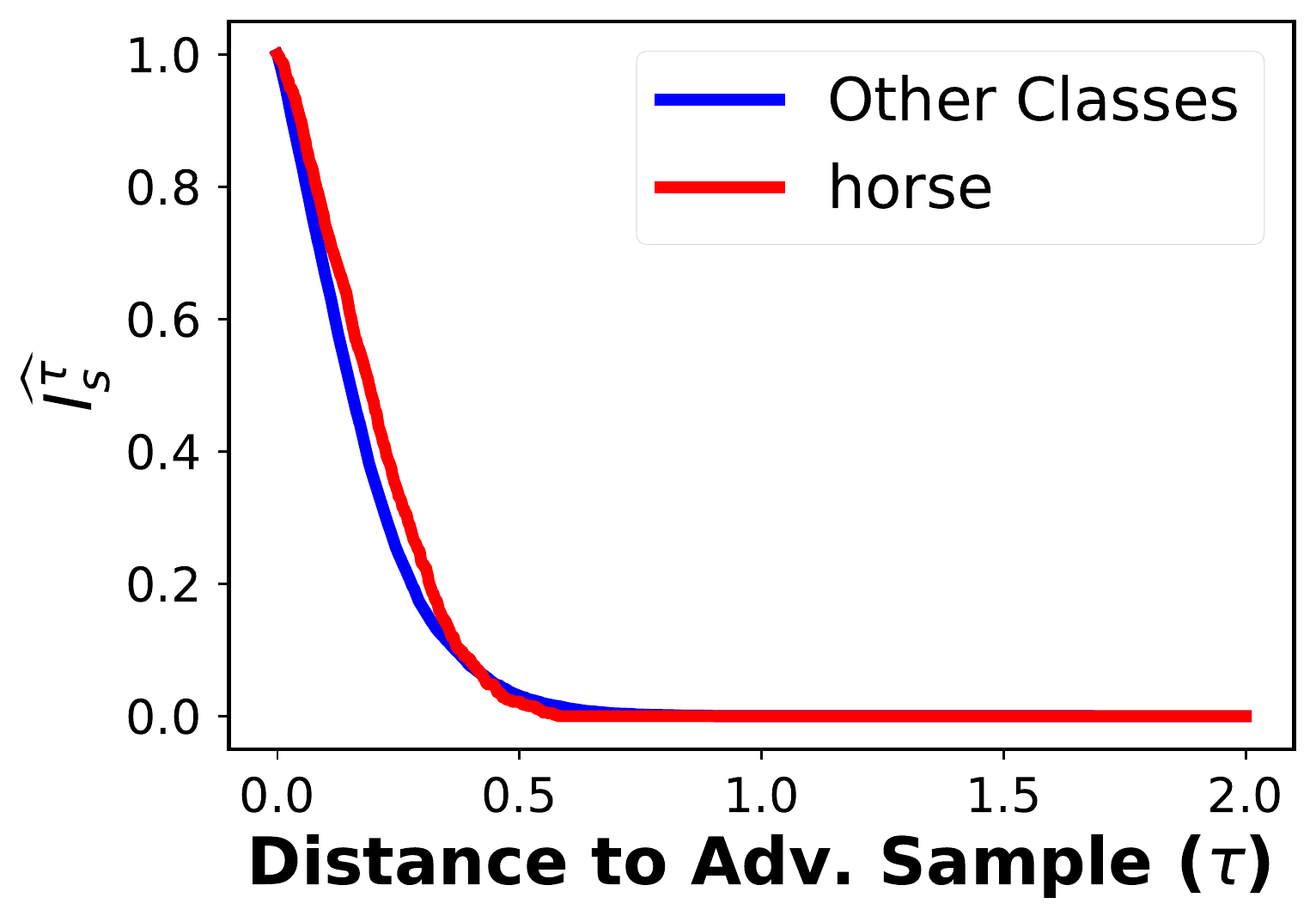}
        \caption{Horse}
        \label{fig:cifar10_horse_vgg}
    \end{subfigure}
    \begin{subfigure}[b]{0.22\textwidth}
        \includegraphics[trim={0cm 0cm 0cm 0cm},clip,width=1\textwidth]{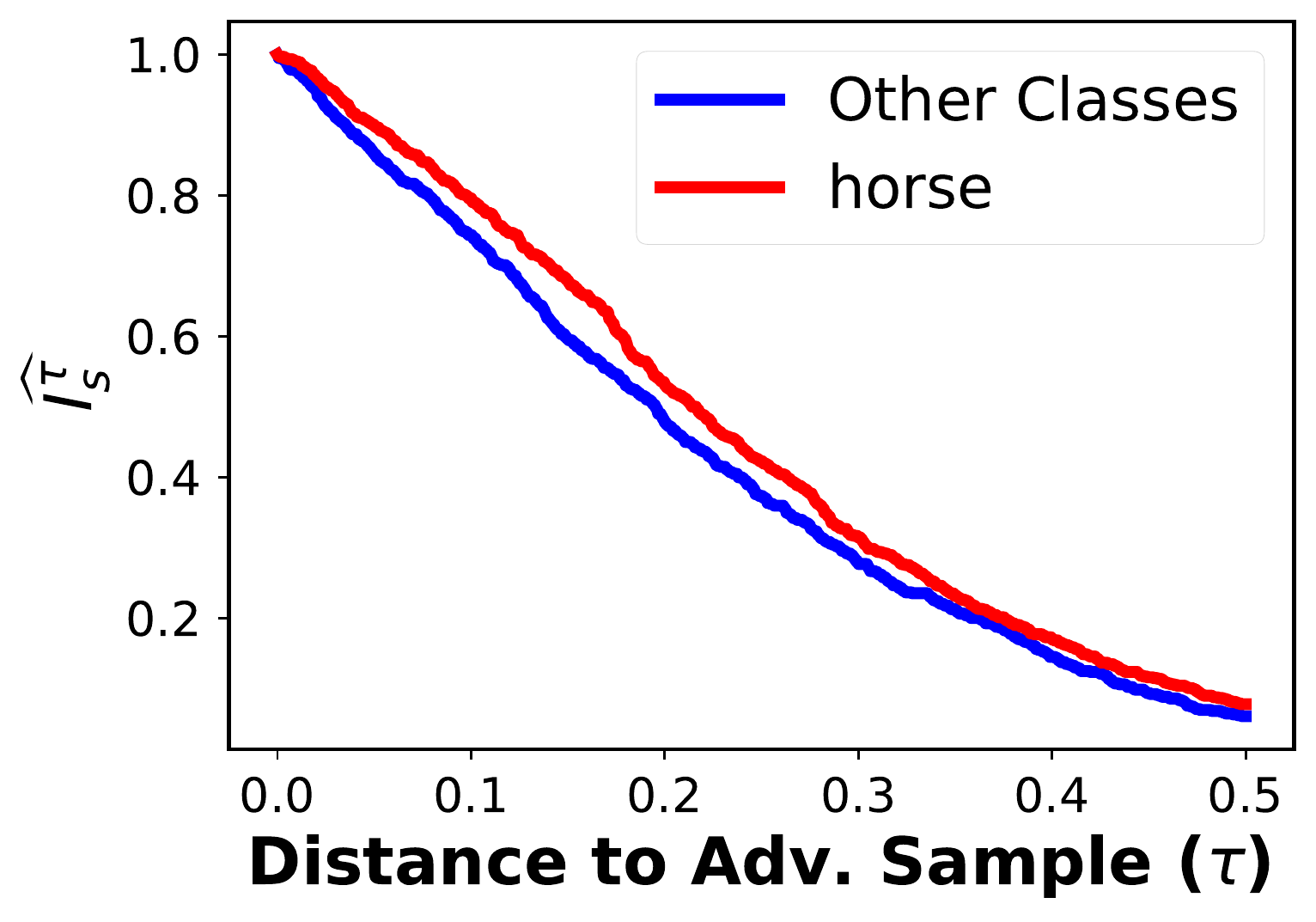}
        \caption{Reg. Horse}
        \label{fig:cifar10_horse_reg_vgg}
    \end{subfigure}
    
    \begin{subfigure}[b]{0.22\textwidth}
        \includegraphics[trim={0cm 0cm 0cm 0cm},clip,width=1\textwidth]{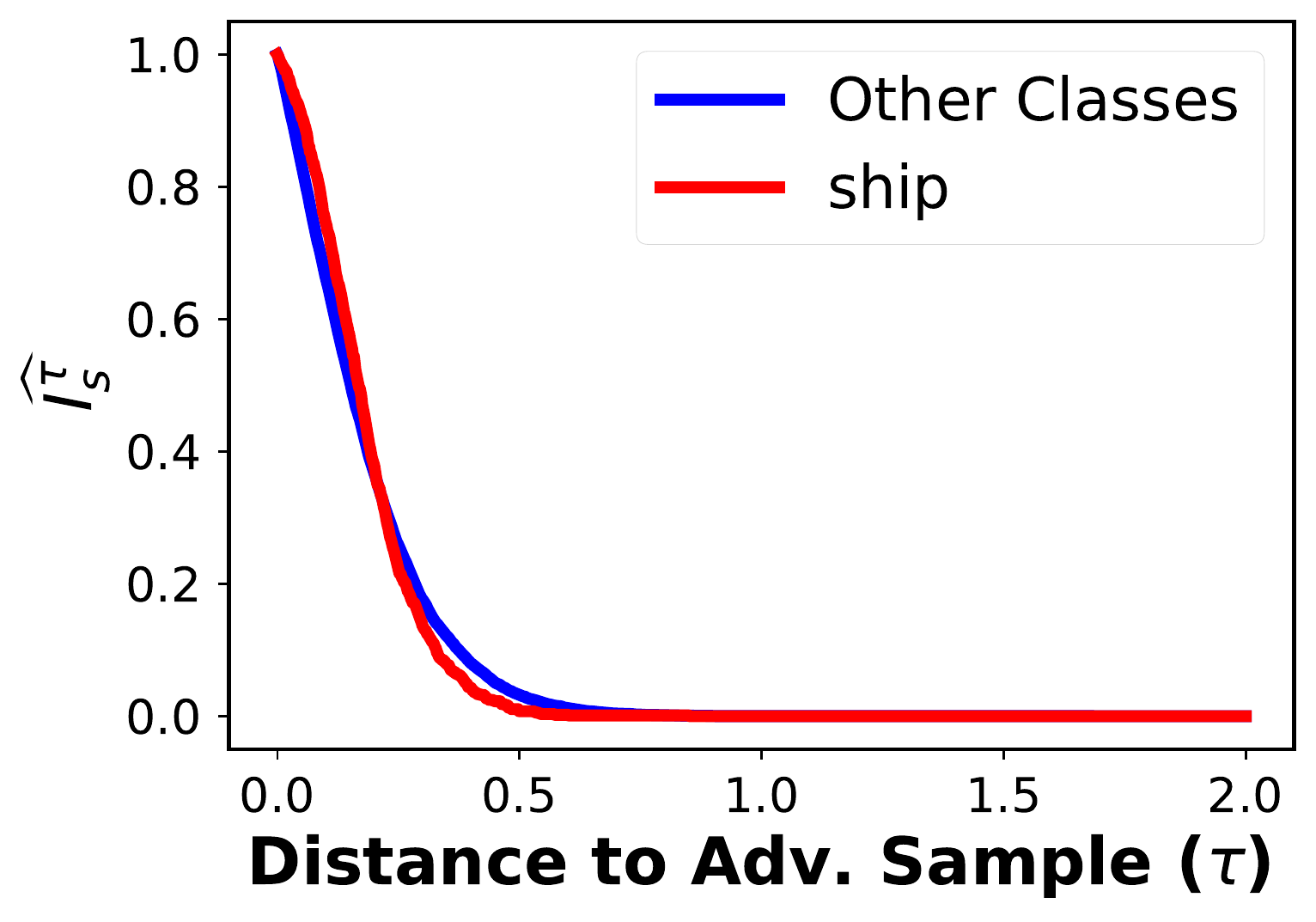}
        \caption{Ship}
        \label{fig:cifar10_ship_vgg}
    \end{subfigure}
    \begin{subfigure}[b]{0.22\textwidth}
        \includegraphics[trim={0cm 0cm 0cm 0cm},clip,width=1\textwidth]{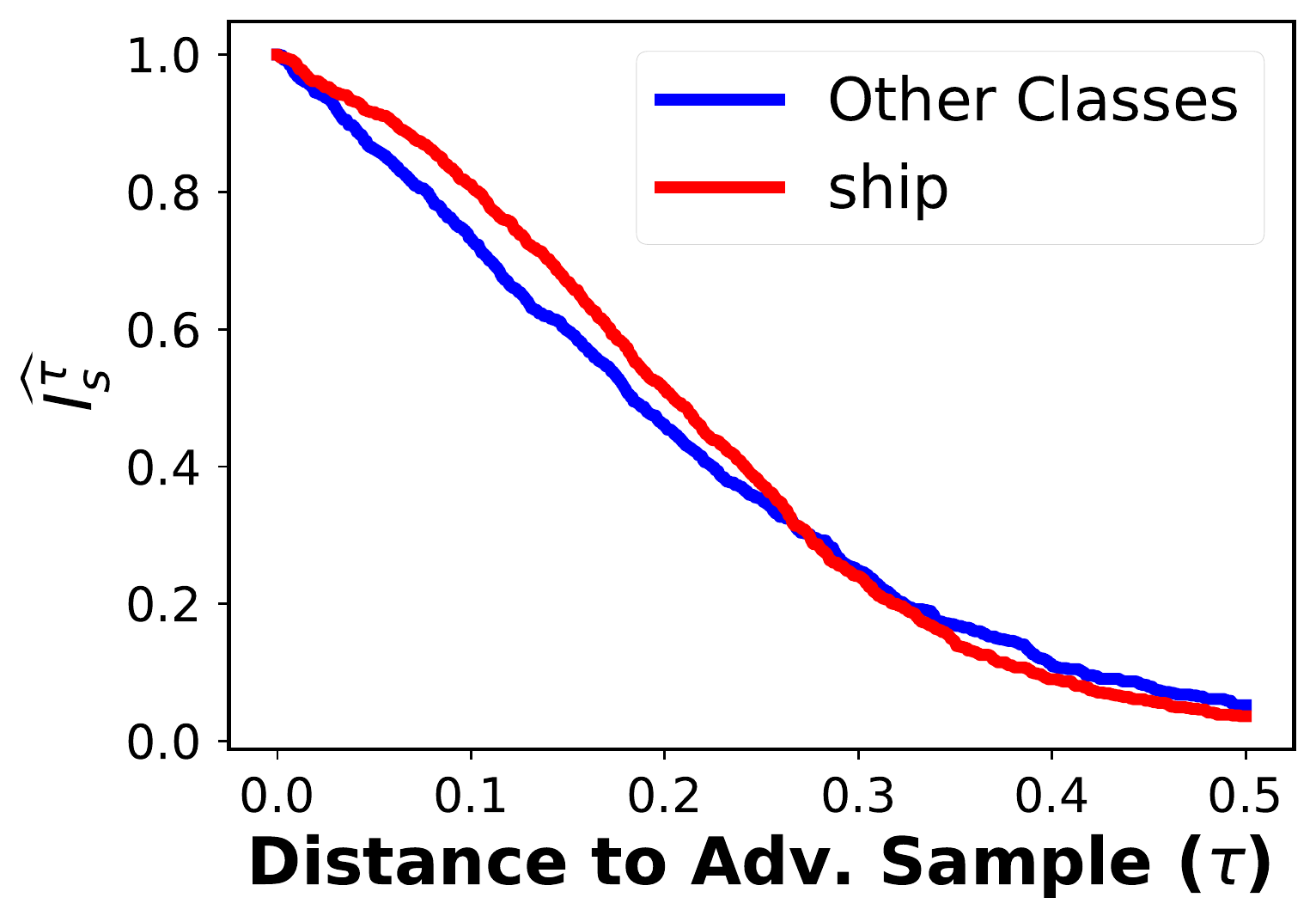}
        \caption{Reg. Ship}
        \label{fig:cifar10_ship_reg_vgg}
    \end{subfigure}
    \begin{subfigure}[b]{0.22\textwidth}
        \includegraphics[trim={0cm 0cm 0cm 0cm},clip,width=1\textwidth]{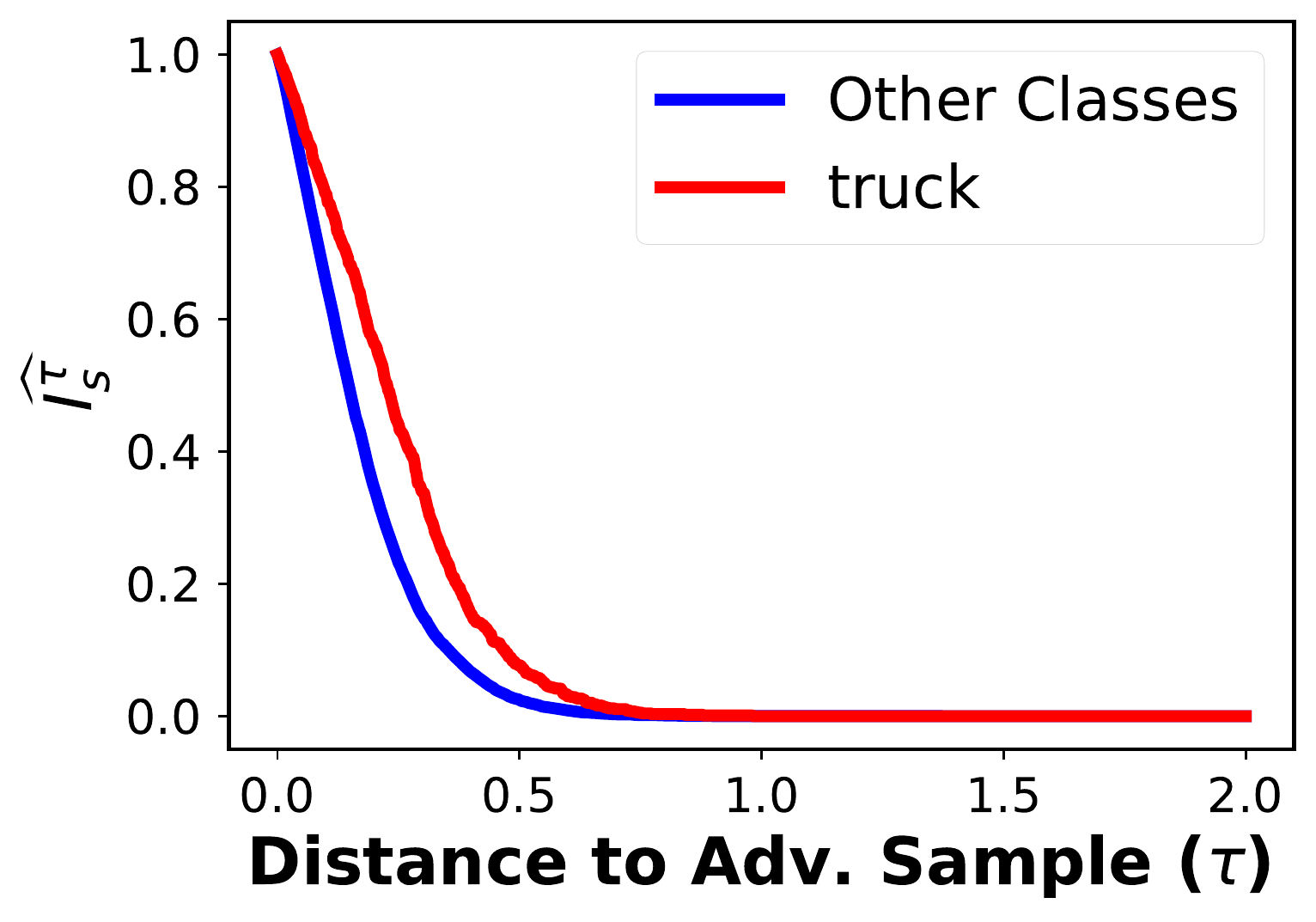}
        \caption{Truck}
        \label{fig:cifar10_truck_vgg}
    \end{subfigure}
    \begin{subfigure}[b]{0.22\textwidth}
        \includegraphics[trim={0cm 0cm 0cm 0cm},clip,width=1\textwidth]{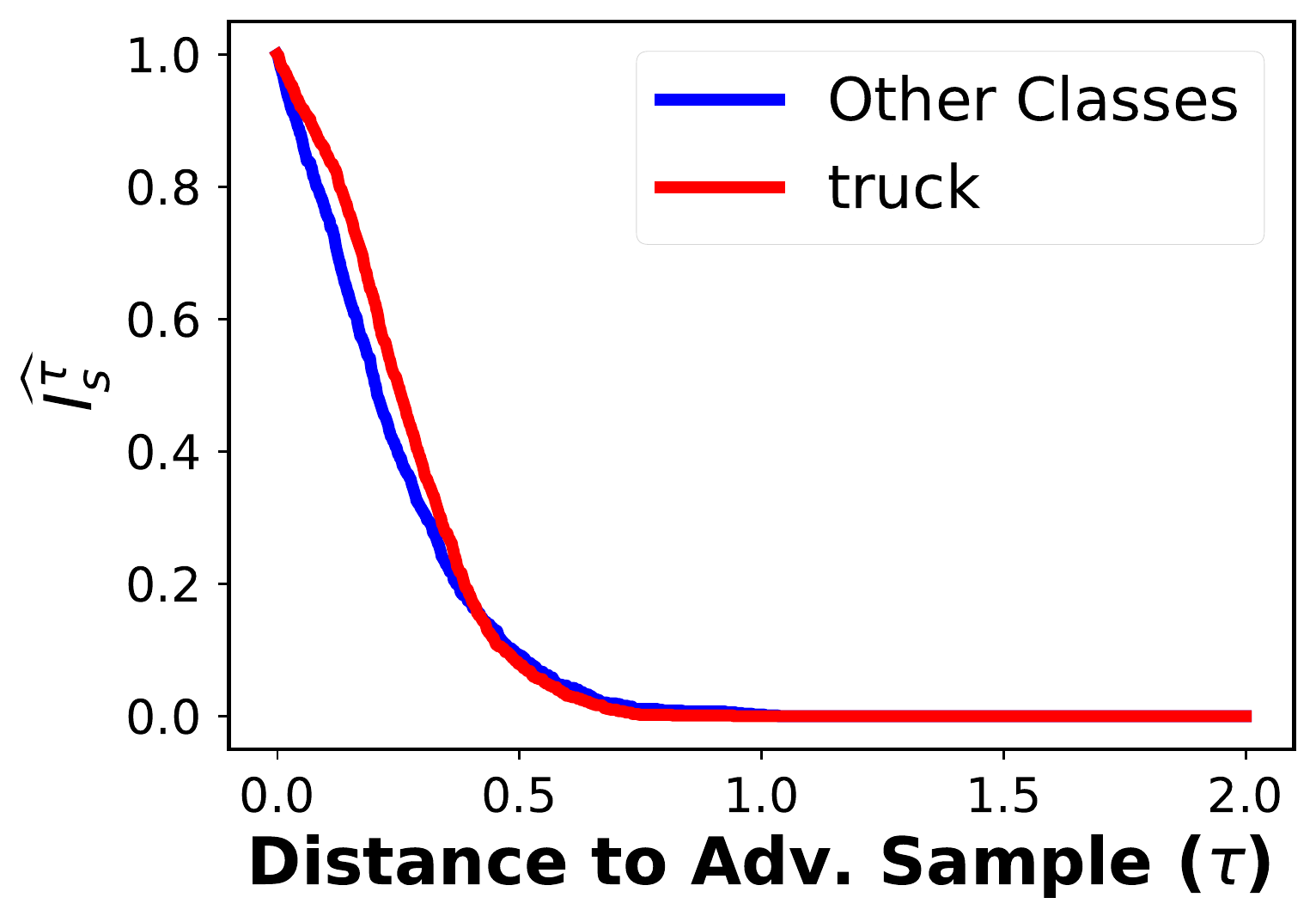}
        \caption{Reg. Truck}
        \label{fig:cifar10_truck_reg_vgg}
    \end{subfigure}

\caption{ [Regularization] CIFAR10 - VGG19
} 
\label{fig:reg_cifar10_vgg}
\end{figure*}

\begin{figure*}[h]
    \centering
    \begin{subfigure}[b]{0.45\textwidth}
        \includegraphics[trim={0cm 0cm 0cm 0cm},clip,width=1\textwidth]{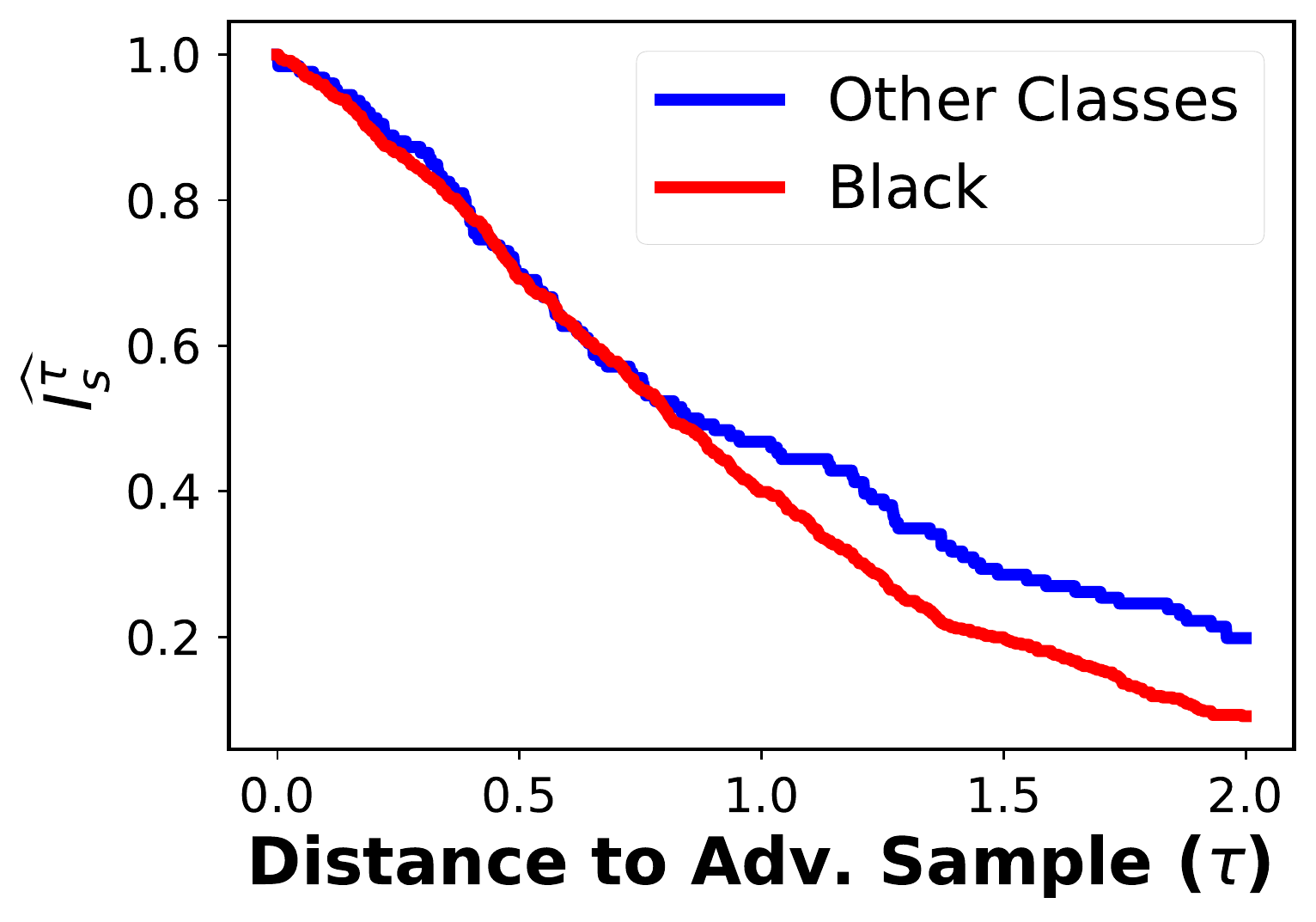}
        \caption{Partitioned by Race}
        \label{fig:utkface_race_utk_classifier}
    \end{subfigure}
    \begin{subfigure}[b]{0.45\textwidth}
        \includegraphics[trim={0cm 0cm 0cm 0cm},clip,width=1\textwidth]{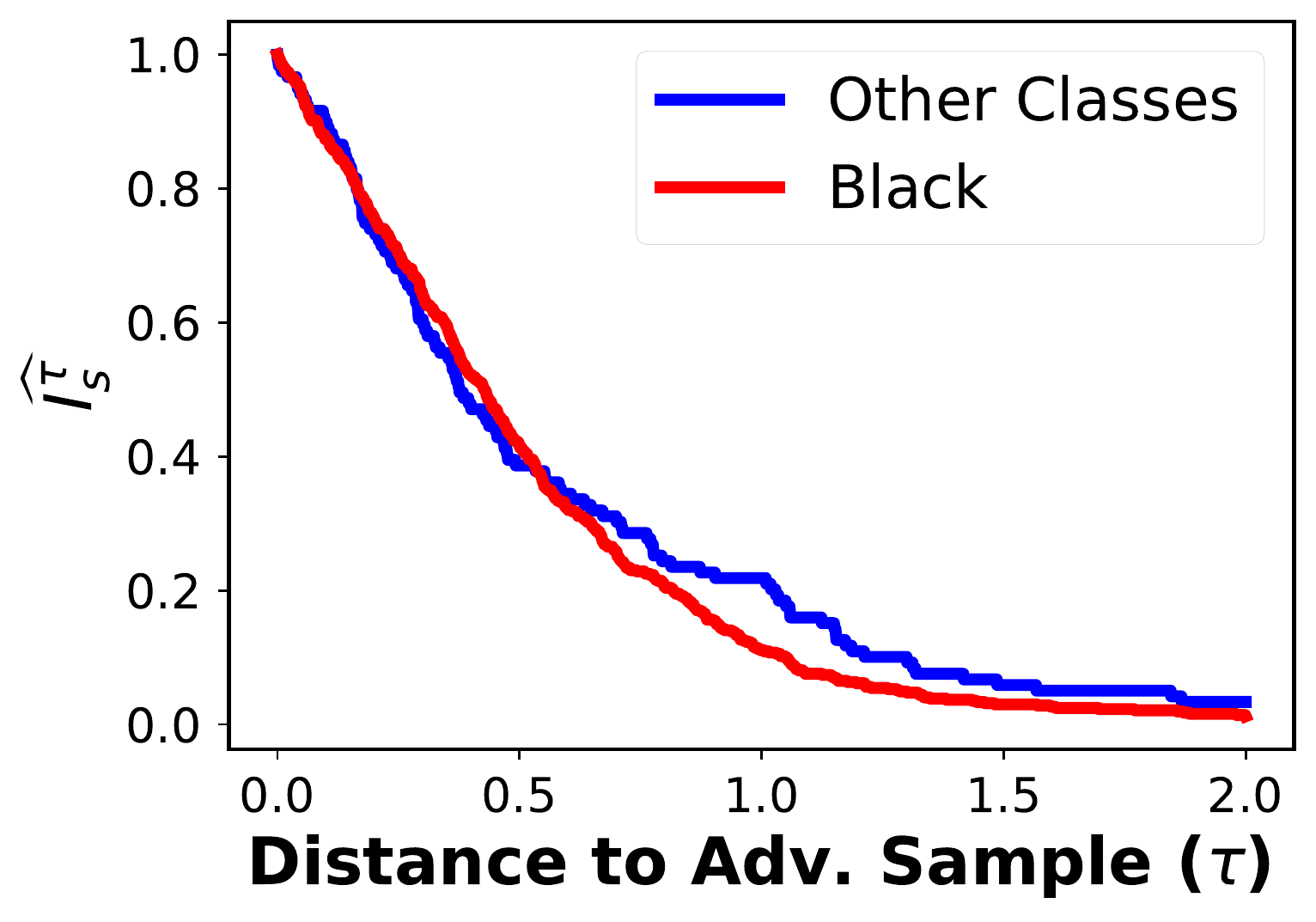}
        \caption{Regularized for Race}
        \label{fig:utkface_race_reg_utk_classifier}
    \end{subfigure}

\caption{ [Regularization] UTKFace partitioned by race - UTK Classifier.
} 
\label{fig:reg_utkface_utk_classifier}
\end{figure*}

\begin{figure*}[h]
    \centering
    \begin{subfigure}[b]{0.45\textwidth}
        \includegraphics[trim={0cm 0cm 0cm 0cm},clip,width=1\textwidth]{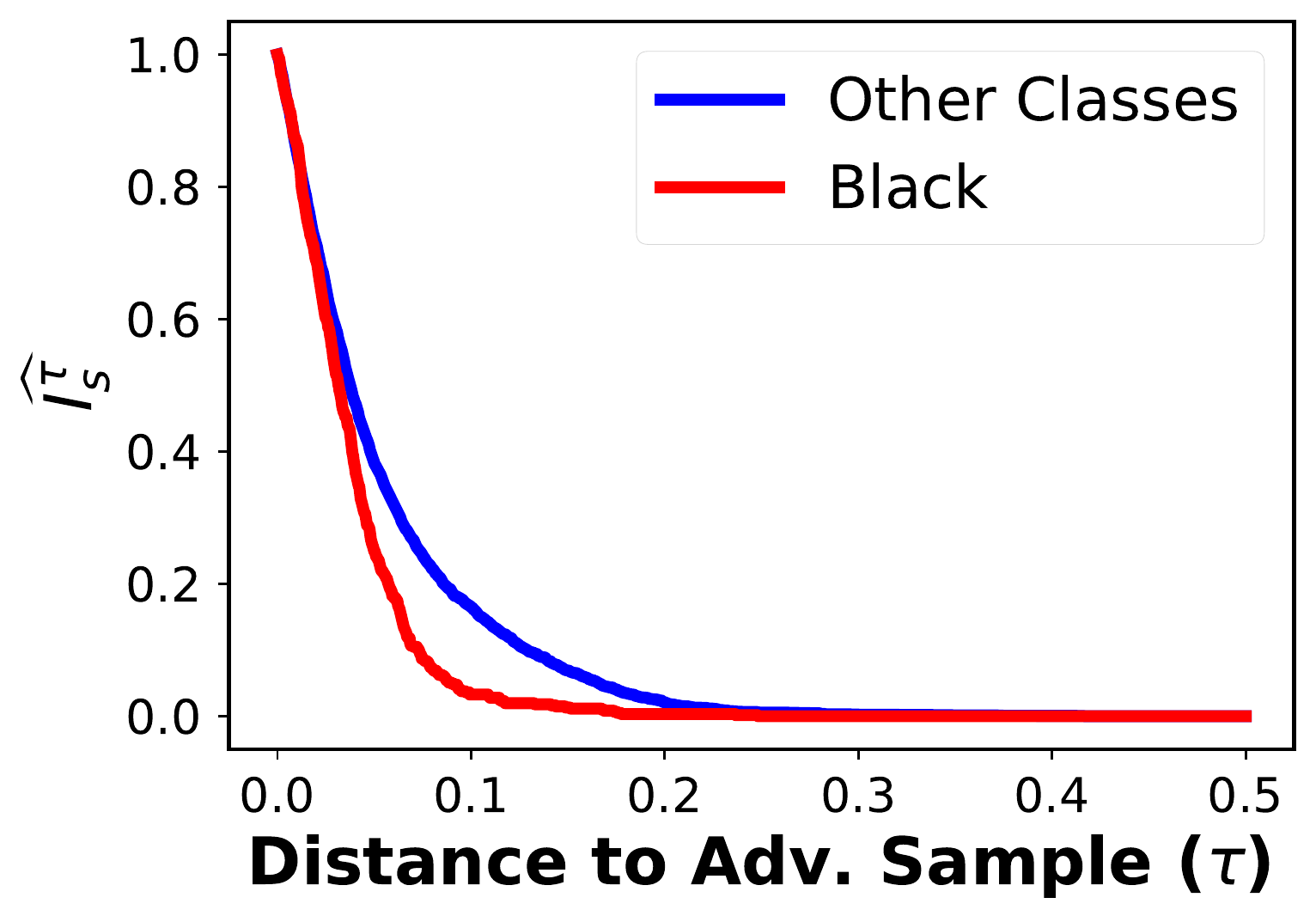}
        \caption{Partitioned by Race}
        \label{fig:utkface_race_resnet}
    \end{subfigure}
    \begin{subfigure}[b]{0.45\textwidth}
        \includegraphics[trim={0cm 0cm 0cm 0cm},clip,width=1\textwidth]{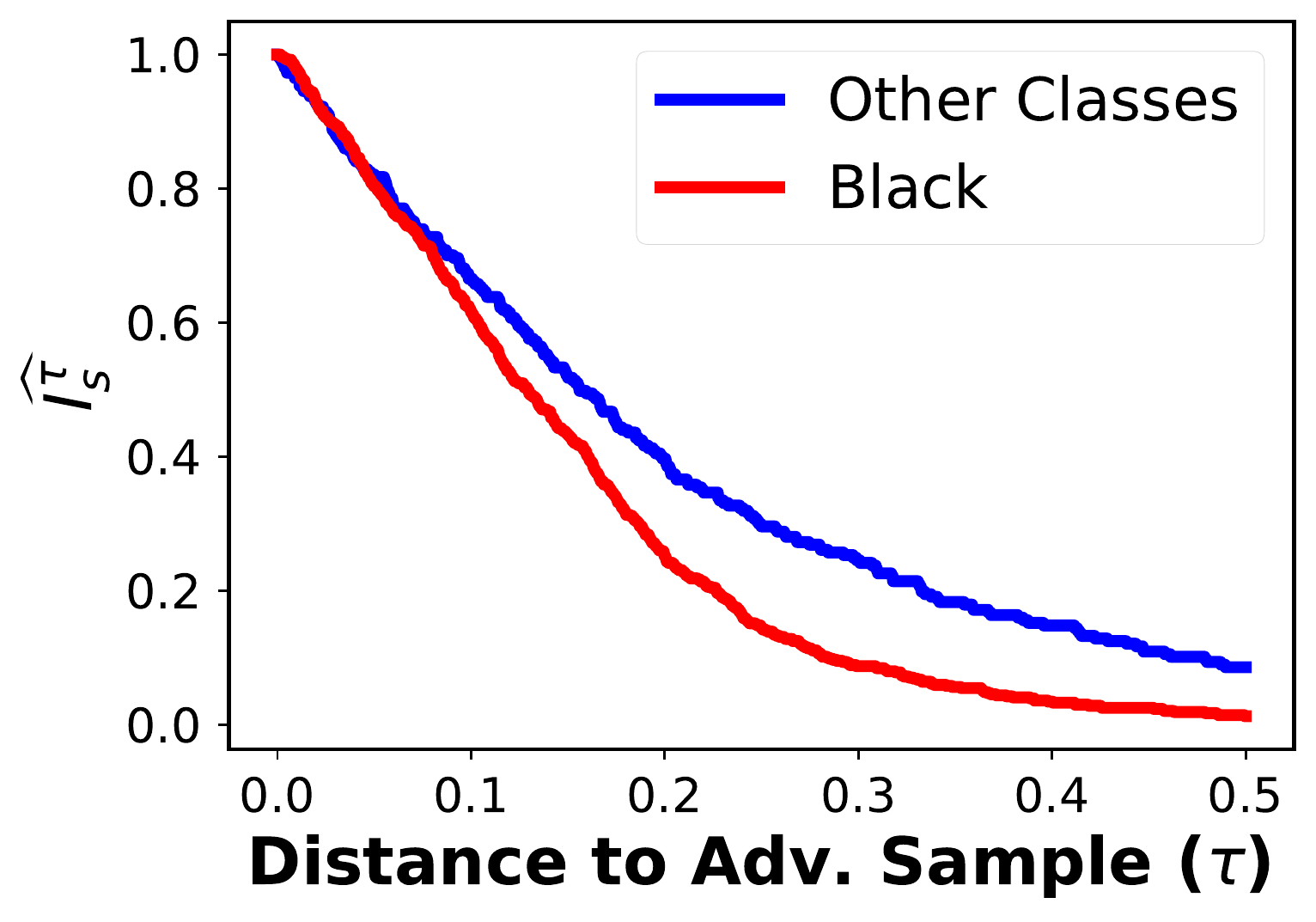}
        \caption{Regularized for Race}
        \label{fig:utkface_race_reg_resnet}
    \end{subfigure}

\caption{ [Regularization] UTKFace partitioned by race - Resnet50.
} 
\label{fig:reg_utkface_resnet}
\end{figure*}

\begin{figure*}[h]
    \centering
    \begin{subfigure}[b]{0.45\textwidth}
        \includegraphics[trim={0cm 0cm 0cm 0cm},clip,width=1\textwidth]{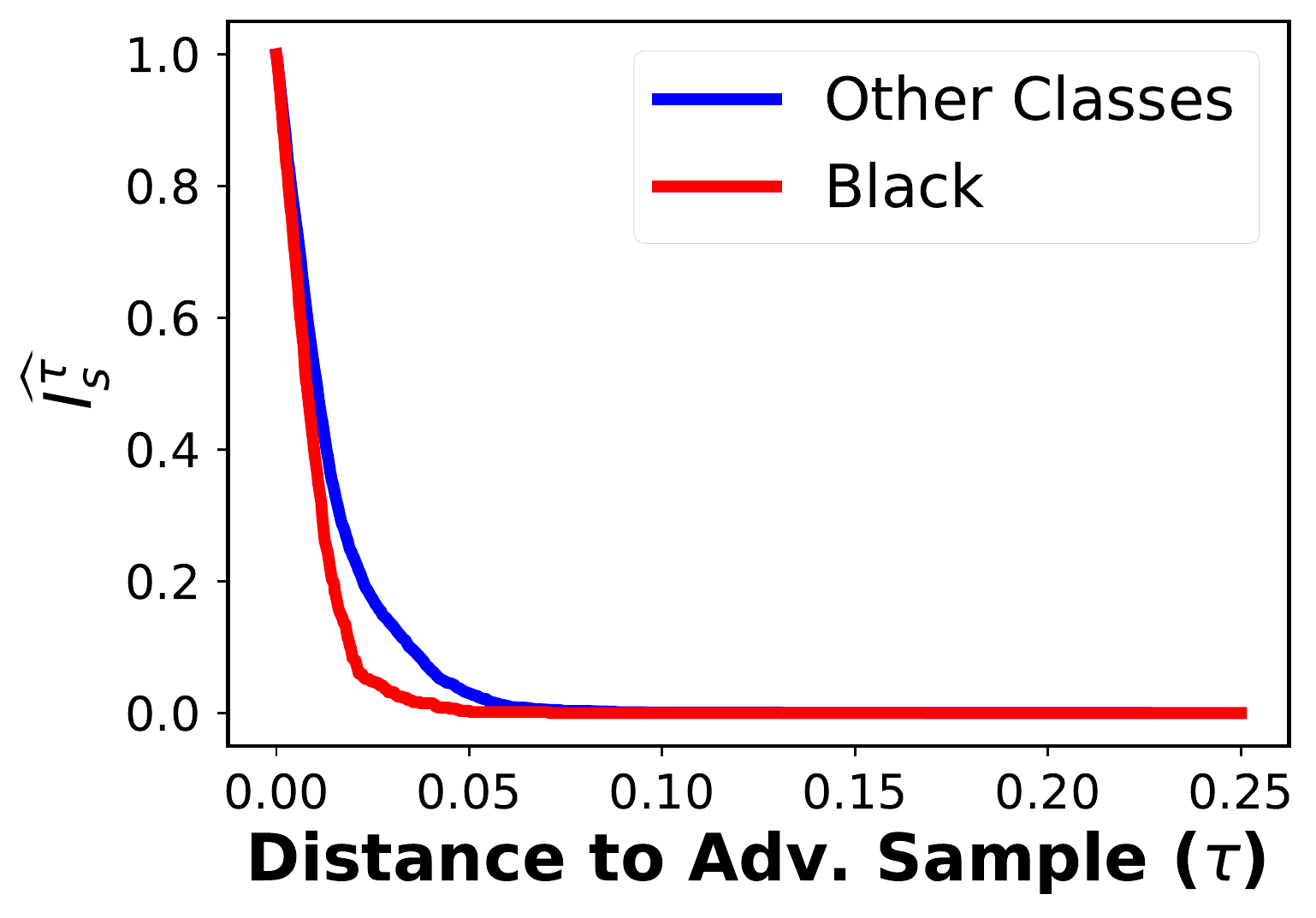}
        \caption{Partitioned by Race}
        \label{fig:utkface_race_vgg}
    \end{subfigure}
    \begin{subfigure}[b]{0.45\textwidth}
        \includegraphics[trim={0cm 0cm 0cm 0cm},clip,width=1\textwidth]{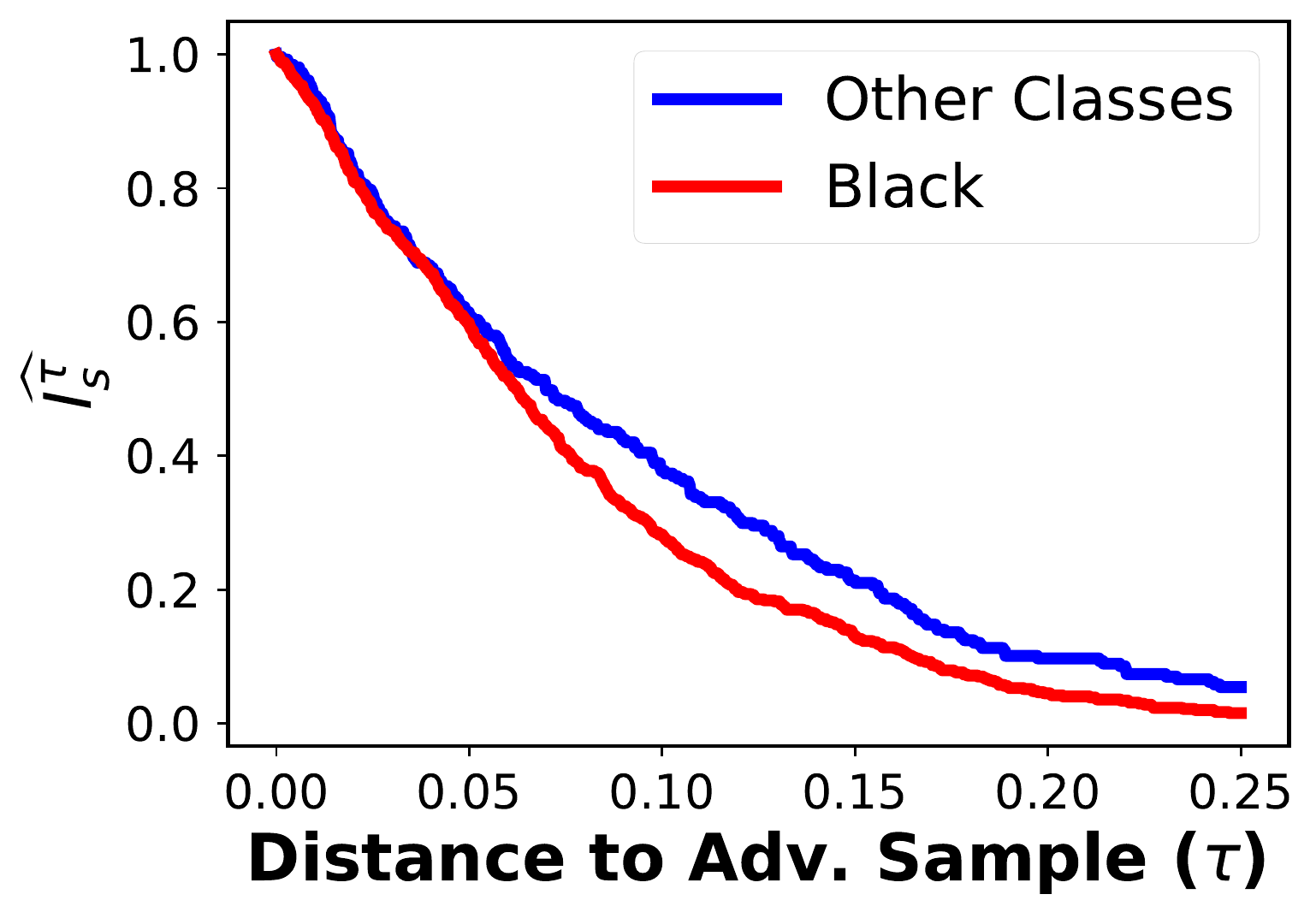}
        \caption{Regularized for Race}
        \label{fig:utkface_race_reg_vgg}
    \end{subfigure}

\caption{ [Regularization] UTKFace partitioned by race - VGG.
} 
\label{fig:reg_utkface_vgg}
\end{figure*}

Figures~\ref{fig:reg_cifar10_deep_cnn},~\ref{fig:reg_cifar10_resnet},~\ref{fig:reg_cifar10_vgg},~\ref{fig:reg_utkface_utk_classifier},~\ref{fig:reg_utkface_resnet},~\ref{fig:reg_utkface_vgg} shows how models trained with our proposed regularization term show lesser robustness bias. Figures~\ref{fig:reg_cifar10_deep_cnn},~\ref{fig:reg_cifar10_resnet}, and~\ref{fig:reg_cifar10_vgg} correspond to CIFAR-10, while Figures~\ref{fig:reg_utkface_utk_classifier},~\ref{fig:reg_utkface_resnet}, and~\ref{fig:reg_utkface_vgg} correspond to UTKFace.  For example, for Deep CNN trained on CIFAR-10, for the partition ``cat'', we see that the distribution of distances become less disparate for the regularized model (Figure~\ref{fig:cifar10_cat_reg}) as compared to the original non-regularized model (Figure~\ref{fig:cifar10_cat}). This trend persists across models and datasets. We provide a PyTorch implementation of the proposed regularization term in our accompanying code.

\section{Comparison of Approximation(s) using Randomized Smoothing and Adversarial Attacks}\label{sec:lower_bounds_experiments}

We present the percent agreement in the signedness of $\sigma^{\text{RS}}$ with $\sigma^{\text{DF}}$ and the signedness of $\sigma^{\text{RS}}$ with $\sigma^{\text{CW}}$. These results are presented in Table \ref{RS-DF-CW}.

\begin{table*}[b]
\centering
\caption{Percentage of agreement with $\sigma^{\text{RS}}$ and $\sigma^{\text{DF}}$ and $\sigma^{\text{CW}}$}
\label{RS-DF-CW}
\begin{tabular}{l|cc}
\hline

        & DeepFool          & CarliniWagner  \\
\midrule
Adience & 83.33 & 79.63\\
UTKFace & 74.24 & 72.73\\
CIFAR10 & 75.00 & 76.67\\
CIFAR100 & 53.83 & 54.00\\
CIFAR100super & 65.83 & 66.67\\
\bottomrule
\end{tabular}
\end{table*}

\setlength{\tabcolsep}{0.5em}
\begin{table*}[t!]
\centering
\caption{Test data performance of all models on different datasets.}
{\renewcommand{\arraystretch}{1.25}
\begin{tabular}{lccccccccc}
\hline

 & \multicolumn{1}{c}{\textbf{\begin{tabular}[c]{@{}l@{}}Deep\\CNN\\(CIFAR100)\end{tabular}}}
 & \multicolumn{1}{c}{\textbf{\begin{tabular}[c]{@{}l@{}}PyramidNet\end{tabular}}}
 & \multicolumn{1}{c}{\textbf{\begin{tabular}[c]{@{}l@{}}Adience\\Classifier\end{tabular}}}
 & \multicolumn{1}{c}{\textbf{\begin{tabular}[c]{@{}l@{}}UTK\\Classifier\end{tabular}}}
 & \multicolumn{1}{c}{\textbf{\begin{tabular}[c]{@{}l@{}}Resnet50\end{tabular}}}
 & \multicolumn{1}{c}{\textbf{\begin{tabular}[c]{@{}l@{}}Alexnet\end{tabular}}}
 & \multicolumn{1}{c}{\textbf{\begin{tabular}[c]{@{}l@{}}VGG\end{tabular}}}
 & \multicolumn{1}{c}{\textbf{\begin{tabular}[c]{@{}l@{}}Densenet\end{tabular}}}
 & \multicolumn{1}{c}{\textbf{\begin{tabular}[c]{@{}l@{}}Squeeze-\\net\end{tabular}}}\\ \hline
\multicolumn{1}{l||}{\textbf{Adience}}
& \multicolumn{1}{c}{-}
& \multicolumn{1}{c}{-}
& \multicolumn{1}{c}{48.80}
& \multicolumn{1}{c}{-}
& \multicolumn{1}{c}{49.75} 
& \multicolumn{1}{c}{46.04}
& \multicolumn{1}{c}{51.41}
& \multicolumn{1}{c}{50.80}
& \multicolumn{1}{c}{49.49} \\
\hline
\multicolumn{1}{l||}{\textbf{UTKFace}}
& \multicolumn{1}{c}{-}
& \multicolumn{1}{c}{-}
& \multicolumn{1}{c}{-}
& \multicolumn{1}{c}{66.25}
& \multicolumn{1}{c}{69.82} 
& \multicolumn{1}{c}{68.09}
& \multicolumn{1}{c}{69.89}
& \multicolumn{1}{c}{69.15}
& \multicolumn{1}{c}{70.73} \\
\hline
\multicolumn{1}{l||}{\textbf{CIFAR10}}
& \multicolumn{1}{c}{86.97}
& \multicolumn{1}{c}{86.92}
& \multicolumn{1}{c}{-}
& \multicolumn{1}{c}{-}
& \multicolumn{1}{c}{83.26} 
& \multicolumn{1}{c}{92.08}
& \multicolumn{1}{c}{89.53}
& \multicolumn{1}{c}{85.17}
& \multicolumn{1}{c}{76.97} \\
\hline
\multicolumn{1}{l||}{\textbf{CIFAR100}}
& \multicolumn{1}{c}{59.60}
& \multicolumn{1}{c}{56.42}
& \multicolumn{1}{c}{-}
& \multicolumn{1}{c}{-}
& \multicolumn{1}{c}{55.81} 
& \multicolumn{1}{c}{71.31}
& \multicolumn{1}{c}{64.39}
& \multicolumn{1}{c}{61.05}
& \multicolumn{1}{c}{40.36} \\
\hline
\multicolumn{1}{l||}{\textbf{CIFAR100super}}
& \multicolumn{1}{c}{71.78}
& \multicolumn{1}{c}{67.55}
& \multicolumn{1}{c}{-}
& \multicolumn{1}{c}{-}
& \multicolumn{1}{c}{67.27} 
& \multicolumn{1}{c}{80.7}
& \multicolumn{1}{c}{76.06}
& \multicolumn{1}{c}{71.22}
& \multicolumn{1}{c}{55.16} \\
\bottomrule
\end{tabular}
}
\label{tab:conv_results_tbl}
\end{table*}

\end{document}